%% file: main.tex
\documentclass[11pt]{article}

\usepackage{jmlr2e}
\usepackage{pifont}
\usepackage{hyperref}
\hypersetup{
    colorlinks=true,
    linkcolor=blue,
    filecolor=magenta,      
    urlcolor=blue,
    citecolor=blue,
}

\usepackage{amsmath}
\usepackage{indentfirst}
\usepackage{amsfonts}
\usepackage{amssymb}
\usepackage{color}
\usepackage[table]{xcolor}
\usepackage{comment}
\usepackage{cleveref}
\usepackage{enumitem}
\usepackage[font=footnotesize,skip=3pt,subrefformat=parens]{subcaption}

\definecolor{darkspringgreen}{rgb}{0.09, 0.45, 0.27}
\definecolor{bistre}{rgb}{0.24, 0.17, 0.12}
\definecolor{auburn}{rgb}{0.43, 0.21, 0.1}
\def\projectpage/{https://trends-in-motion-prediction-2025.github.io/}
\usepackage{pdflscape}
\usepackage{threeparttable}
\usepackage{multirow}
\usepackage{booktabs}
\usepackage{csquotes}
\definecolor{britishracinggreen}{rgb}{0.0, 0.26, 0.15}

\newcommand{\abs}[1]{\left\vert#1\right\vert}
\newcommand{\norm}[1]{\left\lVert#1\right\rVert}
\usepackage{utfsym}
\newcommand{\cmark}{\checkmark}
\newcommand{\xmark}{\scalebox{0.75}{\usym{2613}}}

\usepackage[linesnumbered,ruled,vlined]{algorithm2e}

\firstpageno{1}

\begin{document}



\title{\Large{Trends in Motion Prediction Toward \\Deployable and Generalizable Autonomy:\\
    A Revisit and Perspectives}}







\author{\name Letian Wang \email lt.wang@mail.utoronto.ca \\
       \addr University of Toronto
        \AND
        \name Marc-Antoine Lavoie \email marc-antoine.lavoie@robotics.utias.utoronto.ca \\
       \addr University of Toronto
        \AND
        \name Sandro Papais \email sandro.papais@robotics.utias.utoronto.ca \\
       \addr University of Toronto 
        \AND
        \name Barza Nisar \email barza.nisar@robotics.utias.utoronto.ca \\
       \addr University of Toronto
        \AND
        \name Yuxiao Chen \email yuxiaoc@nvidia.com \\
       \addr NVIDIA Research
       \AND
        \name Wenhao Ding \email wenhaod@nvidia.com \\
       \addr NVIDIA Research
        \AND
       \name Boris Ivanovic \email bivanovic@nvidia.com \\
       \addr NVIDIA Research
       \AND
        \name Hao Shao \email hao.shao@link.cuhk.edu.hk \\
       \addr Chinese University of Hong Kong
       \AND
       \name Abulikemu Abuduweili \email abulikea@andrew.cmu.edu \\
       \addr Carnegie Mellon University
       \AND
        \name Evan Cook \email evan.cook@robotics.utias.utoronto.ca \\
       \addr University of Toronto
       \AND
        \name Yang Zhou \email yang.zhou@robotics.utias.utoronto.ca \\
       \addr University of Toronto 
       \AND
        \name Peter Karkus \email pkarkus@nvidia.com \\
       \addr NVIDIA Research
       \AND
       \name Jiachen Li \email jiachen.li@ucr.edu \\
       \addr University of California, Riverside 
       \AND
       \name Changliu Liu \email cliu6@andrew.cmu.edu \\
       \addr Carnegie Mellon University
       \AND
       \name Marco Pavone \email pavone@stanford.edu, mpavone@nvidia.com \\
       \addr Stanford University, NVIDIA Research
       \AND
       \name Steven L. Waslander \email steven.waslander@robotics.utias.utoronto.ca \\
       \addr University of Toronto}


\maketitle

\noindent Note: This paper has been submitted to journal for review and the current version is not final. Suggestions are welcomed! Personal use of this material is permitted. Permission from the authors must be obtained for all other uses, in any current or future media, including reprinting/republishing this material for advertising or promotional purposes, creating new collective works, for resale or redistribution to servers or lists, or reuse of any copyrighted component of this work in other works.

\newpage
\begin{abstract} 

Motion prediction, recently popularized under the term world models, refers to anticipating the future states of agents or the future evolution of a scene, which is rooted in human cognition to bridge perception and decision-making, enabling us to anticipate, adapt, and act within an ever-changing world. 
It lies at the core of intelligent autonomous systems, such as robotics and self-driving cars, to safely operate in dynamic and human-robot-mixed environments, and also informs broader time-series challenges.
With advances in methods, representations, and datasets, the field has seen rapid progress, reflected in rapidly updated benchmark performance.
However, when state-of-the-art methods are deployed in the real world, they are often found to struggle to generalize to open-world settings and fall short of deployment standards. 
This reveals a gap between reality and benchmarks, which are often idealized or ill-posed, and fail to capture real-world complexity.

To address the pressing need for problem settings that better reflect real-world challenges and guide future research, this paper focuses on revisiting the generalization and applicability of motion prediction models, with an emphasis on robotics, autonomous driving, and human motion applications.
We first provide a comprehensive taxonomy of motion prediction methods, covering representations, modelling methods, application domains, and evaluation protocols.
We then revisit two fundamental problems: 1) how to push motion prediction models to be \textit{deployable} to realistic deployment standards, where motion prediction does not act in a vacuum, but functions as one module of closed-loop autonomy stacks - it takes input from the localization and perception, and informs downstream planning and control. 2) how to \textit{generalize} motion prediction models from limited seen scenarios/datasets to the open-world settings. 
We conclude by highlighting crucial challenges and open problems for future research. By doing so, we aim to recalibrate the community’s efforts, fostering progress that is not only measurable but also meaningful for real-world applications.
The project webpage corresponding to this paper can be found here \href{\projectpage/}{\projectpage/}.

\end{abstract}
\newpage

\begin{figure}[t!]
    \centering
    \includegraphics[width=0.72\textwidth]{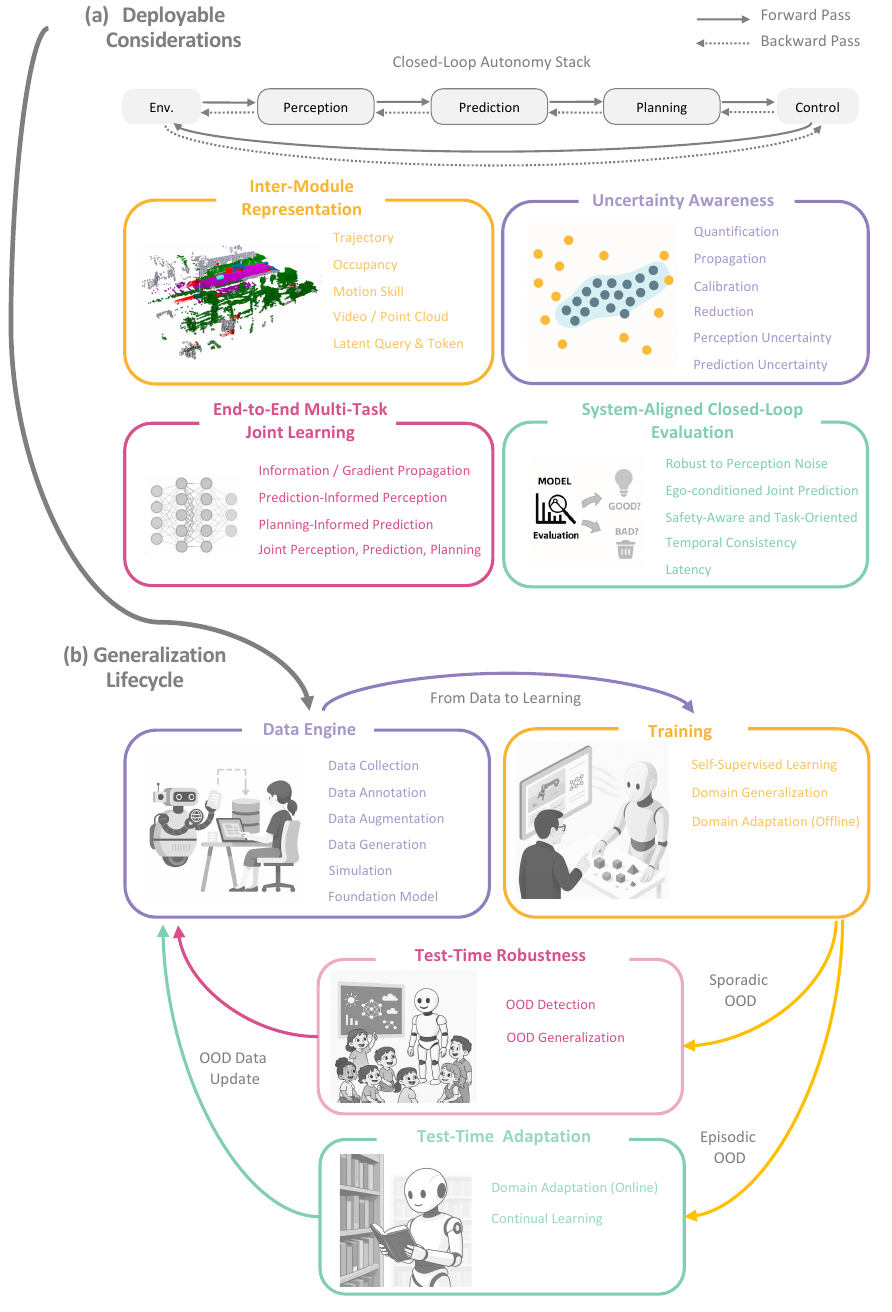}
    \vspace{-1em}
    \caption{\footnotesize{Roadmap toward deployable and generalizable models for autonomous systems. (a) Models intended for real-world deployment should be developed and evaluated under realistic settings, which require: (1) adopting inter-module representations that are informative, efficient, and scalable with data; (2) managing uncertainty throughout the autonomy stack; (3) enabling joint learning across modules to promote information sharing and resolve incompatibilities; and (4) aligning evaluation with the performance of the full closed-loop system.
    (b) The resulting models and evaluation protocols are then integrated into a generalization cycle that: (1) absorbs diverse data to broaden the support of the training distribution; (2) learns representations capable of generalizing across a wide range of operational domains; (3) handles sporadic distribution shifts during deployment to ensure safety; and (4) adapts online to anticipate and respond to episodic distribution changes. The lifelong learning system incrementally updates the database with newly encountered OOD data to expand the data coverage, thereby initiating the next generalization cycle to further expand the operational envelope of the system.}}
    \vspace{-1em}
      \label{fig:vision}
\end{figure}

\tableofcontents
\newpage



\input{section/1_intro}

\input{section/2_taxonomy}

\input{section/3_applicable}
\input{section/4_generalization}

\input{section/5_discussion}

\vskip 0.2in
\bibliography{reference}

\end{document}

%% file: section/1_intro.tex
\section{Introduction}
\label{sec: intro}
\subsection{Background}




When intelligent autonomous systems are deployed in real-world environments, they have to coexist and interact with other users of those spaces. 
This is true for self-driving cars interacting with vehicles and pedestrians on public roads, and for robots collaborating with humans in settings such as manufacturing or home assistance. 
To understand the surroundings and make efficient and safe decisions, these systems need to observe, reason about, and predict the future motion of other space users and the future evolution of the scene.
This is challenging due to the complexity of human behavior and the interdependence between agents and their environments. Factors such as task goals, social norms, interactions between agents, environmental layout, and legal constraints all influence human behavior. The future prediction task is further complicated by the uncertainty and multi-faceted nature of human behavior, the diversity of agents, real-time processing requirements, and sensor limitations.

In light of such complexity, the research community has devoted increasing attention to understanding and modeling the temporal evolution of dynamic objects and scenes, and the field has progressed rapidly, transitioning from early physics- and planning-based paradigms to the prevailing wave of learning-based approaches. This evolution has been accompanied by a growing diversity of representational choices—from compact formats such as trajectories, motion primitives, and flows, valued for their efficiency and interpretability, to sensor-level inputs like video and point clouds, enabled by advances in computational resources and the emergence of the world model concept, and further to latent representations such as tokens and queries. Meanwhile, performance on standard benchmarks has continued to improve, driven by advances in architecture design and representation learning. However, when we apply these motion prediction models to real-world applications, two fundamental limitations become apparent.

First, while performance on public benchmarks has improved rapidly, these evaluations often fail to reflect closed-loop, in-stack performance under real-world deployment standards.
In autonomous systems, motion prediction is one component of a larger system that typically includes localization, perception, planning, and control. The motion prediction module relies on inputs from localization and perception, and in turn, informs planning and control decisions. 
Usually, existing benchmark and methods tend to assume perfect and noise-free upstream perception input data, usually utilize trajectories as the interface between upstream and downstream modules, ignore how the prediction outputs can be compatibaly consumed by downstream planner, and focus on pure open-loop prediction accuracy.
However, such settings are clearly over-simplified and unrealistic when we deploy motion prediction models in practice, where 1) improper interface/representation design between modules can lead to information loss, incompatibility between modules, and difficulty in scaling to large-scale data; 2) uncertainty and errors inevitably arise from both the sensor measurements and perception algorithms, which are propagated and combined with prediction uncertainty/errors to downstream planning and control; 3) performance improvements of the motion prediction module alone do not necessarily translate into the improvement of the whole system, especially when learning and evaluation are conducted in open-loop and in isolation from upstream perception and downstream planning, and the requirements of adjacent modules are ignored.
Due to insufficient consideration of these practical issues, many state-of-the-art prediction methods tend to diverge from real-world conditions and exhibit limited applicability in deployment. Thus, as illustrated in Figure~\ref{fig:vision}, Figure~\ref{fig:applicable_perceptionintegration}, and Figure~\ref{fig:applicable_planningintegration}, we posit that the research community needs to revisit four key challenges in the deployable of motion prediction methods, including the design of inter-module representations, awareness of uncertainty and errors throughout the autonomy stack, joint learning across modules, and aligning the evaluation with the advancement of overall system performance. 

Another core challenge facing motion prediction methods is the need to generalize to open-world settings. All datasets are finite, and no matter how large, they represent only a narrow snapshot of the behaviors encountered in the real world. This limitation is further exacerbated by the relative scarcity of high-quality motion data—especially when compared to the abundance of data in other fields like text and images. Moreover, out of convenience and due to dominance of domain-specific benchmark, most existing works focus on improving prediction accuracy within specific domains—particular scenarios, applications, or datasets—with limited attention paid to the generalization performance on general environments and cases that could deviate from the limited training distribution.
Nevertheless, when autonomous systems are deployed in real-world environments that could be highly diverse and unstructured, they inevitably encounter novel situations that differ from the training distribution, including previously unseen agents, motion behaviors and contexts.
For instance, self-driving cars may encounter unfamiliar road geometries, varying traffic densities, shifting driving norms, and new regulatory constraints; manufacturing and assistive robots may face different factory or home layouts and evolving task specifications. In such scenarios, optimizing for a single performance metric within a fixed domain can come at the cost of generalization capabilities, often resulting in significant performance degradation in unseen settings and ultimately hindering the large-scale deployment of autonomous systems.

In the literature, various approaches have been proposed to advance the generalization capabilities of predictive models, addressing different aspects such as data, learning signals, and model architectures, as illustrated in Figure~\ref{fig:generalization}. In this paper, we introduce a generalization lifecycle that consists of four key stages, as illustrated in Figure~\ref{fig:vision}:
1) \textit{Lifelong data engine} – aims to absorb as much diverse data as possible to broaden the support of the data distribution;
2) \textit{Training} – focuses on accomplishing the task while leveraging generalizable representations and robust model architectures;
3) \textit{Safety-aware deployment} – applies the model in real-world settings with awareness of prediction confidence and system-level safety, including the ability to detect domain shifts, assess their severity, and trigger fallback strategies when necessary to ensure safe operation;
4) \textit{Test-time adaptation} – adapts the model online to anticipate and respond to episodic distribution shifts, or to actively remain within in-distribution operation regimes.
At the end of each generalization cycle, with a quantifiable measure of domain shift, the system can selectively store and annotate detected out-of-distribution (OOD) data to expand the diversity and coverage of the training distribution. The enriched dataset then triggers the next iteration of the cycle, forming a lifelong learning process that continuously improves the model’s ability to generalize across evolving real-world scenarios.

To summarize, this paper aims to revisit two fundamental and relatively under-explored aspects of motion prediction research: developing models that aredeployable under realistic deployment standards, and generalizing models from limited seen scenarios and datasets to open-world settings. 
With particular emphasis on applications of autonomous driving, robotics, and human motion domains, we begin by presenting a comprehensive taxonomy of motion prediction methods in Section~\ref{sec: taxonomy}, covering key aspects such as representation choices, modeling approaches, application domains, and commonly used evaluation strategies. We then review a wide range of existing methods, focusing on how to ensure practical applicability in Section~\ref{sec:applicable}, and how to improve generalization in Section~\ref{sec:generalizable}. Throughout the review, we also offer perspectives that aim to inspire new research directions and provide critical insights into the future development of the field.
We conclude by identifying key challenges and open research questions in Section~\ref{sec: discussion}, with the aim of recalibrating the community’s efforts—shifting the focus toward progress that is not only benchmark-driven, but also aligned with the complexities and demands of real-world deployment.

\subsection{Problem Formulation}
The motion prediction task aims at generating high-fidelity predictions of the future states for the selected agents or the scene in the next $T_f$ seconds, based on observations of the last $T_h$ seconds $\textbf{o}_{t - T_h, t}$:
\begin{equation}
    p(\hat{\textbf{s}}_{t + 1, t + T_f}) = f(\textbf{o}_{t - T_h, t}).
\end{equation}
where $f$ denotes the prediction model, taking $\textbf{o}_{t - T_h, t}$ as input and $\hat{\textbf{s}}_{t + 1, t + T_f}$ as outputs predictions. The output is typically formulated probabilistically to capture the inherent uncertainty and multimodality of human behavior. The representation of $\textbf{o}_{t - T_h, t}$ and $\textbf{s}_{t + 1, t + T_f}$ can take different forms based on the task and sensor setting. As illustrated in Figure~\ref{fig:representation}, these can include object-centric representations such as trajectories, poses, intentions, and motion skills; scene-centric representations such as environment maps and occupancy grids; or sensor-level representations such as point clouds or videos.
In general, the input $\textbf{o}_{t - T_h, t}$ is designed to capture factors influencing future motion, including both internal stimuli (e.g., historical states, goal intentions) and external contextual cues (e.g., interactions with other agents, map layouts). The output $\hat{\textbf{s}}_{t + 1, t + T_f}$ can be formulated deterministically for simplicity or probabilistically to account for multiple plausible futures.
The prediction model $f$ varies in how it absorbs the input data, designs and trains the model, and parameterizes the output. As in Section~\ref{sec: taxonomy}, we categorize motion prediction models into three categories: physics-based models, planning-based models, and deep-learning based models. In recent years, deep-learning-based methods have gained increasing attention within the research community and dominated in motion prediction benchmarks, due to their excellence in learning complex motion patterns from data.

Deployable motion prediction models require development and evaluation under realistic deployment settings, where:
    1) The input $\mathbf{o}_{t - T_h, t}$ should be carefully designed to capture task-relevant information, remain compact for computational efficiency, and scale effectively with large datasets; the output representation $\hat{\mathbf{s}}_{t + 1, t + T_f}$ should be chosen to facilitate seamless integration with downstream planning and control modules.
    2) Uncertainties in both input and output should be quantified, calibrated, and mitigated during model development to enhance robustness to noise and ensure system-level safety.
    3) The prediction model $f$ should be jointly trained with upstream and downstream modules to promote effective information propagation, ensure compatibility with cross-module constraints, and encourage mutual performance benefits.
    4) The evaluation of model $f$ should be conducted in the context of the full autonomy stack, and ultimately aligned with the overall performance of the closed-loop system.
    



Generalizable motion prediction models aim at mitigating or dealing with domain shift that happens when we scale the model to an open-world setting with wider operation domains. 
Essentially, domain shifts can be categorized into three types: \textit{covariate shift}, referring to changes in the input distribution \( p(\mathbf{o}_{t - T_h, t}) \); \textit{label shift}, involving changes in the output distribution \( p(\hat{\mathbf{s}}_{t + 1, t + T_f}) \); and \textit{concept shift}, which denotes changes in the underlying mapping function \( f \) between input and output.
These types of domain shift are not mutually exclusive and may occur simultaneously. To effectively handle these shifts, generalizable models can be designed through coordinated improvements in learning signals, model architectures, and data diversity.





\subsection{Paper's Scope and Framework}
\textbf{Related Tasks} Motion prediction is closely related to two other areas: \textit{motion generation} and \textit{learned planning}, both of which often share similar formulations and methodologies with prediction models.

\begin{itemize}[label={\scriptsize$\bullet$}]
\vspace{-0.5em}
\item \textit{World Models}, a relatively new term that has gained traction in recent years, share many similarities with motion prediction, as both aim to anticipate the future evolution of a scene. One key difference is that world models often place greater emphasis on generation—for example, synthesizing plausible future sensor observations—rather than purely forecasting object-level trajectories. Another important distinction lies in their choice of representation: motion prediction typically relies on object-centric abstractions (e.g., bounding boxes), whereas world models operate directly at the sensor level (e.g., videos, point clouds, voxels).
These differing representations lead to complementary strengths and limitations. Object-centric motion prediction offers stronger object awareness and interpretability, facilitates explicit modeling of agent interactions, and is computationally efficient by abstracting the scene into discrete objects. However, it inevitably loses fine-grained details and struggles to define objects in open-world settings. In contrast, world models can capture rich, detailed information without relying on predefined object classes, enabling a more holistic and open-ended view of the environment. Yet, this comes at the cost of higher computational demands and challenges in maintaining object consistency over time. 
We expand the discussions on the representations in Section~\ref{sec: taxonomy - representation}, Section~\ref{sec: applicable perception-prediction representation}, and Section~\ref{sec: app - prediction-planning representation}. We also specifically discussed self-supervised learning methods including video generation, point cloud forecasting, and scene reconstruction in Section~\ref{sec: SSL - point cloud} and Section~\ref{sec: SSL - video}, which are more closely related to world models.

\item \textit{Motion generation} refers to the synthesis of realistic motion, that spans across diverse domains such as road traffic, human motion, and video generation. With the rise of generative AI, it has become a key tool for simulation, content creation, and data augmentation. While both motion prediction and motion generation aim to produce realistic motion, the former emphasizes \textit{accuracy} with observed behavior, whereas the latter places greater emphasis on \textit{diversity} and \textit{controllability}. Moreover, unlike motion prediction, which must operate within a closed-loop autonomy stack and interface with upstream perception and downstream planning, motion generation---particularly when used outside of autonomous systems (e.g., in animation or generative media)---is unconstrained by such architectural dependencies. This allows for greater flexibility in generation settings, without the need for real-time operation or module compatibility. We expanded the disucssions on motion generation in Section~\ref{sec:data_synthesis}.

\vspace{-0.5em}
\item \textit{Learned planning} focuses on generating the ego agent’s future behavior to accomplish a task. While its architectures and inputs may resemble those used in motion prediction, its objective is fundamentally different: rather than forecasting what other agents will do, learned planning determines what the ego agent \textit{should} do. As a result, compared to motion prediction models that generate multimodal outcomes, the ultimate output of learned planners is often unimodal, predicting a single, task-directed trajectory. Moreover, while motion prediction typically involves joint forecasting for multiple agents, learned planning usually focuses solely on the single ego agent and can therefore be viewed as a form of marginal prediction. Furthermore, their role within the autonomy stack is distinct: learned planners typically sit downstream of prediction and directly inform control, while prediction directly interacts with perception and planning. 
In practice, the boundary between prediction and planning may blur, as prediction models are frequently adapted for use in learned planning---for example, by repurposing predicted trajectories as ego decisions---highlighting their shared foundations despite differing objectives. We discuss the joint learning of prediction and planning, and end-to-end planners in Section~\ref{sec: app - planning-informed prediction}.

\vspace{-0.5em}

\end{itemize}

In this survey, we focus primarily on motion prediction, particularly learning-based methods, given their dominance and proven success in the field, with an emphasis on deployment and generalization challenges. However, motion prediction, world models, motion generation, and learned planning are closely intertwined, often sharing similar model architectures and learning paradigms. Consequently, we also cover representative works from world models, motion generation and learned planning, and many of the insights presented are broadly applicable across these related domains.




\textbf{Application Domains} 
Motion prediction encompasses a wide range of applications. This survey primarily focuses on domains related to robotics and autonomous driving, where prediction plays a critical role in enabling safe and intelligent decision-making. However, many other application areas that involve general time-series prediction, such as surveillance, monitoring, virtual/augmented reality, and sports movement analysis, can benefit from similar motion prediction methodologies and insights. For a comprehensive overview and detailed discussion of these broader applications, readers are referred to Sec~\ref{sec: application domains}.

Moreover, while this survey covers motion prediction across autonomous systems—including both autonomous driving and robotics—we find the literature to be disproportionately focused on driving. This reflects deeper differences in dataset maturity, environmental structure, and task consistency between the two domains.
In Chapter~\ref{sec: taxonomy}, we present a broad range of representations and modelling approaches to cover both self-driving and human-robot interaction (HRI) scenarios, and we compile datasets from both domains. Yet it becomes clear that robotics motion prediction lags behind due to the lack of standardized training splits, prediction horizons, and evaluation protocols. In contrast, autonomous driving benefits from large-scale public benchmarks (e.g., nuScenes, Argoverse, Waymo) that have helped standardize evaluation practices across the field. This asymmetry is also reflected in Chapter~\ref{sec:applicable} and Chapter~\ref{sec: generalization}, where most studies on deployment and generalization focus on driving.

A key reason for this gap lies in the complexity of data collection and task diversity. Driving takes place in structured, homogeneous environments—roads, lanes, traffic rules—with a well-defined and consistent goal: navigating from point A to point B. Robotics, by contrast, operates in unstructured and heterogeneous settings such as homes, offices, and factories, each with varying layouts, objects, and human behaviors. Robot platforms themselves differ in morphology, sensors, and degrees of freedom, making datasets less transferable and harder to scale. As a result, collecting robotics data is often costlier and less reusable, slowing the development of unified benchmarks. 
 Therefore, we view the prevalence of driving literature in this survey not as a limitation of scope, but as an accurate snapshot of the field’s current maturity. Notably, the progress in driving has been driven by data: standardized benchmarks enabled advances in representation learning, integration with perception and planning, and eventually full-stack optimization. One might ask whether a similar trajectory could unfold in robotics, if the community coalesces around shared datasets and benchmarks. We view this as a critical direction, and encourage future work to prioritize scalable, standardized, and platform-agnostic data pipelines to support motion prediction across diverse robotic systems.

\textbf{Related Surveys} This paper aims to comprehensively review the methodologies involved in deployable and generalizable motion prediction. While we also provide an overview of the taxonomy and analysis of motion prediction models in Sec~\ref{sec: taxonomy}, the applicability and generalizability of motion prediction models occupy our core interest in this work as in Sec~\ref{sec:applicable} and Sec~\ref{sec:generalizable}. We suggest readers refer to \cite{rudenko2020human,huang2022survey} for a more extensive digest on the taxonomy and pro-and-cons analysis of different motion prediction methodologies, to \cite{sinha2022system} for perspectives and challenges in dealing with out-of-distribution issues at the robotics system level, to \cite{yang2021generalized} for comprehensive reviews on general methods of out-of-distribution detection, and to \cite{wang2022social} for social interaction modeling in autonomous driving.



\begin{figure}[htbp!]
    \centering
    \includegraphics[width=0.6\textwidth]{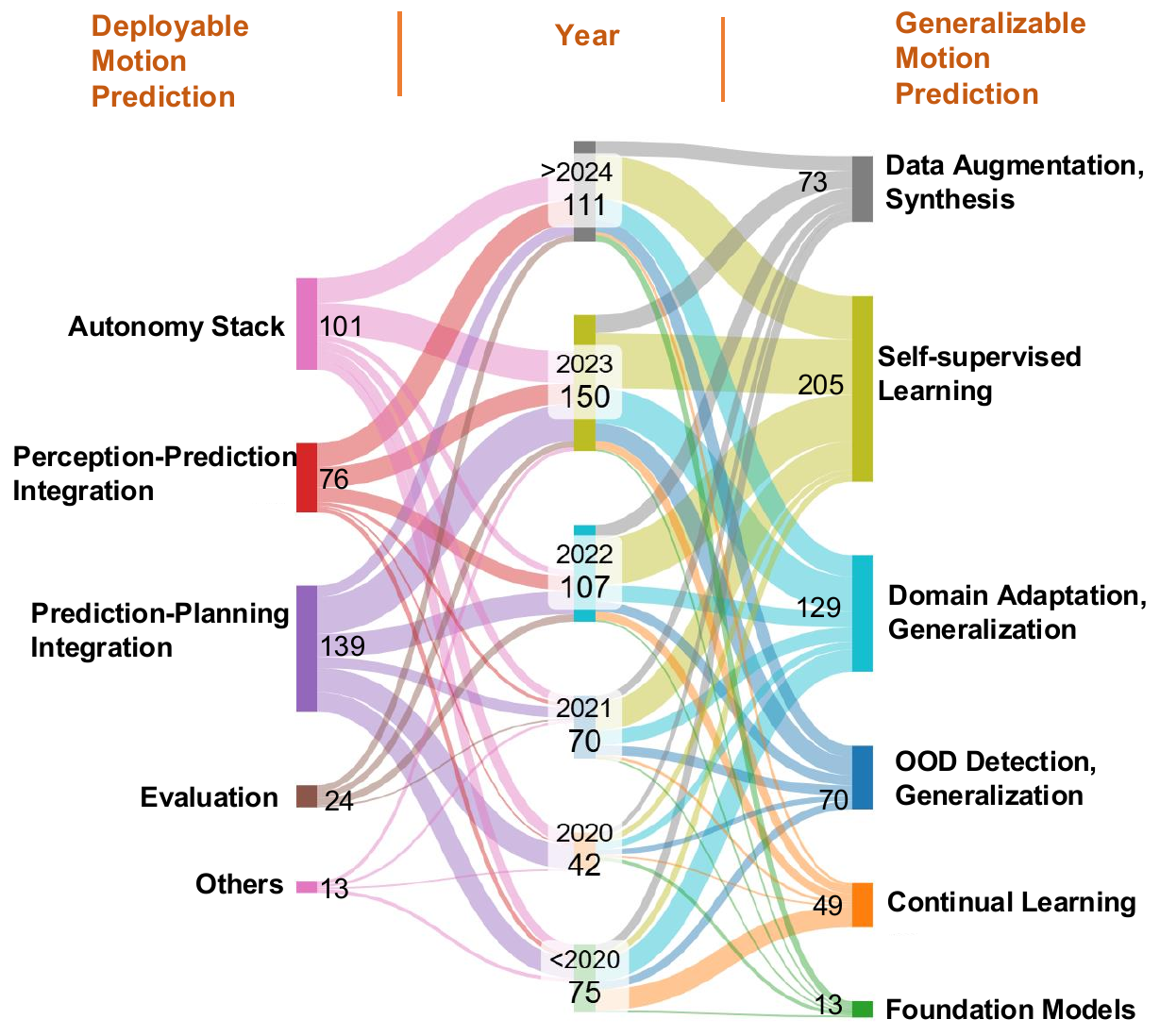}
    \caption{Distribution of literature discussed in this survey across key research topics and publication years, highlighting the evolution and interconnection of deployable motion prediction and generalizable motion prediction research from before 2020 to after 2024.}
    \label{fig:citation_sankey}
\end{figure}
\begin{figure}[htbp!]
    \centering

    \includegraphics[width=1\textwidth]{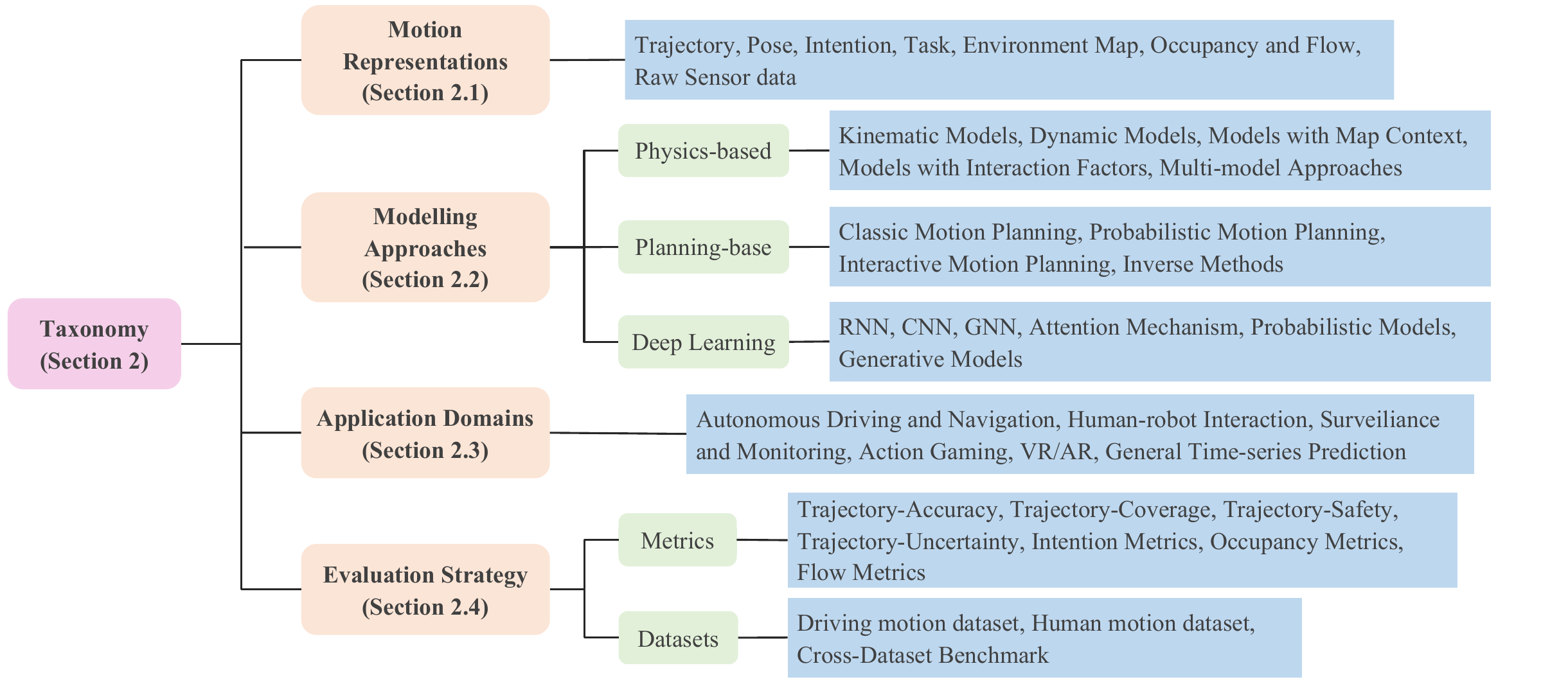}



    \caption{Overview of the content covered in Chapter 2: Taxonomy of Motion Prediction Methods.}

    \label{fig:framework_section2}
\end{figure}
\begin{figure}[htbp!]
    \centering
    \includegraphics[width=1\textwidth]{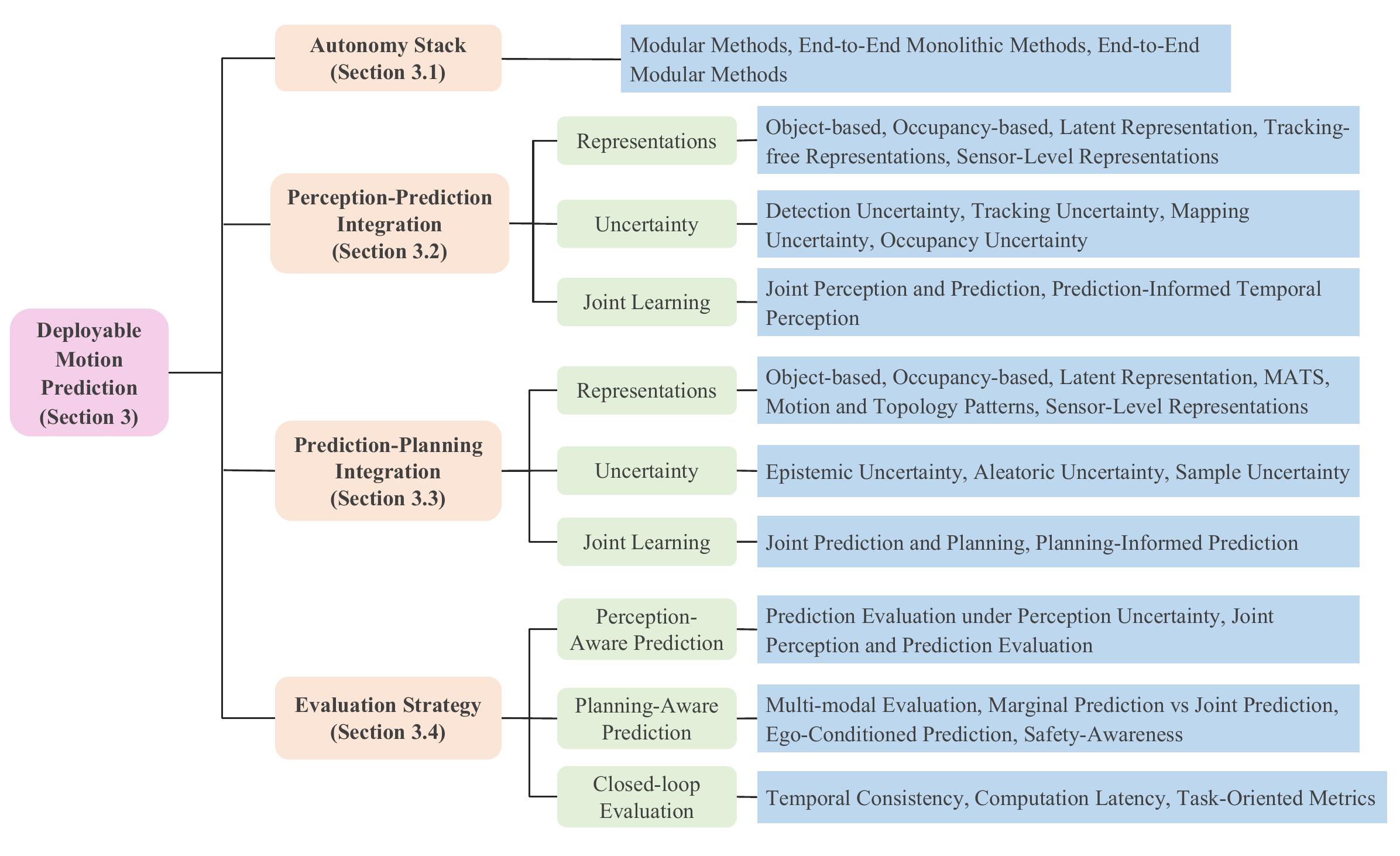}
    \caption{Overview of the content covered in Chapter 3: Deploying Motion Prediction with Real-world Autonomous Systems.}
    \label{fig:framework_section3}
\end{figure}
\begin{figure}[htbp!]
    \centering
    \includegraphics[width=1\textwidth]{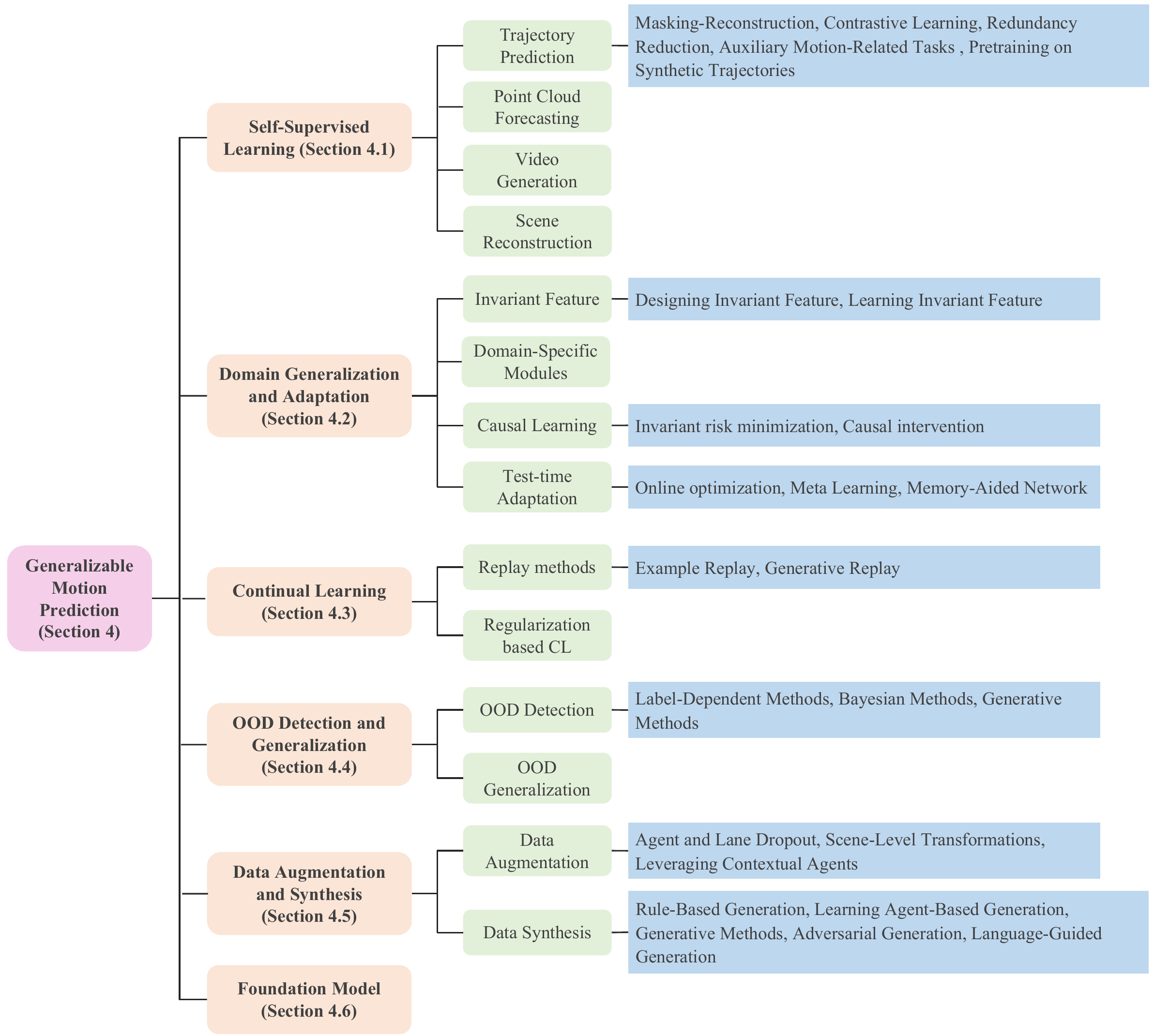}
    \caption{Overview of the content covered in Section 4: Generalizable Motion Prediction.}
    \label{fig:framework_section4}
\end{figure}

\textbf{Paper Framework} Section \ref{sec: taxonomy} briefly discussed the common input and output representation, taxonomy of methodologies, evaluation metrics, and datasets of motion prediction. Section \ref{sec:applicable} covers methodologies and insights to improve the deployability of motion prediction models when they are integrated with other upstream perception and downstream planning in the closed-loop autonomy.
Section \ref{sec:generalizable} introduced the approaches and perspectives to generalize motion prediction models to the open-world setting.  
We highlight insights, challenges, and future directions throughout each section, and conclude in Section \ref{sec: discussion}. The overview of contents covered in each section is illustrated in Figure~\ref{fig:framework_section2}, Figure~\ref{fig:framework_section3}, and Figure~\ref{fig:framework_section4}. We also illustrate the distribution of the literature reviewed in this
paper in Figure~\ref{fig:citation_sankey}, color-coded by topics and the publication year.




%% file: section/2_taxonomy.tex
\newpage
\section{Taxonomy of Motion Prediction Methods}





\label{sec: taxonomy}
In this section, we present a taxonomy of motion prediction methods, organized by representation and modeling approaches, target applications, and evaluation strategies (commonly used metrics and benchmark datasets). While the major focus of this paper is on the deployability and generalizability of motion prediction, we include this brief taxonomy of motion prediction models as supporting background knowledge. Readers are also referred to \citet{rudenko2020human,huang2022survey} and~\citet{gulzar2021survey} for more comprehensive and diverse taxonomies of motion prediction models.

\subsection{Motion Factors and Common Representations}
\label{sec: taxonomy - representation}
Motion prediction relies on a wide range of motion factors, to accurately estimate the future motion of a target agent.  In this section, we discuss the general factors that influence motion behavior, which include the internal factors of the target agent, and external factors from surrounding agents and the environment. We will also discuss common input and output representations utilized by the motion prediction models to ingest motion factors and output motion predictions. 
\subsubsection{Motion Factors}
\textbf{Internal Factors from the Target Agent} Naturally, the target agent's behavior is driven by its internal stimuli. These factors typically include 1) the target agent's historic motion states and articulated poses, such as positions, velocities, headings, body joint states, and gestures; 2) the target agent's intent and ongoing task (e.g. crossing a street, picking up a ball, turning at a corner); 3) the target agent's semantic attributes such as agent type (e.g. vehicle, pedestrian, robot), individual preferences and personality. 

\textbf{External Factors from Surrounding Agents} In many cases, the target agent has to share the public workspace with other agents, which brings influences on an agent's future motion. Therefore, the external factors influencing agent motion first depends on the surrounding agents' internal factors as described above, such as their historic motion, intention, or class. Additionally, external factors from surrounding agents are also shaped by how the target agent reacts to the surrounding agents. The target agent may simply ignore the presence of other agents, or it may engage in interaction, negotiations and coordination with surrounding agents, which may vary in intensity, interaction patterns, and behavioral preferences.


\textbf{External Factors from the Environment} Lastly, the target agent's motion behavior is impacted by the environment it is located in. In the simplest cases, an open-space environment and complete independence from the environment are assumed. More complex forms will consider static obstacles in the environment (e.g. occupancy grids, object primitives), can leverage detailed knowledge of the scene geometry and topology (e.g. HD maps, road geometries), and can even take account of the environment semantics (e.g. forbidden zones, traffic lights, lane markings, motion customs such as driving on the right). 
This rich set of environmental factors significantly shapes agent behavior and presents both challenges and opportunities for developing more structured, context-aware prediction models.


\subsubsection{Input and Output Representation} \label{sec:input-output-representation}
\begin{figure}[t!]
    \centering
    \includegraphics[width=1\textwidth]{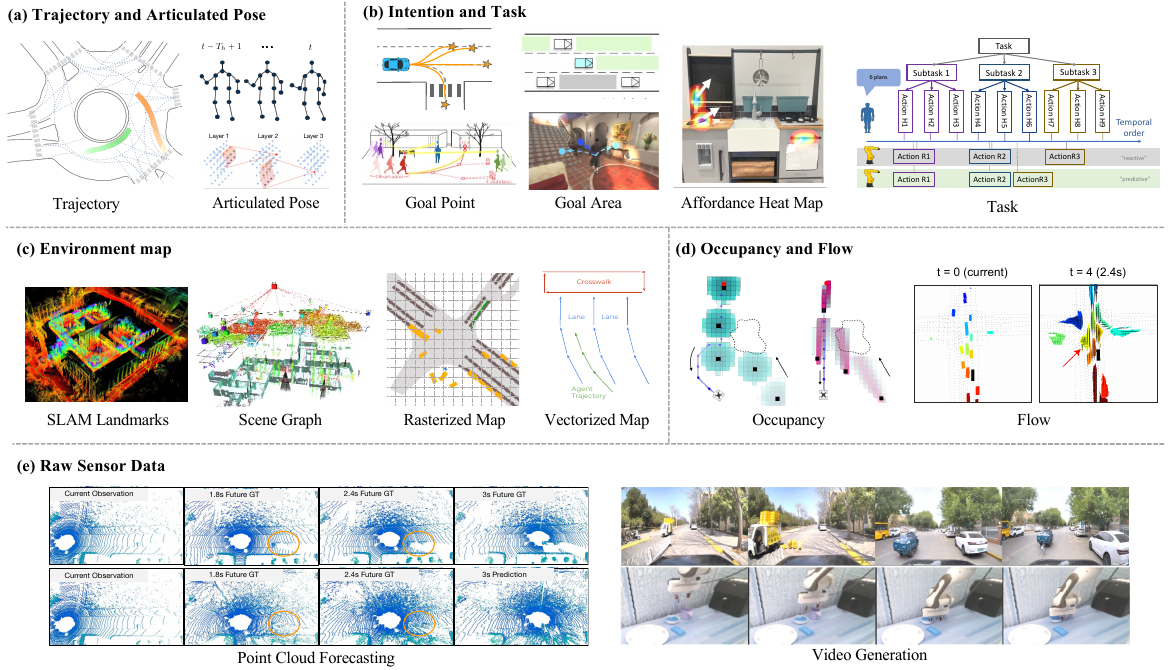}
    \caption{Common representations used in motion prediction, with illustrations adapted from existing papers. (a) trajectory~\cite{wang2021socially} and articulated pose~\cite{li2023pedestrian}; (b) intention and task, including goal point~\cite{zhao2020tnt,bae2023set}, goal area~\cite{hu2020scenario,cao2020long}, affordance heat map~\cite{bahl2023affordances}, and task~\cite{cheng2020towards}; (c) environment map, such as SLAM landmarks~\cite{lin2021r}, scene graph~\cite{
    rosinol20203d,rosen2021advances}, rasterized map~\cite{jiangVADVectorizedScene2023}, and vectorized map~\cite{gao2020vectornet}; (d) occupancy~\cite{fisac2018probabilistically} and flow~\cite{mahjourian2022occupancy}; (e) raw sensor data, such as point cloud~\cite{zhang2023learning}, images and videos~\cite{rietsch2022driver,yang2024video}.}
  \label{fig:representation}
\end{figure}

Ideally, motion prediction models need to consume all the motion factors available to them as inputs in order to produce high-quality predictions. Achieving this requires input and output representations that accurately capture these factors and comprehensively describe the predicted behaviors. This section will introduce the common input and output representations employed by existing works. Figure~\ref{fig:representation} illustrates the common input-output representations used in motion prediction, while Figure~\ref{fig:ad representation} and Figure~\ref{fig:human_motion_represent} depicts the representations tailored for autonomous driving and human motion applications, respectively.

\begin{enumerate}
    \vspace{-0.5em}
    \item \textbf{Trajectories} A widely used and intuitive representation for agent motion prediction is the trajectory, which captures motion behavior with a series of waypoints over a specified time period. For example, works in the autonomous driving domain~\cite{gao2020vectornet,salzmann2020trajectron++,choi2021shared}, human-robot interaction domain~\cite{fisac2018probabilistically,cheng2020towards}, and the video surveillance domain~\cite{antonini2006behavioral,ma2017forecasting} often rely on a series of 2D or 3D agent states, which could include coordinates and velocities at discrete intervals. This representation is frequently used as both the input to capture historic agent motion and the output to represent future motion predictions.
    
    \vspace{-0.5em}
     \item \textbf{Articulated Poses}  For human agents, a more detailed approach beyond 2D or 3D trajectories is the articulated poses, which capture the relative pose and joint angles of body parts such as the head, torso and limbs, offering a more fine-grained depiction of human motion. This representation can be further enriched with additional cues such as eye gaze directions and hand or arm gestures. Articulated poses are particularly valuable in human-robot interaction tasks~\cite{cheng2019human,ma2022multi,li2023pedestrian} and human modeling for VR/AR applications~\cite{majoe2009enhanced,hauri2021multi}.
     
    \vspace{-0.5em}
    \item \textbf{Intention and Task} Beyond trajectories and articulated poses, agent intentions and tasks are critical high-level factors influencing motion behaviors. They can be represented in a variety of forms, including goal point positions~\cite{zhao2020tnt,girase2021loki,shi2022motion,bae2023set,cao2020long}, target point heat maps~\cite{park2018sequence,bahl2023affordances}, target areas~\cite{hu2020scenario,wang2021hierarchical}, maneuvers or skills~\cite{deo2018multi,wang2023efficient}, navigation signals~\cite{codevilla2018end,sun2018probabilistic}, task-specific cost functions~\cite{wang2021socially,wu2020efficient}, current ongoing task~\cite{cheng2020towards}, model choice indicators~\cite{codevilla2018end}, and sampling rejection criteria~\cite{fisac2018probabilistically}. Note that while this information is frequently used as input, it can also serve as an output objective in settings that aim to infer agents' intentions and ongoing tasks, or an intermediate prediction target where it is used to inform fine-grained trajectory forecasting.
    
    
    \vspace{-0.5em}
    \item \textbf{Environment Map} In addition to the agent-centric representation discussed above, environment maps are another crucial representation used in motion prediction. In their simplest form, environments are represented as landmarks~\cite{lin2021r}, which are sparsely distributed collections of points. Landmarks have been a cornerstone of Simultaneous Localization and Mapping (SLAM) systems for over three decades~\cite{rosen2021advances}. To enhance scene understanding, maps can also be represented as a set of obstacles that robots must avoid~\cite{liu2018convex} in a more object-level form, or as hierarchical dynamic-aware scene graphs that further capture relationships among objects in the environment. For more structured environments, such as that in autonomous driving, the road information is usually represented as a rasterized map~\cite{deo2018multi,salzmann2020trajectron++}, which decomposes the scene into a birds-eye-view 2D spatial grid and uses multiple channels to capture semantic information. Each cell is rendered with different colors according to whether it is occupied by agents or map elements such as road curbs and lane lines. Due to the similarity with image pixels, this representation facilitates the application of convolution neural networks. Another popular representation is vectorized maps~\cite{gao2020vectornet,lanegcn}, which describe both geographic entities and agent trajectories as polylines defined by multiple control points. For instance, a lane boundary is represented by multiple control points that form splines, while moving agents can be approximated by polylines based on their motion waypoints. Compared to the rasterized map, the vectorized representation aims to avoid lossy rendering, better capture complex topology relationships, support explicit entity interaction reasoning, and enable efficient computation due to their higher information density.

    \vspace{-0.5em}
    \item \textbf{Occupancy and Flow} In contrast to object-based prediction paradigms that adopt trajectories or poses as the interface,  occupancy representations~\cite{hong2019rules,bansal2018chauffeurnet,toyungyernsub2022dynamics,toyungyernsub2024predicting,wang2025uniocc} adopt an object-free approach.  
    These methods model the scene as a spatio-temporal grid, with each grid cell containing the probability of the cell being occupied by any agent, along with the semantic class of the occupying agent. 
    The occupancy flow field~\cite{mahjourian2022occupancy,agro2023implicit,lange2024scene} representation further extends the occupancy representation to include flow vectors indicating the direction and magnitude of the motion in each grid cell, which enables identification and recovery of agent entities from the occupancy map. 

    \vspace{-0.5em}
    \item \textbf{Sensor Data and Upstream Features} When trajectories, poses, or occupancy serve as input, they are typically obtained from sensor data and upstream perception algorithms. Alternately, raw sensor data such as RGB images~\cite{stocking2023linking} and LiDAR point clouds~\cite{chen2024womd}, as well as upstream perception features~\cite{ivanovic2022heterogeneous,weng2022whose,wang2024cmp,gu2024producing,gu2023vip3d} can be directly presented to motion prediction models as inputs, to provide additional information that could be missed in human-designed explicit interfaces. 
    
    \textit{Raw Sensor Data} With the concept of world models~\cite{lecun2022path}, recent works on point cloud forecasting~\cite{khurana2023point} and video generation~\cite{hu2023gaia} offer novel approaches to directly model the temporal evolution of the scene using unstructured sensor inputs.
    They take in point clouds and 2D images directly from LiDAR and camera sensors, and generate predictions in the same format, bypassing object detection and tracking—thereby capturing scene dynamics in a more unified and end-to-end fashion.
    
    \textit{Upstream Features} Alternatively, motion prediction models can be conditioned on structured upstream perception outputs such as object detection embeddings~\cite{ivanovic2022heterogeneous}, tracking affinity maps~\cite{weng2022whose,wang2024cmp}, online map features~\cite{gu2024producing}, and visual tokens~\cite{gu2023vip3d}. These features encode semantic and spatial information at a mid-level abstraction, offering a compromise between raw sensor signals and high-level trajectory inputs.


\end{enumerate}

\textbf{Deterministic vs. Probabilistic Representations}
In existing works, inputs have commonly been designed to be deterministic. For example, most existing motion prediction benchmarks assume access to curated and noise-free historic trajectories~\cite{sun2020scalability,Argoverse,wilson2023argoverse}. However, inputs can also be provided as probability distribution representations, generated by upstream perception models that are trained to produce not only perception results, but also associated interpretable uncertainties~\cite{ivanovic2022heterogeneous,weng2022whose,gu2024producing}. On the other hand, due to the inherent uncertainties and multi-modal nature of human behavior, the output representations of a prediction model are typically formulated probabilistically. This can take the form of a probabilistic distribution over explicit waypoints, poses, and intentions~\cite{zhao2020tnt}, or a probabilistic distribution over latent features~\cite{salzmann2020trajectron++,lange2024self}, from which samples can be drawn and decoded to define explicit waypoints, poses, and intentions. 
The additional probability information enables uncertainty quantification and calibration for motion prediction models, and is crucial for robust downstream decision-making. We will discuss the strengths and drawbacks of existing representation when they are used to bridge perception, prediction and planning in Section~\ref{sec: applicable perception-prediction representation} and Section~\ref{sec: app - prediction-planning representation}, and how to deal with the associated uncertainties in Section~\ref{sec:deployable_perception_dealing_with_uncertainty} and Section~\ref{sec: prediction uncertainty}. 

\subsection{Modelling Approaches}

As motion prediction receives increasing attention over the years, various approaches have been proposed and studied, which vary not only in the representations discussed above but also in the methodologies employed. In this section, we introduce this body of work organized into three categories: 1) \textit{physics-based methods} that manually define explicit motion models and rules, propagating the model forward in time; 2) \textit{planning-based methods} that model to-be-predicted objects as utility-maximization agents, whose future motion can be predicted by assuming they are optimizing a designed or learned reward function; 3) \textit{deep-learning-based methods} that take a data-driven approach to learn motion patterns and prediction models from data. Note that, many works combine the advantages of multiple categories and may not fall only into a single one.

\subsubsection{Physics-Based Methods}
Physics-based methods are the earliest and simplest methods developed by researchers. This category of methods generates future motion predictions by defining hand-crafted explicit motion models or rules based on Newton’s laws of motion, and propagating the physics model forward in time. 
Inspired by ~\citet{rudenko2020human}, we divide the physics-based models into 1) single-model approaches that rely on a single motion model under an exclusive set of conditions and assumptions, and 2) multi-model approaches that include several motion models to enable adaptive prediction for varied agents in more complex scenarios.

\begin{figure}[t!]
    \centering
    \includegraphics[width=1\textwidth]{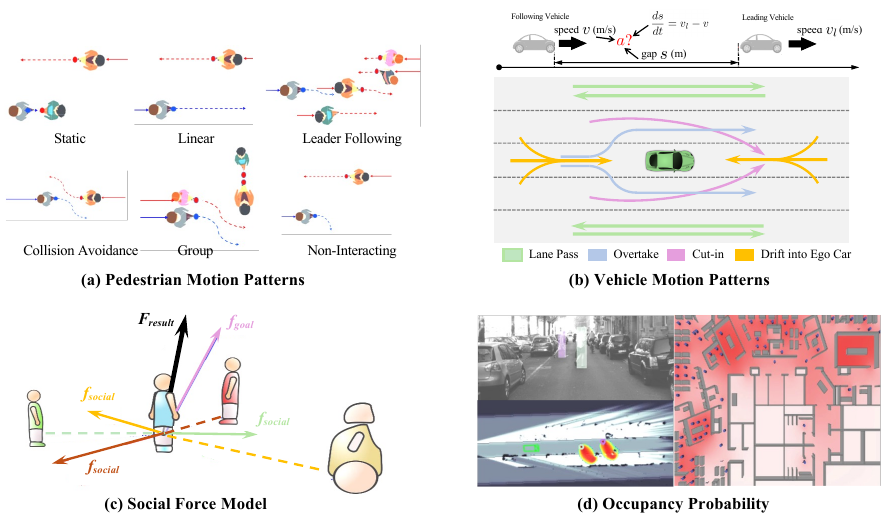}
    \caption{Illustrations of physics-based methods for motion prediction: (a) pedestrian motion patterns~\cite{kothari2021human}; (b) vehicle motion patterns, including intelligent driving model~\cite{zhang2024bayesian}, and driving maneuvers~\cite{deo2018would}; (c) social force model~\cite{ferrer2014behavior}; (d) occupancy probability~\cite{rehder2015goal,schaumann2020join}.}
  \label{fig:physics_methods}
\end{figure}

\paragraph{Single-model approaches}
\textbf{Kinematic Models} Among the many single-model physics-based approaches, kinematic models are perhaps the simplest ones, operating under simplified motion assumptions that often ignore the internal and external forces that govern an agent's motion. Popular examples include the constant velocity model (CV), constant acceleration model (CA), and constant turning rate model (CT) that respectively assume constant velocity, acceleration, and turning rate/speed with noise \cite{ammoun2009real,schubert2008comparison,lytrivis2008cooperative}, respectively. The bicycle model is often used to approximate the kinematics of vehicles with articulated front wheels~\cite {schubert2008comparison,li2019generic}. The human body is frequently modeled as multiple rigid bodies connected by joints, which have multiple degrees of freedom \cite{liu2020human}, whose motion can be propagated by forward and inverse kinematic models \cite{tevatia2000inverse}. Kinematic models are often favored due to their simplicity and acceptable performance, especially for simple prediction problems such as those with limited context influence, low uncertainty, and short prediction horizons.

\textbf{Dynamic Models} While kinematic models offer a simplified view by ignoring forces, dynamic models provide a more comprehensive approach by explicitly accounting for the forces that govern or influence motion, such as actuation forces and external disturbances.
These models can become very complicated when considering detailed actuation forces such as the physics of wheels, engines, or friction effects of terrain, tires or body joints \cite{rajamani2011vehicle}. Such complexity not only leads to more model parameters to be considered, but also brings additional challenges as these forces are sometimes not directly measurable and need to be estimated instead. A number of works also extend deterministic kinematic and dynamic models to include uncertainties and noise on the model parameters, using a mixture of Gaussians \cite{kaempchen2004imm,jin2015switched} or Monte Carlo sampling \cite{broadhurst2005monte,althoff2011comparison}, which can be updated with Kalman filtering methods \cite{ammoun2009real,batz2009recognition}, for example. These extra considerations are beneficial for control-oriented applications and are used extensively for ego motion planning and control purposes, but also introduce extra methodological design requirements and computation burdens that are not always essential for motion prediction purposes. Therefore, for applications such as trajectory prediction, models that simplify or abstract away actuation and disturbance forces are often preferred \cite{lefevre2014survey}. 

\textbf{Models with Map Context} A number of works extend physics-based models to consider information from the map, which vary in how the map factors and constraints are represented and enforced into the prediction model. Some works~\cite{yang2008fusion,batkovic2018computationally,yang2005nonlinear} project the map-free predicted motion onto the map surface, adjusting the predictions to avoid obstacles (e.g. road curbs, walls, sidewalks, and crosswalks). Local frame methods~\cite{zhan2016non,li2020IFAC} transform the motion from the Cartesian frame to the Frenet coordinate frame centered at the agent and aligned with the direction of travel to incorporate the map reference information in a manner consistent across different scenarios. When there are multiple possible future reference paths (e.g. lanes, corridors), multiple predictions can be propagated along these paths to account for multiple possible motions. For motion in open space without explicit map constraints, different representations have been proposed to capture implicit map priors. Grid-based methods such as affordance maps \cite{luber2011place}, belief maps \cite{rehder2015goal}, and polar grids \cite{coscia2018long} can represent the occupancy probability, or walkable/drivable areas over the map, and enable predictions to comply with the map priors. Tree-based approaches~\cite{aoude2010threat} develop a tree of future trajectories, and bias the growth of the tree towards regions that are more likely for the agent to cross.

\textbf{Models with Interaction Factors}
In many cases, the target agent does not exist in isolation and needs to co-exist and interact with other surrounding agents, thus many works have explored different ways to incorporate agent interaction into physics-based prediction models. One popular example is the social force model \cite{helbing1995social}, which defines attractive forces to the goal and repulsive forces from obstacles and other agents for human crowd motion analysis. Many works extend social force models into various domains and consider different problems other than human crowd motion \cite{luber2010people,yan2014modeling}. Some core problems exist across all social force modeling, including the challenge of estimating the target agent's goals or intent when they are not available \citep{elfring2014learning,vasquez2008intentional}, and determining the force parameters for different obstacles and other agents \citep{ferrer2014behavior,oli2013human}. Other interaction-aware methods that are not based on social force models include the intelligent driver model (IDM) \citep{treiber2000congested,liebner2013velocity} for longitudinal vehicle following, and the reciprocal velocity obstacle model \citep{van2008reciprocal} for multi-agent joint prediction.

\paragraph{Multi-model approaches}
While more prior knowledge is incorporated into physics-based models such as agent dynamics and map/interaction factors, a single model can suffer in complex scenarios. Thus many works leverage a mixed set of multiple motion models, each representing the motion of a certain pattern and mode. The members of the model set can be fixed or adapted online \cite{zhang2023bootstrap}. The design of the model set can be diverse depending on the purpose and perspective, which can be as simple as a set of basic motion primitives such as going straight or turning \citep{pool2017using}, or as complex as a set of maneuvers with context semantics such as overtaking and merging \cite{kuhnt2015towards}. As the motion modes of other agents are not immediately observable, techniques such as dynamic Bayesian networks \citep{pentland1999modeling,gu2016motion} are used to estimate the motion mode probabilities based on methods such as intent recognition~\citep{schulz2015controlled} or context compatibility~\citep{kuhnt2015towards,pool2019context}, allowing multi-model approaches to adaptively switch amongst models \citep{schulz2015controlled,pool2017using} or fuse the output from the multiple models \citep{li2005survey} to generate final predictions.

\subsubsection{Planning-Based Methods}
Instead of directly propagating agents' states forward in time with kinematic or dynamics models as in physics-based methods, planning-based methods incorporate the assumption of rational agents making optimal motion decisions. Given a manually-defined or learned objective function, the rational agents are assumed to optimize the objective function, by which their sequence of future actions can be predicted. Such methods can usually be distinguished by what objective function is assumed and what algorithms are leveraged to solve the planning problem. In this section, we divide planning-based methods into 1) forward planning methods that assume immediate access to the objective function for motion prediction and directly perform planning; and 2) inverse planning methods that first infer the objective function from observations and then predict future motion.
\begin{figure}[t!]
    \centering
    \includegraphics[width=1\textwidth]{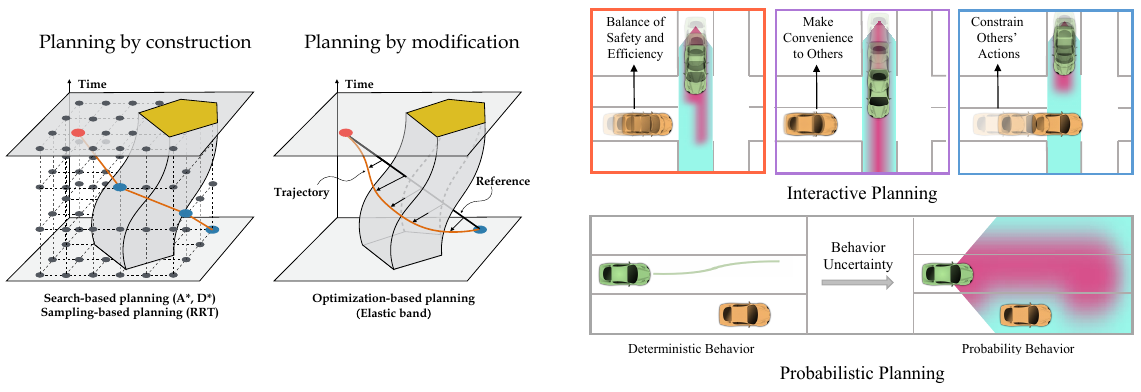}
    \caption{Left: Illustration of three types of planning methods, which can be categorized into planning by construction and planning by modification: search-based planning, sampling-based planning, and optimization-based planning (Figure from \citet{cmu_liu}). Right: Extension to interactive planning and probabilistic planning (Figure from \citet{wang2021socially}).}
    \label{fig:planning}
\end{figure}

\paragraph{Forward Planning Methods}
\textbf{Classic Motion Planning} The simplest and earliest planning-based prediction methods utilize classic path and speed planning techniques with hand-crafted objective functions. The objective function typically aims to promote task completion by minimizing distance to the goal, ensure safety by maximizing or maintaining a minimum distance to unsafe regions, encourage trajectory smoothness by minimizing acceleration and/or jerk, and optionally minimize the magnitude of the control effort. Such objective functions often come in the form of a linear combination of cost terms~\citep{mellinger2011minimum,wang2021socially} or a potential/cost map~\citep{yi2016pedestrian,xie2013inferring}. When resolving the planning problem, approaches generally fall into three categories: 1) The first is the optimization-based methods, which formulate the agent's path and velocity in the continuous space. These methods represent trajectories as polylines parameterized by key control points ~\citep{gu2017improved,li2020IFAC}, or use kinematic and dynamic models parameterized by control inputs~\citep{sadigh2016planning,schwarting2019social}. Techniques such as elastic band algorithms~\citep{gu2017improved,li2020IFAC} and model predictive control methods~\citep{sadigh2016planning,schwarting2019social} can then be employed to iteratively resolve the planning problem in a gradient descent manner. 2) Second, search-based methods discretize the agent's path and speed profile, and then apply search algorithms such as A*~\citep{gong2011multi}, Monte-Carlo tree search~\citep{li2022efficient}, and Dijkstra's~\citep{xie2013inferring} to identify the optimal trajectory. 3) Third, sampling-based methods exploit the problem structure to sample a set of reasonable candidate trajectories, evaluate them using a predefined objective function, and then select the optimal sample~\citep{wang2021socially,fisac2018probabilistically}. 
Notably, all three classes of methods can incorporate dynamic constraints, such as actuation limits or turning feasibility, to ensure physical realizability~\citep{salzmann2020trajectron++,varadarajan2022multipath++}.



\textbf{Probabilistic Motion Planning}
Planning-based methods can also be extended into the probabilistic domain, explicitly modeling the uncertainty of predicted motion to better capture the inherent variability in future behaviors. Two inspiring works \cite{fisac2018probabilistically,fridovich2020confidence} model the confidence of the planning-based motion prediction model within a real-time Bayesian framework, which can capture how “rational” human actions appear to be. This allows the system to quantify uncertainty in human actions, enhancing safety by accounting for deviations from idealized or expected behavior. To account for multiple possible futures, some works set or estimate a set of possible goals or intentions of the target agent according to the semantics or topology of the map~\citep{best2015bayesian}, then predict the future motion and associated probability for each goal~\citep{yen2008goal}, while also recognizing the changes in the estimated intention~\citep{vasquez2008intentional,karasev2016intent}. 

\textbf{Interactive Motion Planning}
Another thread of planning methods propose to consider the presence of other agents and the interactions with them. The simplest method is to divide the problem into a global planning problem with local collision avoidance subproblems~\cite {van2008interactive}: each agent's global path is computed considering only static obstacles, and local collision avoidance along the global path is executed by adjusting each agent's short-term behaviors. For joint prediction of multiple agents, some methods~\cite{rosmann2015timed,bandyopadhyay2013intention} propose topologically distinct candidate trajectories for each agent, and select the best candidate according to metrics related to the joint benefit of the group. To enhance the interaction modeling among agents, game theory concepts such as Nash Games~\citep{schwarting2019social,li2023game}, Stackerberg Games~\citep{sadigh2016planning,wang2021socially}, and K-level Games~\citep{tian2020game} are incorporated. In these methods, when optimizing the ego agent's action, the influence of the ego agent's action on other agents and on the evolution of the entire scene are considered simultaneously. In \citep{chen2023interactive}, the motion plans for the ego agent and relevant non-ego agents are optimized simultaneously, acting as an approximate conditional probabilistic model of the interactive behavior among agents. In interaction-intense settings, various objective functions are proposed to better capture heterogeneous characteristics of different agents, such as social value orientation~\citep{schwarting2019social}, courtesy~\citep{sun2018courteous,wang2021socially}, confidence~\citep{wang2021socially}, and information gathering~\citep{sadigh2016information}. Multiple algorithms have been exploited to resolve multi-agent interaction problems such as bi-level optimization~\citep{sadigh2016planning,sun2018courteous,wang2021socially} and iterative best response~\citep{schwarting2019social}.
Braid theories are also exploited to leverage the topology patterns for multi-agent interaction modeling~\citep{mavrogiannis2020decentralized,mavrogiannis2022analyzing,chen2023interactive} and conflict resolution~\citep{liu2017distributed,chen2017decentralized}. 

\paragraph{Inverse Planning Methods}
In contrast to forward planning approaches which assume immediate access to the agent's objective function, inverse planning methods assume that the objective function is unknown, but can be learned from observations of expert agents' motions with the assumption that these agents are rational. This thread of work is closely related to inverse reinforcement learning (IRL). Initially proposed by ~\citet{kalman1964linear}, the concept of IRL was first formulated in 2000~\cite{ng2000algorithms,abbeel2004apprenticeship} where reinforcement learning and inverse reinforcement learning are conducted iteratively to learn a control policy and reward function simultaneously. To handle uncertainties and noisy observations, IRL was extended with the principle of maximum entropy~\cite{ziebart2008maximum}. This work assumes that agents' actions and behaviors with lower costs are exponentially more probable. Building on this, Levine et al. \cite{levine2012continuous} formulated the continuous IRL algorithm and used it to predict human driving behavior.  A basic reward function can be designed as a linear combination of features~\cite{schwarting2019social,wang2021socially}, where the weighting parameters are learned from demonstrations. This also enables the construction of driving-style-aware objective functions, allowing the model to make adaptive predictions that reflect individual behavioral preferences~\cite{schwarting2019social,wang2021socially}.
A hierarchical IRL framework has also been proposed~\cite{sun2018probabilistic} where the human drivers' intentions and trajectories are divided into different hierarchies for which different reward functions are learned. Compared to s single object function, a set of objective functions can be learned~\cite{rosbach2020driving}, each for a different scenario. By switching between these functions, the model can generate adaptive predictions that better reflect diverse contexts.

\subsubsection{Deep-Learning-Based Methods}
The physics-and-planning-based methods excel in scenarios that align with predefined structures and patterns, such as environment with fixed layouts and routine agent behaviors, where predefined rules and policies can be effectively applied. 
However, they often struggle in more complex, less structured settings that require understanding novel or ambiguous context and interactions.
Examples include crowded roundabouts, multi-lane intersections, pedestrian-heavy public spaces, or cluttered indoor environments with dynamic and adaptive human-robot interaction.
This limitation becomes particularly evident when motion prediction models are deployed in the public open-world spaces, where they frequently encounter new, unforeseen scenarios that fall outside the scope of predefined rules and settings. The resulting performance degradation usually stems from the limited expressiveness of the above models and laborious but ungeneralizable task-specific rule design. In contrast, the success of deep learning has ushered in a variety of effective data-driven methods that excel in learning complex motion patterns from data without requiring extensive human engineering. These deep learning approaches, particularly in the form of neural networks, aim to automatically discover intricate relationships between agents and environmental factors by leveraging large-scale datasets. As a result, deep-learning-based methods have not only gained increasing attention within the research community but have also come to dominate large-scale motion prediction benchmarks in recent years.

Nowadays, neural networks can be really complex and involve multiple submodules, varied losses and diverse training mechanisms. However, there are certain essential types of fundamental neural network building blocks, based on which sophisticated networks can be developed. Different building blocks provide different inductive biases~\citep{battaglia2018relational} on the data and solution space, so researchers select specific building blocks to inject desired inductive biases into the learning process for better performance or reduced computational requirements. 
Inversely, a mismatched inductive bias and problem setting can result in suboptimal or poor performance.  
While the main neural network building blocks may be quite familiar to deep learning practitioners and more complex variants are constantly being developed, we start each subsection below by briefly introducing how different types of neural network layers are used for motion prediction tasks. This helps highlight their respective strengths and weaknesses relative, and clarify the design principles behind modern prediction networks. 

\begin{figure}[t!]
    \centering
    \includegraphics[width=0.9\textwidth]{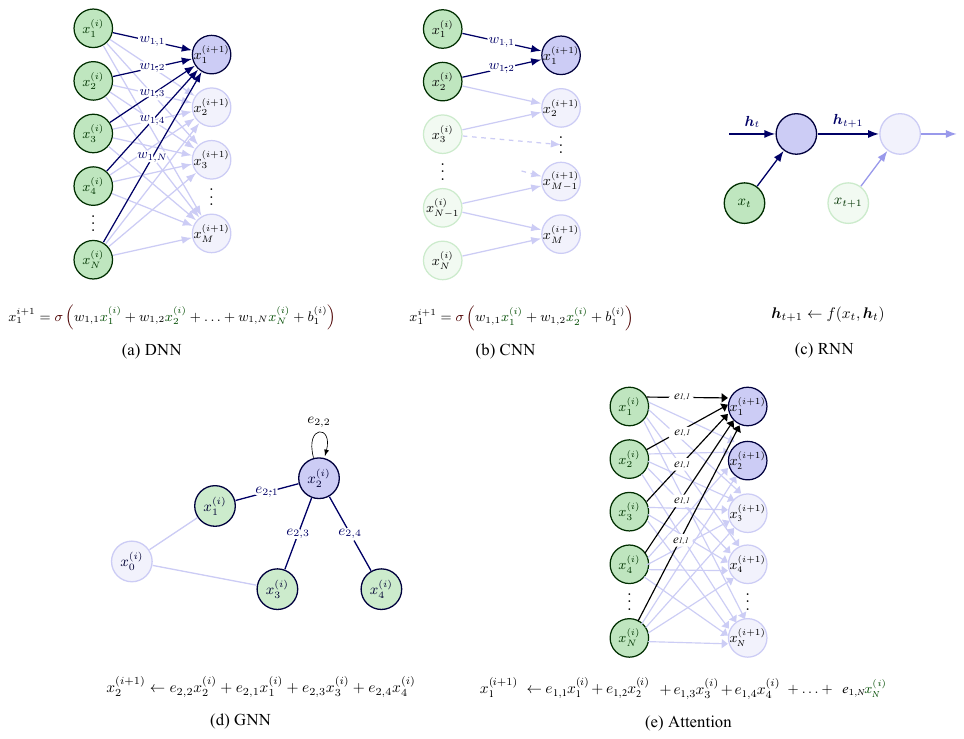}
    \caption{Five fundamental deep learning neural network building blocks for motion prediction. (a) Deep neural networks, (b) Convolution neural networks, (c) Recurrent neural networks, (d) Graph neural networks, and (e) Attention layers. Each inducing different inductive bias into the model. Figure adapted from \citet{wang2022social}.}
  \label{fig:deep_learing_blocks}
\end{figure}

\textbf{Recurrent Neural Network} Due to the temporal nature of the prediction task, recurrent neural networks (RNNs) such as LSTMs~\cite{hochreiter1997long} and GRUs~\cite{cho2014learning} have been a natural and common choice for motion prediction. RNNs maintain an internal hidden state and update that state at each timestep, leading to an inductive bias toward smoothly evolving output dynamics.   Typically, each agent (e.g., vehicle, pedestrian) is modelled with an individual RNN, which encodes and propagates the agent's historical trajectory through the RNN, storing it in hidden states. 
These hidden states are then utilized to generate future trajectories in various forms. One common approach is to directly predict future positions using an LSTM decoder—either deterministically~\cite{altche2017lstm,duan2016travel,chandra2020forecasting} or probabilistically~\cite{zyner2019naturalistic,hu2018framework} (e.g., with a Gaussian Mixture Model). Alternatively, the hidden states can be used to classify the maneuver or skill the agent is performing~\cite{deo2018multi,khosroshahi2016surround}, or to predict the intention and goal point ~\cite{hu2018framework,zyner2017long,rehder2015goal}, future occupancy of the scene~\cite{park2018sequence,dequaire2018deep}, or agent driving style~\cite{xing2019personalized}.  Then the final trajectory prediction is generated through planning-based searching~\cite{park2018sequence} or optimization~\cite{ding2019online,rehder2015goal}, follow-up neural network layers~\cite{xing2019personalized,hu2018framework}, or switching between multiple networks based on the maneuver or skill performed~\cite{xing2019personalized,deo2018multi}.
The seminal work Social-LSTM~\cite{alahi2016social} took a further step by considering the interactions among agents through a social pooling operation that shares information between the hidden state of the nearby agents' LSTMs.
Subsequent works introduced alternative methods for modeling interactions, such as pairwise interaction units~\cite{ding2019predicting}, and surrounding occupancy pooling~\cite{xue2018ss,bartoli2018context}. In addition to interaction modeling, some RNN-based approaches have focused on incorporating map information, such as by embedding map features (e.g. lane, obstacle) via RNN~\cite{Argoverse,kawasaki2020multimodal,bartoli2018context}, or employing map-aware trajectory post-optimization~\cite{ding2019online,rehder2015goal}.

\textbf{Convolution Neural Network} 
While RNNs excel in modeling temporal dependencies, their ability to model complex spatial contexts and interactions is limited. For example, simply feeding map data into an RNN~\cite{Argoverse,kawasaki2020multimodal,bartoli2018context} usually struggles to capture the rich spatial context and agent-map interaction in complex scenarios; the social pooling operation in Social-LSTM~\cite{alahi2016social} is computationally expensive and slow, as it cannot be parallelized and needs to be performed at each RNN step.
As an alternative, convolution neural networks (CNNs), which excel at modeling spatial structures and interactions via their inductive bias of locality and translation invariance~\cite{battaglia2018relational}, have been applied to motion prediction, building on their well-established success in computer vision~\cite{krizhevsky2012imagenet,he2016deep}.
Due to the advantage that CNNs exhibit in processing grid-like data, many motion prediction methods have formulated inputs as a top-down-view occupancy grid, and fed them into the CNN for spatial reasoning. 
For example, to account for \textit{agents' interactions}, some CNN methods~\cite{su2021convolutions,thomas2022learning} populate a grid cell map with the physical state of the detected agent that lies in a grid cell, and then apply a series of convolutional layers to learn the spatial inter-dependencies of agents.
Instead of using raw coordinates to populate the grid, other CNN methods~\cite{deo2018convolutional,messaoud2019non,messaoud2020attention,zhang2020multi} encode each vehicle's historical dynamics into a hidden state vector using a shared LSTM module. These hidden state vectors then populate the spatial grid cell map, where convolutional operations are subsequently applied.
To account for \textit{map information}, top-down-view rasterized map images, which contain semantic map information, are extensively exploited by CNN-based methods in multiple ways: 1) The first approach uses CNNs to extract scene context from the rasterized maps and then concatenates the embedded map features with object features~\cite{salzmann2020trajectron++,lee2017desire,gilles2021home,cui2019multimodal}. 2) The second approach spatially aligns the coordinates of agents and maps to formulate agent-augmented rasterized images~\cite{chai2019multipath,ye2021tpcn,mahjourian2022occupancy,hong2019rules}, or spatially aligns agent features and context features to form top-down-view feature grids~\cite{zhao2019multi}. Both types of alignment allow CNNs to explicitly express and better reason about the spatial relationship between agents and the contexts.
Finally, due to CNN's natural compatibility with sensor data such as camera images and LiDAR point clouds, many works apply CNNs to directly process raw sensor data, which enables motion prediction to be performed along with other perception tasks in an end-to-end manner~\cite{luo2018fast,su2021convolutions}.

Usually, convolutional interaction encoding could better leverage the spatial relationships among agents and maps for deeper interaction reasoning especially when compared to RNN methods. However, depending on the specific scenario, interactions can occur not only between agents and map features in close proximity, but also among all agents and map features within a larger region of interest, or even between distant yet relevant agents and map elements. For instance, on highways, drivers may need to look much further ahead than they would on local streets due to the vehicle's high speed.  This multi-scale challenge is not easy to solve in the CNN setting, and leads to a non-trivial trade-off between the receptive field size and computation efficiency. To this end, \citet{su2021convolutions} carefully explored how the grid map size, layout, and resolution can impact interaction modeling performance with CNNs. Further works \cite{messaoud2019non, messaoud2020attention} extended the convolutional interaction encoding by adding a non-local social pooling module with a multi-head attention mechanism to consider long-distance interdependencies. A dilated convolution social pooling architecture was also proposed to capture global context with larger receptive field \citet{zhang2020multi}.

\begin{figure}[t!]
    \centering
    \includegraphics[width=1\textwidth]{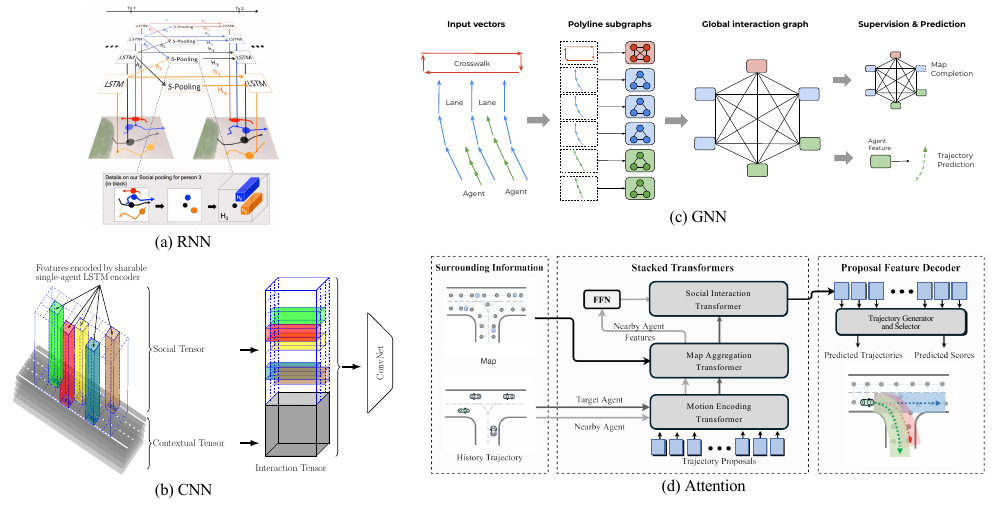}
    \caption{Illustrations of deep learning methods for motion prediction: (a) recurrent neural networks (figure from Social-LSTM~\cite{alahi2016social}), (b) convolution neural networks (figure from \citet{wang2022social}), (c) graph neural networks (figure from VectorNet~\cite{gao2020vectornet}), (d) Attention (figure adapted from MTR~\cite{shi2022motion} and mmTransformer~\cite{liu2021multimodal}).}
  \label{fig:deep_learing_methods}
\end{figure}

\textbf{Graph Neural Network} 
While methods using RNNs and CNNs have made significant progress in temporal and spatial modeling, several challenges remain. First, explicit reasoning about relationships amongst agents remains limited: RNNs are primarily designed for temporal modeling for each individual agent, and CNNs typically model spatial relationships in terms of an occupancy grid, which represents space rather than individual objects. Second, capturing long-range interactions via convolutions with small receptive fields is ineffective, while increasing the receptive field dramatically raises computational burdens. Finally, The number of interacting agents in a scene varies, and in most cases, agents are sparsely occupying the scene. This makes convolutions less suitable for prediction than image processing, as convolutions often end up wasting significant resources by extensively applying convolution operations to empty space. 

To address these challenges, another approach is to represent the scenes as interaction graphs. Each scene can be viewed as an irregular graph where agents are represented as graph nodes with attributes (e.g., coordinates, velocity, object class), and relationships among agents are captured by undirected or directed edges between nodes. The nodes and edges together form a spatial-temporal graph, enabling more efficient relational reasoning across agents and map elements. Given these graph-based representations, Graph Neural Networks~\cite{battaglia2018relational} (GNNs) naturally become a powerful choice for encoding the relationships between agents and the environment. A graph neural network (GNN) is well-suited for encoding inter-agent relationships, as they can learn expressive node and graph representations by aggregating neighborhood information through the graph’s topological structure and node features \cite{battaglia2018relational, wu2020comprehensive, wang2021property}.
Specifically, the constructed graph is passed into GNN layers for interaction learning via a message-passing scheme, where each node aggregates the features of its neighboring nodes to update its own attributes. Various variants of GNNs have been explored in the motion prediction community, such as Graph Convolution Networks~\cite{li2019grip,dax2023disentangled,li2019grip++,jeon2020scale,mohamed2020social,cao2021spectral,zhou2022grouptron}, and Graph Attention Networks~\cite{huang2019stgat,sun2022interaction,diehl2019graph,kosaraju2019social,li2021rain,li2021spatio}.
In these works, researchers usually combine RNNs and GNNs to process temporal information more effectively. For example, MFP~\cite{tang2019multiple} designed the attributes of each node as the hidden embedding of each agent's history information. Context and map information is sometimes taken into account by creating a context node as in EvolveGraph~\cite{li2020evolvegraph}, or embedding a context feature from images as in Trajectron++~\cite{salzmann2020trajectron++}. Instead of directly formulating vehicles as nodes,  SGN~\cite{hu2020scenario} and HATN~\cite{wang2021hierarchical,wang2022transferable,wang2021online} formulated the dynamic insertion area (DIA), a slot that vehicles can insert into. DIAs can be extracted from the driving scenes and then regarded as nodes to construct semantic graphs. In this way, the context and scene information is naturally incorporated into the graph representation, leading to easier learning and more effective generalizability and transferability. Attention mechanisms are also exploited to demonstrate agents' relative importance or interaction intensity \citep{hu2020scenario,wang2022transferable}. Heterogeneous encoding can also be added by applying different embedding functions to different types of agents \citep{li2020evolvegraph}, incorporating an online adaptation module \citep{wang2021online}, or designing an extra category layer \citep{ma2019trafficpredict}.

To better capture agent-map relationship reasoning, recent works have extended graph representations to include map elements, commonly referred to as vectorized maps. These maps use polylines with multiple control points and attributes to represent road structures such as lane lines, road curbs and crosswalks with the polylines forming groups of vectors treated as nodes and incorporated into a scene graph.
Early efforts like VectorNet and LaneGCN~\cite{gao2020vectornet} adopted the vectorized map representation and applied GNNs to these maps, showing enhanced performance and efficiency in motion prediction. 
Building on VectorNet, TNT~\cite{zhao2020tnt} introduced target-driven approaches, defining sparse goal anchors and selecting optimal trajectories toward the target. DenseTNT~\cite{gu2021densetnt}  further improved by estimating dense goal candidates, leading to more accurate predictions. This vectorized map representation has since inspired numerous follow-up works and now plays a dominant role in several driving motion prediction benchmarks~\cite{zhou2023query,shi2022motion,zhou2022hivt,varadarajan2022multipath++}.

\textbf{Attention Mechanism}
Another common approach to relational reasoning for motion prediction is the use of attention -- a mechanism that quantifies and amplifies how one feature influences others, thereby representing the relationship amongst features.  Having achieved immense success in the natural language processing domain since their introduction in 2017~\cite{vaswani2017attention}, attention mechanisms have rapidly spread to many other domains, including motion prediction. 
Formally, consider we have $m$ attending features, for which we generate $m$ queries respectively $\textbf{q}=\{q_1, \dots, q_m\}$, and $n$ attended features, for which we generate $n$ pairs of key and value $\{(k_1, v_1), \dots, (k_n, v_n)\}$. 
In the attention mechanism, each attending feature aggregates information from $n$ attended features via $\sum_{i=1}^{n} \alpha(\mathbf{q}, \mathbf{k}_i) \mathbf{v}_i$, where the attention measure function $\alpha(\mathbf{q}, \mathbf{k}_i) \in \mathbb{R} \ (i=1, \dots, m)$ captures the relationship between attended features. 
In motion prediction, the attention mechanism is frequently used to quantify the influence between different types of features over the \textit{temporal dimension} and \textit{spatial dimensions}. We provide here five common ways to leverage the attention mechanism in motion prediction.

\begin{enumerate}
    
\vspace{-0.5em}\item \textit{Temporal Attention.} Historic information is crucial for predicting the future behaviors of dynamic agents. Thus, the attention mechanism is employed in the temporal space to attend to information from different historic time steps~\citep{chen2019attention}. To this end, agent $i$'s information in the time step $t$ is encoded into a hidden vector $\boldsymbol{h}_{i}^t$. Attention can be applied to all hidden vectors of different time steps. For example, predicting agent $i$'s behavior for the future time step $t+1$ needs to consider its own historic behaviors in the past $r$ time steps:
$\boldsymbol{h}_{i}^{t+1} = \sum_{t'=t-r}^{t} \alpha_i^{t,t'} \boldsymbol{h}_i^{t'}$, 
where $\alpha_i^{t,t'}$ is current time step $t$'s normalized attention on the historic time step $t'$ for agent $i$. Note that such a scheme can be easily extended to temporal attention between different agents by considering the attention between agent $i$'s state at time step $t$ and agent $j$'s state at time step $t'$ \citep{wu2021hsta,ngiam2022scene}. Applying attention to temporal data is generally more efficient than using RNNs, as attention mechanisms are inherently parallelizable, whereas RNNs rely on sequential calculations.

\item \textit{Agent Attention.} In real-world multi-agent settings, agents pay attention to others in varying degrees and aspects while interacting with them. Thus an intuitive way to leverage the attention mechanism is to formulate each agent as an entity and compute attention between them \citep{ngiam2022scene,ding2019predicting,leurent2019social,li2022important,mcallister2022control,zhou2024smartrefine,chandra2019traphic}. To this end, each agent's trajectory (or historical trajectory) is independently encoded into a hidden vectorized representation, $\boldsymbol{h}_{i}$, through a shared encoder such as an RNN or temporal attention layers. The feature of target agent $i$ is then updated as an attention-weighted sum of all other agents $j$ in the scene:
$\boldsymbol{h}_{i}^{\prime} = \sum_{j} \alpha_{i,j} \boldsymbol{h}_{j}$
, where $\alpha_{i,j}$ is the normalized attention weight capturing the agent $j$'s effect on the target agent $i$. Such an update can capture the target agent's relationship and interaction with its neighbor agents, considering the attention to each neighbor agent at different levels. 
There are also works injecting graph structure or masks into attention computation, for example, only attending to agents connected in the graph  \citep{velivckovic2017graph} or unmasked agents \citep{ngiam2022scene}.

\item \textit{Agent-Map Attention.} Map information provides crucial geometric and semantic context for motion behavior. Therefore, it has become common to formulate each map element (e.g., lanes, road boundaries, traffic signs, obstacles, target object to manipulate) as an entity and apply attention mechanisms to model their interactions~\citep{wang2023prophnet,chen2022scept,zhou2022hivt,zhou2023query,forecast_mae,liu2021multimodal}. Usually, both self-attention among map elements themselves, and cross-attention between agents and map elements are performed to aggregate information from both agents and maps. 
Self-attention allows each map element to interact with others by computing attention weights based on their relevance to one another, enabling the model to capture dependencies between different parts of the map, such as intersections or road curvatures. For example, under the vectorized representation, each map element $m_i$ is represented by a feature embedding $\boldsymbol{h}_{m_i}$, and is then updated by $\boldsymbol{h}_{m_i}^{\prime} = \sum_{j} \alpha_{i,j} \boldsymbol{h}_{m_j}$.
Regarding cross-attention, agent trajectories and map elements are treated as separate entities. 
Each agent’s trajectory embedding $ \boldsymbol{h}_{a_i} $ is updated by attending to the embeddings of relevant map elements $ \boldsymbol{h}_{m_j}$ via $\boldsymbol{h}_{a_i}^{\prime} = \sum_{j} \alpha_{i,j} \boldsymbol{h}_{m_j}$, where the attention weights $\alpha_{i,j}$ capture the importance of different map elements (e.g., lanes or traffic signals) for the agent’s behavior. 
This cross-attention mechanism allows the model to integrate both spatial relationships and semantic cues from the map, enabling more accurate predictions of agent motion based on environmental constraints.
Note that temporal attention, agent attention, and agent-map attention are usually combined together to comprehensively reason about the scene information from both spatial and temporal dimensions.

\item \textit{Space-Based Attention.} In addition to formulating agents and map elements as individual entities, another practice is to formulate spatial areas as entities, which inherently captures the relationship between agents and maps. Two approaches to space-based attention have been proposed: grid-based attention and insertion-area attention. 
The idea behind grid-based attention is to rasterize the scene as an $M\times N$ spatial grid $H$, where each cell is populated with hidden states of the agent and map information~\citep{messaoud2021trajectory,chen2022intention,guo2022vehicle}. The target agent $\mathbf{h}$ then aggregates information from the spatial grids via attention $\mathbf{h}_{i}^{\prime} = \sum_{m,n}^{M,N} \alpha_{m,n} H_{m,n}$, where $\alpha_{m,n}$ denotes the normalized attention weight between an agent's hidden vector and the cell $(m,n)$ of the embedded spatial grid.

Alternatively, the insertion-area methods aim to capture more delicate contexts in the scene. The main idea is that, in clustered dynamic scenes, agents constantly recognize and pay attention to areas and gaps that they can move into, referred to as dynamic insertion areas (DIAs) \citep{hu2018probabilistic,hu2020scenario,wang2021hierarchical,wang2022transferable}, and then choose one such DIA to drive through. In driving scenes, such decisions occur commonly. For example, when performing unprotected left turns, a driver must wait for a sufficiently large gap, or DIA, between oncoming vehicles before executing the maneuver. Thus, the insertion-area attention scheme views DIAs as entities among which the attention is formulated to model the relationship among potential DIAs, rather than among agents. In practice, DIAs are formed by agent and map markings, thus the relationship amongst DIAs can explicitly represent the interaction between agents and the environment, and cover diverse types of topology patterns. Specifically, each DIA is extracted from the scene and then encoded into a hidden vectorized $\boldsymbol{h}_{i}$. The $i$-th DIA's feature is then updated with the attention-weighed sum of all other DIAs in the scene: $\boldsymbol{h}_{i}^{\prime} = \sum_{j} \alpha_{i,j} \boldsymbol{h}_{j}$, where $\alpha_{i,j}$ are the normalized attention weights, capturing the $j$-th DIA's effect on the $i$-th DIA. The idea of insertion area also extends beyond the DIA, incorporating concepts such as lane homotopies~\cite{chen2023categorical} and topological braids~\cite{mavrogiannis2022analyzing}.
\end{enumerate}

\textbf{Multimodality - Anchors, Probabilistic and Generative Models}
To account for the inherent multimodality and uncertainty in human motion, many methods employ pre-defined anchors~\citep{chai2019multipath,shi2022motion}, learned anchors~\citep{varadarajan2022multipath++,nayakanti2023wayformer,ngiam2022scene}, goal points~\citep{zhao2020tnt,gu2021densetnt}, or occupancy heatmaps~\citep{gilles2021home,gilles2022gohome} to sample multimodal trajectories. They may also generate cost functions to rank candidate trajectories~\citep{casasMP3UnifiedModel2021,zeng2019end}. Moving forward, probabilistic and generative models are further employed.

Probabilistic models, such as Gaussian Mixture Model (GMM)~\cite{wiest2012probabilistic} and Mixture Desntiy Network (MDN)~\cite{hu2018probabilistic}, focus on directly modeling the conditional probability distribution of future motion based on historic data. These models are particularly effective at handling uncertainty by providing probability distributions over possible outcomes rather than single deterministic predictions. 

Generative models, on the other hand, are a class of machine learning techniques designed to learn the underlying data distribution and generate new samples that are consistent with this learned distribution. In the context of motion prediction, generative models excel at capturing the inherent uncertainty and multimodality of future motions. Unlike probabilistic models, the generative models typically use latent variables to represent the underlying multi-modality of data, enabling them to generate diverse future trajectories, each representing different plausible outcomes. 
Popular methods include 1) conditional variational autoencoders (CVAE)~\cite{salzmann2020trajectron++,tang2019multiple,casas2020implicit,rhinehart2018r2p2}, and Wasserstein Auto-Encoder~\cite{ma2019wasserstein}, which parameterize the probabilistic distribution of latent trajectory embeddings conditioning on historic motion and surrounding contexts; 2) generative adversarial networks (GANs)~\cite{gupta2018social,li2019interaction,li2019conditional,sadeghian2019sophie,zhao2019multi}, where a generator produces predicted trajectories and a discriminator distinguishes between real and generated trajectories, with both networks trained adversarially to improve one another; and 3) diffusion models~\cite{barquero2023belfusion,jiang2023motiondiffuser,saadatnejad2023generic,gu2022stochastic,abuduweili2025enhancing}, a latent-variable-free approach that learns a denoising function using noisy data to approximate the underlying data distribution. They generate samples by iteratively removing noise and refining predictions, starting from pure Gaussian noise.
With a focus on diversity, generative models are also used to go beyond motion prediction to motion generation~\cite{zhang2024motiondiffuse,tan2021scenegen,ding2023realgen,tan2023language}, a valuable tool for simulation and data generation.

Together, anchor-based methods, probabilistic models, and generative models offer complementary approaches to understanding and predicting diverse future motions. Anchor-based methods leverage domain knowledge in motion prediction to formulate the problem as a classification task, providing a structured and interpretable framework for trajectory prediction. Probabilistic models emphasize precise probability estimation, while generative models focus on flexible sampling and representation of complex multimodal distributions. This combined capability is crucial for tasks like autonomous driving or human motion analysis, where accurately accounting for uncertainty and multiple potential outcomes is essential.

\subsection{Application Domains}
\label{sec: application domains}
Motion prediction is a common challenge across multiple domains, which has resulted in overlapping methodologies that are shared across application domains, and also many methods that are specific to the unique requirements of each domain.  In this subsection, we outline the key differences between application domains and how they affect prediction methodology. 

\subsubsection{Autonomous Driving and Navigation}

\begin{figure}[t!]
    \centering
    \includegraphics[width=0.8\textwidth]{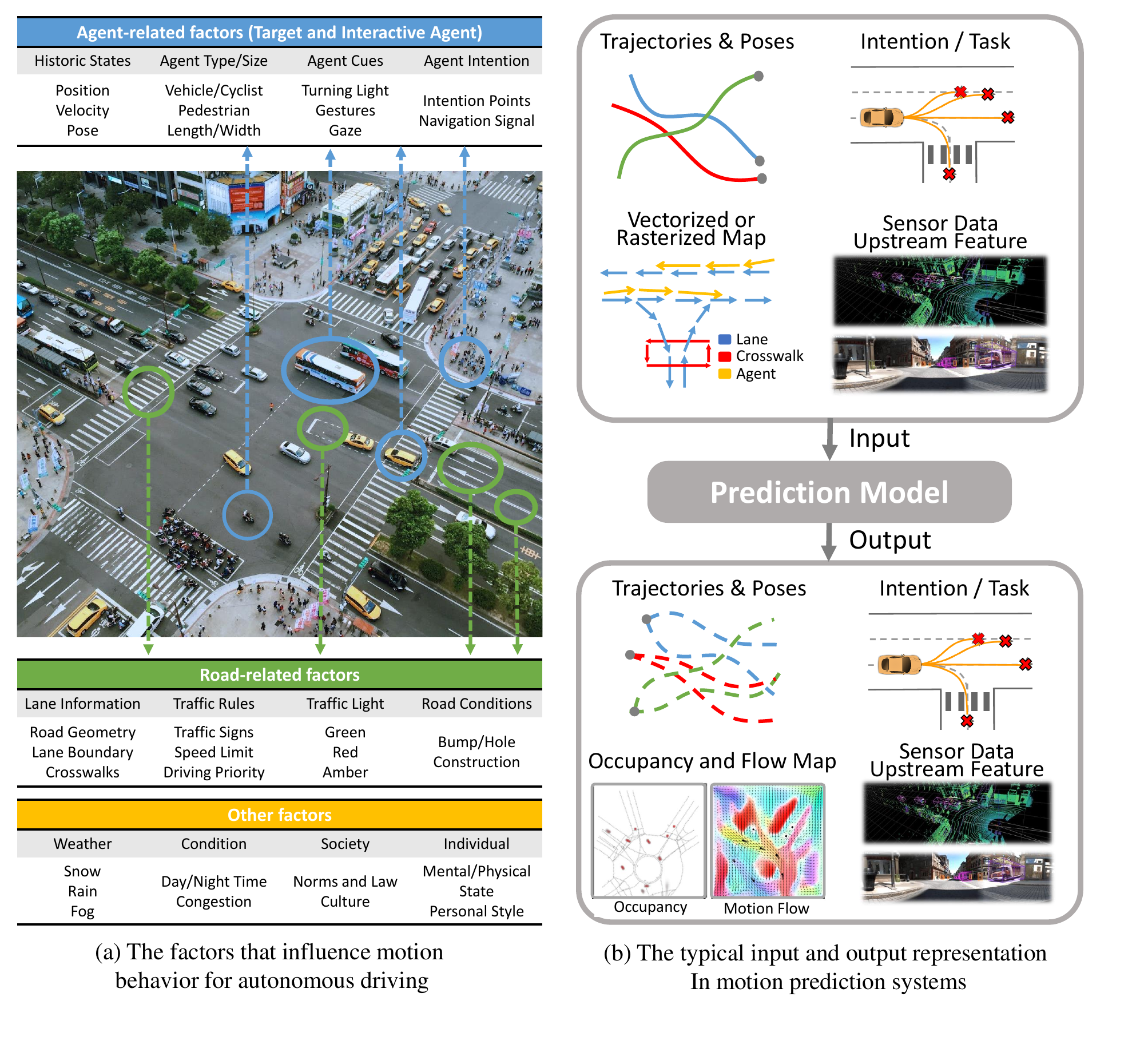}
    \caption{For autonomous driving, we illustrate (a) the motion factors that influence agent behavior, (b) the common input and output representations employed in existing methods. Note, however, that the actual motion factors influencing motion behavior extend beyond what existing works have covered to date. Figure adapted from ~\cite{deziel2021pixset,mahjourian2022occupancy}. }
    \label{fig:ad representation}
\end{figure}

Autonomous driving seeks to automate the process of navigating safely along roadways through a wide range of vehicles, pedestrians, environmental conditions and regulations. The long-standing goal of autonomous driving has had an outsized impact on the development of motion prediction methods, due to the immense importance of high-accuracy, real-time predictions for the safety of the autonomous vehicle, and to the vast investment that has occurred in the domain over the last 20 years. Examples of motion prediction in autonomous driving are illustrated in Figure \ref{fig:ad representation}(a).  The industry can be traced back to the pioneering works of Ernst Dickmanns~\cite{dickmanns} among others, who demonstrated navigation via computer vision in 1985, and laid the foundation for future motion prediction works through his efforts on Dynamic Vision. The goal of prediction for autonomous driving is to extend detections and tracks of vehicles, pedestrians, cyclists and other road agents a short period into the future to enable conflict-free driving actions to be planned. 
This usually involves predicting these road users' ground-level 2D states, which typically includes essential information like position, velocity, and heading. To achieve accurate motion prediction, autonomous driving systems consider many crucial factors in their predictive models. First and foremost, the historical states of ego agents and observed surrounding agents are taken into account. Additionally, motion prediction systems incorporate critical environmental information, including road maps, lane markings, curbs, and nearby structures to understand the spatial constraints and rules of the road. Traffic regulations, like traffic signs and traffic lights, are also factored in to anticipate how traffic participants are likely to proceed in a scenario.
In the literature, the full gamut of input and output representations (e.g. trajectory~\cite{gao2020vectornet,salzmann2020trajectron++}, intention~\cite{zhao2020tnt,cheng2020towards,fisac2018probabilistically}, occupancy and flow~\cite{mahjourian2022occupancy,agro2023implicit,hong2019rules,bansal2018chauffeurnet}), and methodologies (e.g. physics~\cite{helbing1995social,treiber2000congested}, planning~\cite{wang2021socially,schwarting2019social} and learning-based~\cite{shi2024mtr++,zhou2023query,zhou2022hivt} prediction techniques) have been used for this task. Many of these methods have been successfully deployed in real-world vehicle autonomy stacks, enabling navigation in some of the most complex and busiest streets and intersections in the world. Notably, the recent performance of Waymo’s self-driving taxi service has demonstrated parity—or even superiority—relative to human drivers, thanks in no small part to accurate predictions of the future motions of all road participants~\cite{waymosafety}. 



\subsubsection{Human-Robot Interaction}
\begin{figure}[ht!]
    \centering
    \includegraphics[width=0.8\textwidth]{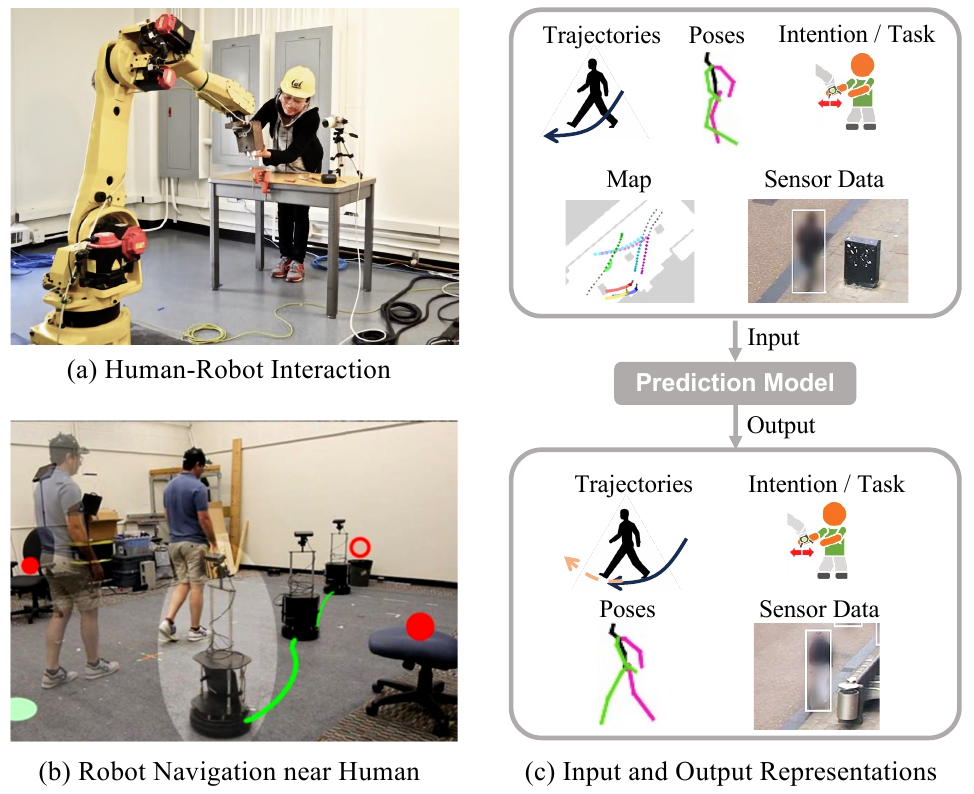}
    \caption{Human motion prediction is crucial for human-robot interaction applications, where we need to (a) predict the 3D human body pose for robots to seamlessly collaborate with humans \cite{liu2016algorithmic}, (b) predict BEV 2D position and optionally 3D human pose for robots to navigate in crowds \cite{bajcsy2020robust}. In (c), we illustrate the common input and output representations employed in existing human motion prediction methods.}
  \label{fig:human_motion_represent}
\end{figure}
In human-robot interaction (HRI)~\cite{kim2017anticipatory,liu2023proactive}, motion prediction also plays a central role in enabling robots to plan and execute tasks, navigate dynamic and complex environments, and interact seamlessly with humans and other objects in shared human-centric space. Examples of motion prediction in human-robot interactions are illustrated in figure \ref{fig:human_motion_represent}(a,b). To enable safe HRI, robots must be able to perceive human activities and reason about human intentions/trajectories \cite{thomaz2016computational,yu2024robustifying}, in order to avoid collision during close operations and collaboration with humans \cite{abuduweili2019adaptable,cheng2020towards}.This can encompass a wide range of robotic applications: from manufacturing robots streamlining production lines \cite{jahanmahin2022human,liu2016algorithmic}, to warehouse robots efficiently managing logistics \cite{ng2020adaptive}, to home assistant robots providing personalized services \cite{szot2021habitat,abuduweili2024wehelp}, to medical robots assisting in delicate surgeries \cite{taylor2016medical}, and even to cooking robots automating culinary tasks \cite{bollini2013interpreting}. In the world of robotics, motion prediction not only involves predicting the ground-level 2D states of surrounding or interesting agents (e.g. humans, mobile devices), but also the articulated 3D poses (robotic arm configurations, end-effector states, human poses and joint angles).  The specific inputs may vary depending on the type of robot and its application, but some common inputs include surrounding agents' current and historic trajectories and poses, environmental data such as maps and scene semantics, and agents' tasks/intents. 

Human motion prediction faces different challenges and requirements in local planning versus global planning. Local motion planning and collision avoidance of robots is the first concern in HRI. To avoid collision between robots and people, the human motion prediction model should have fast and efficient inference for predicting short-term future trajectories for several people around the robot \cite{balan2006real}.  In the simplest case, even linear velocity projections are sufficient for safeguarding and smoothing the robot’s local planning solution\cite{chen2017decentralized}. More advanced methods handle human-human interaction \cite{pellegrini2009you,alahi2016social,gupta2018social,li2024multi}, model the influence of the robot’s presence and actions on human motion \cite{oli2013human,schmerling2018multimodal,rhinehart2019precog}, and perceive high-level body cues from human motion for disambiguating their short-term intentions \cite{unhelkar2015human,kooij2019context}. In safety-critical applications, reachability-based methods provide a guarantee of local collision avoidance \cite{bansal2020hamilton,tian2022safety}. 

For global path and task planning in HRI, long-term and multi-hypothesis human-motion predictions, or high-level intention predictions are desired, posing a considerably more challenging task for the prediction system in terms of modelling the inherent multi-modality and uncertainty of human behaviors.  Real-time reactivity requirements are usually relaxed in this setting, while understanding both dynamic \cite{ma2017forecasting} and static contextual cues \cite{kitani2012activity,chung2010mobile} which influence motion in the long-term, reasoning over the map of the environment \cite{karasev2016intent} and inferring intentions of observed agents \cite{vasquez2016novel,best2015bayesian,rehder2018pedestrian} become more important.

\subsubsection{Visual Motion Understanding}
\begin{figure}[h!]
    \centering
    \includegraphics[width=0.95\textwidth]{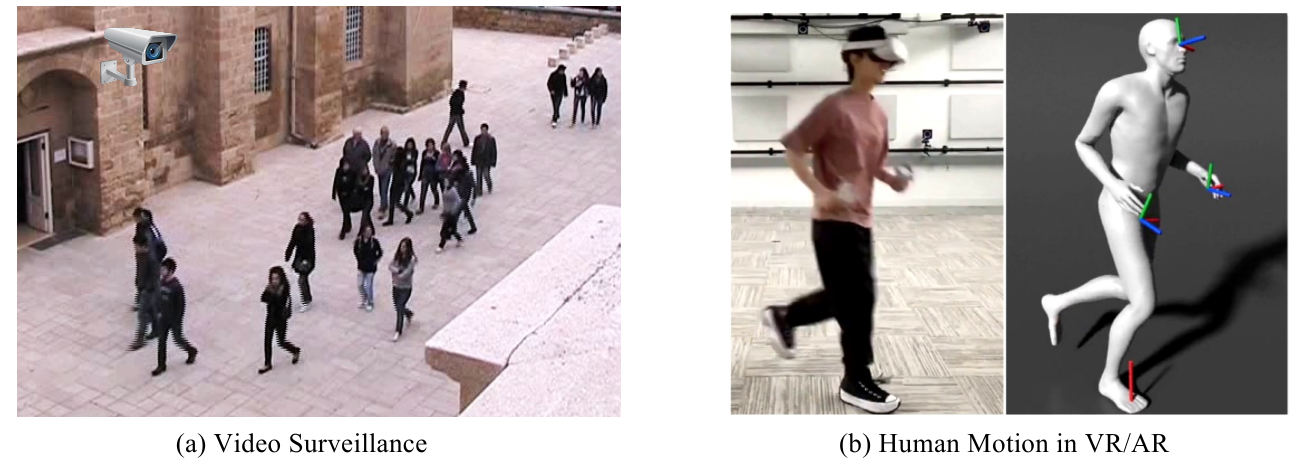}
    \caption{In the application of visual motion understanding, motion prediction is also widely applied in (a) video surveillance and monitoring \cite{lerner2007crowds}, and (b) VR/AR and computer graphics applications \cite{winkler2022questsim}.}
    \label{fig:visual_motion_app}
\end{figure}

\textbf{Surveilance and Monitoring} The classification of goals and behaviors as well as the accurate prediction of human motion is of great importance for surveillance applications such as operation and construction monitoring, retail analytics, or crowd control. An example of motion prediction in visual motion understanding is depicted in Figure \ref{fig:visual_motion_app}(a).  Common installations for these applications use stationary sensors to monitor the environment. In any industrial environment with humans in the workspace, understanding and predicting human motions are critical building blocks for creating a safer and better working environment \cite{rashid2017coupling}. For example, in-situ worker motion predictions can provide a warning system for reducing falling risks based on the relationship between the environment and the worker motion features \cite{novin2021development}. Early injury prevention for critical events can also be realized via human motion simulations and predictions \cite{bataineh2016neural}.
Traffic monitoring or behavior monitoring applications can benefit from long-term prediction models, as they help tracking systems to associate new observations with existing tracks \cite{pellegrini2009you,yamaguchi2011you,luber2010people}. Furthermore, prediction enables the analysis and control of customer flows in populated areas such as malls and airports, by gathering extensive information on human motion patterns
\cite{yoo2016visual,tay2008modelling}, understanding crowd movement in light and dense scenarios, tracking individuals within them, and making future predictions of individuals or crowds under different scenarios (e.g. crowd density prediction for emergency egress). Often these methods benefit from employing sociological methods which include social interaction, behavior analysis, as well as group and crowd mobility modeling \cite{antonini2006behavioral,zhou2015learning,ma2017forecasting,li2022evolvehypergraph}.

\textbf{Action Gaming and VR/AR} Recently, a new generation of full-body action games, such as dance and sports games, has increased the appeal of gaming to family members of all ages \cite{jones2012codename}.
Understanding human actions, therefore, has become a core requirement of interactive or full-body action games \cite{majoe2009enhanced,hauri2021multi}. 
To enable accurate perception of human actions, these games use cost-effective RGB-D sensors \cite{shotton2012efficient}, which provide color and depth data to the game engine \cite{yang2014super,hadfield2013hollywood}. This depth data encodes rich structural information of the entire scene and facilitates motion prediction and action recognition tasks as it simplifies intra-class motion variations and reduces cluttered background noise \cite{kong2015bilinear,jia2014latent}. The motion prediction task enables games to react more quickly to player actions, and to resolve tracking and detection ambiguities when a player self-occludes or moves rapidly. Furthermore, understanding and modeling articulated human poses is essential for creating realistic in-game avatars that move naturally in response to the player's movements~\cite{hou2019head}. In VR/AR environments, motion prediction enhances immersion by enabling real-time interaction, allowing virtual objects or characters to respond fluidly to human gestures and movements.  


\subsubsection{General Time-Series Prediction}
Motion prediction is a specialized case of time-series prediction, where the temporal dynamics pertain to physical agent motion in space. Consequently, many methods developed for motion prediction can be applied interchangeably with general time-series prediction tasks, 
whose applications are vast, ranging from weather forecasting where meteorological variables (e.g. temperature, precipitation, wind speed/direction) are predicted~\cite{lam2023learning,gneiting2005weather}, to financial market forecasting which models and predicts stock prices and market trends~\cite{plummer2009forecasting,hsu2016bridging}. In healthcare, time-series models help monitor patient vitals, predict disease progression, and manage chronic conditions like diabetes~\cite{kaushik2020ai,morid2023time}.Similarly, energy demand forecasting enhances electricity generation and grid management~\cite{islam2020energy,ahmad2018comprehensive}, while businesses utilize such models to improve supply chain efficiency and inventory control~\cite{kochak2015demand,gayam2021optimizing}. 
In robotics and physical systems, time-series prediction also plays a central role. Models are employed to learn robot dynamics~\cite{rueckert2017learning,chen2024korol,li2024continual} and to predict the behavior of physical objects in motion~\cite{han2022learning}, enabling more adaptive and data-driven control strategies. Likewise, time-series models are used for anomaly detection and predictive maintenance in industrial IoT settings~\cite{dandyala2020predictive,dong2021effective,fahim2019anomaly}, and in smart transportation systems to anticipate traffic congestion and improve urban planning~\cite{tascikaraoglu2018evaluation,neelakandan2021iot}. 
Lastly, environmental monitoring and climate change prediction utilize statistical methods and machine learning models to forecast air quality, pollution levels, and long-term climate trends~\cite{heydari2022air,arsov2021multi}, aiding policymakers in making informed decisions. Each application presents unique challenges, including data complexity, external influences, and the need for real-time predictions.


\subsection{Evaluation Strategy}
Due to the broad application domains and various input/output representations in motion prediction, numerous evaluation metrics have emerged in the literature, each measuring a slightly different aspect of prediction performance, such as accuracy, the proportion of predictions within acceptable ranges, safety, and uncertainty.
Datasets and benchmarks have also been standardized for different applications, such as autonomous driving and human motion prediction, allowing practitioners to fairly assess advances. We will discuss both performance metrics and datasets in the remainder of this section.

\subsubsection{Performance Metrics} \label{sec:metrics}
As mentioned earlier, existing metrics focus on different aspects of prediction performance, such as accuracy, the proportion of predictions within acceptable ranges, safety, and uncertainty. Moreover, these metrics are closely tied to the output representation used in each task, such as trajectory, intention, occupancy, and flow. In this section, we provide an overview of commonly used metrics in motion prediction, summarized in Table~\ref{tab:metrics} and elaborated below.

\begin{table}[t!]
\centering
\resizebox{0.8\textwidth}{!}{%
\begin{tabular}{@{}c|c|c|c|c|c|c|c@{}}
\toprule \toprule
& \multicolumn{4}{c|}{\textbf{Trajectory}} & \multirow{2}{*}{\textbf{Intention}} & \multirow{2}{*}{\textbf{Occupancy}} & \multirow{2}{*}{\textbf{Flow}} \\ \cmidrule(lr){2-5}
& \multicolumn{1}{c|}{\textbf{Accuracy}} & \multicolumn{1}{c|}{\textbf{Coverage}} & \textbf{Safety} & {\textbf{Uncertainty}} &  &  \\ \midrule
\multirow{3}{*}{\textbf{\shortstack{Confidence\\Agnostic}}} & ADE/MSE & MissRate & Collision Rate & N/A & Binary Accuracy & AUC & EPE \\ 
 & minFDE & DAC &  &  &  &  Soft-IoU  \\ 
 & minADE  &  &  &  &  &  mAP \\ \midrule
\multirow{4}{*}{\textbf{\shortstack{Confidence\\Aware}}} & p-minFDE & p-MissRate & - & KL Divergence & - &  Soft-IoU & - \\ 
 & p-minADE & mAP &  & Likelihood &  & mAP \\ 
 & Brier-minFDE & Soft mAP &  &  &  & Soft mAP  \\ 
 & Brier-minADE & & & & & \\ \bottomrule \bottomrule
\end{tabular}%
}
\caption{Overview of the categorization of prediction evaluation metrics, which is closely tied to the output representation used in the task, such as trajectory, intention, occupancy, and flow. They emphasize different aspects of prediction performance, including accuracy, the proportion of predictions within acceptable ranges, safety, and uncertainty.}
\label{tab:metrics}
\end{table}

We first introduce evaluation metrics under the trajectory representation, as these are the most extensive and widely used. Considering one agent in one scenario at a certain time step, we denote $\mathbf{s}\in \mathbb{R}^{T\times D}$ as its ground-truth future trajectory over the next $T$ time steps, where $D$ denotes the state dimension of the waypoint at each time step, including positions and headings. For human pose prediction tasks, $D$ also includes the number of key points, such as $3 \times 17$ in the case of 17-keypoint 3D pose prediction. We use $\hat{\mathbf{s}}\in \mathbb{R}^{K\times T \times D}$ to represent the $K$ multi-modal predicted trajectories over $T$ time steps, and use $\mathbf{p}\in \mathbb{R}^{K}$ to denote the probability for each of the $K$ predicted trajectories. 
We denote $\hat{\mathbf{s}}^{k'}$ as the most accurate trajectory, among all the predicted trajectories.
For single-mode motion prediction tasks, we set $K=1$. Metrics under this representation can be categorized into four sub-groups: accuracy metrics, coverage metrics, safety metrics, and uncertainty metrics.

\textbf{Trajectory-Accuracy Metrics} This group of metrics focuses on the accuracy of predictive trajectories. At their core, these metrics measure the accuracy of the predicted endpoint or of each waypoint in the entire trajectory. They are then extended to account for multi-modality and prediction confidence. We will introduce seven metrics in this group: ADE, FDE, minFDE, minADE, p-minFDE, p-minADE, Brier-minFDE and Brier-minADE. The first two metrics are specifically designed for unimodal prediction tasks.
\begin{itemize}
    \item  \textit{Average Displacement Error} (ADE) assumes only unimodal future motion is predicted, and calculates the mean L2 (Euclidean) distance between each point in the predicted unimodal trajectory $\hat{\mathbf{s}}_t$ and the corresponding ground truth trajectory points.
\begin{equation}
    \text{ADE} = \frac{1}{T} 
 \sum_{t=1}^{T} \| \hat{\mathbf{s}}_t - \mathbf{s}_t \|_2
\end{equation}
In some studies, the L1 (Manhattan) distance is used instead of the L2 distance to calculate trajectory prediction error. When using the L1 distance, ADE is equivalent to the \textit{Mean Absolute Error} (MAE). Additionally, some studies evaluate trajectory prediction error using the squared L2 distance, which results in the \textit{Mean Squared Error} (MSE): 
\begin{equation}
    \text{MAE} = \frac{1}{T} 
 \sum_{t=1}^{T} \| \hat{\mathbf{s}}_t - \mathbf{s}_t \|_1, ~  \text{MSE} = \frac{1}{T} 
 \sum_{t=1}^{T} \| \hat{\mathbf{s}}_t - \mathbf{s}_t \|_2^2
\end{equation}
\end{itemize}

\begin{itemize}
\item  \textit{Final Displacement Error} (FDE) similarly assumes only unimodal future motion is predicted, and calculates the mean L2 (Euclidean) distance between the final waypoint in the predicted unimodal trajectory $\hat{\mathbf{s}}_t$ and the corresponding ground truth trajectory points.
\begin{equation}
    \text{FDE} = \| \hat{\mathbf{s}}_T - \mathbf{s}_T \|_2
\end{equation}
\end{itemize}

The following two metrics extend beyond unimodal predictions to capture multi-modality, but only evaluate on the most accurate prediction without accounting for the confidence of multi-modal predictions:
\begin{itemize}
    \item \textit{Minimum Final Displacement Error} (minFDE) quantifies the L2 distance between the endpoint of the most accurate predicted trajectory and the ground truth. In this context, the most accurate trajectory is identified as the one with the lowest endpoint error, resulting in a metric that highlights the precision of the prediction in its final position.
    \begin{equation}
    \text{minFDE} = \min_{k \in \{1, \dots, K\}} \| \hat{\mathbf{s}}_T^k - \mathbf{s}_T \|_2
    \end{equation}     
    For simplicity, in later equations, we denote $\hat{\mathbf{s}}^{k'}$ as the most accurate trajectory, among all the predicted trajectories.
    \item \textit{Minimum Average Displacement Error} (minADE) calculates the mean L2 distance between all waypoints in the most accurate predicted trajectory and the ground truth trajectory.
\begin{equation}
    \text{minADE} = \frac{1}{T} 
 \sum_{t=1}^{T} \| \hat{\mathbf{s}}_t^{k'} - \mathbf{s}_t \|_2
\end{equation}
\end{itemize}

Building on minFDE and minADE, the latter four metrics incorporate prediction confidence into their evaluation. These metrics are divided into two sets, distinguished by the prefixes "p-" and "Brier-", based on how the confidence is calculated. \label{eqs:probabilistic_metrics}
\begin{itemize}
    \item \textit{Probabilistic minimum Final Displacement Error} ($p$-minFDE) extends the concept of minFDE by incorporating the probability of the most accurate predicted trajectory. The metric adds the minimum of negative logarithms of the most accurate trajectory's probability $p^{k'}$ and negative logarithms of a probability threshold value $p_{thre}$ (e.g. 0.05) to the L2 distance at the endpoint. This adaptation acknowledges the confidence in the trajectory prediction.
    \begin{equation}
        p\text{-minFDE} = \| \hat{\mathbf{s}}_T^{k'} - \mathbf{s}_T \|_2 + \min \left( -\log(p^{k'}), -\log(p_{thre}) \right)
    \end{equation}
    \item \textit{Probabilistic minimum Average Displacement Error} ($p$-minADE) builds upon the minADE metric by adding a probabilistic component. Similar to $p$-minFDE, it involves adding the minimum of negative logarithms of the most accurate trajectory' probability ($p^{k'}$) and negative logarithms of a probability threshold value $p_{thre}$ (e.g. 0.05) to the average L2 distance. 
    \begin{equation}
    p\text{-minADE} = \frac{1}{T} \sum_{t=1}^{T} \| \hat{\mathbf{s}}_t^{k'} - \mathbf{s}_t \|_2 + \min \left( -\log(p^{k'}), -\log(p_{thre}) \right)
    \end{equation}

    \item \textit{Brier minimum Final Displacement Error} (Brier-minFDE) adapts the minFDE metric by adding a different probabilistic measure. It involves augmenting the L2 distance at the endpoint with $(1.0 - p^{k'})^{2}$, where $p^{k'}$ is the probability of the most accurate predicted trajectory. This metric thus combines the final displacement error with the Brier score measure of the uncertainty in the trajectory's prediction.
    \begin{equation}
        \text{Brier-minFDE} = \| \hat{\mathbf{s}}_T^{k'} - \mathbf{s}_T \|_2 + (1.0 - p^{k'})^2
    \end{equation}

    \item \textit{Brier minimum Average Displacement Error} (Brier-minADE) is an extension of the minADE metric, incorporating the Brier score probability factor. Similar to Brier-minFDE, it calculates the average L2 distance, adding to it $(1.0 - p)^{2}$. This version of minADE hence evaluates the average displacement error while factoring in the uncertainty of the trajectory forecast.
    \begin{equation}
        \text{Brier-minADE} = \frac{1}{T} \sum_{t=1}^{T} \| \hat{\mathbf{s}}_t^{k'} - \mathbf{s}_t \|_2 + (1.0 - p^{k'})^2
    \end{equation}
    
\end{itemize}

\textbf{Trajectory-Coverage Metrics} This group of metrics emphasizes the percentages of predictions that fall within some acceptable ranges from the ground truth and drivable area. 
We will introduce five metrics in this group: MR, DAC, p-MR, mAP, and soft mAP. The first two metrics address the multi-modality but do not factor in the confidence of the predictions:
\begin{itemize}
    \item \textit{Miss Rate} (MR) is defined as the proportion of agents for which none of the predicted trajectories fall within a specific range of the ground-truth trajectory. In the simplest form, as used in Argoverse dataset~\cite{Argoverse}, the success range is directly set as a fixed value, such as 2.0 meters, and the evaluation is based solely on the endpoint error. 
    \begin{equation}
        \text{MR} = \frac{1}{N} \sum_{i=1}^{N} \mathbb{I} \left( \| \hat{\mathbf{s}}_T^{k'} - \mathbf{s}_T^i \|_2 > r \right)
    \end{equation}
    where $N$ agents are considered for evaluation, $\mathbb{I}(\cdot)$ denotes the indicator function, which returns 1 if the condition inside is true and 0 otherwise, and $r$ denotes the success radius threshold. In more recent formulations, such as in the Waymo dataset~\cite{sun2020scalability}, the success range can vary based on factors such as lateral and longitudinal directions, the agent's initial speed, and the measurement step. This allows for a more nuanced evaluation, where the thresholds for success adapt dynamically to the agent's behavior and context, and cover the entire prediction horizon\footnote{See details in https://waymo.com/open/challenges/2024/motion-prediction/}, rather than only relying on a fixed range for endpoint error.
    \item \textit{Drivable Area Compliance} (DAC) measures a model's capability to produce viable future trajectories. It is calculated as the ratio of the number of trajectories that remain within the drivable area over the total number of trajectories. DAC thus reflects the proportion of trajectories that adhere to the constraints of the drivable space.
    \begin{equation}
    \text{DAC} = \frac{1}{K} \sum_{k=1}^{K} \mathbb{I} \left( \hat{\mathbf{s}}^k_T \in \text{Drivable Area}\right)
    \end{equation}
    where we consider $K$ possible predictions for one agent, and use the endpoint of each predicted trajectory to determine whether it falls within the drivable area, as used in Argoverse dataset~\cite{Argoverse}.
\end{itemize}
The next three metrics incorporate prediction confidence into their evaluation.
\begin{itemize}
    \item \textit{Probabilistic Miss Rate} ($p$-MR), as utilized in Argoverse~\cite{Argoverse}, modifies the original Miss Rate by incorporating the probability factor of the most accurate predicted trajectory. 
    \begin{equation}
        p\text{-MR} = \frac{1}{N} \sum_{i=1}^{N} \Big( \mathbb{I} \big( \| \hat{\mathbf{s}}_T^{k'} - \mathbf{s}_T^i \|_2 > r \big) + \mathbb{I} \big( \| \hat{\mathbf{s}}_T^{k'} - \mathbf{s}_T^i \|_2 \leq r \big) \cdot (1.0 - p_{k'}) \Big)
    \end{equation}
    In this metric, in addition to the original Miss Rate metric (the first term), a contribution of (1.0 - $p^{k'}$) is added when the endpoint error of the most accurate predicted trajectory is within the threshold range $r$ of the ground truth endpoint. And 0.0 is added if no predicted trajectory falls within the $r$ threshold. Here, $p^{k'}$ represents the probability of the most accurate predicted trajectory. This metric integrates the confidence level of the prediction into the miss rate calculation. Unlike the original MissRate, which does not reflect the nuanced quality for predictions that fall in a threshold range, this $p$-MR metric penalizes the low confidence for those predictions.  

    \item \textit{Mean Average Precision} (mAP) is a trajectory prediction metric that emphasizes the effect of the confidence threshold on the predicted coverage. 
    In the case of multiple possible predictions for one agent, this metric first sorts and filters trajectory predictions by a series of confidence thresholds. For the remaining trajectories under one confidence threshold, any trajectory predictions classified as a miss are assigned a false positive, and any that are not considered a miss are assigned a true positive. The definition of miss aligns with the Miss Rate introduced above. 
    Consistent with object detection mAP metrics, only 1 true positive is allowed for each prediction - it is assigned to the highest confidence prediction, all other predictions for the object are assigned a false positive\footnote{See illustrations in https://waymo.com/open/challenges/2024/motion-prediction/}. 
    With the recall and precision values at different confidence thresholds, the AP is calculated as the area under the precision-recall curve. 
    \begin{equation}
     \text{AP} = \sum_{m=1}^{M} \left( R(m) - R(m-1) \right) P_{\text{interp}}(m)
    \end{equation}
    where, we consider $M$ different thresholds. $R(m)$ is the recall at the $m$-th threshold. $P_{\text{interp}}(m)$ is the interpolated precision at the $m$-th threshold. 
    Note that the ground truth trajectory is categorized into different shape buckets, including straight, straight-left, straight-right, left, right, left u-turn, right u-turn, and stationary. After an mAP metric has been computed for all trajectory shape buckets, an average across all buckets is computed as the overall mAP metric.
    \begin{equation}
        \text{mAP} = \frac{1}{C} \sum_{c=1}^{C} \text{AP}_c
    \end{equation}
    where $C$ denotes the total number of shape buckets.
    

    
    \item \textit{Soft mAP} is similar to mAP but differs in how it handles multiple matching predictions for a given ground-truth trajectory. In both metrics, the highest confidence match is counted as a true positive. However, while the standard mAP treats additional matching predictions as false positives, soft mAP ignores these additional matches and does not penalize them, making it a more lenient metric that matches better with the multi-hypothesis prediction task.
\end{itemize}

\textbf{Trajectory-Safety Metrics} In addition to accuracy and coverage, there are also metrics that evaluate the safety aspects. This category has been crucial for autonomous systems but has been relatively less explored, with just one metric identified in this category. 

\textit{Actor Collision Rate (actorCR)}, sometimes referred to as \textit{Overlap Rate}, measures how often predicted trajectories of different objects collide/overlap spatially at the same point in time. It selects the most confident prediction~\cite{sun2020scalability} or the most likely prediction~\cite{wilson2023argoverse} for each object and checks if the agent' trajectory is within a certain close range with other agents' trajectories over the prediction horizon. The rate is calculated as the total number of overlaps/collisions divided by the total number of objects predicted. This metric assesses the model's accuracy and consistency in predicting trajectories in a shared space, since collisions are usually rare in data.
\begin{equation}
    \text{Overlap Rate} = \frac{1}{N}\sum_{i=1}^{N} \sum_{j=i+1}^{N} \mathbb{I}\left( \min_{t \in \{1, \dots, T\}} \|\hat{\mathbf{s}}_t^i - \hat{\mathbf{s}}_t^j\|_2 < \delta \right)
\end{equation}
where we assume $N$ agents in the scene, and $\delta$ denotes the distance threshold for overlap/collision detection between two waypoints.

\textbf{Trajectory-Uncertainty Metrics} One of the drawbacks of the above metrics is their insufficiency in directly measuring the uncertainty of the model predictions, which has been crucial in assessing the level of trust for prediction and determining if the current scene goes out-of-distribution from the training data. 

\begin{itemize}
    \item \textit{Kullback-Leiber (KL) Divergence.} The dissimilarity of two distributions is commonly measured using Kullback-Leiber (KL) divergence~\cite{kullback1951information}. In motion prediction, the KL divergence between the probability distribution of ground-truth human behavior, $p(\mathbf{s}_{1:T})$, and the probability distribution of the predicted behavior, $q(\hat{\mathbf{s}}_{1:T})$ is computed as follows :
    \begin{equation}
        d_{KL}(p \parallel q) \simeq \sum_{\mathbf{s}_{1:T} \in \mathcal{S}} \left\{-p(\mathbf{s}_{1:T}) \log q(\mathbf{s}_{1:T}) + p(\mathbf{s}_{1:T}) \log p(\mathbf{s}_{1:T}) \right\}
    \end{equation}
    where $\mathcal{S}$ denotes the space of all trajectories. A smaller $d_{KL}(p \parallel q)$ indicates less discrepancy between the ground-truth and predicted distributions, meaning more accurate and confident predictions.
    \item \textit{Likelihood.} In practice, directly measuring the probability of the ground-truth trajectory distribution $p(\mathbf{s}_{1:T})$ can be challenging. As a result, several variants of KL divergence are more often used~\cite{salzmann2020trajectron++,ivanovic2022heterogeneous}, such as \textit{negative log likelihood}, expressed as $\sum_{s_{1:T} \in \mathbb{D}} -\log q(s_{1:T})$, or \textit{the predicted probability}, $\sum_{s_{1:T} \in \mathbb{D}} q(s_{1:T})$, where $\mathbb{D}$ denotes the dataset of ground-truth trajectories.
\end{itemize}

\textbf{Intention Metrics} In addition to the trajectory-based metrics mentioned above, which focus on evaluating predictions in the trajectory space, there are also metrics specifically designed to assess the accuracy of intention prediction. These metrics are closely tied to the definition of intention. Frequently, intention prediction is framed as a classification task, such as determining a driving maneuver to be executed (e.g., turning left or right, changing lanes) or identifying the task being performed by an agent (e.g., grabbing an object, or exiting a room). In such cases, \textit{classification accuracy} is a straightforward and commonly used metric to evaluate performance. Additionally, target point prediction is another common task associated with intention prediction, where the trajectory-based metrics discussed earlier can be directly applied. This allows for a more comprehensive assessment of both the predicted intention and the resulting motion.

\textbf{Occupancy Metrics} For occupancy evaluation, the metrics are similar to those used in object detection and segmentation tasks. Commonly used metrics include Intersection over Union (IoU). Advanced metrics are further introduced to account for the prediction confidence and thresholding, such as Area Under the Curve (AUC), Soft Intersection over Union (Soft-IoU), and mean Average Precision (mAP)\footnote{See https://waymo.com/open/challenges/2024/occupancy-flow-prediction/ for more details.}:
\begin{itemize}
    \item \textit{Area Under the Curve} (AUC) is typically used to evaluate the performance of binary classification tasks, utilizing Receiver Operating Characteristic (ROC) curve or other curves like the Precision-Recall curve. In the context of occupancy prediction, it measures the model's ability to distinguish between occupied and unoccupied space. It is computed as the area under the Receiver Operating Characteristic (also referred to as AUROC), which plots the true positive rate (TPR) against the false positive rate (FPR) at various threshold settings: 
    \begin{equation}
    \text{AUC} = \int_{0}^{1} \text{TPR}(\text{FPR}) \, d(\text{FPR})
    \end{equation}
    where $\text{TPR} = \frac{\text{TP}}{\text{TP} + \text{FN}}$ denotes the true positive rate, and $\text{FPR} = \frac{\text{FP}}{\text{FP} + \text{TN}}$ denotes the false positive rate, under one certain threshold. A higher AUC value indicates better performance, with 1 being a perfect classifier and 0.5 representing a random classifier.
    \item \textit{Soft Intersection over Union} (Soft-IoU) is a relaxed version of the traditional Intersection over Union (IoU), commonly used in segmentation tasks. In the context of occupancy, it measures how well the predicted occupied space overlaps with the ground truth, allowing for partial matches by using soft probabilities instead of hard classifications.
    \begin{equation}
        \text{Soft-IoU} = \frac{\sum_{i} p_i \cdot g_i}{\sum_{i} \left( p_i + g_i - p_i \cdot g_i \right)}
    \end{equation}    
    where $p_i$ is the predicted probability of occupancy for voxel $i$, $g_i$ is the ground truth occupancy for voxel $i$.
    Higher Soft-IoU indicates better alignment between predicted and ground-truth occupancy distributions.
    \item \textit{mean Average Precision} (mAP) is is commonly used in object detection tasks to evaluate precision and recall across multiple thresholds. In occupancy prediction, mAP quantifies how well the predicted occupancy matches the ground truth across varying thresholds.
    \begin{equation}
        \text{AP} = \int_{0}^{1} P(R) \, dR
    \end{equation}
    Where \( P(R) \) is the precision as a function of recall. The mean average precision is the mean of the AP values for all objects or classes in the dataset:
    \begin{equation}
    \text{mAP} = \frac{1}{C} \sum_{c=1}^{C} \text{AP}_c
    \end{equation}
    where $C$ is the number of classes (or objects) in the dataset. $\text{AP}_c$ is the average precision for class/object $c$.
    mAP provides a comprehensive evaluation by considering both precision and recall across different thresholds.
\end{itemize}

These metrics provide a quantitative assessment of how well the predicted occupancy aligns with the ground-truth occupancy, allowing for a comprehensive evaluation of the model's performance in occupancy prediction tasks.

\textbf{Flow Metrics} In addition to occupancy forecasting, it is also common to predict the motion flow of each occupancy grid cell, typically represented as a 2D vector. In such motion flow prediction tasks, flow metrics are designed to capture how well the predicted flow matches the ground truth in terms of direction, speed, and magnitude. Below are some common flow metrics used in such tasks:
\begin{itemize}
    \item \textit{Endpoint Error}  (EPE) is a widely used metric to evaluate the accuracy of predicted flow. It measures the Euclidean distance between the predicted flow vector and the ground truth flow vector at each point:
    \begin{equation}
        \text{EPE} = \frac{1}{N} \sum_{i=1}^{N} \|\mathbf{f}_i^{\text{pred}} - \mathbf{f}_i^{\text{gt}}\|
    \end{equation}
    where
    $\mathbf{f}_i^{\text{pred}}$ is the predicted flow vector at voxel $i$,
    $\mathbf{f}_i^{\text{gt}}$ is the ground truth flow vector at voxel $i$,
    $N$ is the total number of voxels. A lower EPE indicates better prediction accuracy.
\end{itemize}

\begin{figure}[t!]
    \centering
    \includegraphics[width=0.95\textwidth]{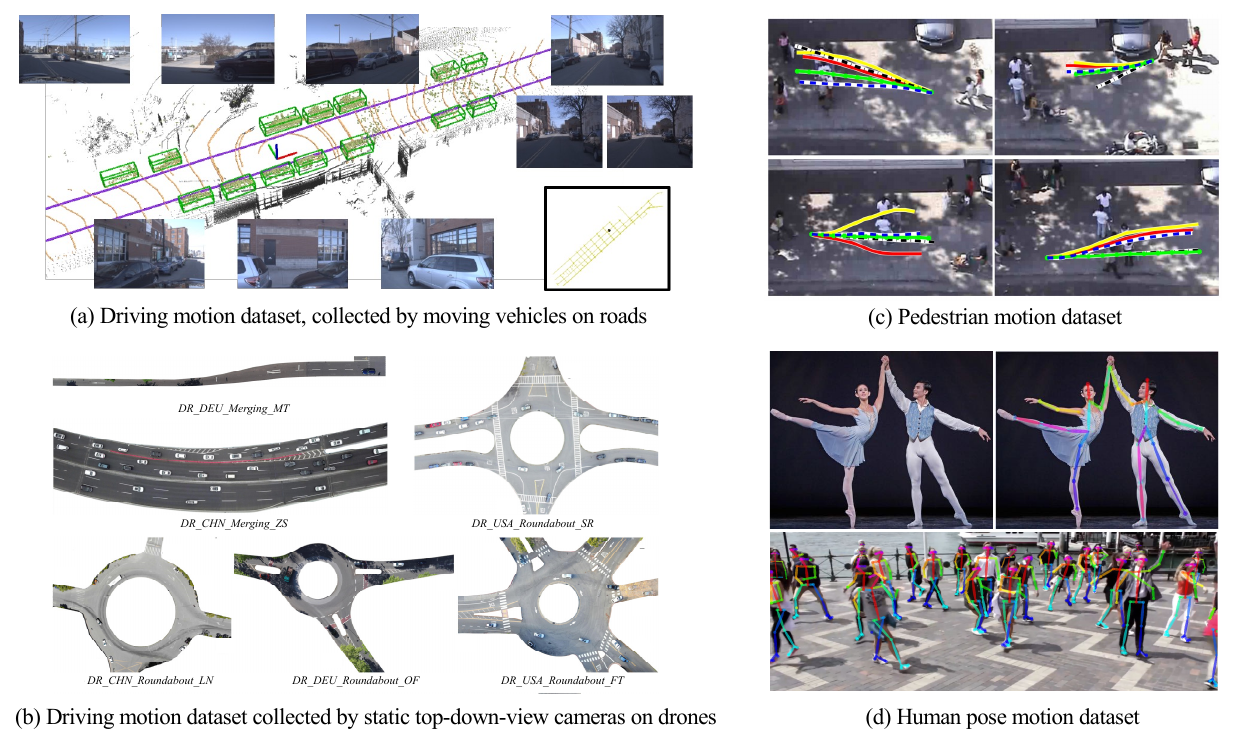}
    \caption{Four common types of motion datasets: (a) Driving motion dataset, collected by vehicles driving across scenes (figure from the Argoverse dataset~\cite{Argoverse}). (b) Driving motion dataset collected by static top-down-view cameras on drones (figure from the INTERACTION dataset\cite{zhan2019interaction}). (c) Pedestrian 2D motion dataset
    (figure from the UCY dataset~\citep{lerner2007crowds}).
    (d) Human pose motion dataset
    (figure from the OpenPose dataset~\cite{cao2017realtime}).
    }
  \label{fig:dataset}
\end{figure}

\subsubsection{Datasets}

\begin{table}[!t]
    \centering
    \resizebox{\linewidth}{!}{%
    \begin{tabular}{ c | c | c | c | c | c | c | c | c | c | c}
    \toprule
    \toprule

    {\multirow{2}{*}{\textbf{Dataset}}} & 
    \multicolumn{4}{|c|}{\textbf{Data Collection}} & 
    \multicolumn{6}{|c}{\textbf{Motion Prediction - Data Format and Representation}} \\ \cmidrule(lr){2-11}

    &  
    {\begin{tabular}[c]{@{}c@{}}\textbf{Countries}\\ \textbf{(Cities)}\end{tabular}} &  
    {\textbf{Scenarios}} &  
    {\textbf{Sensors}} & 
    {\begin{tabular}[c]{@{}c@{}}\textbf{Duration}\\ \textbf{Scenes or Tracks} \\\textbf{Scene Length} \\ \textbf{Frequency} \end{tabular}} & 
    {\begin{tabular}[c]{@{}c@{}}\textbf{Target}\\ \textbf{Agent}\\ \textbf{Type}\end{tabular}} & 
    {\begin{tabular}[c]{@{}c@{}}\textbf{Train/Val/}\\ \textbf{Test Split}\\ \textbf{Size}\end{tabular}}  & 
    {\begin{tabular}[c]{@{}c@{}}\textbf{Historic/}\\ \textbf{Future} \\\textbf{Horizon} \\\textbf{\& Modality}\end{tabular}}  & 
    {\begin{tabular}[c]{@{}c@{}}\textbf{Motion}\\ \textbf{Representation}\end{tabular}} & 
    {\begin{tabular}[c]{@{}c@{}}\textbf{Map}\\ \textbf{Format/}\\ \textbf{Representation}\end{tabular}} & 
    {\begin{tabular}[c]{@{}c@{}}\textbf{Additional}\\ \textbf{Signal}\end{tabular}} \\ \midrule

    \multicolumn{11}{c}{\textbf{Prediction datasets with official benchmark, providing standardized data split, historic/future horizon, and future modality.}} \\ \midrule
    
    {\begin{tabular}[c]{@{}c@{}}Waymo \\ Open Dataset \\ \citep{sun2020scalability}\end{tabular}} &
    {\begin{tabular}[c]{@{}c@{}}USA (SF,\\ Phoenix,\\ Mountain View)\end{tabular}} &
      {\begin{tabular}[c]{@{}c@{}}Urban\\ Suburban\end{tabular}} &
      {\begin{tabular}[c]{@{}c@{}}(Moving) \\ LiDAR\\ Camera\\ Radar\end{tabular}} &
      \begin{tabular}[c]{@{}c@{}}574 hours\\103k scenes\\ 20 sec \\10 Hz \end{tabular} &
      {\begin{tabular}[c]{@{}c@{}}vehicles\\ cyclists\\ pedestrian\end{tabular}} &
      {\begin{tabular}[c]{@{}c@{}}487k\\ 44k\\ 44k\end{tabular}} &
      {\begin{tabular}[c]{@{}c@{}}1 sec / \\ 8 sec \\ 6 \end{tabular}} & 
      {\begin{tabular}[c]{@{}c@{}}2D BEV Traj \\ Occ. and Flow \end{tabular}} & 
      {\begin{tabular}[c]{@{}c@{}}Vectorized \\ Sample Points \end{tabular}} & 
      LiDAR\\ \midrule
    
    {\begin{tabular}[c]{@{}c@{}}Argoverse\\ \citep{Argoverse}\end{tabular}} &
    {\begin{tabular}[c]{@{}c@{}}USA (Miami,\\Pittsburgh)\end{tabular}} &
      {\begin{tabular}[c]{@{}c@{}}Urban\end{tabular}} &
      {\begin{tabular}[c]{@{}c@{}}(Moving)\\ LiDAR\\ Camera\end{tabular}} &
      \begin{tabular}[c]{@{}c@{}}320 hours\\325k scenes\\5 sec\\10 Hz \end{tabular} &
      {\begin{tabular}[c]{@{}c@{}}vehicles\\ cyclists\\ pedestrian\end{tabular}} &
      {\begin{tabular}[c]{@{}c@{}}206k\\ 39k\\ 78k\end{tabular}} &
      {\begin{tabular}[c]{@{}c@{}}2 sec / \\ 3 sec \\ 6 \end{tabular}} & 
       2D BEV Traj & 
       \begin{tabular}[c]{@{}c@{}}Vectorized\\Rasterized\end{tabular} & - \\ \midrule

      {\begin{tabular}[c]{@{}c@{}}Argoverse 2\\ \citep{wilson2023argoverse}\end{tabular}} &
      {\begin{tabular}[c]{@{}c@{}}USA (Austin,\\Detroit, Pitts.,\\Palo Alto, Miami, \\Washington D.C)\end{tabular}} &
      {\begin{tabular}[c]{@{}c@{}}Urban\end{tabular}} &
      {\begin{tabular}[c]{@{}c@{}}(Moving)\\LiDAR\\ Camera\end{tabular}} &
      \begin{tabular}[c]{@{}c@{}}763 hours\\250k scenes\\11 sec\\10 Hz \end{tabular} &
      {\begin{tabular}[c]{@{}c@{}}vehicles\\ cyclists\\ pedestrian\end{tabular}} &
      {\begin{tabular}[c]{@{}c@{}}200k\\ 25k\\  25k\end{tabular}} & 
      {\begin{tabular}[c]{@{}c@{}}5 sec / \\ 6 sec\\  1, 6\end{tabular}} & 
       2D BEV Traj & 
       \begin{tabular}[c]{@{}c@{}}Vectorized\\Rasterized\end{tabular} &
       -
       \\ \midrule

    {\begin{tabular}[c]{@{}c@{}}nuScenes\\ \citep{caesar2020nuscenes}\end{tabular}} &
    {\begin{tabular}[c]{@{}c@{}}USA (Boston)\\Singpore\end{tabular}} &
    {Urban} &
    {\begin{tabular}[c]{@{}c@{}}(Moving)\\LiDAR\\ Camera\\ Radar\end{tabular}} &
    \begin{tabular}[c]{@{}c@{}}5.5 hours\\1k scenes\\20 sec\\2 Hz\end{tabular} &    
    {\begin{tabular}[c]{@{}c@{}}vehicles\\ pedestrian\end{tabular}} &
    {\begin{tabular}[c]{@{}c@{}}32k\\ 9k\\ 9k\end{tabular}} &
      {\begin{tabular}[c]{@{}c@{}}2 sec/\\ 6 sec\\  5, 10\end{tabular}} & 
       2D BEV Traj & 
       \begin{tabular}[c]{@{}c@{}}Vectorized\\Rasterized\end{tabular} & 
       {\begin{tabular}[c]{@{}c@{}} -\end{tabular}} \\ \midrule
       
    \multicolumn{11}{c}{\textbf{Motion Planning Datasets with official benchmark.}} \\ \midrule

    {\begin{tabular}[c]{@{}c@{}}nuPlan 1.1 \\ \citep{caesar2021nuplan}\end{tabular}} &
    {\begin{tabular}[c]{@{}c@{}}USA (Boston, \\Pittsburgh, \\Las Vegas)\\Singpore\end{tabular}} &
    {Urban} &
    {\begin{tabular}[c]{@{}c@{}}(Moving)\\LiDAR\\ Camera\end{tabular}} &
    \begin{tabular}[c]{@{}c@{}}1500 hours\\16733 logs\\-\\20 Hz\end{tabular} &    
    {\begin{tabular}[c]{@{}c@{}}vehicles\\ pedestrian\\bicyle\\Barrier...\end{tabular}} &
    {\begin{tabular}[c]{@{}c@{}}~7M\\ 0.6M\\ 0.75M\end{tabular}} &
      {\begin{tabular}[c]{@{}c@{}}2 sec\\ 8 sec\\ -\end{tabular}} & 
       2D BEV Traj & 
       \begin{tabular}[c]{@{}c@{}}Vectorized\\Rasterized\end{tabular} & 
       {\begin{tabular}[c]{@{}c@{}} 75 Scenarios Types \\ 10\% Sensor Data\end{tabular}} \\ \midrule

    \multicolumn{11}{c}{\textbf{Prediction datasets without an official benchmark, and just providing raw motion and map data.}} \\ \midrule

    {\begin{tabular}[c]{@{}c@{}}highD\\ \citep{krajewski2018highd}\end{tabular}} &
   \begin{tabular}[c]{@{}c@{}}German\\(Cologne)\end{tabular} &
      {6 Highways} &
      \begin{tabular}[c]{@{}c@{}}(Static)\\Camera\end{tabular} &
      \begin{tabular}[c]{@{}c@{}}16.5 hours\\110k tracks\\-\\25 Hz\end{tabular}& 
      \begin{tabular}[c]{@{}c@{}}cars\\trucks\end{tabular} &
      - & 
      - &
      2D BEV Traj & 
      Lane Number/Width & 
      Maneuver Class \\ \midrule

    {\begin{tabular}[c]{@{}c@{}}inD\\ \citep{bock2020ind}\end{tabular}} &
    \begin{tabular}[c]{@{}c@{}}German\\(Aachen)\end{tabular}  &
      {4 Intersections} &
      \begin{tabular}[c]{@{}c@{}}(Static)\\Camera\end{tabular} &
      \begin{tabular}[c]{@{}c@{}}10 hours\\11.5k tracks\\-\\25 Hz\end{tabular}&
      \begin{tabular}[c]{@{}c@{}}car/bus\\truck\\pedestrian\\cylist\end{tabular} &
      - & 
      - &
      2D BEV Traj & 
      \begin{tabular}[c]{@{}c@{}}Lanelet2\\OpenDRIVE\end{tabular} &
      -\\ \midrule

    {\begin{tabular}[c]{@{}c@{}}roundD\\ \citep{krajewski2020round}\end{tabular}} &
    \begin{tabular}[c]{@{}c@{}}German\\(Aachen)\end{tabular}&
      {3 Roundabouts} &
      \begin{tabular}[c]{@{}c@{}}(Static)\\Camera\end{tabular} &
      \begin{tabular}[c]{@{}c@{}}6 hours\\13.7k tracks\\-\\10 Hz\end{tabular} &
      \begin{tabular}[c]{@{}c@{}}Car/Van/Bus\\Truck\\Pedestrian\\Bicylist\end{tabular} &
      - & 
      - &
      2D BEV Traj & 
      \begin{tabular}[c]{@{}c@{}}Lanelet2\\OpenDRIVE\end{tabular} &
      - \\ \midrule


    {\begin{tabular}[c]{@{}c@{}}NGSIM\\ \citep{coifman2017critical}\end{tabular}} &
    {\begin{tabular}[c]{@{}c@{}}USA (I-80,\\ US-101)\end{tabular}} &
      {Highway} &
      {\begin{tabular}[c]{@{}c@{}}(Static)\\ Camera\end{tabular}} &
      \begin{tabular}[c]{@{}c@{}}45 mins\\2.9k tracks\\-\\10 Hz\end{tabular} & 
      {vehicles} &
      - & 
      - &
      2D BEV Traj & 
      Lane Number/Width & 
      - \\ \midrule

    {\begin{tabular}[c]{@{}c@{}}INTERACTION\\ \citep{zhan2019interaction}\end{tabular}} &
    {\begin{tabular}[c]{@{}c@{}}Multiple\\Countries\end{tabular}} &

      {\begin{tabular}[c]{@{}c@{}}Roundabout\\ Intersection\\ Merge\end{tabular}} &
      {\begin{tabular}[c]{@{}c@{}}(Static)\\  Camera\end{tabular}} &
      \begin{tabular}[c]{@{}c@{}}36 hours\\40k tracks\\-\\10 Hz\end{tabular} &
      {\begin{tabular}[c]{@{}c@{}}vehicles\\ cyclists\\ pedestrian\end{tabular}} &
      - &
      - &
      2D BEV Traj &
      Lanelet2 &
      - \\ \midrule
    {\begin{tabular}[c]{@{}c@{}}KITTI\\ \citep{geiger2013vision}\end{tabular}} &
    {\begin{tabular}[c]{@{}c@{}}Germany\\ (Karlsruhe)\end{tabular}} &
      {Urban} &
      {\begin{tabular}[c]{@{}c@{}}(Moving)\\ Camera \\ LiDAR \end{tabular}} &
      \begin{tabular}[c]{@{}c@{}}-\\-\\-\\10 Hz\end{tabular} & 
      {\begin{tabular}[c]{@{}c@{}}vehicles\\ cyclists\\ pedestrian\end{tabular}} &
      - & 
      - &
      2D BEV Traj & 
      - &
      LiDAR, RGB cameras\\ \midrule

    {\begin{tabular}[c]{@{}c@{}}Apolloscapes\\ \citep{ma2019trafficpredict}\end{tabular}} &
    China &
      {Urban} &
      {\begin{tabular}[c]{@{}c@{}}(Moving)\\ LiDAR\\ Camera\end{tabular}} &
      \begin{tabular}[c]{@{}c@{}}105 mins\\-\\-\\2 Hz\end{tabular} & 
      {\begin{tabular}[c]{@{}c@{}}vehicles\\ cyclists\\ pedestrian\end{tabular}} &
      - & 
      {\begin{tabular}[c]{@{}c@{}}3 sec / \\ 3 sec\\ 1\end{tabular}} &
      2D BEV Traj & 
      - & 
      {\begin{tabular}[c]{@{}c@{}}RGB cameras, LiDAR, \\ Semantic Segmentation \end{tabular}} \\ \midrule

    {\begin{tabular}[c]{@{}c@{}}Stanford Drone\\ \citep{robicquet2016learning}\end{tabular}} &
    {\begin{tabular}[c]{@{}c@{}}USA\\ (Stanford Campus)\end{tabular}} &
      {Urban} &
      {\begin{tabular}[c]{@{}c@{}}(Static)\\  Camera\end{tabular}} &
      \begin{tabular}[c]{@{}c@{}}9 hours\\ 19k tracks\\-\\2.5 Hz\end{tabular} & 
      {\begin{tabular}[c]{@{}c@{}}vehicles\\ cyclists\\ pedestrian\end{tabular}} &
      - & 
      {\begin{tabular}[c]{@{}c@{}}2.4 sec / \\ 4.8 sec\\ 1\end{tabular}} &
      2D BEV Traj
       & - & -
      \\ 

      
      \bottomrule\bottomrule

    \end{tabular}}
    \caption{Overview of the commonly used motion prediction datasets for driving scenes.}
    \label{tab: driving motion dataset}
    \end{table}

In this section, we review several widely used datasets for motion prediction tasks, divided into two main categories. Table \ref{tab: driving motion dataset} provides an overview of trajectory prediction datasets in driving scenes, with raw data collected primarily through cameras or LiDAR systems. Table \ref{tab:human_motion_dataset} lists widely-used human motion prediction datasets, including human (3D/2D) trajectory and 3D human pose prediction tasks, where the data is generally acquired via cameras or motion capture (MoCap) systems.

\textbf{Motion Prediction Dataset in Driving Scenes} 
Table \ref{tab: driving motion dataset} summarizes a variety of datasets curated for motion prediction in driving domains. These datasets encompass diverse geographical locations, including various countries and cities, and capture a range of driving environments such as urban, suburban, and highway scenarios. Data collection methods generally fall into two categories: (1) dynamic data captured by vehicles equipped with LiDAR and camera sensors as they traverse driving scenes, and (2) static top-down data collected by cameras mounted on drones.
Datasets captured by moving vehicles, such as Waymo, Argoverse, Argoverse 2, and nuScenes, typically provide the data in the form of sequential driving scenes, where each scene spans a fixed time interval and includes varying numbers of agents. In contrast, datasets gathered using static drone-mounted cameras provide agent trajectories within a designated region, with each agent's time interval being dynamic and determined by the agent's entry into and exit from the captured area. These datasets incorporate diverse agent types, including vehicles, cyclists, and pedestrians.

Several datasets, including Waymo, Argoverse, Argoverse 2, and nuScenes, offer official splits for training, evaluation, and testing, predefined temporal horizons for historical and future data, and future prediction modalities. These standardized configurations enable fair and consistent benchmarking of motion prediction methods. 
Most datasets provide agents' 2D trajectories in bird’s-eye view (BEV), while Waymo extends this by including occupancy and flow tasks. Map representations vary across datasets, encompassing formats such as vectorized maps, rasterized maps, sample points, Lanelet2, OpenDrive, or simplified lane configurations (e.g., number and width of lanes). Some datasets also include auxiliary signals, such as LiDAR data, RGB camera feeds, and agent maneuver classifications, enriching their applicability for perception tasks together with the prediction task.

\textbf{Human Motion Prediction Dataset}
Table \ref{tab:human_motion_dataset} provides a detailed summary of datasets widely used for human motion prediction, including both the trajectory and the pose prediction tasks. The datasets cover a wide range of geographical locations, reflecting their diversity. For example, UCY was collected in Nicosia, Cyprus, while ETH originates from Zurich, Switzerland. Others, like AMASS, combine data from multiple countries, ensuring diverse environmental and cultural contexts. Scenarios vary greatly across the datasets, from urban environments captured in ETH and UCY to campus settings in Edinburgh. Indoor lab environments dominate datasets like HumanEva and Human 3.6M, which are used for detailed motion capture. Unique scenarios such as shopping centers (ATC) and simulation-based urban scenes (ForkingPaths) further expand the breadth of application domains. The sensors used for data collection range from static cameras, as seen in ETH and UCY, to motion capture systems like those employed in HumanEva and CMU Mocap. Other datasets integrate multimodal approaches, such as THÖR, which uses a combination of cameras, LiDAR, and motion capture systems. The duration and frequency of the recording are also related to the sensors. Generally, motion capture system captures more high-frequency data. For instance, CMU Mocap achieves high precision with data collected at 120 Hz, while ATC provides extensive temporal coverage with observations spanning 92 days from visual camera. Certain datasets provide official the historic and future horizons, such as UCY, ETH, and ForkingPaths, while official training/val/test data splits are usually not provided in these datasets, and thus it has been common for papers to customize splits for comparison and evaluation, making it less convenient to reuse other work's results and perform fair comparison.

Motion representation formats are equally diverse, ranging from 2D trajectories in bird’s-eye view (UCY, ETH) to detailed 3D poses (HumanEva, AMASS) and body meshes (AMASS). These formats cater to different research goals, whether they involve high-level trajectory forecasting or fine-grained pose estimation. To provide spatial context, many datasets include map representations. As an instance, pixel-level maps are commonly used in datasets like UCY and ETH. Auxiliary signals enrich many datasets, offering additional layers of information. UCY and ATC include gaze directions, while PIE provides intention and behavior annotations. Bounding boxes and video frames appear in datasets like VIRAT and Town Center, making them versatile for tasks that combine motion analysis with visual data.    

It is also worth noting that some studies in human-robot interaction have introduced specialized human motion prediction datasets tailored for specific tasks. These datasets are often designed for specific applications rather than general motion prediction and include examples like the human pose prediction dataset for human-robot co-assembly tasks developed by \cite{abuduweili2019adaptable, abuduweili2023online}.

\begin{table}[t!]
    \centering
    \resizebox{\linewidth}{!}{%
    \begin{tabular}{ c | c | c | c | c | c | c | c | c | c}
    \toprule\toprule
        {\multirow{2}{*}{\textbf{Dataset}}} & 
        \multicolumn{4}{|c|}{\textbf{Data Collection}} & 
        \multicolumn{5}{|c}{\textbf{Motion Prediction - Data Format and Representation}} \\ \cmidrule(lr){2-10}

    &  
        {\begin{tabular}[c]{@{}c@{}}\textbf{Countries}\\ \textbf{(Cities)}\end{tabular}} &  
        {\textbf{Scenarios}} &  
        {\textbf{Sensors}} & 
        {\begin{tabular}[c]{@{}c@{}}\textbf{Duration}\\ \textbf{Tracks} \\ \textbf{Frequency} \end{tabular}} & 
        {\begin{tabular}[c]{@{}c@{}}\textbf{Target}\\ \textbf{Agent}\\ \textbf{Type}\end{tabular}} & 
        {\begin{tabular}[c]{@{}c@{}}\textbf{Historic/}\\ \textbf{Future} \\\textbf{Horizon} \end{tabular}}  & 
        {\begin{tabular}[c]{@{}c@{}}\textbf{Motion}\\ \textbf{Representation}\end{tabular}} & 
        {\begin{tabular}[c]{@{}c@{}}\textbf{Map}\\ \textbf{Format/}\\ \textbf{Representation}\end{tabular}} & 
        {\begin{tabular}[c]{@{}c@{}}\textbf{Additional}\\ \textbf{Signal}\end{tabular}} \\ \midrule

    \begin{tabular}[c]{@{}c@{}} UCY\\ \citep{lerner2007crowds}\end{tabular} &
     \begin{tabular}[c]{@{}c@{}}Cyprus \\  (Nicosia) \end{tabular} &
        \begin{tabular}[c]{@{}c@{}}Urban, \\ Campus \end{tabular} &
      \begin{tabular}[c]{@{}c@{}}(Static)\\ Camera\end{tabular} &
        \begin{tabular}[c]{@{}c@{}}16.5 min \\ over 700 tracks \\ 2.5 Hz\end{tabular} &
        {Pedestrian} &
        \begin{tabular}[c]{@{}c@{}}3.2 sec / \\ 4.8 sec \end{tabular} & 
        2D BEV Traj &
       \begin{tabular}[c]{@{}c@{}}Pixel-level Map \end{tabular} &
        {Gaze directions}  \\ \midrule
        
    \begin{tabular}[c]{@{}c@{}} ETH (EWAP) \\ \citep{pellegrini2009you}\end{tabular} &
        \begin{tabular}[c]{@{}c@{}}Switzerland \\ (Zurich)\end{tabular} &
        {Urban} &
        \begin{tabular}[c]{@{}c@{}}(Static)\\ Camera\end{tabular} &
        \begin{tabular}[c]{@{}c@{}}25 min \\ 650 tracks \\ 2.5 Hz\end{tabular} &
        {Pedestrian} &
        \begin{tabular}[c]{@{}c@{}}3.2 sec / \\ 4.8 sec \end{tabular} & 
        2D BEV Traj &
        {Pixel-level Map} &
        \begin{tabular}[c]{@{}c@{}}Velocities,\\ groups, obstacles \end{tabular}  \\ \midrule

    \begin{tabular}[c]{@{}c@{}} VIRAT\\ \citep{oh2011large}\end{tabular} &
        \begin{tabular}[c]{@{}c@{}}USA\\(Multiple Sites)\end{tabular} &
        {Diverse} &
        \begin{tabular}[c]{@{}c@{}}(Static)\\ Camera\end{tabular} &
        \begin{tabular}[c]{@{}c@{}} 12.5 hours (annotated) \\ - \\ 2-30 Hz\end{tabular} &
        \begin{tabular}[c]{@{}c@{}}Pedestrian\\ Vehicles\end{tabular} &
        - &
        \begin{tabular}[c]{@{}c@{}} Video frames \end{tabular} &
        Pixel-level Map &
        \begin{tabular}[c]{@{}c@{}} Video frames, \\ bounding boxes, \\ intention, events \end{tabular} \\ \midrule

    \begin{tabular}[c]{@{}c@{}} Daimler\\ \citep{schneider2013pedestrian}\end{tabular} &
        \begin{tabular}[c]{@{}c@{}}Germany\end{tabular} &
        {Urban Road} &
        \begin{tabular}[c]{@{}c@{}}(Moving)\\ Camera \\ LiDAR\end{tabular} &
        \begin{tabular}[c]{@{}c@{}}5 minutes \\ 68 tracks \\ 16 Hz\end{tabular} &
        {Pedestrian} &
        - &
        2D BEV Traj & 
        {Pixel-level Map}  &
        \begin{tabular}[c]{@{}c@{}} Stereo images, \\ bounding boxes, \\ calibration data \end{tabular} \\ \midrule

    \begin{tabular}[c]{@{}c@{}} Central Station (GC)\\ \citep{zhou2012understanding}\end{tabular} &
        \begin{tabular}[c]{@{}c@{}}USA\\(New York)\end{tabular} &
        {Indoor} &
        \begin{tabular}[c]{@{}c@{}}(Static)\\ Camera\end{tabular} &
        \begin{tabular}[c]{@{}c@{}}33 minutes \\ over 10k tracks \\ 25 Hz\end{tabular} &
        {Pedestrian} &
        - &
         2D BEV Traj &
        {Pixel-level Map}  &
        Video frames \\ \midrule

    \begin{tabular}[c]{@{}c@{}} Town Center\\ \citep{benfold2011stable}\end{tabular} &
        \begin{tabular}[c]{@{}c@{}}UK\\(Oxford)\end{tabular} &
        {Urban} &
        \begin{tabular}[c]{@{}c@{}}(Static)\\ Camera\end{tabular} &
        \begin{tabular}[c]{@{}c@{}}5 minutes \\ 230 tracks \\ 25 Hz\end{tabular} &
        {Pedestrian} &
        - &
        2D BEV Traj &
        {Pixel-level Map}  &
        \begin{tabular}[c]{@{}c@{}} Video frames, \\ bounding boxes \end{tabular} \\ \midrule
        
    \begin{tabular}[c]{@{}c@{}} Edinburgh\\ \citep{majecka2009statistical}\end{tabular} &
        \begin{tabular}[c]{@{}c@{}}UK\\(Edinburgh)\end{tabular} &
        {Campus} &
        \begin{tabular}[c]{@{}c@{}}(Static)\\ Camera\end{tabular} &
        \begin{tabular}[c]{@{}c@{}} Over 100 days  \\ 92k tracks \\ 9 Hz \end{tabular} &
        {Pedestrian} &
        - &
         2D BEV Traj &
        {Pixel-level Map} &
        {Bounding box} \\ \midrule
      
    \begin{tabular}[c]{@{}c@{}}PIE\\ \citep{Rasouli2019PIE}\end{tabular} &
        \begin{tabular}[c]{@{}c@{}}Canada\\ (Toronto)\end{tabular} &
        {Urban} &
        \begin{tabular}[c]{@{}c@{}}(Moving) Camera \\ Onboard Diagnostics sensor\end{tabular} &
        \begin{tabular}[c]{@{}c@{}}6 hours \\ 1842 tracks \\ 30 Hz\end{tabular} &
        {Pedestrian} &
        - &
        2D BEV Traj &
        - &
        \begin{tabular}[c]{@{}c@{}} Bounding boxes \\ intentions, behaviors \\ GPS \end{tabular} \\ \midrule

    \begin{tabular}[c]{@{}c@{}}ForkingPaths\\ \citep{liang2020garden}\end{tabular} &
        \begin{tabular}[c]{@{}c@{}}Simulation\\ (3D Urban Scenes)\end{tabular} &
        \begin{tabular}[c]{@{}c@{}}Urban\\ simulation\end{tabular} &
        \begin{tabular}[c]{@{}c@{}}(Synthetic)\\  Camera\end{tabular} &
        \begin{tabular}[c]{@{}c@{}}4 hours \\ 750 tracks \\ 25-30 Hz\end{tabular} &
        {Pedestrian} &
        \begin{tabular}[c]{@{}c@{}}4.8 sec / \\ 10.4 sec \end{tabular} &
        2D BEV Traj &
        \begin{tabular}[c]{@{}c@{}}Semantic   Map\end{tabular} &
        \begin{tabular}[c]{@{}c@{}}Bounding boxes, \\ video frames\end{tabular} \\ \midrule

    \begin{tabular}[c]{@{}c@{}} ATC\\ \citep{brvsvcic2013person}\end{tabular} &
        \begin{tabular}[c]{@{}c@{}}Japan\\(Osaka)\end{tabular} &
        \begin{tabular}[c]{@{}c@{}}Shopping \\ center\end{tabular} &
        \begin{tabular}[c]{@{}c@{}}3D range sensors\end{tabular} &
        \begin{tabular}[c]{@{}c@{}}92 days \\ - \\ 20 Hz\end{tabular} &
        {Pedestrian} &
        - &
         3D Traj & 
        \begin{tabular}[c]{@{}c@{}}Occupancy \\ Map \end{tabular} &
        \begin{tabular}[c]{@{}c@{}}Orientations,\\ velocities, gaze directions\end{tabular} \\ \midrule

    \begin{tabular}[c]{@{}c@{}} TH{\"O}R  \\ \citep{thorDataset2019} \end{tabular} &
        \begin{tabular}[c]{@{}c@{}}Sweden\\(Örebro)\end{tabular} &
        {Indoor} &
        \begin{tabular}[c]{@{}c@{}}Camera\\ LiDAR \\ Mocap\end{tabular} &
        \begin{tabular}[c]{@{}c@{}}60 min \\ 600 tracks \\ 100 Hz\end{tabular} &
        {Pedestrian} &
        - &
        {3D Traj} &
        \begin{tabular}[c]{@{}c@{}}Semantic   Map\end{tabular} &
        \begin{tabular}[c]{@{}c@{}}Orientation, \\ eye gaze \end{tabular} \\  \midrule

    \begin{tabular}[c]{@{}c@{}}Aschaffenburg (APD) \\ \citep{kress2021pose} \end{tabular} &
        \begin{tabular}[c]{@{}c@{}}Germany\end{tabular} &
        {Urban} &
        \begin{tabular}[c]{@{}c@{}}(Moving)\\ Camera\end{tabular} &
        \begin{tabular}[c]{@{}c@{}} - \\ 6526 tracks \\ 25 Hz\end{tabular} &
        \begin{tabular}[c]{@{}c@{}}Cyclists\\ Pedestrian\end{tabular} &
        - &
        {3D Poses} &
        {Pixel-level Map} &
        \begin{tabular}[c]{@{}c@{}}2D poses, \\ behaviors\end{tabular} \\ \midrule
        
    \begin{tabular}[c]{@{}c@{}} CMU Mocap  \\ \citep{cmumocap} \end{tabular} &
        \begin{tabular}[c]{@{}c@{}}USA\\ (Pittsburgh)\end{tabular} &
        \begin{tabular}[c]{@{}c@{}}Indoor, \\ Campus \end{tabular} &
        \begin{tabular}[c]{@{}c@{}}Motion Capture\\ system\end{tabular} &
        \begin{tabular}[c]{@{}c@{}} - \\ 2605 tracks \\ 120 Hz\end{tabular} &
        {Human} &
        - &
        {3D Poses} &
        - &
        {\begin{tabular}[c]{@{}c@{}} Video frames, \\ intention, behaviors \end{tabular}} \\ \midrule

        \begin{tabular}[c]{@{}c@{}} HumanEva\\ \citep{sigal2010humaneva} \end{tabular} &
        \begin{tabular}[c]{@{}c@{}}USA \\ (Indoor lab)\end{tabular} &
        {Indoor} &
        \begin{tabular}[c]{@{}c@{}}Motion Capture \\ Cameras\end{tabular} &
        \begin{tabular}[c]{@{}c@{}} - \\ 56 tracks \\ 120 Hz\end{tabular} &
        {Human} &
        - &
        {3D poses} &
        {- } &
        {2D poses} \\ \midrule

        \begin{tabular}[c]{@{}c@{}} Human 3.6M\\ \citep{ionescu2013human3} \end{tabular} &
        \begin{tabular}[c]{@{}c@{}}Romanian \\ (Indoor lab)\end{tabular} &
        {Indoor} &
        \begin{tabular}[c]{@{}c@{}}Motion Capture \\ Cameras\end{tabular} &
        \begin{tabular}[c]{@{}c@{}} - \\ - \\ 50 Hz\end{tabular} &
        {Human} &
        - &
        {3D Poses} &
        {-} &
        {Intentions, behavior} \\ \midrule

        \begin{tabular}[c]{@{}c@{}} AMASS \\ \citep{mahmood2019amass} \end{tabular} &
        \begin{tabular}[c]{@{}c@{}} Multiple \\ countries \end{tabular} &
        {Indoor/Outdoor} &
        \begin{tabular}[c]{@{}c@{}} Motion Capture  \end{tabular} &
       \begin{tabular}[c]{@{}c@{}} 40 hours \\ Over 10k \\ 30-120 Hz \end{tabular} &
        {Human} &
        - &
        {3D Poses} &
        {-} &
        {3D body mesh} \\ \bottomrule
        
        \bottomrule
    \end{tabular}}
    \caption{Overview of human motion datasets for prediction and analysis.}
    \label{tab:human_motion_dataset}
\end{table}

\textbf{Cross-Dataset Benchmarking}
Beyond these common datasets, some researchers have introduced comprehensive motion prediction benchmarks by combining multiple datasets. For instance, Trajnet \citep{sadeghian2018trajnet} aggregates five datasets to create a benchmark for human trajectory prediction. Similarly, Trajnet++ \citep{Kothari2020HumanTF} extends this approach, focusing on agent-agent interaction scenarios with an expanded dataset incorporating five sources. OpenTraj \citep{amirian2020opentraj} broadens the scope by offering a benchmark that encompasses both vehicle and human motion trajectories, drawing on data from more than 20 datasets.
Trajdata~\cite{ivanovic2024trajdata} provides a unified interface to multiple human/vehicle trajectory datasets. A related work for the task of vehicle trajectory planning is ScenarioNet~\cite{li2024scenarionet}, a simulator aggregating multiple real-world datasets into a unified format and providing a planning development and evaluation framework. UniTraj~\cite{feng2025unitraj} and SmarPretrain~\cite{zhou2024smartpretrain} unified the data formats of multiple driving datasets and conduct large-scale training and cross-dataset evaluation.

%% file: section/3_applicable.tex
\clearpage
\section{Deploying Motion Prediction within Real-World Autonomous Systems}
\label{sec:applicable}

In autonomous systems that operate in dynamic environments, such as autonomous vehicles and robotics, motion prediction does not act in isolation but functions as one module of closed-loop autonomy stacks by receiving upstream localization and perception and informing downstream planning and control. 
Since the 2004 DARPA Grand Challenges, decomposing autonomous systems into separate modules has become a well-established paradigm. 
Under this practice, benchmarks have been created to evaluate these modules independently, allowing researchers to develop novel and effective methods for each component. 
Specifically, for motion prediction, existing benchmarks often provide curated and noise-free upstream perception input, usually utilize trajectories as the interface between upstream and downstream modules, and focus solely on open-loop prediction accuracy without considering the impact on downstream planning and control.
This paradigm has been highly successful, fostering the development of innovative methodologies and representations.

However, when deployed in real-world settings that can deviate from such idealized training conditions, state-of-the-art motion prediction models can come with reduced effectiveness and reliability in deployment. In this section, we discuss the approaches, challenges, and perspectives of developing motion prediction models under realistic deployment standards in autonomous systems, which can be categorized into the following key areas:

\textbf{Issue 1 - Representation} 
Motion prediction is a critical middle ground that bridges perception and planning, making it essential to design and select a representation that seamlessly propagates information through the autonomy stack, is computationally efficient, is compatible with perception and planning modules, and can scale effectively with large datasets. The benefits and drawbacks of common representations, such as object-centric, occupancy-centric, and sensor-level representations, and the fusion of these representations need further exploration. 

\textbf{Issue 2 - Uncertainty/Error Awareness} While prediction accuracy is a primary focus of motion prediction, uncertainty and error awareness are essential for autonomous systems to be aware of their prediction confidence and thus make safer, more reliable decisions in dynamic environments. It enables the system to account for imperfect or incomplete information, sensor limitations, and model capabilities. Moreover, uncertainty and error can arise and propagate across the entire autonomy stack, originating from sensors, perception algorithms, prediction models, and planning algorithms. Ignoring these uncertainties and errors can not only lead to overconfident decision-making but also cause their accumulation throughout the stack, leading to system failures. However, existing benchmarks typically assume ideal or noise-free upstream perception inputs and fail to address how multi-modal predictions impact downstream planning. The challenges of \textit{propagating, calibrating, and reducing uncertainty and error} throughout the autonomy stack remain insufficiently explored.

\textbf{Issue 3 - Joint Learning}
While the classic autonomy stack typically follows a sequential perception, prediction, and planning process, \textit{joint learning} of these modules enables end-to-end optimization by propagating gradients through the entire stack. This approach enhances overall performance by accounting for interdependencies and information sharing between modules, improves system robustness by accounting for and reducing errors and uncertainties, and aligns the performance of motion prediction to that of the overall system. Additionally, such joint learning fosters the exploration of different architecture choices, such as sequential, parallel, or hybrid. For example, \textit{passing module outputs back upstream} provides notable benefits: feeding motion prediction into perception~\cite{liMoDARUsingMotion2023} improves perception accuracy, especially in dealing with sensor limitations, degradation, and occlusions; and incorporating motion planning into prediction~\cite{wang2021socially} accounts for the ego agent's influence on surrounding agents’ future motions, while also enabling more efficient computation by focusing on regions and agents that the ego agent will traverse.

\textbf{Issue 4 - Joint Evaluation} 
While joint learning of multiple modules offers various benefits in realistic settings, as discussed above, conducting a fair and comprehensive evaluation of such models remains challenging. Existing benchmarks tend to fall into two extremes: 1) Standalone modular motion prediction benchmarks, such as Argoverse~\cite{Argoverse} and Waymo~\cite{sun2020scalability}, provide curated perception inputs (e.g., historical trajectories, HD maps) and focus on open-loop prediction accuracy without considering downstream planning. In these cases, motion prediction models are isolated from raw upstream sensor data and perception outputs, and downstream planning, making the joint evaluation with other modules infeasible; 2) End-to-end benchmarks, such as nuPlan~\cite{caesar2021nuplan} and CARLA~\cite{dosovitskiy2017carla}, prioritize assessing ultimate driving performance, often at the expense of evaluating motion prediction itself, if not completely ignoring it. While there have been efforts proposing customized evaluations for joint perception-prediction~\cite{gu2023vip3d} and joint prediction-planning~\cite{huang2023dtpp}, a unified open-source evaluation protocol remains elusive—one that allows for the evaluation of motion prediction models under consistent conditions or across different combinations of perception and planning methods to assess the general effectiveness of motion prediction.

\textbf{Issue 5 - Closed-Loop Evaluation} In addition to joint evaluation, closed-loop evaluation~\cite{phongWhatTrulyMatters2023} is another under-studied area. In closed-loop autonomous systems, the prediction algorithm influences the behavior of the ego agent, which, in turn, influences the behaviors of other agents nearby and then further influences the ego agent's perception in the next timestep. 
This introduces a series of challenges that must be considered, including ensuring temporal consistency across predictions at adjacent time steps, balancing computational latency and accuracy under real-time constraints, and aligning evaluation protocols with system-level safety and task performance.
These interdependencies are often overlooked in current evaluation protocols.

\textbf{Summary and Chapter Outline} Facing these issues, research in motion prediction must emphasize realistic deployment standards by designing expressive and efficient interfaces between modules, dealing with the uncertainties/errors through the autonomy stack, alignment with overall system performance through joint learning and evaluation with other modules,  and promoting closed-loop evaluation. In the remainder of this section, we first introduce the autonomy stack and end-to-end autonomy methods that include explicit or implicit motion prediction in Section~\ref{sec:applicable_end-to-end}. We then discuss how to integrate prediction with perception in Section~\ref{sec:applicable_perception}, and how to integrate prediction with planning in Section~\ref{sec:applicable_planning}, where we discuss the implications of representation design, uncertainty awareness, and joint learning. 
Lastly, Section~\ref{sec:applicable_evaluation} expands the discussion regarding the last two issues, namely joint and closed-loop evaluation. 

\subsection{Motion Prediction and the Autonomy Stack}
\label{sec:applicable_end-to-end}
\subsubsection{Introduction to the Autonomy Stack}
\begin{figure}[!t]
\centering
\includegraphics[width=\textwidth]{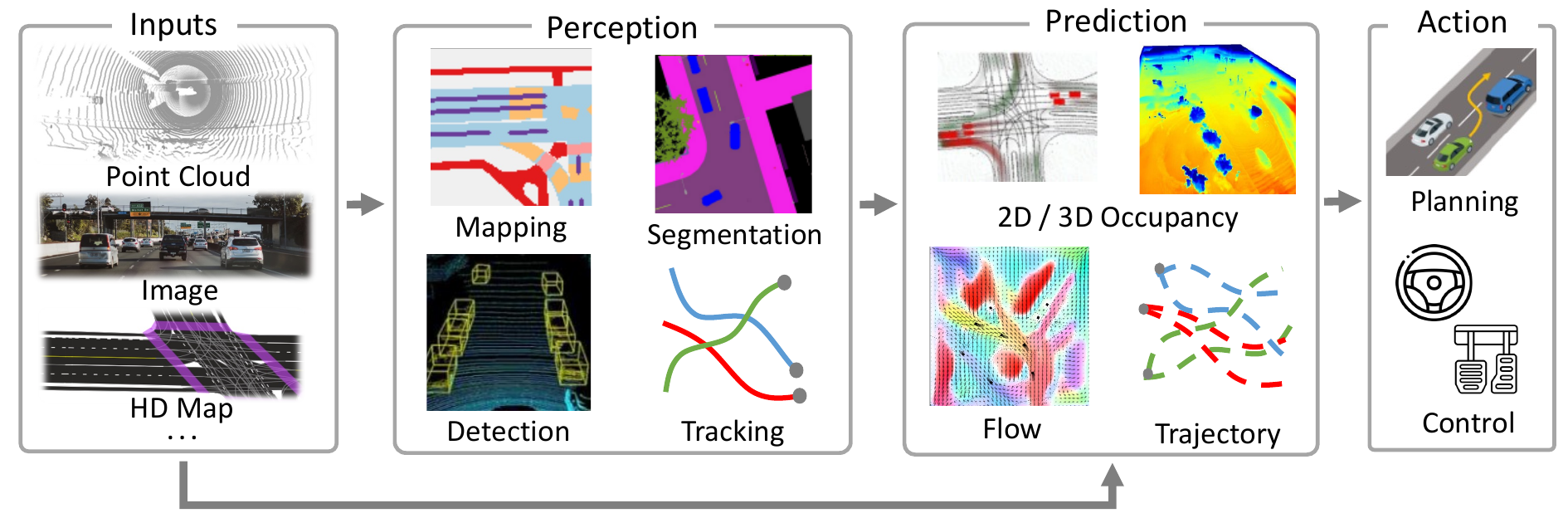}
\caption{End-to-end autonomy can be decomposed with several possible inputs and intermediate tasks. Diagram adapted from past works~\cite{hu2023planning, khurana2023point, liu2023bevfusion, mahjourian2022occupancy}.}
\label{fig:applicable_e2e_decomp}
\end{figure}

\textbf{Input Modalities} 
As shown in Fig~\ref{fig:applicable_e2e_decomp}, autonomous systems collect the environmental information from sensors, such as multi-view camera images, LiDAR points, or both. 
Less common modalities also include radar, which is robust to various weather and lighting conditions, and event cameras, which capture changes in the scene at a much higher temporal resolution than traditional cameras.
As additional inputs, some methods may rely on offline constructed HD maps as input to the model for additional environment context. In contrast, other methods may create the map online as a perception task using real-time sensor inputs. 
It has been shown in the literature that combining multiple sensor inputs usually leads to better and more robust performance, albeit with higher computational demands, increased system costs, and challenges of sensor fusion.

\textbf{Autonomy Stack} 
Given the inputs, various tasks are designed within the autonomy stack to enable autonomous systems to operate in dynamic environments. As shown in Fig~\ref{fig:applicable_e2e_decomp}, focusing on motion prediction, we can categorize these tasks into three main areas: perception, prediction, and action. The typical perception tasks include detection, segmentation, tracking, and mapping, while the prediction tasks usually include 
trajectory prediction, occupancy prediction, and flow prediction. Finally, the action task is planning and control. 
To this day, data-driven learnable approaches are demonstrated to dominate benchmarks for most tasks, while heuristic methods are still applied in certain areas. For instance, simple heuristics are sometimes used for tracking~\cite{luo2018fast}, and optimization techniques are commonly employed in control tasks to account for the kinetics and dynamics of the system in closed-loop.

\textbf{Inter-Module Representations}
In Figure~\ref{fig:representation} and Section~\ref{sec:input-output-representation}, we introduced and reviewed five common types of motion prediction representations. However, in autonomous systems that consider full-stack tasks beyond motion prediction, the representations between perception, prediction, and action are commonly designed as object-based, occupancy-based, or a combination of both~\cite{hu2023planning, vidar}: 

\begin{itemize}[label={\scriptsize$\bullet$}]
    \vspace{-0.5em}
    \item In the object-based representation, the perception module performs object detection and tracking to provide historical trajectories of objects. These trajectories, along with the map data, are then processed by the prediction module to predict future trajectories of agents. Finally, the action module uses these predicted trajectories to plan and execute the ego agent's movements. The trajectory serves as a compact interaction between modules. 
    \vspace{-0.5em}
    \item In the occupancy-based representation, the perception module divides the environment into a grid, estimating the likelihood of occupancy in each cell and identifying the class of objects present. The prediction module then predicts the future evolution of the occupancy grid. At the same time, the planning module uses the occupancy grid to make decisions and plan the ego agent’s actions based on the predicted environment. 
    Some occupancy prediction methods, such as \citet{mahjourian2022occupancy}, ImplictO\cite{agro2023implicit}, and StreamingFlow\cite{shi2024streamingflow}, further predict the flow of the occupancy grid, which provides additional information and facilitates object identity recovery.
    \vspace{-0.5em}
\end{itemize}

In this chapter, we will discuss the pros and cons of these representations for perception in Section~\ref{sec:applicable_perception} and planning in Section~\ref{sec:applicable_planning}. There are also other representations such as queries~\cite{casas2024detra,gu2023vip3d}, affinity maps~\cite{wengMTPMultihypothesisTracking2022,weng2022whose}, raw sensor data~\cite{yang2023learning,yang2024video,wang2024driving}, and mixtures of affine time-varying systems~\cite{ivanovic2020mats}, which we will discuss in Section~\ref{sec:applicable_perception} and Section~\ref{sec:applicable_planning}. These representations can be in 3D, commonly used in robotics for tasks that require detailed spatial understanding and interaction with the environment, such as robot manipulation and drone navigation in cluttered spaces. In contrast, bird's-eye-view (BEV) 2D representations are typically applied to autonomous vehicles and mobile robots, where they simplify the environment to a top-down view, making it easier to model and plan for navigation, obstacle avoidance, and path planning on flat surfaces.

\subsubsection{Autonomy Tasks Decomposition}
\label{sec: applicable autonomy stack}

\textbf{Modular Methods} Since the 2004 DARPA Grand Challenges, it has been a predominant paradigm to decompose the autonomy stack, separately design or train standalone modules, and integrate them into a complete autonomous system. This modular approach enables the training of intermediate tasks, enhances system interpretability, and enables a better understanding of failures, thus facilitating rapid improvement of each module. 
For motion prediction, most benchmarks and methods have fallen in this category to focus on the motion prediction task itself alone, such as Argoverse~\cite{Argoverse} and Waymo~\cite{sun2020scalability} motion prediction dataset that takes trajectory as the prediction representation, and the Waymo Occupancy and Flow Dataset~\cite{sun2020scalability} that promotes the occupancy as the prediction representation.
However, as discussed in \cite{weng2024drive}, this approach faces significant integration challenges during deployment: information bottlenecks between modules are common, with potential loss of information due to thresholding and filtering during inter-module communication;
Additionally, the separate training of modules leads to misaligned objectives, resulting in upstream tasks not being optimally tailored for downstream-aware learning. This recognition that traditional modular pipelines introduce inefficiencies and error propagation has led the autonomous driving field to move toward jointly trained end-to-end architectures.

\begin{figure}[!t]
\centering
\includegraphics[width=\textwidth]{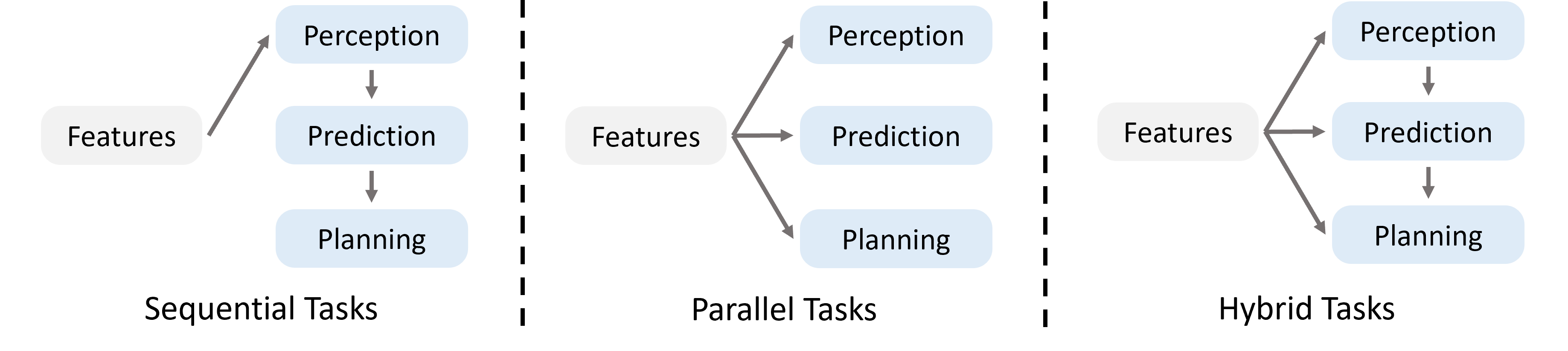}
\caption{End-to-end yet modular architecture design variations differ in the inter-module connections and the information flow between modules.}
\label{fig:applicable_connectivity}
\end{figure}

\textbf{End-to-End Monolithic Methods} To overcome these challenges, there have been many recent advancements toward end-to-end models, the fully differentiable approach with a single monolithic model that directly maps sensor data to planning or control without intermediate tasks, as pioneered by prior works such as \cite{bojarski2016end,pomerleau1988alvinn}. 
These approaches are particularly appealing as they avoid hand-crafted module interfaces, eliminate integration challenges and information bottlenecks, enable gradients to be propagated through the entire model, and align the optimization of each module to the whole system performance.
Moreover, end-to-end learning offers the advantage of rapidly scaling with large amounts of data by eliminating the need to annotate intermediate outputs, such as ground truth object histories.
Recent work in this direction has further bolstered the success of end-to-end approaches in closed-loop driving~\cite{chen2020learning,codevilla2018end,prakash2021multi,wu2022trajectory,kim2024openvla}. Despite these advancements, end-to-end planning approaches still face significant challenges regarding interpretability, verifiability, and safety for real-world deployment. 
For our interest, such monolithic models do not explicitly output motion predictions, which are crucial for understanding surrounding agents' behaviors and ensuring safe decision-making in dynamic environments. Without explicit predictions, it becomes challenging to assess the model's reasoning and ensure it handles edge cases, uncertainties, and unforeseen scenarios that could compromise safety.

\textbf{End-to-End Modular Methods} A second approach to end-to-end methods is the end-to-end modular approach, which seeks to leverage the benefits of both modular and end-to-end approaches.
This hybrid approach keeps multiple modules within a fully differentiable autonomy stack and 
optimizes all modules together. As a result, it maintains safety and interpretability while simultaneously optimizing all modules for downstream planning. 
In practice, a key consideration in the end-to-end modular pipeline is determining the appropriate level of task decomposition and integration between modules. Some studies~\cite{weng2024drive} suggest that certain modules may be redundant for downstream planning tasks, and their removal can reduce computational overhead, resulting in more compact yet practical models. Moreover, the optimal way to integrate modules remains an open research question. As illustrated in Figure~\ref{fig:applicable_connectivity}, existing approaches explore various inter-module connectivity strategies, including \textit{sequential}, \textit{parallel}, and \textit{hybrid} designs.
1) In the sequential approach, detection, tracking, motion prediction, and planning are connected in sequence~\cite{liangPnPNetEndtoEndPerception2020a, hu2023planning, gu2023vip3d}, where the output of one module is the direct input of the next module.
2) Recent studies, including PARA-Drive \cite{weng2024drive}, have demonstrated that it is feasible to implement a fully parallel architecture for perception, prediction, and planning. In this approach, all functions operate from a common bird’s-eye-view (BEV) feature embedding, which is collectively optimized for all tasks.
3) Hybrid approaches~\cite{kimenhanced, jiangVADVectorizedScene2023, hu2023planning} use hybrid inter-module connectivity by passing both the outputs of previous models and a shared BEV feature representation to each module. 
In light of these challenges, the following sections revisit how motion prediction is integrated with perception and planning within the autonomy stack.

\subsubsection{Motion Prediction in End-to-End Modular Methods}
\label{sec: app - Motion Prediction in End-to-End Modular Methods}

Centered around motion prediction, numerous efforts have been dedicated to exploring joint learning of motion prediction with other modules, which are discussed below and also summarized in Table~\ref{tab:applicable_e2e}:

\textbf{Joint Perception and Prediction} 
Instead of taking perception and prediction as independent standalone modules, early end-to-end works unify detection, tracking, and short-horizon forecasting into a fully differentiable \textit{CNN architecture} without compromising real-time performance. For example, 
FaF~\cite{luo2018fast} showed that a voxelized LiDAR backbone can jointly optimize 3D boxes detection and three-second motion forecasts with heuristic-based tracking, supporting 60Hz inference.
Expanding on this, IntentNet~\cite{casasIntentNetLearningPredict2018} added an auxiliary “intent” branch to predict discrete maneuvers (e.g., turn, go straight, yield). PnPNet~\cite{liangPnPNetEndtoEndPerception2020a} further advanced end-to-end modeling by making the tracking module differentiable and integrating it into the model for joint learning. This joint paradigm soon extended to camera-input pipelines, among which FIERY~\cite{hu2021fiery} forecasted future BEV instance segmentation with calibrated uncertainty.

\textit{Attention-based models} have since emerged as the dominant approach for joint perception–prediction.
PF-Track~\cite{pangStandingFutureSpatioTemporal2023a} and ViP3D~\cite{gu2023vip3d} use spatio-temporal transformers and adopt 3D agent queries to pass information among perception and prediction to preserve end-to-end differentiability. Recent approaches mitigate the information bottlenecks among modules: DeTra~\cite{casas2024detra} and ForeSight~\cite{papais2025foresight} remove non-differentiable tracking by formulating detection and forecasting as a joint trajectory refinement problem to avoid information loss from intermediate tracking; ViRR~\cite{kimenhanced} improves prediction by building a visual scene graph to model agent interactions and integrating it into the prediction module. Some works, such as P2D~\cite{kim2023predict}, demonstrate that prediction can benefit perception, especially for long-term occlusion and tensor degradation.
Building on attention mechanisms, recent methods further represent the scene as a continuous \textit{4D occupancy space} rather than as discrete objects. ImplicitO~\cite{agro2023implicit} predicts occupancy and flow over time with a single neural network, allowing unified occupancy estimation and forecasting. Self-supervised methods like 4D-Occ~\cite{ma2024cam4docc} and UnO~\cite{agro2024uno} learn future occupancy with self-supervision from LiDAR data, avoiding occupancy annotation.

\begin{table}[!t]
\centering
\caption{Taxonomy of methods that perform joint learning of motion prediction with other tasks. The inputs to the system include lidar point clouds (Ldr), cameras (Cam), and HD maps (Map). The tasks are detection (Det), BEV segmentation or current occupancy prediction (OccP), tracking (Trk), mapping (Map), trajectory prediction (Traj), future occupancy forecasting (OccF), flow prediction (Flow), planning (Pln), and Control (Ctl), where $\cmark$ indicates a learnable task, $\protect\xmark$ indicates a task that is not learned, and $-$ specifically denotes planning tasks that are not learned but instead solved through optimization or sampling-based methods.}
\label{tab:applicable_e2e}
\resizebox{0.85\linewidth}{!}{
\begin{tabular}{@{}l|l|ccc|cccc|ccc|cc@{}}
\toprule\toprule
           &                                                           & \multicolumn{3}{c|}{\multirow{2}{*}{Inputs}} & \multicolumn{9}{c}{Tasks}                                                                                                                                       \\ \cmidrule(l){6-14} 
Conference & Method                                                    & \multicolumn{3}{c|}{}                        & \multicolumn{4}{c|}{Perception}                                        & \multicolumn{3}{c|}{Prediction}                           & \multicolumn{2}{c}{Action} \\ \cmidrule(l){3-14} 
           &                                                           & Ldr           & Cam           & Map          & Det        & OccP        & Trk        & \multicolumn{1}{c|}{Map}        & Traj       & OccF        & \multicolumn{1}{c|}{Flow}       & Pln          & Ctl         \\ \midrule
\multicolumn{14}{c}{Joint Perception and Prediction} \\ \midrule
CVPR'18 & FaF \cite{luo2018fast}                         & \cmark &        &        & \cmark &        & \xmark &        & \cmark &        &        &        &        \\
CORL'18 & IntentNet \cite{casasIntentNetLearningPredict2018}        & \cmark &        & \cmark & \cmark &        & \xmark &        & \cmark &        &        &        &        \\
CVPR'20 & PnPNet \cite{liangPnPNetEndtoEndPerception2020a}          & \cmark &        & \cmark & \cmark &        & \cmark &        & \cmark &        &        &        &        \\
CVPR'23 & PF-Track \cite{pangStandingFutureSpatioTemporal2023a}     &        & \cmark &        & \cmark &        & \cmark &        & \cmark &        &        &        &        \\
CVPR'23 & ViP3D \cite{gu2023vip3d}                    &        & \cmark & \cmark & \cmark &        & \cmark &        & \cmark &        &        &        &        \\
ECCV'24 & ViRR \cite{kimenhanced}                                   &        & \cmark &        & \cmark &        & \cmark & \cmark & \cmark &        &        &        &        \\
ECCV'24 & DeTra \cite{casas2024detra}                               & \cmark &        & \cmark & \cmark &        &        &        & \cmark &        &        &        &        \\
CVPR'23 & ImplictO \cite{agro2023implicit}                          & \cmark &        & \cmark &        & \cmark &        &        &        & \cmark & \cmark &        &        \\
CVPR'23 & 4D-Occ \cite{khurana2023point}            & \cmark &        &        &        & \cmark &        &        &        & \cmark &        &        &        \\
CVPR'24 & UnO \cite{agro2024uno}                                    & \cmark &        &        &        & \cmark &        &        &        & \cmark &        &        &        \\
ICCV'25 & ForeSight \cite{papais2025foresight}                                    &  & \cmark       &    \cmark    &   \cmark     &     &        &        &    \cmark    &  &        &        &        \\
\midrule
\multicolumn{14}{c}{Joint Prediction and Planning} \\ \midrule

ECCV'20 & Trajectron++ & & & & & & & & \cmark & & & \xmark & \\
CoRL'20 & MATS \cite{ivanovic2020mats} & & & & & & & & \cmark & & & - & \\
RAL'21 & UAPP \cite{wang2021socially} & & & & & & & & \cmark & & & - & \xmark \\
ICRA'23 & TPP \cite{chen2023tree} & & & & & & & & \cmark & & & - & \xmark \\
ICRA'24 & DTPP \cite{huang2023dtpp} & & & & & & & & \cmark & & & - & \xmark \\

RAL'24 & IJP \cite{chen2023interactive} & & & & & & & & \cmark & & & - & \xmark \\

ICCV'23 & GameFormer \cite{huang2023gameformer}& & & & & & & & \cmark & & & \cmark & \xmark\\
NeurIPS'24 & BeTop \cite{liu2024reasoning} & & & & & & & & \cmark & & & \cmark & \xmark\\

\midrule
\multicolumn{14}{c}{Joint Perception, Prediction, and Planning} \\ \midrule
arXiv & DAVE-2 \cite{bojarski2016end} & & \cmark & & & & & & & & & \cmark  \\
CoRL'19 & LBC \cite{chen2020learning} & & \cmark & & & & & & & & & \cmark & \xmark  \\

ECCV'20 & P3 \cite{sadat2020perceive}                               & \cmark &        & \cmark &        & \cmark &        &        &        & \cmark &        & \cmark &        \\
CVPR'21 & MP3 \cite{casasMP3UnifiedModel2021}                       & \cmark &        &        &        & \cmark &        & \cmark &        & \cmark & \cmark & \cmark &        \\
ICCV'21 & FIERY \cite{hu2021fiery}                                  &        & \cmark &        &        & \cmark &        &        &        & \cmark &        &        &        \\

CORL'22 & InterFuser \cite{shaoSafetyEnhancedAutonomousDriving2023} & \cmark & \cmark &        &    \cmark    &  &        &        &      \xmark  &  &        & \cmark & \xmark \\
TPAMI'22 & Transfuser \cite{chitta2022transfuser} & \cmark & \cmark &        &    \cmark    &  &        &        &      \xmark  &  &        & \cmark & \xmark \\

ECCV'22 & ST-P3 \cite{huSTP3EndtoEndVisionBased2022}                &        & \cmark &        &        & \cmark &        & \cmark &        & \cmark &        & \cmark & \xmark \\
CVPR'23 & ReasonNet \cite{shao2023reasonnet}                        & \cmark & \cmark &        & \cmark &  &        &        &    \xmark    &  &        & \cmark & \xmark \\

CVPR'23 & UniAD \cite{hu2023planning}                               &        & \cmark &        & \cmark & \cmark & \cmark & \cmark & \cmark & \cmark &        & \cmark &        \\
ICCV'23 & VAD \cite{jiangVADVectorizedScene2023}                    &        & \cmark &        & \cmark &        &        & \cmark & \cmark &        &        & \cmark & \xmark \\
CVPR'24 & PARA-Drive \cite{weng2024drive}                           &        & \cmark &        & \cmark & \cmark & \cmark & \cmark & \cmark & \cmark &        & \cmark &        \\
CVPR'24 & ViDAR \cite{vidar}                                        &        & \cmark &        & \cmark & \cmark & \cmark & \cmark & \cmark & \cmark &        & \cmark &  



      \\ \bottomrule
\bottomrule
\end{tabular}
}
\end{table}

\textbf{Joint Prediction and Planning} 
Compared to the standard sequential paradigm, where prediction outputs are passed to the planner, joint prediction and planning approaches aim to integrate these two traditionally separate components. A prominent strategy in this direction is \textit{planning-informed prediction}, where the planned ego trajectory is incorporated into the prediction process to account for how the ego agent’s actions may influence surrounding agents.
A common form of planning-informed prediction is \textit{ego-conditioned prediction}, in which the ego trajectory is used as additional input to predict the future behaviors of other agents. This enables the model to explicitly capture interaction dynamics.
This conditioning can be implemented in various ways: through traditional planning frameworks (e.g., UAPP~\cite{wang2021socially}), where other agents' future behavior is optimized or sampled based on ego plans; deep learning models (e.g., Trajectron++~\cite{salzmann2020trajectron++}), where the ego trajectory serves as an additional input to the network; or occupancy-based representations (e.g., ImplicitO~\cite{agro2023implicit}), which predict the occupancy of grid cells traversed by the planned ego trajectory, offering higher efficiency compared to full-grid prediction.
\textit{Policy planning} methods further leverage ego-conditioned prediction to evaluate and select optimal ego plans. These approaches include:
sampling-based methods, such as UAPP~\cite{wang2021socially}, TPP~\cite{chen2023tree}, DTPP~\cite{huang2023dtpp}, and \citet{sun2018courteous}, where multiple ego trajectories are sampled, and ego-conditioned prediction is performed for each, followed by cost evaluation and ranking to select the best ego plan;
optimization-based methods, such as IJP~\cite{chen2023interactive} and MATS~\cite{ivanovic2020mats}, exploit differentiable models to compute gradients with respect to the ego trajectory, enabling direct optimization.
Finally, prediction and planning modules can also be jointly trained as in GameFormer~\cite{huang2023gameformer} and BeTop~\cite{liu2024reasoning}, allowing end-to-end optimization of the full decision-making pipeline.
    
\textbf{Joint Perception, Prediction, and Planning}
Recent work further advances the joint learning of perception, prediction, and planning within a unified framework, either open-loop or closed-loop. 
This can be instantiated in a fully end-to-end approach, which directly maps sensor data to final control commands without intermediate tasks, as in early works DAVE-2~\cite{bojarski2016end} and LBC~\cite{chen2020learning}. 
This can also be formulated as a unified differentiable architecture with explicit intermediate tasks. For example, P3~\cite{sadat2020perceive} forecasts future semantic occupancy given LiDAR and offline map inputs, based on which sampling-based planning can be performed. 
MP3~\cite{casasMP3UnifiedModel2021} extends this by performing online mapping, which is jointly optimized with other perception tasks.
InterFuser~\cite{shaoSafetyEnhancedAutonomousDriving2023} fuses camera, radar, and LiDAR data using a transformer into an interpretable intermediate representation, where detection, traffic light recognition, and waypoint prediction can be performed. 
UniAD~\cite{hu2023planning} further exemplifies this unified approach, where planning serves as the supervisory signal to guide perception and prediction. 

Another line of end-to-end work focuses on temporal reasoning and rich contextual understanding.
ST-P3~\cite{huSTP3EndtoEndVisionBased2022} aggregates multi-frame dense BEV features to perform joint perception, prediction and planning. 
ReasonNet~\cite{shao2023reasonnet} uses a sparse memory bank, combining a short-term buffer for high-resolution recent features and a long-term buffer that retains important features selectively.
Efforts to improve computation efficiency and scalability include VAD~\cite{jiangVADVectorizedScene2023}, which vectorizes scene elements to reduce computation; PARA-Drive~\cite{weng2024drive}, which proposes a parallelized architecture for real-time processing; and ViDAR~\cite{vidar}, which investigates scaling with sensor data via self-supervised pretraining of point cloud forecasting.
Together, these advances reflect a shift toward integrated, interpretable, and planning-centric frameworks that avoid traditional module boundaries, paving the way for scalable and robust autonomous systems.

\textbf{Challenges in End-to-End and Joint Learning Models}  
End-to-end joint learning of multiple tasks offers significant advantages for motion prediction, such as 1) facilitating more efficient information flow throughout the pipeline, allowing the prediction model direct access to both upstream raw sensor data and perception features, and downstream planning and control; 2) aligning with system-level objectives through multi-task information sharing and the use of shared network parameters; and 3) mitigating the need for annotating intermediate outputs, thus enabling better scalability with large datasets. However, challenges that go beyond what is mentioned at the beginning of this chapter remain: 1) as the number of tasks increases, multi-task learning becomes more complex, requiring models to optimize potentially conflicting objectives; 2) this coupling hinders interoperability, making it difficult to update or replace individual components without retraining the entire model. Addressing these trade-offs requires careful integration and robust evaluation to realize the full potential of end-to-end motion prediction systems. In the coming sections, we will elaborate on challenges in integrating prediction with perception, integrating prediction with planning, and evaluation. 

\subsection{Perception-Prediction Integration}
\label{sec:applicable_perception}

\begin{figure}[!htb]
\centering
\includegraphics[width=0.88\textwidth]{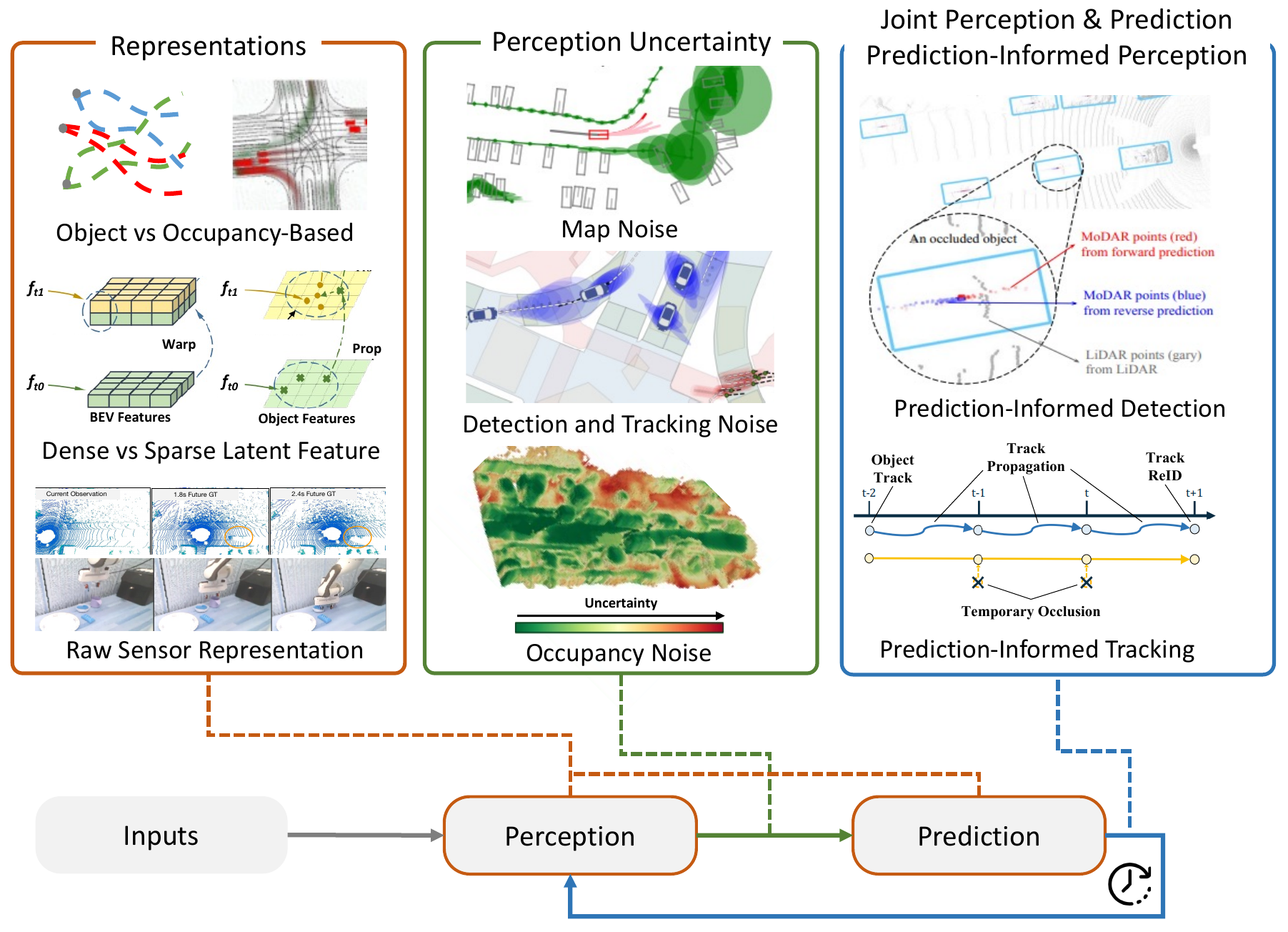}
\caption{Integration between perception and prediction considers the alignment of representations \cite{wang2023exploring}, leveraging perception uncertainty for prediction \cite{gu2024producing,ivanovic2022propagating,cao2024pasco} and temporal feedback of prediction for perception \cite{liMoDARUsingMotion2023, pangStandingFutureSpatioTemporal2023a}.}
\label{fig:applicable_perceptionintegration}
\end{figure}

In typical autonomy stacks, perception provides essential inputs for motion prediction to infer the future evolution of objects or a scene. The seamless integration between perception and prediction is crucial for efficient information propagation and overall system safety and performance. This integration requires 1) proper design of representations and the interface between the two modules that enable smooth information propagation and efficient computation; 2) propagating, calibrating, and eventually reducing the uncertainties and errors from upstream perception to enable robust prediction; 3) joint learning and temporal information pass of the two modules to facilitating information sharing and align the optimization to the performance of the overall system. These concepts are illustrated in Figure~\ref{fig:applicable_perceptionintegration} and will be explored in greater detail in the following subsections.

\subsubsection{Representations}
\label{sec: applicable perception-prediction representation}

Multiple representations have been proposed between perception and prediction. As mentioned in Section~\ref{sec:applicable_end-to-end}, two widely used approaches are object-based representation~\cite{pangStandingFutureSpatioTemporal2023a,gu2023vip3d} and occupancy-based representation~\cite{hu2021fiery, ma2024cam4docc}. These representations can complement each other, differing in object priors, granularity, computation expense, interpretability, and information completeness through the pipeline. While some applications integrate both methods~\cite{hu2023planning}, their fusion remains to be further explored. Some other methods also include latent representation~\cite{gu2023vip3d,casas2024detra}, tracking-free representation~\cite{weng2022whose,wengMTPMultihypothesisTracking2022}, and raw sensor representations~\cite{khurana2023point,hu2023gaia},. Each representation possesses its strengths and limitations, as discussed below.

\textbf{Object-Based Representation} Taking the object states (e.g. bounding boxes, poses) as the representation, the object-based representation is usually the most common approach and possesses the advantages of 1) \textit{clear object understanding}: it provides explicit identification of individual objects and a detailed knowledge of their motion, making it easier to leverage object-specific motion patterns (e.g., acceleration, turning), model interactions between objects (e.g., collisions, coordination). This also enables better interpretability, which makes debugging, performance analysis, and system updates more intuitive. 
2) \textit{efficient scene representation}: objects serve as compact representations of the surrounding environment, making the system more efficient.
However, it also suffers from 1) \textit{error accumulation}: errors in object detection or tracking can accumulate through the system, leading to inaccuracies in predictions and potentially hazardous planning decisions; 2) \textit{information bottleneck}: sensor data and perception features/uncertainties are not fully accessible to motion prediction due to the non-differentiable operations, such as thresholding and non-maximum suppression in detection, and hard associations in tracking; 3) \textit{limited for unknown or irregular objects}: typical object-centric representations rely on predefined classes and geometric assumptions, making it difficult to accurately perceive and predict open-set or irregular objects that do not fit these patterns. 

\textbf{Occupancy-Based Representation} Focus on the occupancy of each cell in a 2D or 3D space, the occupancy-based representation show benefits of 1) \textit{representation of fine details and uncommon objects}: occupancy grids effectively model detailed spatial distributions, allowing them to accommodate uncommon classes of objects with irregular shape and boundaries; 2) \textit{compatibility with grid-space features}: occupancy grids integrate seamlessly with grid-based features like BEV (Bird’s Eye View) feature maps and rasterized maps, enabling smooth fusion of data and feature from diverse sources such as LiDAR and cameras. Meanwhile, it also faces weaknesses of 1) \textit{loss of object identity}: The occupancy representation does not provide explicit information on individual objects (e.g., type, size, velocity) and require unique designs to recover object identities from predicted occupancy and flow, limiting its ability to model object interactions, leverage motion patterns, and offer interpretability; 2) \textit{high and inefficient computation}: Occupancy-based representation requires discretizing the environment into a large number of grid cells (2D or 3D) and estimating the likelihood of occupancy in each cell. Predicting the future evolution of the scene extends this grid into an additional temporal dimension, resulting in significant memory and computational demands. Occupancy estimation is especially challenging in large-scale environments or when using high-resolution grids. Moreover, as many scenes are sparsely populated with objects, many of the grid cells represent space, making the approach computationally inefficient.
3) \textit{information bottleneck}: similar to object-based representations, non-differentiable operations (e.g., thresholding) prevent motion prediction models from accessing upstream perception features, making them agnostic to and sensitive to perception uncertainties and errors.

\textbf{Latent Representation} 
Explicit representations, such as bounding boxes and occupancy grids mentioned above, carry clear physical meanings and thus offer high interpretability and easy integration between perception and prediction. Their interpretability makes them well-suited for both rule-based and learning-based methods.
They also promote modularity, allowing perception and prediction modules to be developed independently, which simplifies and accelerates the development process. 
Additionally, bounding boxes and low-resolution occupancy grids provide computational efficiency by compressing the scene into a compact representation, reducing processing overhead, and supporting faster real-time execution.
However, the main drawback lies in the potential information bottleneck~\cite{casas2024detra}. By compressing the scene into a compact representation, subtle but valuable details may be omitted, limiting the richness of contextual information. In addition, due to non-differentiable operations such as thresholding, filtering, and hard associations, sensor data and perception features are not fully accessible to the motion prediction module.
This loss of information can propagate through the pipeline, potentially degrading the performance of downstream tasks that depend on fine-grained context, such as trajectory prediction and planning under uncertainty.

To this end, latent representations have been explored as an alternative interface between modules. These representations use the embeddings or intermediate features produced by neural networks as the interface between perception and prediction. In the context of transformer-based architectures, they are also often referred to as token- or query-based representations.
By avoiding non-differentiable operations, such latent representations enable the continuous propagation of gradients and uncertainty throughout the pipeline, supporting joint learning across modules that can be aligned with the overall system objective. Furthermore, by allowing downstream modules to access fine-grained contextual information from upstream, latent representations facilitate more informed decision-making and nuanced behavior modeling.
However, latent representations come with several drawbacks. They typically offer lower interpretability compared to explicit representations and often require more data due to their reduced structural priors. Additionally, they tend to be more computationally intensive, lacking the compactness and efficiency of explicit formats like bounding boxes or occupancy grids.
Moreover, integration complexity and limited interoperability can pose significant challenges. Latent features often require careful alignment across modules, especially when combining multiple sensor modalities or deep learning architectures. If a component needs to be replaced or updated, the entire system may require retraining to ensure compatibility, as the latent features are not standardized or modular by default.

\textbf{Tracking-Free Representations} 
Another noteworthy approach is the tracking-free paradigm. Conventional autonomy systems typically rely on object detection to produce per-frame object states, followed by a tracking module that associates detections across frames to maintain object identities, providing historical context for motion prediction.
In contrast, tracking-free methods are motivated by the observation that this conventional pipeline is inherently error-prone. The tracking step can compound detection inaccuracies and introduce additional errors and uncertainties, particularly through incorrect associations such as identity switches.
When such tracking errors occur, motion prediction models may be conditioned on noisy or incorrect historical states, ultimately degrading the quality of future trajectory predictions.
To this end, tracking-free approaches aim to remove the tracking module and circumvent the information and uncertainty bottlenecks introduced by hard associations. By avoiding explicit tracking, they preserve a richer, temporally consistent feature space without enforcing rigid identity assignments across frames. Performing trajectory prediction directly from object detections simplifies the perception pipeline and mitigates the impact of noisy tracking inputs.
While this approach may sacrifice certain historical cues, it eliminates tracking-induced errors and improves overall system robustness, especially in complex scenarios involving occlusion, dense traffic, or ambiguous agent interactions.

In practice, tracking-free approaches usually adopt a latent representation. Two approaches for tracking-free prediction include soft associations and detection-based methods. 1) Taking a soft association approach, AffiniPred~\cite{weng2022whose} leverages per-frame detections and aggregates features over time using affinity matrices instead of rigid tracking assignments. These matrices, rather than hard-associated tracks, are then fed into motion prediction models, enhancing robustness to occlusions and identity misassociations. 2) The detection-based approaches go a step further by aiming to eliminate tracking, even soft associations, relying solely on per-frame detections for motion prediction. DeTra~\cite{casas2024detra} introduced an attention-based architecture that encodes motion cues from sensor data and detections without predefined tracks, jointly learning detection and prediction for improved performance while removing the tracking bottleneck. Similarly, VAD~\cite{jiangVADVectorizedScene2023} and PPAD~\cite{chenppad} predict trajectories directly from detection inputs using a vectorized representation, bypassing tracking altogether. These models can store and leverage short-term historical context by aggregating past feature maps, such as with BEVFormer~\cite{li2024bevformer}, without requiring hard associations, allowing them to recover from past errors as new information becomes available. However, a key challenge for tracking-free methods remains: maintaining long-term temporal consistency without explicit identity linking. Learning representations that implicitly capture object permanence and motion continuity over time is essential to address this. Ultimately, tracking-free motion prediction presents a promising direction for building more scalable, resilient, and accurate autonomous driving systems.

\textbf{Sensor-Level Representations.} With the rise of the world model concept, another emerging approach involves representing the scene directly using raw sensor inputs such as RGB images and LiDAR point clouds, and predicting the scene’s future evolution through tasks like point cloud forecasting~\cite{khurana2023point} and video generation~\cite{hu2023gaia}. This approach enables models to directly consume sensor data, including internet-scale video sources, making it highly scalable in terms of data accessibility. However, it incurs significantly higher computational costs due to the absence of scene compression into lower-dimensional representations. Additionally, it often lacks explicit object-level awareness, which can limit interpretability and downstream integration. This approach is closely related to self-supervised learning, which will be further discussed in Section~\ref{sec:gen_pred_ssl}.

\subsubsection{Dealing with Perception Uncertainty}
\label{sec:deployable_perception_dealing_with_uncertainty}

\begin{table}[!htb]
\centering
\caption{Summary of representative approaches for modeling uncertainty in perception and prediction. The inputs include lidar point clouds (Ldr), cameras (Cam), HD maps (Map), and ground truth labeled objects (GT). The tasks are detection (Det), BEV segmentation or present occupancy (OccP), tracking (Trk), mapping (Map), trajectory prediction (Traj), future occupancy prediction (OccF), flow prediction (Flow), planning (Pln), and Control (Ctl), where $\cmark$ indicates a probabilistic output that includes uncertainty.}
\label{tab:applicable_uncertainty}
\resizebox{0.85\linewidth}{!}{
\begin{tabular}{@{}l|l|llll|llll|ll@{}}
\toprule \toprule
\multicolumn{1}{l|}{\multirow{2}{*}{Conference}} & \multicolumn{1}{l|}{\multirow{2}{*}{Method}} & \multicolumn{4}{c|}{Inputs} & \multicolumn{4}{c|}{Perception}    & \multicolumn{2}{c}{Prediction}             \\ \cmidrule(l){3-12} 
\multicolumn{1}{c|}{}                            & \multicolumn{1}{c|}{}                        & Ldr   & Cam   & Map  & GT  & Det & OccP & Trk & Map & Traj & OccF \\ \midrule
\multicolumn{12}{c}{Detection Uncertainty} \\ \midrule
ITSC'18    & TSAD~\cite{feng2018towards} &\xmark&     &     &    &\cmark&     &     &     &      &     \\ 
CVPR'19    & LaserNet~\cite{meyer2019lasernet}&\xmark&     &     &    & \cmark   &     &     &     &      &     \\
ICRA'20    & BayesOD~\cite{harakeh2020bayesod}&\xmark&     &     &    & \cmark   &     &     &     &      &     \\
IROS'22    & HAICU~\cite{ivanovic2022heterogeneous}&     &     &     &\xmark&\cmark&     &     &     &\cmark&     \\ \midrule
\multicolumn{12}{c}{Tracking Uncertainty} \\ \midrule
ICRA'24    & SWTrack~\cite{papais2024swtrack}&\xmark&\xmark&     &    &\xmark&     &\cmark&     &      &     \\
ICRA'24    & UncertaintyTrack~\cite{lee2024uncertaintytrack}&     &\xmark&     &    &\cmark&     &\cmark&     &      &     \\ CVPR'22    & AffiniPred~\cite{weng2022whose}&\xmark&     &\xmark&    &\xmark&     &\cmark&     &\xmark&     \\
ICRA'22    & PSU-TF~\cite{ivanovic2022propagating}&\xmark&     &     &    &\xmark&     &\cmark&     &\cmark&      \\ 
IV'23      & MTP~\cite{wengMTPMultihypothesisTracking2022}&\xmark&     &\xmark&    &\xmark&     &\cmark&     &\xmark&      \\ \midrule
\multicolumn{12}{c}{Mapping Uncertainty} \\ \midrule
CVPR'24    & MUP~\cite{gu2024producing}&     &\xmark&     &    &     &     &     &\cmark&\cmark&      \\ \midrule
\multicolumn{12}{c}{Occupancy Uncertainty} \\ \midrule
CVPR'24    & PaSCo~\cite{cao2024pasco}&\xmark&     &     &    &     &     \cmark &    &     &      &      \\
ICRA'25    & OCCUQ~\cite{heidrich2025occuq}&     &\xmark&     &    &     &  \cmark   &    &     &      &      \\
ICCV'21    & FIERY~\cite{hu2021fiery}&     &\xmark&     &    &  &\cmark  &      &     &      &\cmark\\
 \bottomrule \bottomrule
\end{tabular}
}
\end{table}

\textbf{Perception Uncertainty and Errors - What? Why? How?}
Integrating perception and prediction introduces significant challenges due to the inherent uncertainty in perception outputs, which arise from both sensor noise and limitations in perception algorithms. As discussed in the previous section, different inter-task representations vary in how effectively they propagate information through the stack. While \textit{latent-based representations} preserve information flow and can implicitly propagate uncertainty by avoiding non-differentiable operations (e.g., thresholding, non-maximum suppression, hard associations), more commonly used explicit inter-task representations without latent embeddings often suffer from information loss. These representations typically rely on deterministic perception outputs such as object bounding boxes, trajectory histories, and occupancy maps. Such deterministic outputs may fail to capture the ambiguity and uncertainty inherent in real-world environments. For example, detection errors (e.g., mislocalization, misclassification), tracking failures (e.g., identity switches), and mapping inaccuracies (e.g., false or imprecise map elements) can propagate downstream and degrade prediction quality. These issues are especially pronounced in challenging scenarios, such as adverse weather, poor lighting, or temporary sensor shutdowns, where perception performance deteriorates and uncertainty becomes crucial for enabling cautious, risk-aware decision-making.

Therefore, \textit{propagating, calibrating, and ultimately reducing} these uncertainties is critical for improving the robustness and reliability of prediction models and the entire autonomy stack. Existing efforts to address uncertainty in detection, tracking, occupancy, and mapping generally fall into three categories: 1) adopting the latent represetnation and directly propagating perception features to prediction~\cite{pangStandingFutureSpatioTemporal2023a, hu2023planning}, 2) bypassing errors in traditional sequential pipelines by removing modules such as tracking or introducing parallel pathways~\cite{casas2024detra, weng2024drive}, 2) modeling perception uncertainty using probabilistic representations and providing them as prediction inputs~\cite{ivanovic2022heterogeneous, ivanovic2022propagating, gu2024producing, hu2021fiery}, and. 
While promising, these approaches face key trade-offs: in particular, balancing the fidelity of uncertainty modeling with the computational efficiency required for real-time inference remains an open and pressing challenge.
In the following, we focus specifically on the third category—probabilistic uncertainty modeling, as the other approaches were discussed in the previous Section~\ref{sec: applicable perception-prediction representation}.

\textbf{Detection Uncertainty} 
Managing uncertainty, namely a model’s confidence in its outputs, is essential for robust and reliable autonomous systems. In a typical autonomy stack, the object detection module provides presence likelihoods, object classes, and bounding boxes to the motion prediction module, and uncertainty can arise from each of these outputs. Several works have focused on quantifying and calibrating detection uncertainties.
TSAD~\cite{feng2018towards} models both aleatoric and epistemic uncertainty using Bayesian deep learning, BayesOD~\cite{harakeh2020bayesod} reformulates detector inference and non‑maximum suppression from a Bayesian perspective to produce well‑calibrated per‑instance covariance estimates, and LaserNet~\cite{meyer2019lasernet} predicts full probability distributions over 3‑D bounding boxes from LiDAR range images. These approaches underscore the importance of uncertainty modeling in improving detection reliability and overall system safety. Building on this, several studies explore how to appropriately propagate these uncertainties into motion prediction, as discussed below.

\begin{itemize}[label={\scriptsize$\bullet$}]
    \vspace{-0.5em}
\item 
For the \textit{object state} uncertainties, standard motion prediction models usually take the deterministic output from upstream detections, which can be vulnerable to errors and noise in the detection. To this end, PSU-TF \cite{ivanovic2022propagating} propagates detection uncertainties by incorporating state uncertainty distributions as additional object state features to motion prediction. These uncertainties can be calibrated by introducing a new statistical distance-based loss function, encouraging prediction uncertainties to better match upstream perception uncertainties.

\vspace{-0.5em}
\item 
For the \textit{object class}, motion prediction typically takes the most likely class for an object, which can introduce errors in cases of ambiguous classification, such as when distinguishing between similarly looking bicycles and motorcycles. To this end, the work HAICU~\cite{ivanovic2022heterogeneous} showed that directly incorporating class probabilities from upstream perception systems as additional input features into a state-of-the-art trajectory prediction method effectively improves the prediction performance in the presence of uncertainty (i.e., object classification error). The authors further illustrate how the perception class and its uncertainty affect prediction results and confidence. Additionally, they introduce PUP, a challenging new real-world autonomous driving dataset that includes unfiltered agent class probabilities to encourage further investigation into the effect of class uncertainty on prediction.

\vspace{-0.5em}
\item 
A key but underexplored source of uncertainty in motion prediction is the \textit{object presence likelihood}—the model’s confidence that a detected object truly exists. 
This presence likelihood can be represented in different ways depending on the detector architecture.
In anchor-free detectors~\cite{yin2021center}, the model explicitly predicts a separate likelihood score to infer object existence. In contrast, anchor-based detectors~\cite{zhou2018voxelnet} do not directly output existence probabilities; instead, they incorporate a background class into the object classification output to account for non-object regions.
Presence likelihood is particularly useful in occluded or cluttered scenes, where detection confidence varies; incorporating it into prediction could help models better handle false positives and missed detections by adjusting forecast uncertainty accordingly. 
However, most prediction models rely on binary detections and overlook presence uncertainty, which reduces robustness in cases of missed detections or over-detections. Effectively representing and propagating this uncertainty remains an open research challenge. A promising direction is to pass rich perception feature maps to the prediction module, rather than depending solely on deterministic object presence.
\end{itemize}

\textbf{Tracking Uncertainty} Object tracking uncertainty and errors arise from ambiguity in object associations across frames and errors in estimated object states, such as position and velocity. These errors can significantly degrade trajectory prediction performance. To address this, two main approaches have been explored in the literature.

\begin{itemize}[label={\scriptsize$\bullet$}]
    \vspace{-0.5em}
\item 
\textit{Probabilistic tracking representation}: 
Accurately modeling tracking uncertainty is critical for maintaining consistent object identities and supporting reliable prediction and planning in dynamic environments. Two key strategies to study tracking uncertainty include addressing association ambiguity through multiple hypothesis tracking (SWTrack~\cite{papais2024swtrack}), and propagating detection and localization uncertainty to enable robust tracking (UncertaintyTrack~\cite{lee2024uncertaintytrack}). Building on these ideas, recent methods propagate tracking uncertainty to prediction by adopting probabilistic representations that explicitly capture tracking uncertainty. AffiniPred~\cite{weng2022whose} uses a soft affinity matrix to represent the pairwise probabilities of objects belonging to the same track, allowing the model to marginalize over associations during prediction. In contrast, MTP~\cite{wengMTPMultihypothesisTracking2022} enumerates multiple track hypotheses per object and aggregates over them, enabling explicit reasoning about identity switches and association errors. In addition, similar to object detection state uncertainties, track localization uncertainty can be propagated from detection or estimated with a Kalman filter and used to improve motion prediction, such as in PSU-TF~\cite{ivanovic2022propagating}.

\vspace{-0.5em}
\item 
\textit{Tracking-free representations}: As an alternative to modeling and propagating tracking uncertainty, another approach to mitigating the impact of noisy tracking inputs is to perform trajectory prediction directly from object detections. While removing multi-object tracking from the perception pipeline sacrifices valuable information, it eliminates errors that can arise within the tracking subsystem, as previously discussed in \ref{sec: applicable perception-prediction representation}.

\end{itemize}

\textbf{Mapping Uncertainties} Maps play a crucial role in motion prediction by providing essential spatial context, such as roads, lanes, and obstacles. This information allows models to anticipate object movements and make safer, more reliable decisions. Traditionally, static, pre-built offline maps were utilized. However, with the emergence of online mapping systems capable of covering unmapped or rapidly changing areas, the challenge of ensuring the accuracy of these online maps and managing their uncertainties has become increasingly significant. By estimating uncertainty in online map generation, autonomous systems can assess the reliability of the map data they use, thereby enhancing decision-making and safety. Unfortunately, most existing online map estimation methods do not provide associated uncertainty or confidence information. 
To address this issue, the pioneering work~\cite{gu2024producing} presents a vectorized map uncertainty formulation and extends multiple state-of-the-art online map estimation methods to include uncertainty estimates without compromising mapping performance. This research analyzes uncertainty sources like occlusion, sensor range, lighting/weather, and motion. By incorporating the estimated map distribution into advanced motion prediction approaches, the study demonstrates that incorporating map uncertainty improves prediction accuracy by up to 15\% and accelerates training convergence by 50\%.

\textbf{Occupancy Uncertainty} 
While the above discussion focuses primarily on object-based representations, uncertainties in occupancy-based representations have also been actively studied, as highlighted by two recent works,
OCCUQ~\cite{heidrich2025occuq} and PaSCo~\cite{cao2024pasco}. OCCUQ~\cite{heidrich2025occuq} introduces a method that dynamically calibrates model confidence using epistemic uncertainty estimates, effectively identifying unseen data and enhancing robustness against sensor corruptions like fog or missing cameras. Similarly, PaSCo~\cite{cao2024pasco} extends semantic scene completion by integrating instance-level information and employs an efficient ensemble strategy to estimate both voxel-wise and instance-wise uncertainties, thereby improving performance and reliability in 3D scene understanding.
However, while these advancements have improved occupancy predictions at the current time, the role of uncertainty in future occupancy forecasting remains underexplored. Incorporating uncertainty estimates into occupancy forecasting models could enhance their reliability, particularly in dynamic environments where factors like moving objects and occlusions introduce variability. By quantifying the confidence of predictions over time, autonomous systems can make more informed decisions, anticipate potential hazards, and adapt to unforeseen changes. This presents a promising avenue for future research to develop forecasting models that not only predict future occupancy states but also provide calibrated uncertainty measures, ultimately contributing to safer and more robust autonomous systems.

\subsubsection{Joint Learning and Prediction-Informed Perception}
\label{sec: app - prediction-informed perception}

\textbf{Joint Perception and Prediction} While perception and prediction modules are typically developed independently, following the conventional task decomposition in autonomy and supported by well-established task-independent benchmarks, joint learning approaches have gained increasing attention for their potential to optimize both tasks within a unified framework. By modeling the module interdependencies, joint learning mitigates information bottlenecks, facilitates richer feature sharing, and reduces uncertainty and error accumulation between modules. As discussed previously in Section~\ref{sec: applicable perception-prediction representation}, the effectiveness of such joint models is closely tied to the nature of the chosen inter-module representations. In explicit representations, where object trajectories or occupancy grids serve as the intermediate interface, joint training and information sharing are often hindered by non-differentiable operations such as thresholding, filtering, and hard association. These disrupt the gradient flow between perception and prediction components. Recent work addresses this by eliminating non-differentiable bottlenecks~\cite{casas2024detra} and using continuous, differentiable latent representations~\cite{hu2023planning} to maintain end-to-end trainability.
In practice, this latent representation can take the form of dense occupancy feature maps~\cite{ma2024cam4docc} or sparse query tokens~\cite{gu2023vip3d}, both of which enable joint learning and information sharing between modules, with varying computational costs.

\begin{figure}[!htb]
\centering
\includegraphics[width=0.95\textwidth]{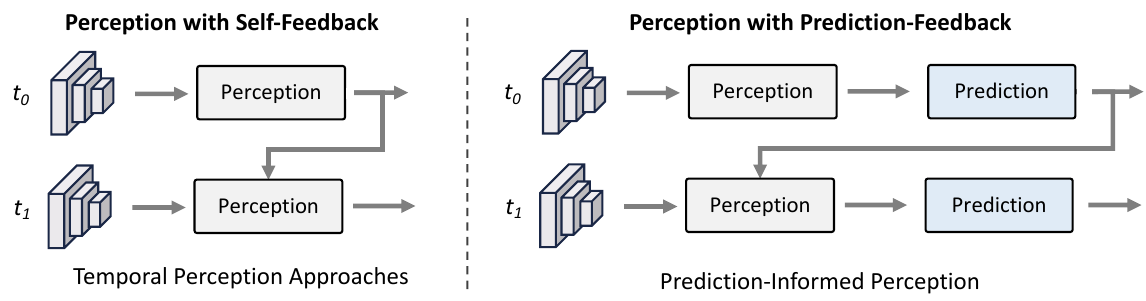}
\caption{Temporal perception approaches enhance single-frame perception models by aggregating information across historic timesteps into the current step. Prediction-informed perception approaches take a step further to leverage the predicted motion of surrounding objects during aggregating the historic features. These designs help address challenges such as long-period occlusion of critical objects, sensor limitation (e.g., insufficient coverage of surroundings, over-sparse point cloud), sensor degradation in extreme weather and lightning conditions, and tracking errors.
}
\label{fig:applicable_prediction_informed_perception}
\end{figure}

\textbf{Prediction-Informed Temporal Perception} In addition to typical joint perception and prediction approaches, another joint learning strategy, known as prediction-informed temporal perception, has been explored. Unlike traditional pipelines, where perception outputs are passed downstream to prediction, prediction-informed temporal perception feeds prediction outputs back into the perception module of subsequent cycles. This creates a temporal feedback loop that improves perception accuracy by using predicted object motion and scene dynamics to accumulate historical observations. Particularly, it helps address challenges such as long-period occlusion of critical objects, sensor limitation (e.g. insufficient coverage of surroundings, over-sparse point cloud), sensor degradation in extreme weather and lightning conditions, and tracking errors. This unlocks the potential for more robust and accurate perception systems, particularly in dynamic and uncertain environments. The following discusses three categories of perception works that exploit the historical information.

\begin{itemize}[label={\scriptsize$\bullet$}]
\vspace{-0.5em}
    \item \textit{Temporal Detection.} Before exploiting prediction priors, temporal object detection methods such as StreamPETR~\cite{wang2023exploring}, PETRv2~\cite{liu2023petrv2}, Sparse4Dv2~\cite{lin2023sparse4d}, BEVFormer~\cite{li2024bevformer}, BEVDet4D~\cite{huang2022bevdet4d}, AUTOLights~\cite{wu2023autolights} have demonstrated the potential and improved performance of propagating and accumulating historic features and queries across frames to the current time.
    By accumulating temporal information and employing a temporal alignment operation (warping) to compensate for ego-motion, these methods initialize detection queries based on the last known positions.
    Such temporal information can be useful as a prior for objects' potential positions to enhance single-frame detection models by aggregating information across multiple timesteps. 
    However, their lack of object prediction priors and explicit supervision for predicting future states results in the reliance on implicit modeling of temporal dynamics, which limits their ability to fully exploit prediction information. Integrating explicit prediction supervision could further enhance temporal learning, enabling these models to better anticipate occlusions, motion patterns, and other dynamic scene challenges, as discussed in the prediction-informed temporal detection methods below.
    
    \vspace{-0.5em}
    
    \item \textit{Prediction-Informed Temporal Detection.} 
    Compared to temporal detection methods that aggregate historic features simply via alignment operation (warping) to compensate for ego-motion, prediction-informed temporal detection methods take a step further to leverage the future motion of surrounding objects during the aggregation of historic features.
    MoDAR~\cite{liMoDARUsingMotion2023} adopts motion prediction models to \textit{propagate point clouds} observed in a historic window to the current time, which are taken as an additional modality and accumulated together to densify point clouds of the current timestep. Feeding these augmented point clouds to the LiDAR-based 3D object detection model is shown to improve general detection performance and mitigate occlusion.
    In addition to the raw point cloud, motion prediction can also be used to \textit{propagate encoded features} over time. 
    P2D~\cite{kim2023predict} predicts current object positions from historic bird's-eye-view features, which are utilized to initialize sparse object queries. Specifically, deformable attention is employed to aggregate historical and current features into the object queries. Then the 3D detection head takes the aggregated spatio-temporal feature and outputs the final detection results, which shows a more accurate and reliable 3D object detection via explicitly exploiting multi-frame inputs and object motion.

    \vspace{-0.5em}
    \item \textit{Prediction-Informed Tracking.} 
    Tracking methods can also benefit from improved short-term predictions compared to trying to implicitly learn them or relying on simple motion models such as constant velocity. For example, PF-Track~\citep{pangStandingFutureSpatioTemporal2023a} demonstrates that integrating prediction with tracking by passing predictions to the next tracking step can enhance tracking performance. 
    Compared to typical emphasis on long-term prediction accuracy (e.g., future horizons around 6 seconds),
    prediction-informed tracking here highlights the importance of balancing both long-term and short-term prediction accuracy to ensure comprehensive tracking performance, with short-term predictions playing a key role in maintaining robust tracking performance, especially during brief occlusions.
    \vspace{-0.5em}
\end{itemize}

\subsection{Prediction-Planning Integration}
\label{sec:applicable_planning}

\begin{figure}[!htb]
\centering
\includegraphics[width=0.88\textwidth]{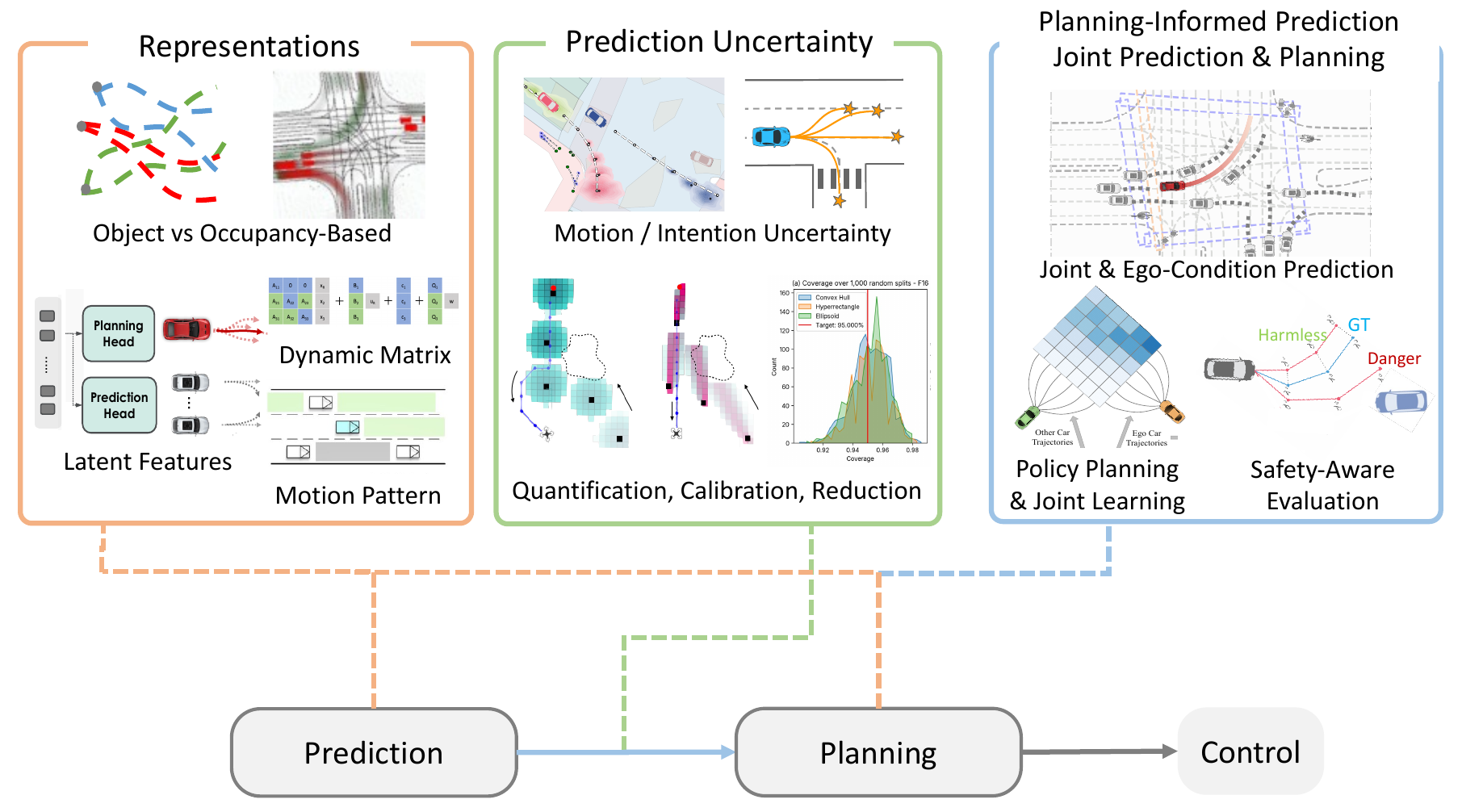}
\caption{Integration between prediction and planning considers 1) the choices of representations~\cite{ivanovic2020mats,hu2020scenario}, 2) quantifying, calibrating, and reducing prediction uncertainty~\cite{salzmann2020trajectron++,zhao2020tnt}, and 3) planning informed prediction and joint prediction and planning~\cite{liu2024reasoning,wang2021socially,he2019rethinking}.}
\label{fig:applicable_planningintegration}
\end{figure}

Having discussed the integration between perception and prediction, we now move further along the autonomy stack to examine the integration between prediction and planning. In typical autonomy architectures, motion prediction provides essential inputs for motion planning to generate safe and efficient ego trajectories. As we approach the end of the autonomy stack, it introduces additional design desiderata beyond accuracy, placing greater emphasis on real-world closed-loop performance, robustness, and compatibility with planning algorithms. Achieving seamless integration with planning involves 1) choosing suitable representations at the prediction–planning interface that enable effective information propagation, computational efficiency, and compatibility with downstream modules; 2) it also requires properly propagating, calibrating, and ultimately reducing the uncertainties and errors from prediction to support robust and risk-aware planning; 3) furthermore, joint learning of prediction and planning encourages the design of prediction outputs that are better structured for direct consumption by planning algorithms.
These concepts are illustrated in Figure~\ref{fig:applicable_planningintegration} and will be explored in greater detail in the following subsections.

\subsubsection{Representations}
\label{sec: app - prediction-planning representation}
As discussed in Section~\ref{sec:applicable_end-to-end}, among various representations, two widely used approaches for planning to consume prediction outputs are object-based representation~\cite{salzmann2020trajectron++,varadarajan2022multipath++,nayakanti2023wayformer,luo2023jfp,lee2017desire} and occupancy-based representation~\cite{salzmann2020trajectron++,varadarajan2022multipath++,nayakanti2023wayformer,luo2023jfp,lee2017desire}. Latent representations also excel in joint learning paradigms~\cite{huang2023gameformer,liu2024reasoning}. In Section~\ref{sec: applicable perception-prediction representation}, we examined the advantages and limitations of these three representations in the context of perception-prediction integration, considering factors such as object priors, granularity, computational cost, interpretability, and information completeness. These discussions also extend naturally to prediction-planning integration.
Beyond these common representations, other approaches tailored for planning have been explored in the literature, such as mixtures of affine time-varying systems~\cite{ivanovic2020mats}, and motion/topology patterns~\cite{wang2023efficient}. Each representation comes with its own strengths and limitations, as discussed below.

\textbf{Object-Based Representation} takes the object states (e.g. bounding boxes, poses) as the representation and demonstrates the advantage of 1) \textit{clear object understanding}: the explicit object structure makes it easier for planners to apply rule-based constraints (e.g., collision avoidance, lane-keeping) and incorporate structured decision-making strategies; 2) \textit{efficient scene representation}: Object trajectories provide a compact and interpretable representation of prediction outputs, making them well-suited for a wide range of planning methods. Their efficiency is particularly advantageous in multimodal prediction scenarios, where agents' future actions could be multiple hypotheses, and managing multiple potential prediction outcomes is crucial for robust decision-making.
However, if also suffered from 1) \textit{information bottleneck and error accumulation}: Due to non-differentiable operations in detection (e.g., thresholding, non-maximum suppression) and hard associations in tracking, crucial information and uncertainties from sensor data, perception, and prediction are not fully accessible to motion planning. As a result, planning becomes heavily dependent on the accuracy of upstream components and is less robust to compounding errors and uncertainties that propagate through the pipeline. 2) \textit{Limited expressiveness for open-world objects and behaviors}: Object-based representations impose structured assumptions on agent classes and motion, limiting the ability to anticipate novel objects and their associated behaviors in open-world scenarios. When novel-class agents exhibit unpredictable motion, predefined object categories and models may fail to capture critical nuances, resulting in rigid, suboptimal, or unsafe planning decisions.

\textbf{Occupancy-Based Representation} focuses on the occupancy of each cell in a 2D or 3D space, and offers advantage of \textit{representing diverse object class and motion}: 
Occupancy-based representations are more expressive in capturing fine spatial distributions and arbitrary motion patterns, making them well-suited for scenarios involving unknown or irregular objects. Unlike object-based methods, they do not rely on predefined categories, allowing greater flexibility in modeling dynamic and open-world environments.
However, it also presents notable challenges:
1) \textit{loss of object identity and planning compatibility}: 
Since occupancy grids do not explicitly encode individual object attributes, planners must compensate for this lack of structured object information. This can impact interaction-aware motion planning, rule-based constraints, and structured behavioral reasoning, which often rely on object-centric understanding. Recovering object identities from occupancy grids requires additional processing steps, adding complexity to the planning pipeline.
2) \textit{high and inefficient computation}: 
The dense nature of occupancy-based representations makes them less computationally efficient than object-centric approaches. Low-resolution grids result in coarse obstacle representations, limiting precision in planning, while high-resolution grids impose significant computational overhead, making real-time planning more challenging. The computational demands are even higher for multimodal prediction and ego-conditioned prediction, both of which are crucial for planning. Such high computational demands may necessitate adaptive resolution strategies or selective attention mechanisms to balance efficiency with accuracy.

\textbf{Latent Representation} Compared to object-based and occupancy-based representations, which carry explicit physical meanings and offer high interpretability but often suffer from information bottlenecks due to non-differentiable operations, latent representations use embeddings within neural networks to support information sharing and enable joint optimization of prediction and planning. These representations have been widely adopted in joint prediction and planning frameworks~\cite{huang2023gameformer,liu2024reasoning} as well as in end-to-end planning methods~\cite{hu2023planning,kim2024openvla,black2410pi0}. In the era of foundation models, latent representations are particularly effective at leveraging large-scale data without requiring carefully designed modular structures or interfaces. However, it remains challenging to ensure safety given the limited interpretability of these systems, and to balance performance with the demands of real-time computation.

\textbf{Mixtures of Affine Time-Varying Systems (MATS)} have been proposed as a representation more amenable to downstream planning and control applications~\cite{ivanovic2020mats}. Unlike object-based or occupancy-based representations, MATS are designed to align more naturally with system dynamics, as most planning and control algorithms operate within a dynamical systems framework rather than directly reasoning about future trajectory or occupancy.
Specifically, inspired by linear dynamic systems, MATS generate future motion in the form of matrices representing dynamic system parameters, allowing planners to employ classic control-theoretic principles and classic optimization when making decisions. This structured representation enables direct integration with model-based control approaches, facilitating smoother and more interpretable trajectory planning.

\textbf{Motion and Topology Patterns}
Another class of methods adopts a ``divide and conquer" strategy by decomposing motion prediction into two hierarchical levels: high-level prediction of motion or topology patterns, followed by low-level trajectory generation.
Various designs for these high-level representations have been proposed. For instance, driving skills~\cite{wang2023efficient,deo2018multi} and robot plans~\cite{cheng2020towards} are used as intention representations to enhance learning efficiency and generalization across tasks and environments.
Other approaches aim to model topological relationships between agents and the environment. HATN~\cite{hu2020scenario,wang2021hierarchical,wang2022transferable}, for example, introduces the concept of a Dynamic Insertion Area (DIA)—an invariant high-level representation that captures complex road geometries and interaction patterns. When combined with a low-level Frenet Coordinate-based representation, this framework demonstrates strong zero-shot generalization across diverse driving scenarios.
Similarly, the concept of Topology-aware Scene Modes~\cite{chen2023categorical,mavrogiannis2022analyzing,chen2022scept,rowe2023fjmp} explicitly encodes spatial and topological relationships between agents and map elements using either continuous values or discrete categories. For instance, the relative positioning between two agents may be captured via free-end homotopy~\cite{chen2023categorical,chen2023interactive} (e.g., clockwise, counterclockwise), while the spatial relationship between an agent and a map element can be categorized into labels such as ON, AHEAD, BEHIND, LEFT OF, RIGHT OF, or MISALIGN.
Further discussion on the generalizability of these representations is provided in Section~\ref{sec: generalization DA/DG invariant feature}.
    
\textbf{Sensor-Level Representations} As similarly discussed in Section~\ref{sec: applicable perception-prediction representation} on perception–prediction integration, and resonating with the concept of a world model, raw sensor inputs such as video images~\cite{hu2023gaia} and point clouds~\cite{khurana2023point} have also been explored as proxies for the temporal evolution of the scene during integrating prediction and planning.
We discuss these approaches in detail in Section~\ref{sec:gen_pred_ssl}.
Among these, video generation has garnered particular interest due to several compelling advantages: 1) video provides a unified representation capable of capturing rich physical world information; 2) a wide range of tasks in computer vision, embodied AI, and the sciences can be formulated or supported through video generation models; and 3) the vast availability of video data from sources like the internet.
Recent works have demonstrated proof-of-concept results using action-conditioned video generation to simulate future scene evolution based on robot actions~\cite{yang2024video} or driving commands~\cite{wang2024driving}. By visually anticipating multiple future outcomes and receiving feedback from these imagined trajectories before real-world decision-making, such models enable more informed and rational planning, ultimately improving the generalization and safety of robot or driving systems.
Looking forward, key challenges for this line of research include ensuring geometric consistency in generated videos, accurately conditioning on actions, and developing effective methods to directly evaluate planning performance from generated visual predictions.

\subsubsection{Deal with Prediction Uncertainty in Planning}
\label{sec: prediction uncertainty}

\textbf{Prediction Uncertainty/Errors - What? Why? How?}
Integrating prediction and planning poses challenges in handling the inherent uncertainties in prediction outputs, which stem from different sources and demand distinct handling strategies. 
Effectively \textit{quantifying, calibrating, and ultimately reducing} these uncertainties is critical for confidence-aware and thus reliable prediction, and safe and efficient downstream planning. 
In theory, these uncertainties would include several sources:

\begin{itemize}[label={\scriptsize$\bullet$}]
    \vspace{-0.5em}
    \item \textit{Epistemic Uncertainty.} This refers to the uncertainty arising from \textit{a lack of knowledge or information} about the true underlying system or environment. This type of uncertainty is particularly prominent in scenarios where 1) the training data is sparse, incomplete, or unrepresentative of the full range of possible situations the system may encounter. For example, in autonomous driving, if the vehicle encounters an unusual construction site layout that was never seen during training, the prediction model may exhibit high epistemic uncertainty. 2) The \textit{input design} is limited, where important factors, such as road conditions, weather, or interactions between objects, are not captured in the input features fed to the model. 3) Additionally, even if relevant data and features are available, epistemic uncertainty may arise from \textit{limited model capacity}, where the model lacks sufficient size or expressive power to learn the complex mappings between inputs and outputs. Without adequate capacity, the model may oversimplify relationships, fail to generalize to unseen cases, or ignore critical interactions between factors, further amplifying epistemic uncertainty. In other words, epistemic uncertainty captures the model's \textit{ignorance} about parts of the environment it has not observed, understood, or been given access to through its inputs or architecture.
    Epistemic uncertainty is \textit{reducible} in principle—as more data is collected, particularly from diverse and rare scenarios, and as the input design and model capacity is improved to capture critical and more critical factors, the model can refine its understanding and reduce its epistemic uncertainty.
    \vspace{-0.5em}
    \item \textit{Aleatoric Uncertainty.} This instead captures \textit{the inherent randomness} in the system or environment itself. This type of uncertainty persists even if the model is fully informed and perfectly trained, as it represents the inherent stochasticity in the system, where the same initial conditions can lead to different outcomes.
    Aleatoric uncertainty in human behavior can be further divided into two aspects: 1) \textit{Intention uncertainty}, which refers to the uncertainty about the target object's destination to reach or task to be performed. For instance, different pedestrians or vehicles may have different goals, and these goals lead to different trajectories accordingly. 2) \textit{Motion uncertainty}, which captures the variability in how a human executes a particular movement even when the goal is fixed. Human motion is often only approximately rational with respect to an objective function and can be affected by personal choices, distractions, or momentary deviations, leading to variations in movement. 
    \vspace{-0.5em}
    \item \textit{Sample Uncertainty.} A perhaps surprising source of uncertainty is the sampling process itself, which is needed in many prediction models with a probabilistic latent space. While the sampling process is necessary for prediction diversity, it could be problematic with a downstream planner: it is demonstrated in \cite{chen2023tree} that high-frequency sampling of the latent variable leads to poor temporal consistency, i.e., predicted trajectories tend to oscillate between time steps, which can significantly reduce the planning performance. Readers are also referred to Section~\ref{sec:CL_eval} for more discussions regarding the importance of temporally consistent predictions.
    \vspace{-0.5em}
\end{itemize}
In essence, multimodal prediction is critical for capturing the diverse and stochastic nature of human and agent behaviors. Predictors must not only output likely trajectories, but also reflect a calibrated range of plausible futures. A common failure mode is mode collapse, where models produce overconfident and overly conservative outputs that fail to represent meaningful alternatives—limiting planner flexibility and potentially leading to unsafe or suboptimal decisions. Conversely, generating over-diverse or uncalibrated predictions—such as including implausible or irrelevant futures—can overwhelm downstream modules and reduce system reliability. These challenges highlight the importance of balancing diversity and precision, ensuring that predictive uncertainty is both informative and actionable in real-world deployment.

\textbf{Quantifying Prediction Uncertainty}
 As a first step, modelling and quantifying these uncertainties have been crucial for reliable systems, and multiple approaches have been proposed.

 \begin{itemize}[label={\scriptsize$\bullet$}]
    \vspace{-0.5em}
    \item To model the \textit{motion uncertainties}, it is common to learn a distribution (e.g. Gaussian distribution~\cite{alahi2016social,djuric2020uncertainty,salzmann2020trajectron++}) to represent the uncertainty of the predicted waypoint of each timestep. However, such unimodal approaches, outputting a single trajectory per agent and the Gaussian uncertainty per timestep, are often unable to adequately capture multiple possibilities of human intentions, and typically average over behaviors (e.g., halfway between a right turn and going straight); 
    \vspace{-0.5em}
    \item To capture such multiple-future \textit{intention uncertainties}, multimodal approaches have been proposed to output multiple possible trajectories and their associated likelihoods. This is usually accomplished by 1) either formulating the prediction problem as a classification problem over target goal points~\cite{zhao2020tnt,gu2021densetnt}, a fixed or adaptive trajectory set~\cite{chai2019multipath,phan2020covernet}, motion patterns~\cite{deo2018multi,codevilla2018end}, topology modes~\cite{chen2023categorical,wang2022transferable,hu2020scenario}, occupancy grids~\cite{hong2019rules,mohajerin2019multi}, and learnable queries~\cite{zhou2023query,shi2022motion,shi2024mtr++}; 2) or encoding the agent history and map information into a discrete or continuous latent space from which multimodal predictions can be sampled~\cite{salzmann2020trajectron++,zhou2022hivt}. 
    \vspace{-0.5em}
    \item Moreover, various \textit{classic uncertainty quantification (UQ)} methods have been applied for motion prediction. Bayesian inference and Kalman filter are applied to estimate the uncertainty regarding the human intent/rationality~\cite{fisac2018probabilistically}, human psychological characteristics~\cite{wang2021socially}, and deep-learning model parameters~\cite{wang2022transferable}. 
Meanwhile, since the epistemic and aleatory uncertainties are observed together, uncertainty estimation methods can often be overly conservative. To tell the two sources apart, a model ensemble \cite{tang2022prediction,min2019rnn,huang2019uncertainty} is used where multiple models are trained/used and their difference reflects the source of uncertainty, and \citet{acharya2023learning} adds a Bayesian flavor to the model ensemble approach. Monte Carlo dropout~\cite{eltouny2024tgn} is also applied to estimate the uncertainty from multiple model inference with different dropout masks. 
\end{itemize}

\textbf{Calibrating Prediction Uncertainty}
While uncertainty quantifying techniques, as mentioned above, focus on estimating the magnitude and type of uncertainty associated with predicted motion, uncertainty calibration instead concerns the degree to which these \textit{estimated uncertainties accurately reflect the true likelihood of future outcomes}. In other words, a well-calibrated prediction model not only outputs diverse and plausible future motions
but also assigns uncertainty values (e.g., confidence intervals or predicted variances) that correctly match the actual error rates observed in practice. For example, if a prediction system claims 90\% confidence for a particular trajectory, a perfectly calibrated model would ensure that the true future lies within the associated uncertainty bounds exactly 90\% of the time.

\begin{itemize}[label={\scriptsize$\bullet$}] 
\vspace{-0.5em}
\item \textit{Common Uncertainty Calibration Methods.} Traditional calibration techniques, mostly developed for classification tasks, focus on aligning predicted confidence scores with actual empirical accuracy. Notable methods include Platt scaling~\cite{platt1999probabilistic}, isotonic regression~\cite{zadrozny2002transforming}, and the widely used temperature scaling~\cite{guo2017calibration}, which rescales logits to achieve better alignment between predicted and actual confidence. In regression or distributional settings, techniques such as conformal prediction~\cite{vovk2005algorithmic} and quantile regression calibration extend these ideas to ensure that predicted uncertainty intervals cover the true outcomes at the desired confidence levels.

\vspace{-0.5em}
\item \textit{Limited Exploration in Motion Prediction.} 
Despite the growing attention to uncertainty quantification in motion prediction, particularly for autonomous driving and robotics scenarios, the calibration aspect has received \textit{far less attention}. Most existing works focus on generating diverse multimodal trajectories with associated likelihoods or variances, but rarely evaluate whether these likelihoods or variances are properly calibrated with respect to actual future outcomes. Notably, a few works apply conformal prediction techniques~\cite{yao2024sonic,lindemann2023safe,tumu2024multi,luo2024sample} to construct prediction intervals for trajectories, partially because existing benchmarks do not reward methods doing so. However, these works primarily focus on estimating a \textit{guaranteed coverage rate} in a post-hoc manner (uncertainty quantification in a broad sense), rather than explicitly calibrating the predicted uncertainties output by the underlying prediction models themselves. As a result, current motion prediction systems often \textit{overestimate uncertainty in familiar scenarios and underestimate uncertainty in novel, complex, or interactive environments}, making \textit{uncertainty calibration an underexplored and critical challenge for future research}.
Combining calibration-aware training objectives (e.g., explicitly penalizing miscalibrated uncertainties during training) with post-hoc calibration techniques (e.g., temperature scaling or conformal correction layers) may provide a promising path towards more trustworthy and risk-aware motion prediction systems, especially in safety-critical applications like robotics and autonomous driving.
\end{itemize}

\textbf{Reducing Prediction Uncertainty}
Beyond quantifying and calibrating prediction uncertainties, an equally important aspect and the ultimate goal is to reduce such uncertainties. Existing works have explored various uncertainty reduction approaches, such as better leveraging posteriors, active exploration, and tighter integration with downstream planning and control, as discussed below.


\begin{itemize}[label={\scriptsize$\bullet$}]
    \vspace{-0.5em}
    \item \textit{Leveraging Posteriors from Historical Observations.} 
    As autonomous systems are deployed in the real world, they continuously observe the behavior of surrounding objects. The more we observe, the more we can learn about these objects, especially humans, whose internal states are often uncertain.
    By incorporating such posterior knowledge from historical observation, we can refine the estimation of uncertain internal states of humans, such as human intent and behavioral characteristics, over time. To this end, Bayesian inference and Kalman filtering techniques have been widely applied to continuously estimate and update the human goals and rationality~\cite{fisac2018probabilistically}, social and psychological characteristics~\cite{wang2021socially}, as well as the parameters of deep-learning models used for prediction~\cite{wang2022transferable}. By iteratively leveraging posteriors as new observations arrive, these methods gradually reduce epistemic uncertainty and improve prediction fidelity.

    \vspace{-0.5em}
    \item \textit{Active Information Gathering.}
    While the estimation of human internal states (e.g., goals, intentions, preferences) mentioned above is \textit{passive}, where the prediction system merely observes human actions and updates its estimates accordingly, a promising complementary approach is to enable the autonomous agent to engage in \textit{active information gathering}, where it proactively selects actions to elicit informative responses from other agents~\cite{sadigh2016information}. For example, an autonomous vehicle may intentionally \textit{nudge into a neighboring lane} to probe the surrounding human driver’s level of assertiveness or driving style. By purposefully interacting with uncertain agents, the ego agent can gather clarifying information, thus reducing prediction uncertainty in future interactions.

    \vspace{-0.5em}
    \item \textit{Ego-grounded and Scene-consistent Prediction.}
    A third strategy reduces prediction uncertainty by leveraging how other agents' behavior is influenced by the ego agent, as well as ensuring scene-wide consistency across predicted trajectories for multiple agents. We term this \textit{Ego-grounded Scene-consistent Prediction}, which reduces uncertainty through two complementary mechanisms. They are also closely related to joint prediction and planning approaches, which we will further elaborate on in the next subsection.  
    \begin{itemize}
        \item \textit{Ego Conditioning}, where the prediction model conditions the future trajectories of surrounding agents on the planned behavior of the ego vehicle~\cite{salzmann2020trajectron++,song2020pip,liu2022interaction,chen2022scept}. Since the ego agent’s policy (e.g., yielding, accelerating, nudging) directly influences how others respond, explicitly modeling this coupling reduces uncertainty stemming from unaccounted ego-agent interactions.
        \item \textit{Joint Scene-consistent Prediction}, where the prediction model jointly reasons about the plausible futures of all agents in the scene~\cite{yuan2021agentformer,ngiam2022scene,luo2023jfp,chen2022scept}, ensuring that multi-agent trajectories form a coherent, interaction-aware, and collision-free set. By jointly enforcing spatial feasibility, behavioral plausibility, and game-theoretic rationality across all predicted trajectories, the prediction model can rule out physically or socially inconsistent futures, thus reducing overall uncertainty. 
    \end{itemize}
\end{itemize}

\subsubsection{Planning-Informed Prediction and Joint Learning}
\label{sec: app - planning-informed prediction}

\textbf{Planning-Informed Prediction} 
In the classic autonomy stack, the motion prediction and planning modules are typically applied sequentially — the prediction module generates predictions of other agents’ future motions, which are then passed to the planning module, enabling the ego agent to formulate its trajectory based on these predictions. However, in actual application, the relationship between prediction and planning is inherrently bidirectional - 1) the planning poses certain requirements that prediction needs to meet, e.g. joint predictions of all agents in the scene; 2) the ego agent’s planned motion also simultaneously influences the future behavior of surrounding agents, creating a dynamic feedback loop between prediction and planning. To better capture these bidirectional interactions, the concept of planning-informed prediction has been introduced and actively explored. This concept can be implemented with varying levels of complexity, aiming to achieve more realistic and behaviorally-consistent predictions in highly interactive environments.

\begin{itemize}
    \vspace{-0.5em}
    \item \textit{Marginal Prediction.}  
    Early prediction models often produced marginal predictions (also known as agent-centric predictions)~\cite{salzmann2020trajectron++,varadarajan2022multipath++,zhao2020tnt,nayakanti2023wayformer}, where the model predicts the future motion of one selected agent at a time, and the future trajectories of all agents are obtained by running the model independently for each agent and combining these prediction together. However, downstream planners need to reason about the motions of all agents in the scene simultaneously, which requires understanding their joint motions. Marginal predictions, by design, do not explicitly evaluate whether the combination of all predicted trajectories results in a globally consistent and plausible scene-level outcome.
    Such inconsistency issues become even more pronounced in the presence of multimodal predictions, as there is no principled way to meaningfully combine the marginal distributions of different agents from different modes into a globally coherent scene.

    Notably, many existing benchmarks primarily evaluate prediction models based on marginal predictions, where each agent’s future is predicted independently. These benchmarks typically adopt agent-level metrics and only assess the accuracy of each agent’s predicted trajectories in isolation, without considering whether the combination of all agents’ predictions forms a globally consistent and plausible scene. As a result, high-quality marginal prediction performance in these benchmarks does not necessarily translate into practical utility for downstream planners.
    For further discussion regarding the evaluation, please refer to Section \ref{sec:metrics}.

    \vspace{-0.5em}
    \item \textit{Joint Prediction.}
    To better support downstream planning, models that generate joint predictions (also known as scene-centric predictions) for the entire scene involving multiple agents have been developed \cite{yuan2021agentformer,ngiam2022scene,luo2023jfp,chen2022scept}. These joint predictions are expected to be self-consistent, meaning the predicted motions of all agents should not lead to collisions between them \cite{chen2022scept} and should also comply with the surrounding map geometry. However, this introduces the challenge of properly defining joint patterns/modes that capture the topology relationships among agents and map elements.
    
    While some works define scene modes as a latent space to be sampled from \cite{yuan2021agentformer,ngiam2022scene}, these approaches do not explicitly model the underlying relational and topological structures between agents and the map, which can limit their performance in complex scenes. To address this, the concept of dynamic insertion area~\cite{wang2022transferable,wang2021hierarchical,hu2020scenario} and Topology-aware Scene Modes~\cite{chen2023categorical,mavrogiannis2022analyzing,chen2022scept,rowe2023fjmp} has been proposed, explicitly encoding spatial and topological relationships between agents and map elements into continuous values or discrete categories. For example, the relative positioning between two agents can be represented using free-end homotopy~\cite{chen2023categorical,chen2023interactive} (e.g., clockwise, counter-clockwise), while the spatial relationship between an agent and a map element can be categorized into predefined labels such as ON, AHEAD, BEHIND, LEFT OF, RIGHT OF, or MISALIGN. Each unique combination of these spatial and topological relations defines a possible Topology-aware Scene Mode, enabling structured reasoning over the joint scene configuration. However, this approach suffers from over-reliance on designed topology structure, and exponential growth in the number of scene modes as the number of agents and map elements increases — a challenge further exacerbated when accounting for multimodal predictions.
    
    \vspace{-0.5em}
    \item \textit{Ego-Conditioned Prediction.}
    As mentioned above, scene-centric prediction models face significantly higher complexity and computation, which becomes a bottleneck when enumerating joint scene modes. Ego-conditioned prediction was proposed as a remedy, not only reducing computational cost, but also lowering uncertainty and better aligning with how planning systems consume predictions in practice. With a downstream planner in mind, its core idea is that the motion prediction should focus its computational resources on parts that matter to the planning task. 
    In practice, such planning conditioning can be achieved in two ways.
    \begin{itemize}
        \vspace{-0.5em}
        \item \textit{Explicitly}, through approaches that directly condition prediction on hypothetical future trajectories of the ego vehicle~\citep{wang2021socially,sun2018probabilistic,agro2023implicit,salzmann2020trajectron++,song2020pip,liu2022interaction,chen2022scept}, identify and focus only on agents most relevant to the ego vehicle~\citep{messaoud2020attention,zhang2022ai,messaoud2021trajectory}, or only perform occupancy prediction to regions that will be traversed by the planned ego trajectory~\citep{agro2023implicit}. During training, the ground-truth ego future motion is used as the conditioning input; during inference, a trajectory sampler generates a set of ego-motion candidates, each of which is used to condition the prediction model and evaluate the resulting planning outcome. This ego-conditioning strategy has been shown to improve both prediction accuracy (when using ground-truth ego motion) and downstream planning performance~\citep{salzmann2020trajectron++,chen2023tree}.  
        \item \textit{Implicitly}, via end-to-end joint training of the prediction and planning modules~\citep{hu2023planning,karkus2023diffstack,huang2023dtpp,huang2023gameformer}, where the model learns to prioritize high-impact agents by minimizing the overall planning loss.
    \end{itemize}

    \vspace{-0.5em}
    \item \textit{Policy Planning.} 
    With ego-conditioned prediction, the planner can generate ego plans that account for their influence on other agents’ future motions. Such a decision-making process, often known as policy planning or contingency planning, contrasts with traditional planning methods that rely on unconditioned motion predictions and could be biased due to ignoring the impact of ego plans on the scene. By modeling the bidirectional interaction between the ego vehicle and its environment, policy planning enables safer and more socially compliant behavior.
    In practice, policy planning can be implemented using either sampling-based methods or optimization-based methods:
    
    \begin{itemize}
        \vspace{-0.5em}
        \item \textit{Sampling-based methods} is probably the more common approach, where a discrete set of ego trajectories is sampled, while other agents’ motions are predicted conditioned on each sampled ego trajectory, and the optimal ego action is then selected via dynamic programming. The sampling process usually accounts for the map geometry and the ego state~\citep{wang2021socially}, and can be extended into a tree structure~\citep{chen2023tree} to better capture multi-step interactions between the ego and surrounding agents, where multimodal behaviors are modeled as branching decisions across stages.
        However, as pointed out in \citet{chen2023tree}, as the number of ego-motion samples and ego-conditioning predictions gets large (for better planning performance), policy planning becomes increasingly expensive, and one needs to strike a balance between performance and run-time. To avoid running ego-conditioned prediction on a large number of samples, several tricks have been proposed. 
        DTPP~\citep{huang2023dtpp} adopts different sampling granularities at different stages to reduce the number of ego trajectory candidates. It further prunes low-score nodes in the trajectory tree using a learned cost model, allowing the planner to focus on high-value branches. The authors show that node pruning significantly reduces runtime (by nearly half) without compromising planning performance.
        In addition, unlike prior approaches that perform prediction separately for each sampled ego trajectory, DTPP incorporates the entire ego trajectory tree into a single prediction model, further improving efficiency by avoiding redundant computations.
        \item \textit{Optimization-based methods.} Another line of work bypasses the high computational cost of sampling-based policy planning by adopting optimization-based approaches, though this comes with the overhead of solving a potentially costly optimization problem. IJP~\citep{chen2023interactive} achieves this by removing the need for repeated ego-conditioned predictions and requiring only a single unconditioned prediction. It first uses a joint trajectory predictor to generate unconditioned future trajectories of surrounding agents, then employs a route planner to produce multiple ego trajectories representing different interaction homotopy classes. For each ego trajectory, the method jointly optimizes the motions of the ego vehicle and nearby agents to construct plausible scene modes. The ego plan with the lowest overall cost is selected as the final decision.
        MATS~\citep{ivanovic2020mats} enables differentiable policy planning through a novel prediction representation. It formulates each agent's motion as a dynamic system influenced by its own actuation, other agents, and the map context. Instead of directly predicting trajectories, the model predicts parameters that govern the future dynamics of surrounding agents. This formulation allows for direct optimization of the ego vehicle’s future motion while explicitly accounting for its influence on the behavior of others.
    \end{itemize}
    Notably, another line of work incorporates notions of rational and reactive behavior into the policy planning process by leveraging concepts such as game theory~\citep{huang2023gameformer} and cumulative prospect theory~\citep{sun2019interpretable,sun2018courteous}. These frameworks provide useful inductive biases that can guide the learning process. However, for the same reason, they could potentially be constrained by the assumptions of their underlying notions and may lack the expressiveness compared to purely data-driven approaches.    
\end{itemize}

\textbf{Joint Prediction and Planning} In addition to planning-informed prediction, an alternative and increasingly popular approach is joint prediction and planning~\cite{huang2023dtpp,huang2023gameformer,liu2024reasoning}, which learns a joint model for predicting the future behaviors of surrounding agents and generating feasible ego trajectories simultaneously, conditioned on perception outputs. The motivation for this design stems from two key considerations:

\begin{itemize}[label={\scriptsize$\bullet$}]

    \vspace{-0.5em}
    \item \textit{Overcoming limitations of rule-based planning.}
    Traditional rule-based planning approaches, while interpretable and safety-guaranteed in well-defined scenarios, often rely on extensive hand-engineering, including the design of complex reward functions, hierarchical planning trees, and exhaustive corner case handling rules. As the complexity of rule-based systems increases, it becomes increasingly challenging to manage rule interactions, extend the system to handle new scenarios, and avoid brittle behavior due to rule conflicts. This results in diminishing returns, where further improving performance becomes increasingly difficult. In contrast, joint prediction and planning methods—sharing a similar motivation with fully end-to-end approaches discussed in Section~\ref{sec: applicable autonomy stack}—are fully data-driven and capable of learning nuanced, human-like driving behaviors directly from expert demonstrations. For instance, when navigating a gently curving road, rule-based planners often attempt to follow the exact centerline, while data-driven systems tend to learn the more human-like behavior of smoothing the path into a straighter line when possible. However, the absence of explicitly injected human knowledge, as used in rule-based systems, means that such data-driven joint prediction and planning algorithms highly rely on and pose higher requirements of the diversity and richness of the training data.

    \vspace{-0.5em}
    \item \textit{Decouple from perception pipelines.} 
    Unlike fully end-to-end systems that directly map raw sensor data to control commands as discussed in Section~\ref{sec: applicable autonomy stack}, joint prediction and planning approaches retain a decoupling from the perception system, offering enhanced interpretability and transparency. This separation facilitates inspection and debugging of the system’s reasoning process and makes it easier to assess safety and robustness, particularly in safety-critical domains like autonomous driving and robotics, where regulatory and legal standards often demand clear explanations for system behavior.

    Moreover, this decoupling enables the reuse of mature perception pipelines. Modern perception algorithms—especially for visual tasks—have benefited from large-scale annotated datasets, standardized benchmarks, and rigorous evaluation protocols. The abundance of internet-scale visual data has also led to the development of highly robust and transferable perception backbones. In contrast, datasets for motion prediction and planning, particularly in complex interactive scenarios, remain relatively limited in both volume and diversity. As a result, several recent works adopt a hybrid paradigm: treating perception as a standalone module trained with established techniques and large-scale data, while focusing end-to-end learning efforts on jointly optimizing prediction and planning, where data is more scarce and the benefits of joint optimization are more pronounced.

\end{itemize}
\vspace{-0.5em}

\noindent

Joint prediction and planning hold great potential for enabling human-like and adaptive driving behavior in complex, interactive environments. However, realizing its full potential requires overcoming several key challenges. These include developing scalable methods for acquiring diverse, interaction-rich training data; designing model architectures that can reason effectively about multi-agent interactions while remaining computationally tractable; and integrating uncertainty estimation and safety constraints into the learning objective to support reliable and interpretable decision-making. While such challenges are shared across other approaches, they remain relatively underexplored within the context of joint prediction and planning, highlighting important opportunities for future research.

\subsection{Evaluation}
\label{sec:applicable_evaluation}

\begin{figure}[!t]
\centering
\includegraphics[width=0.8\textwidth]{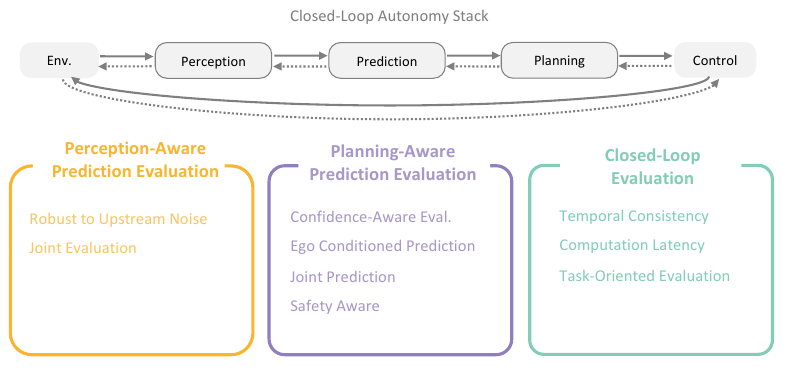}
\caption{To more accurately reflect real-world performance, a realistic evaluation framework must capture implications from upstream perception, downstream planning, and closed-loop systems.}
\label{fig:applicable evaluation}
\end{figure}

In realistic deployment, motion prediction does not operate in isolation, but functions as a critical module within the closed-loop autonomy stack, receiving inputs from upstream localization and perception, and informing downstream planning and control. However, existing benchmarks typically adopt a simplified evaluation setting, where prediction is assessed independently of other modules and in an open-loop manner. Specifically, they often assume idealized, curated perception inputs without accounting for associated noise and errors, and disregard how prediction results are consumed by downstream planning. Moreover, evaluations are typically conducted in open-loop settings, overlooking important aspects such as temporal consistency, ultimate task performance, and computation efficiency.
To more accurately reflect real-world performance, a realistic evaluation framework must incorporate \textbf{perception-aware metrics}, \textbf{planning-oriented metrics}, and \textbf{closed-loop dynamics}, which we illustrate in Figure~\ref{fig:applicable evaluation} and will discuss in the following subsections, respectively.

\subsubsection{Perception-Aware Prediction Evaluation.} 
\label{sec: app - Perception-Aware Prediction Evaluation}

Existing benchmarks typically assume curated perception results to serve as inputs to motion prediction models. However, motion prediction models trained on these curated inputs often struggle to transfer effectively to real-world deployment settings, where perception errors and uncertainties are unavoidable and will accumulate in the stack if not properly managed. As a result, performance on these benchmarks may fail to reflect true real-world prediction performance and degrade significantly in deployment. To this end, there have been two threads of efforts investigating perception-aware prediction evaluation:

\begin{itemize}[label={\scriptsize$\bullet$}]
    \vspace{-0.5em}
    \item \textbf{Prediction Evaluation when Injecting Perception Uncertainties into Inputs} As discussed in Section~\ref{sec:deployable_perception_dealing_with_uncertainty}, various approaches propose to introduce uncertainties from object detection (class~\cite{ivanovic2022heterogeneous} and state~\cite{ivanovic2022propagating}), object tracking~\cite{weng2022whose,wengMTPMultihypothesisTracking2022}, and mapping~\cite{gu2024producing}, to the inputs of the prediction model during evaluating and training. In these works, prediction methods are evaluated using the standard prediction metrics, but taking probabilistic perception inputs instead of deterministic ones. When these uncertainties are taken into the training process of prediction models, improved prediction performance and robustness are observed.

    \vspace{-0.5em}
    \item \textbf{Joint Perception and Prediction Evaluation} 
    While the approaches discussed above inject perception uncertainties into prediction inputs, such as object classes or historical states, the deeper coupling aspects in joint detection and prediction are not considered. 
    First, in joint perception–prediction paradigms, the prediction module can also influence perception, as explored in prediction-informed perception (Section~\ref{sec: app - prediction-informed perception}), which is not accounted for in the evaluation protocol above.
    Second, the above evaluation setting falls short in capturing the issue of input-output correspondence in joint perception and prediction. In standard trajectory prediction tasks, as framed by most existing benchmarks, an oracle input–output correspondence is assumed: each predicted trajectory is supervised or evaluated against a known and fixed ground-truth trajectory.
    In contrast, object detection does not assume such a direct input-output correspondence. For each detected object, the corresponding ground-truth label must be dynamically determined through a matching process (e.g., Hungarian matching), where detected objects are associated with the closest ground-truth instances.
    In joint perception and prediction, each trajectory prediction is conditioned on a detected object, which means evaluation must follow the detection paradigm and include an online matching step to associate predicted trajectories with the closest ground-truth counterparts. This introduces a critical dependency: since each prediction is conditioned on a detected object whose identity may not be aligned with the ground-truth instance, prediction performance becomes tightly coupled with detection quality, as errors or misses in detection directly affect which ground-truth trajectory is used for evaluation, potentially skewing the results.
    
    To this end, with the rise of end-to-end approaches, recent methods have increasingly advocated for jointly evaluating prediction alongside other tasks. In the simplest form, prediction metrics are reported together with perception metrics, as seen in UniAD~\cite{yang2023unipad}. 
    Beyond this, several works have proposed novel metrics specifically designed to evaluate joint detection and prediction performance. For example, 
    ~\citet{peri2022forecasting,xu2024towards} propose the Mean
    Forecasting Average Precision ($mAP_f$), which extends beyond standard detection AP to consider a true positive as trajectories with correct first frame detections and correct final forecast position. In their formulation, a true positive requires a correct match at both the current timestamp (detection) and a future timestep (forecast); otherwise, the forecast is treated as a false positive.
    Similarly, ViP3D~\cite{gu2023vip3d} proposes End-to-End Prediction Accuracy (EPA), which adopts a similar definition of true positives as $mAP_f$, but further penalizing over-detections (False Positives). 

    
    Note that a systematic analysis and comparison of these metrics remains lacking in the current literature, and different works could adopt different evaluation protocals. For example, during the matching process, DeTra \cite{casas2024detra} uses an IoU threshold of 0.5 between detection and ground truth to associate with the closest box and then evaluate prediction, while ViP3D \cite{gu2023vip3d} uses a distance threshold of 2m with an L2 distance bipartite matching to associate ground truth objects. A unified evaluation protocol for joint perception and prediction is still needed within the community to ensure fair and consistent comparisons.    
    
\end{itemize}

\subsubsection{Planning-Aware Prediction Evaluation.}
\label{sec: app - Planning-Aware Prediction Evaluation}

Targeting applications in robotics and autonomous driving, the primary goal of motion prediction is to support downstream motion planning. Therefore, it is natural to evaluate prediction quality through the lens of planning, where metrics can reflect how predictions impact the overall safety, comfort, and efficiency of the autonomous system.

As the motion prediction community continues to evolve, various evaluation metrics have been proposed, as discussed in Section~\ref{sec:metrics}, and the dominant metrics adopted by popular benchmarks have shifted over time. Early evaluation relied heavily on displacement-based metrics, such as Final Displacement Error (FDE) and Average Displacement Error (ADE), which measure the distance between predicted and ground-truth trajectories. These were later extended to minFDE and minADE to better accommodate multi-modal predictions by evaluating only the best-matching trajectory. To account for uncertainty, brier-minFDE was introduced, incorporating a likelihood term into the minFDE score. In parallel, mean Average Precision (mAP) was adopted to better capture the coverage of multiple plausible futures.
While these metrics represent meaningful progress, they still fall short in capturing how prediction quality translates to the practical utility in downstream planning. Bridging this gap remains a critical direction for future research. In this section, we will discuss some key issues with the current evaluation of motion prediction when downstream planning is taken into account.

\begin{itemize}[label={\scriptsize$\bullet$}]
    \vspace{-0.5em}
    \item \textbf{Minimum vs. Most Likely} A natural question arises: why do current metrics evaluate the best-case prediction rather than the most likely one? From a planning perspective, particularly when using deterministic planning, we typically care about the most likely future motion of surrounding agents to generate safe and efficient ego trajectories. While Brier-minFDE partially addresses this concern by incorporating a likelihood term into the error metric, a more direct alternative would be to report FDE/ADE for the most likely predicted trajectory, which would better reflect downstream planning utility.

    Besides, a side topic regarding multi-modality is that, while human motion is inherently uncertain, for any given data point, only a single ground-truth trajectory is available, even though multiple future motions could be equally reasonable. This inherent ambiguity makes the evaluation of motion prediction more challenging than deterministic tasks like object detection. To achieve a statistically meaningful evaluation under such uncertainty, a larger number of evaluation samples is required.
    
    \vspace{-0.5em}
    \item \textbf{Marginal Prediction vs. Joint Prediction} As previously discussed in Section~\ref{sec: app - prediction-informed perception}, most existing benchmarks usually consider single-agent prediction, which is often called marginal prediction. This has led to the emergence of methods that optimize marginal accuracy alone, where the future trajectories of all agents are obtained by running the model independently for each agent and combining the predictions together.
    However, downstream planners need to reason about the motions of all agents in the scene simultaneously, which requires understanding their joint motions, such as inter-agent interactions, social compliance, and global feasibility. 
    Marginal predictions, by design, do not explicitly evaluate whether the combination of all predicted trajectories results in a globally consistent and plausible scene-level outcome.
    Such inconsistency issues become even more pronounced in the presence of multimodal predictions, as there is no principled way to meaningfully combine the marginal predictions of different agents from different predicted modes into a globally coherent scene.
    As a result, high-quality marginal prediction performance does not necessarily translate into practical utility for downstream planners. Despite this limitation, most existing benchmarks only require forecasting the trajectory of a single agent, which has, to some extent, encouraged an ill-posed focus on marginal prediction within the community.
    On the other hand, although some datasets—such as the Waymo Open Motion Dataset—emphasize multi-agent forecasting, and several works have explored joint motion prediction~\cite{luo2023jfp, chen2022scept, yuan2021agentformer, ngiam2022scene}, the majority of approaches still default to marginal prediction due to its relative simplicity and ease of implementation.
    Looking ahead, future benchmarks should place greater emphasis on and offer stronger incentives for joint prediction, better aligning with the practical demands of real-world autonomous systems that must reason about the collective behavior of multiple agents.
    Prioritizing joint prediction will also encourage the development of new modeling paradigms that explicitly capture inter-agent dependencies, social dynamics, and scene-level feasibility—shifting the field toward more holistic and coordinated motion modeling. At the same time, this shift also brings light to several important challenges. Joint prediction models often require significantly more computation, especially under multimodal settings where the joint output space grows combinatorially with the number of agents and modes. This can lead to increased inference latency and scalability issues in real-time systems. Moreover, ensuring diversity, plausibility, and consistency across joint samples remains a non-trivial modeling and evaluation problem. Addressing these challenges calls for innovations in model architectures, sampling strategies, and benchmark design to make joint motion prediction both practical and effective for deployment.
    
    \vspace{-0.5em}
    \item \textbf{Ego-Conditioned Prediction} Existing benchmarks typically evaluate unconditioned prediction, where the predicted trajectories of other agents are generated independently of the ego vehicle's future planning motion. However, in the context of downstream planning, this assumption is limiting. In reality, the future motion of surrounding agents not only influences the ego vehicle's behavior, but is also influenced by the ego vehicle’s planned actions.
    This mutual dependency calls for ego-conditioned prediction, where the prediction model explicitly accounts for the ego vehicle’s intended plan. Ego-conditioned prediction allows the planner to simulate how different planning candidates may influence the behavior of other agents, thereby enabling plan-aware prediction and more informed decision-making. It plays a crucial role in evaluating and selecting the most favorable plan.
    
    \vspace{-0.5em}
    \item \textbf{Safety-Awareness} Traditional accuracy-based metrics evaluate predictions by measuring their similarity to ground-truth values, such as trajectory distances or bounding box overlaps, without considering the consequences of prediction errors in downstream decision-making. However, outputs with similar error magnitudes can have vastly different implications for safety and planning. For instance, an inaccurate prediction to an agent that veers away from the ego agent may have little to no impact on planning, while a similarly inaccurate prediction to an agent that crosses into the ego path could lead to a catastrophic outcome.
    This asymmetry highlights the need for planner-aware metrics that reflect how prediction quality impacts the ego agent’s motion plan and safety. In other words, in high-stakes applications like robotics and autonomous driving, not all actors are equally important since only a small subset of actors will directly impact the vehicle's planning and decision-making.
    Therefore, it is essential to explicitly distinguish between harmless and hazardous prediction errors and to penalize the latter more heavily. For example, prediction algorithms should produce more accurate and nuanced forecasts for objects that are close to the ego agent, compared to those that are far away or unlikely to interact.
    
    Several recent works have proposed augmentations to standard metrics that incorporate 1) planning sensitivity~\cite{ivanovic2021rethinking}, quantifying the extent to which prediction errors affect the ego agent’s chosen trajectory while keeping the metric general and not limited to specific planner, and 2) safety awareness~\cite{dauner2024navsim}, placing greater emphasis on resulting safety and comfort rather than relying solely on distance-based metrics such as ADE or FDE. These metrics help bridge the gap between offline evaluation and real-world utility, encouraging the development of models that are not only accurate but also robust and safe when deployed in interactive environments. However, such metrics are still underutilized in the literature and have yet to become standard in benchmark evaluations.

\end{itemize}

\subsubsection{Closed-loop Evaluation} \label{sec:CL_eval}
Most motion prediction methods are evaluated in an open-loop setting, where predicted trajectories are compared to ground-truth futures. While this setup is simple and widely used in benchmarks, it fails to capture how predictions affect the system in deployment. In real-world closed-loop scenarios, predictions influence downstream planning and control decisions, which in turn affect the subsequent perception and overall system behavior.
Recent studies have shown that open-loop prediction and planning metrics are often poor proxies for actual closed-loop performance~\cite{phongWhatTrulyMatters2023,dauner2024navsim,dauner2023parting}. 
To properly assess prediction performance in closed-loop, additional key considerations should be captured.

\begin{itemize}[label={\scriptsize$\bullet$}]
    \vspace{-0.5em}
    \item \textbf{Temporal Consistency} 
    Temporal inconsistency in motion prediction refers to abrupt changes in predicted trajectories over consecutive time steps, even when the input varies only slightly. This instability is not captured by standard open-loop evaluation and can cause significant issues in downstream planning and closed-loop settings. For instance, 1) fluctuating predictions may lead to frequent and distinct replanning, resulting in unstable ego behaviors such as unnecessary braking, sudden steering adjustments, or trajectory jittering—ultimately degrading ride comfort and safety. 2) Moreover, inconsistent predictions may cause the planner to misinterpret the intent of surrounding agents, leading to overly conservative or erratic decisions. 3) In interactive environments, the problem is compounded: inconsistent prediction leads to the ego vehicle’s inconsistent behavior, which can further confuse other agents, increasing other agents' hesitation or unsafe maneuvers, especially in dense or highly interactive scenarios. 4) In systems that perform joint learning of prediction with other modules, such inconsistencies can further propagate through the autonomy stack, undermining the overall robustness and reliability of the system. These challenges underscore the importance of evaluating temporal consistency, particularly in closed-loop settings where prediction quality directly impacts system-level performance.
    Considering metrics that quantify temporal consistency, such as frame-to-frame deviation and smoothness penalties, represents an important next step.

    \vspace{-0.5em}
    \item \textbf{Computation Latency} For motion prediction in real-world systems, such as autonomous driving or robotics, inference latency is a critical performance factor. Despite this, current benchmarks tend to focus almost exclusively on accuracy, often ignoring latency entirely in evaluation protocols. This oversight has several important consequences.
    \begin{itemize}
        \item \textit{Real Time Computation.} First, neglecting latency can lead to models that achieve strong benchmark scores but are impractical for deployment, particularly in real-time applications where predictions must be generated at high frequency and low latency. Many state-of-the-art methods are not well-suited for this, as discussed in ProphNet~\cite{wang2023prophnet}.
        Notably, one effective strategy to reduce latency is reusing historical embeddings through streaming encoders. For example, QCNet~\cite{zhou2023query} and DriveTransformer~\cite{jia2025drivetransformer} maintain a queue of past embeddings, reusing embeddings for seen elements and updating it incrementally as new frames arrive. This avoids redundant computation while preserving temporal context, making such models more suitable for deployment in latency-sensitive or resource-constrained environments.
        \item \textit{Unfair Comparison without Accounting for Model Size.} Second, the absence of latency constraints has encouraged an unbounded growth in model size. Larger and more complex architectures—featuring extensive fusion modules, multi-stage refinement~\cite{shi2024mtr++,shi2022motion,zhou2024smartrefine}, or multiple specialized heads—can improve accuracy but often at the cost of substantial increases in inference time. This leads to unfair comparisons, where gains in performance may stem primarily from increased computational budgets rather than algorithmic innovation, and risks overshadowing work that seeks to improve efficiency-performance trade-offs.
        \item \textit{Favouring Marginal Prediction.} Third, the absence of latency considerations in current benchmarks has inadvertently favoured marginal prediction methods, where the model predicts each agent’s future independently. While simple to implement and easy to scale in accuracy-focused benchmarks, marginal prediction introduces duplicated computation—since each agent is processed separately, often in its own coordinate frame—and lacks scene-level consistency. While batch computation can be performed for acceleration, additional computation is still needed. In contrast, joint prediction methods reason about all agents simultaneously, enabling globally consistent and interaction-aware predictions while sharing computation across the scene, making them inherently more efficient. However, without latency as part of the evaluation, these efficiency advantages of joint prediction methods are not captured or rewarded. As a result, research has skewed toward marginal approaches, despite their limitations in real-time deployment and multi-agent reasoning.

    \end{itemize}

    To address these issues, latency should be treated as another critical evaluation criterion. Benchmarks should introduce latency constraints or reporting requirements, and future work should focus on exploring the latency–performance frontier. This includes investigating scalable architectures that can adapt across hardware profiles, optimizing joint prediction pipelines, and rethinking designs that balance accuracy with deployability. Only through such efforts can we ensure that advances in prediction research translate into practical, deployable systems.

    \vspace{-0.5em}
    \item \textbf{Task-Oriented Metrics} 
    Beyond accuracy and safety, prediction quality should also be evaluated based on whether it enables successful ultimate task completion, such as a vehicle reaching its destination or a robot successfully assembling an object in the given time. This is conceptually related to the safety-aware evaluation discussed earlier, but places greater emphasis on task success as the ultimate objective.
    This helps align the evaluation of motion prediction with the performance of the overall system and can reveal issues that open-loop metrics might miss. 
    
    For example, a prediction model that is overly conservative or underconfident may generate a wide range of possible futures to ensure that at least one aligns with the ground truth. While this may yield favorable results under open-loop metrics like minADE, which consider only the best-matching trajectory, such predictions may paralyze the ego agent to be overconservative in practice, as no feasible path can be safely planned through overly uncertain environments.
    To avoid such behavior, an effective prediction model should produce confident, high-quality predictions that allow the ego agent to make progress efficiently, rather than being overly cautious. At the same time, the model should not be overly confident—underestimating uncertainty in highly interactive or ambiguous scenarios may lead to unsafe decisions and increased collision risk. 
    In this context, task-oriented evaluation is closely related to prediction calibration: only well-calibrated models that express uncertainty faithfully can support planners in making safe and efficient decisions under uncertainty.

    More broadly, task-oriented metrics may surface challenges that go beyond prediction calibration, especially in end-to-end systems where aligning intermediate outputs with ultimate system performance is critical.
    
\end{itemize}

\subsection{Perspectives}
\label{sec:applicable_perspectives}
In this section, we explored the integration of motion prediction into the closed-loop autonomy stack and deploying it to real-world applications, particularly focusing on autonomous robots operating in dynamic environments. 
The interplay between perception, prediction, and planning demonstrates that motion prediction cannot be treated in isolation. Instead, it requires holistic approaches that account for inter-module information sharing and compatibility, the cascading effects of errors, and alignment with system-level performance. Despite significant advancements in applying motion prediction to real-world robotics applications, several challenges persist that present opportunities for future research.

\textbf{Real-World System-Level Evaluation and Benchmarking} One of the most pressing challenges in motion prediction is bridging the gap between idealized open-loop evaluations and realistic closed-loop settings. 
During \textit{perception–prediction integration}, as discussed in Section~\ref{sec:deployable_perception_dealing_with_uncertainty} and Section~\ref{sec: app - Perception-Aware Prediction Evaluation}, the propagation of upstream uncertainties—such as tracking and detection errors—must be explicitly accounted for in motion prediction evaluation to avoid cascading degradation and to promote robustness against noisy perception inputs. To fully capture the tight coupling between perception and prediction, where the learning of one module influences the other and prediction is conditioned on detected objects, it is essential to design evaluation metrics that jointly assess both modules.
In \textit{prediction–planning integration}, as discussed in Section~\ref{sec: prediction uncertainty} and Section~\ref{sec: app - Planning-Aware Prediction Evaluation}, evaluation must reflect downstream implications to ensure that benchmark performance meaningfully translates into practical utility for planners. This includes assessing the model’s ability to quantify and calibrate prediction uncertainty to enable confidence-aware decision-making by offering actionable and confident outputs that reduce conservativeness while maintaining safety. It also calls for evaluation protocols to place greater emphasis on joint prediction, ego-conditioned prediction, policy planning, and safety-aware prediction, better aligning the evaluation criteria with how motion prediction outputs are actually consumed by downstream planners.
For \textit{closed-loop evaluation}, as discussed in Section~\ref{sec:CL_eval}, several factors must be prioritized to better align prediction evaluation with the performance of the full autonomy stack. These include temporal consistency, which encourages stable planning and coherent agent interactions; computation latency, which promotes real-time applicability and incentivizes efficient innovation; and task-oriented metrics, which directly measure ultimate task performance.

To this end, existing benchmarks tend to fall into two extremes: (1) standalone modular motion prediction benchmarks, such as Argoverse~\cite{Argoverse} and Waymo~\cite{sun2020scalability}, which fully isolate prediction evaluation from other modules; and (2) end-to-end benchmarks, such as nuPlan~\cite{caesar2021nuplan} and CARLA~\cite{dosovitskiy2017carla}, which prioritize assessing overall driving performance—often overlooking the evaluation of motion prediction entirely.
A middle ground between these two paradigms remains elusive. What is clearly needed is a benchmark that enables rigorous evaluation of motion prediction models under realistic conditions, capturing the complexities of real-world deployment discussed above rather than relying on idealized settings. Such a framework should also support integration with diverse perception and planning modules, allowing for a comprehensive assessment of prediction robustness, generalization, and practical utility across system configurations. 
Achieving these goals will require building new datasets and evaluation metrics. But such efforts are well justified and have the potential to recalibrate the community’s focus toward more practical and deployable solutions.

\textbf{End-to-End Systems} The integration of perception, prediction, and planning presents an opportunity to develop more unified frameworks. 
Promising strategies, such as joint and end-to-end learning that allows information propagation and feature sharing across tasks
offer promising pathways to improve overall system coherence and performance. 
This trend toward end-to-end systems mirrors the shift seen in computer vision, where models like AlexNet revolutionized image classification in 2012 by eliminating hand-crafted features in favor of fully data-driven learning. A similar shift is now emerging in autonomous systems, where approaches that depart from the traditional autonomy task decomposition (i.e., perception, prediction, and planning) are showing promising results—Tesla’s ability to drive on public roads being a notable example.

However, an interesting observation is that even in well-developed fields like image classification, such as face recognition systems, errors persist, often requiring multiple attempts to succeed. This highlights that end-to-end autonomy, which is significantly more complex and safety-critical, still offers a vast space for exploration in terms of performance, reliability, interpretability, and robustness.
To this end, multiple key challenges remain, and considerable effort has been devoted to areas such as designing representations that are both informative and efficient (Section~\ref{sec: applicable perception-prediction representation}, Section~\ref{sec: app - prediction-planning representation}), optimizing multi-task objectives (Section~\ref{sec: app - prediction-informed perception}, Section~\ref{sec: app - planning-informed prediction}), ensuring the robustness of individual modules, and balancing computational trade-offs.
As discussed in Section~\ref{sec: app - Motion Prediction in End-to-End Modular Methods}, the community has also explored various integration paradigms, including different forms of task coupling—such as joint perception–prediction, joint prediction–planning, and fully integrated perception–prediction–planning—as well as diverse connectivity structures, including sequential, parallel, and hybrid configurations.
These back-and-forth explorations across system design, learning strategies, and modular integration are expected to give rise to the next generation of stronger, more capable autonomous models, with immense potential ahead.

\textbf{Large Models}
We are in the era of foundation models—large-scale neural networks trained on massive and diverse datasets—that exhibit strong generalization and emerging capabilities across a wide range of tasks. This trend not only applies to modular approaches, but are also extended to end-to-end systems.

\begin{itemize}[label={\scriptsize$\bullet$}]
    \vspace{-0.5em}
    \item Within the \textit{modular} autonomy stack, there exists a significant gap between vision and motion data. Compared to motion data, visual data is available at a vastly larger scale, often at internet scale, making it well-suited for pretraining large perception models. In contrast, collecting diverse, high-quality motion data for prediction and planning is costly, logistically difficult, and limited in coverage. To take advantage of this imbalance, as discussed at the end of Section~\ref{sec: app - planning-informed prediction}, some approaches decouple perception from decision-making, treating it as a standalone module. This allows direct use of large-scale pre-trained vision backbones that are highly robust and transferable. 

    Developing large models for motion prediction places a premium on data collection, which is especially challenging and expensive given that motion data is far less abundant than image or text data. For real-world deployment, capturing rare yet safety-critical scenarios, such as near-collisions, complex interactions, or edge-case behaviors, is essential for building robust and generalizable models.
    In the following Chapter~\ref{sec:generalizable}, we delve into strategies for developing generalizable motion prediction models, covering a range of approaches including data design, learning signals, and model architectures, as well as considerations across different stages of model development.
    
    \vspace{-0.5em}
    \item The use of large pre-trained models is also being actively explored in \textit{end-to-end} autonomous systems, which offer the promise of more streamlined and fully data-driven pipelines. Recent advances in vision-language models (VLMs) and video foundation models are opening up new possibilities for tighter integration across the autonomy stack. As discussed at the end of Section~\ref{sec: app - Motion Prediction in End-to-End Modular Methods}, several emerging approaches aim to adapt these foundation models to autonomous driving tasks, leveraging their ability to process complex temporal and multi-modal information. By unifying traditionally separate modules, these end-to-end systems seek to simplify interfaces, reduce reliance on hand-engineered components, and improve scalability and adaptability across diverse environments.
    
\end{itemize}

While these large pre-trained models offer powerful capabilities, deploying them in autonomous systems requires careful consideration of real-time constraints and computational resources.
To improve efficiency, pruning techniques remove redundant parameters to reduce computational cost without significant accuracy loss. Lightweight architectures—such as efficient CNNs and real-time attention mechanisms—further help balance performance and latency for deployment in dynamic environments.
Model distillation and compact neural designs also enable large models to retain key capabilities with reduced runtime demands. These strategies collectively help meet real-world requirements by optimizing the trade-off between accuracy and efficiency.

\textbf{Outlook} Motion prediction lies at the heart of autonomous systems, serving as a critical bridge between perception and decision-making. Advancing this field, through improvements in efficiency, robustness, and seamless integration, will be key to unlocking the full potential of autonomy. As we confront and overcome these challenges, motion prediction will evolve from a standalone component into a foundation for reliable, adaptable, and intelligent behavior in complex environments. These efforts will not only enable safer and more efficient deployment across diverse domains but also help shape the next generation of autonomous technologies that can learn, interact, and thrive in the real world.

%% file: section/4_generalization.tex
\newpage
\section{Generalizable Motion Prediction}
\label{sec:generalizable}






\begin{figure}[h!]
    \centering
    \includegraphics[width=1\textwidth]
    {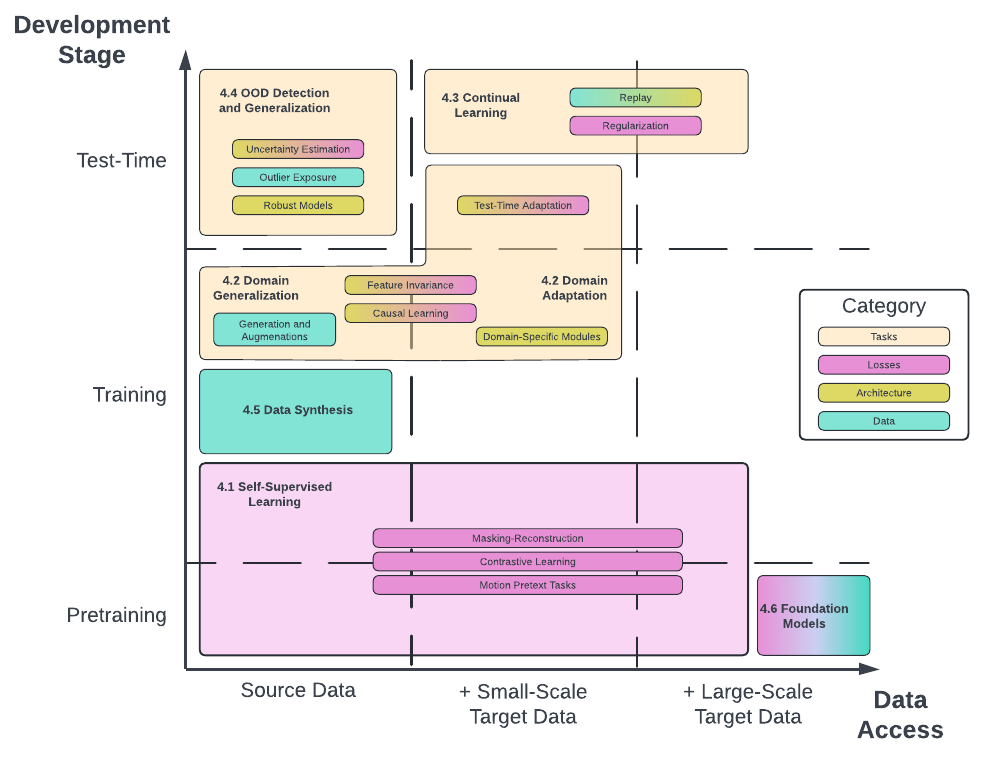}
    \caption{An overview of approaches for generalizable motion prediction covered in Chapter~\ref{sec:generalizable}. We organize them according to how they differ in terms of 1) data access assumptions as in the x-axis: access to only source data and no target data, access to small-scale target data that is yet smaller than source data such as few-shot or streaming data, and access to large-scale target data that is equal to or larger than source data;
    2) their development stage in the y-axis: the approach is applied during pretraining, training, or test time;
    and 3) the primary focus in the legend—whether they target different generalization tasks tailored to particular use cases, learn generalizable representations through modified or additional learning signal, modify the model architecture or add additional parameters for more robustness and adaptability, or achieve generalization by scaling up the data.
    }
    \label{fig:generalization}
\end{figure}

When autonomous systems, such as robots and self-driving cars, are deployed in the real world, they will encounter diverse scenarios varying in many factors such as different environmental geometries, new types of agents, unexpected events (weather, accidents) and adversarial attacks. In such contexts, the models can often be exposed to data that is outside the model's training distribution, where the model can have a significant drop in the performance or even fail to generate reasonable results. 
A straight-forward solution is to extend the training dataset to cover a wide range of operating scenarios through extensive data collection and labeling. While data collection has proven highly effective in fields like computer vision and natural language processing, where large-scale datasets can be scraped relatively easily, collecting and annotating data for robots and self-driving cars remains costly, inefficient, and often unable to capture the full diversity of possible events. This challenge highlights the need for efficient data collection pipelines along with other complementary approaches to enhance generalization and fast adaptation—just as humans \textit{generalize} from past experiences and \textit{quickly} adapt to new environments with minimal examples, without forgetting previously acquired skills.
Ultimately, robust generalization hinges on the seamless interplay between data, learning signals, and model architectures.

As discussed in Chapter~\ref{sec: intro} and illustrated in Figure~\ref{fig:vision}, in this paper, we introduce a generalization lifecycle that consists of four key stages to ultimately obtain models that generalize and adapt across distribution shifts: 
1) \textit{Data engine} that aims to absorb as much diverse data as possible to broaden the support of the data distribution. This includes methods such as data augmentation, data generation, and simulation; 
2) \textit{Training universal models} that learns transferable representations and adopts robust model architectures to harness strong zero-shot performance in a wide range of test settings, and methods like self-supervised learning, and domain generalization/adaptation can all be useful; 
3) \textit{Safety-aware deployment} that applies the model in real-world settings with awareness of prediction confidence and system-level safety, including methods such as OOD detection/generalization that identify domain shifts, assess their severity, and trigger fallback strategies when necessary to ensure safe operation; 
4) \textit{Test-time adaptation} that adapts the model online to anticipate and respond to episodic distribution shifts and optionally also keep the performance in the
original domain, or to actively remain within in-distribution operation regimes.
At the end of each generalization cycle, with a quantifiable measure of domain shift, the system can selectively store and annotate detected out-of-distribution (OOD) data to expand the diversity and coverage of the training distribution. The enriched dataset then triggers the next iteration of the cycle, forming a lifelong learning process that continuously improves the model’s ability to generalize across evolving real-world scenarios.

To achieve such goals, various approaches in the literature have been proposed to advance the generalization capabilities of predictive models, addressing different aspects such as data, learning signals, and model architectures, as illustrated in Figure~\ref{fig:generalization}. 
This section discusses different settings of the generalization problem (e.g. varying in target task, use case, and access to target domain data), and various methods that are used to improve generalizability:

\begin{itemize}[label={\scriptsize$\bullet$}]
    \vspace{-0.5em}
    \item 
\textbf{Self-Supervised Learning}
Self-supervised (SSL) methods use proxy tasks, such as trajectory reconstruction and contrastive learning, to learn informative and transferable representations in addition to the original prediction supervision. 
This can either be done sequentially, where SSL methods are used as a pretraining step, or simultaneously in multi-task learning, where the model is optimized for both SSL and original supervised losses. In addition to the trajectory representation, SSL methods are also applied to other representations such as point cloud forecasting, video generation, and scene reconstruction.
    \vspace{-0.5em}
    
    \item 
\textbf{Domain Generalization and Adaptation}
\textit{Domain generalization} is the most straightforward distribution shift setting and considers the case where a model trained one or more source datasets is tested on one or more different target datasets in zero-shot. This corresponds to dropping the source pre-trained model in a new environment. A common training setting involves multiple source datasets, and models, that learn generalizable features and are robust to distribution shits across the source domains, can perform well on the target domain. In contrast to domain generalization, \textit{domain adaptation} considers the case where some target data is also available in addition to the source data. The target data is often available only in a limited fashion e.g. only a handful of examples in the form of offline batch data or online streaming data, or without prediction labels. This allows exploiting the target data, and the key challenge is to effectively leverage the target information to maximize performance without overfitting.
    \vspace{-0.5em}

\item 
\textbf{Continual Learning}
Continual learning extends domain adaptation to a sequence of new datasets or to a continually changing distribution instead of a single shift. Moreover, compared to domain adaptation that ignores performance in the source domain, continual learning methods also evaluate and aim to keep the performance on all previously seen domains, where the challenge is to avoid forgetting past experiences as additional ones are learned.
    \vspace{-0.5em}

\item 
\textbf{Out-of-Distribution Detection and Generalization}
A slightly different test case considers the presence of rare, difficult, or adversarial examples in the test set. These are the out-of-distribution (OOD) examples. The assumption is that OOD data is so far from standard inlier data that a learned model would not generalize properly as in domain generalization, and so we consider two strategies. The first is OOD detection, where the objective is to identify the OOD data and reject it instead of using it as a prediction input. The second is OOD generalization, which still attempts to generate outputs from the OOD input data, but often requires additional assumptions or models. In practice, this can mean training with representative outliers or using additional unlearned models as a fall-back strategy to supplement the learned model when faced with OOD inputs. This can be done in conjunction with OOD detection by routing detected OOD inputs to the robust model. 
    \vspace{-0.5em}

\item 
\textbf{Data Augmentation and Synthesis}
A promising direction to improve model generalization is to enhance existing datasets through data augmentation and synthesis.
Data augmentation aims to increase data diversity by applying transformations or perturbations to existing examples, thereby improving the model's robustness and generalization ability.
In motion prediction, augmentation techniques include perturbing agent trajectories, altering scene contexts, modifying initial conditions, or simulating alternative agent intentions.
Beyond augmentation, data synthesis further expands dataset diversity by generating new examples that capture a broader range of scenarios. Synthesis can introduce challenging and rare situations—such as collisions or adversarial agents—that are under-presented in the original datasets but are crucial for real-world deployment, particularly in new or unseen test settings.
Unlike image-based tasks, motion prediction data is subject to significant structural constraints, such as agent interactions, initial configurations, environment layout, and temporal dynamics. Generated motions must remain physically and behaviorally plausible, respecting map constraints and interaction dynamics over time. As a result, a key challenge in both data augmentation and synthesis is ensuring realism—that is, generating or modifying data in ways that are consistent with plausible human or agent behavior. 

    \vspace{-0.5em}

\item 
\textbf{Foundation Models:}
Training on massive datasets is shown to greatly improved models and exhibit emerging capabilities in natural language processing and computer vision, and this will probably also be the case in motion prediction. However, current academic efforts are limited by the size of existing public motion datasets and the fact that these datasets often possess discrepant formats (e.g. input and output sequence lengths), making it non-trivial to effectively leverage multiple datasets simultaneously. Meanwhile, in the indutry where much more data is available, it remains an open problem regarding how to quantify the motion data distribution, measure data quality and diversity, and improve data balance, which are all critical in efficiently scaling up training data. Alternatively, instead of directly developing motion foundation models, adapting existing foundation models (e.g. leveraging the reasoning capabilities of large language models) to the motion prediction domain is another promising approach to explore.
    
\end{itemize}


\begin{figure}[t]
    \centering
    \includegraphics[width=0.7\textwidth]{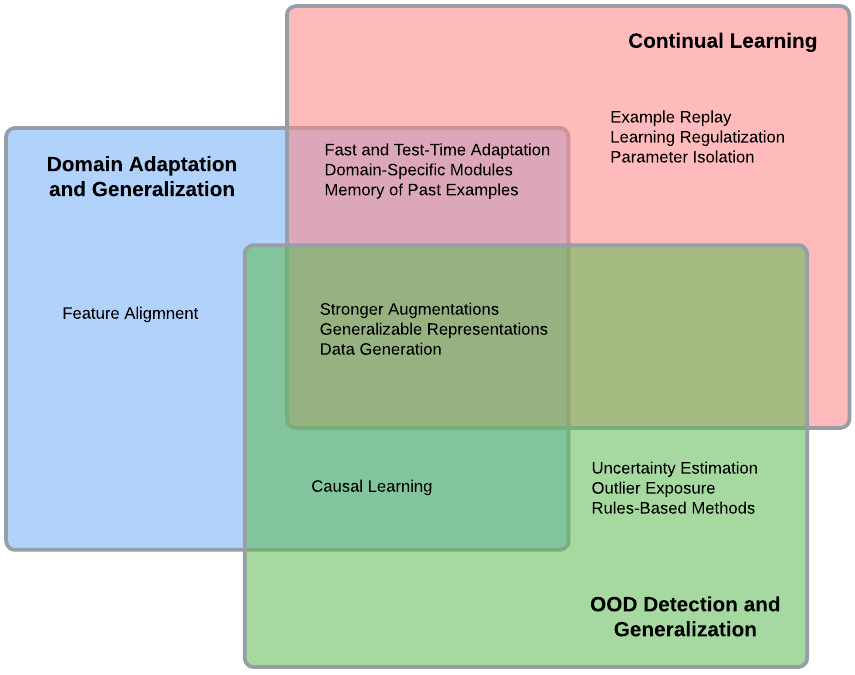}
    \caption{While different generalization tasks—such as domain adaptation, domain generalization, continual learning, and out-of-distribution (OOD) detection and generalization—involve distinct settings, assumptions, and use cases, there is considerable overlap in their underlying methodologies.}
    \label{fig:methods_gen}
\end{figure}

As illustrated in Fig.~\ref{fig:generalization}, these approaches differ in terms of data access assumptions (e.g., source-only, limited target data, or large-scale target data), development stage (pretraining, training, or test-time), and primary focus—whether they target different generalization scenarios tailored to particular use cases, or achieve generalization by leveraging different perspectives, such as data, loss functions, or model architectures.
In this section, we will proceed to discuss about details of each setting and method. First, we present \textit{self-supervised learning} methods that leverage proxy tasks to learn informative and generalizable representations in Section \ref{sec:gen_pred_ssl}. 
The following three sections discuss three categories of settings/tasks for generalizability and their associated approaches in the literature. 
First, we discuss \textit{domain adaptation and generalization} in Section~\ref{sec:domain-adaptation-and-generalization}, which focuses on transferring from a source domain to a new target domain and solely focus on the performance on the target domain. Next, we introduce \textit{continual learning} in Section \ref{sec:continual-learning}, which considers adapting to a sequence of new domains while maintaining performance on previous domains. After that, we cover \textit{out-of-distribution detection and generalization} in Section \ref{sec:OOD_det_gen}, considering how to identify and deal with rare and unexpected samples that are out of the training distribution. These three tasks have significant overlap in their methodologies, and Figure \ref{fig:methods_gen} presents some common approaches and how they are shared between tasks. Following this, we summarize current \textit{data augmentation and synthesis} approaches that can be used to obtain rare or adversarial data in Section \ref{sec:data_synthesis}. Finally, we present a discussion on the use of \textit{foundational models} in motion prediction in Section \ref{sec:foundation_models}, which have shown incredible generalization and open-world performance in other communities and have the potential to do the same for motion prediction.

\subsection{Self-Supervised Learning} \label{sec:gen_pred_ssl}
Deep learning models have achieved significant success in trajectory forecasting benchmarks through data-driven supervised learning approaches. However, the process of collecting and annotating trajectory data remains highly challenging and resource-intensive. Typically, trajectory data is gathered using self-driving vehicles equipped with advanced sensor systems. Subsequently, annotators are required to label detected objects, associate their positions across frames, and generate trajectories, which often necessitate further refinement and smoothing to ensure accuracy. Reliance on annotations can limit the model's generalizability and make the training difficult to scale large datasets. On the other hand, the NLP and CV communities have developed remarkable advances in self-supervised learning (SSL). As evidenced by the success of models such as BERT~\citep{devlin2019bertpretrainingdeepbidirectional} and Masked Autoencoders~\citep{he2022masked}, SSL is demonstrated to acquire expressive representations/features by pre-training on unlabelled data with well-defined pretext tasks, leading to enhanced downstream performance after fine-tuning on the desired task.  Recognizing the importance of self-supervised representation learning, recent works are exploring SSL techniques to improve the performance of motion prediction. Generally, the success of SSL methods depends on designing SSL pretext tasks that provide additional signals to learn rich and generalizable features that effectively align with the downstream prediction task.   

As discussed in Section~\ref{sec:input-output-representation}, various representations are employed in motion prediction tasks. One of the most common approaches involves using trajectories to represent motion. In this representation, unlike computer vision (CV) or natural language processing (NLP), where SSL can be directly applied to raw input data without requiring annotations, motion trajectories require annotation to determine agent coordinates. Consequently, the primary goal of SSL in trajectory prediction is to learn more effective feature representations, rather than to eliminate the need for annotations entirely. 
Conversely, emerging object-free tasks, such as point cloud forecasting~\cite{khurana2023point} and video generation~\cite{hu2023gaia}, leverage raw sensor data representations like RGB images and LiDAR point clouds directly. These approaches offer innovative methods for motion prediction by directly modeling the temporal evolution of the world through point clouds and images. In this section, we explore SSL techniques applied to the representations of trajectories in Section~\ref{sec: SSL - traj}, point clouds in Section~\ref{sec: SSL - point cloud}, and images in Section~\ref{sec: SSL - video}.

\subsubsection{Self-Supervised Learning for Trajectory Prediction}
\label{sec: SSL - traj}

\begin{table}[!htb]
\centering
\caption{Taxonomy of self-supervised-learning (SSL) approaches for motion prediction. Abbreviations used are: MR: Masking-Reconstruction, CL: Contrastive Learning, RR: Redundancy Reduction, AT: Auxiliary Tasks, ST: Pretraining on Synthetic Trajectories, Transf: Transformer, GNN: Graph Neural Network, RNN: Recurrant Neural Network, Map Repr.: Map Representation, RM: Rasterized Map, Vec: Vector Representation, Frenet: Frenet Coordinate}
\label{tab:ssl}
\resizebox{1.0\linewidth}{!}{
\begin{tabular}{@{}ll|ccccc|cc|ccc|ccc}
\toprule\toprule
\multirow{2}{*}{Method} & \multirow{2}{*}{Venue} & \multicolumn{5}{c|}{SSL Task} & \multicolumn{2}{c|}{Training Strategy} & \multicolumn{3}{c|}{Architecture} & \multicolumn{3}{c}{Map Repr.}\\ \cmidrule(l){3-15} 
 & & MR & CL & RR & AT & ST & \begin{tabular}{c}Pretrain\\Finetune\end{tabular} & \begin{tabular}{c}Multi-Task\\Learning\end{tabular} & Transf. & GNN & RNN & RM & Vec. & Frenet \\ \midrule
Forecast-MAE \cite{forecast_mae}  & ICCV'23 & \checkmark & & & & & \checkmark & & \checkmark & & & & \checkmark & \\
Traj-MAE \cite{traj_mae}  & ICCV'23 & \checkmark & & & & & \checkmark & & \checkmark & & & & \checkmark & \\
SEPT \cite{sept}  & ICLR'24 & \checkmark & & & & & \checkmark & & \checkmark & &  & & \checkmark & \\
SmartPretrain~\cite{zhou2024smartpretrain}  & ICLR'25 & \checkmark & \checkmark & & & & \checkmark & & \checkmark & & & & \checkmark & \\
Pre-TraM \cite{pretram} & ECCV'22 & & \checkmark & & & & \checkmark & & \checkmark & & \checkmark & \checkmark & & \\
\citet{ma2021multi} & ITSC'21 & & \checkmark & & & & \checkmark & & & \checkmark & \checkmark & & & \checkmark \\
Pre-CLN \cite{precln} & WWW'23 & \checkmark & \checkmark & & \checkmark & & & \checkmark & \checkmark &  &  &  & \checkmark & \\
SSL-Lanes \cite{ssllanes} & CoRL'22 & \checkmark & & & \checkmark & & & \checkmark & & \checkmark  &  & & \checkmark & \\
Social-SSL \cite{socialssl} & ECCV'22 & \checkmark & & & \checkmark & & \checkmark & & \checkmark & & & & & \\
ReLe-GCN~\cite{SSPNFMR} & T-ETCI'23 & & \checkmark & & \checkmark & & \checkmark & & & \checkmark & & & & \\
\citet{li2023pedestrian} & ICRA'23 & & & & \checkmark & & & \checkmark & & \checkmark & & & \checkmark & \\
RedMotion \cite{redmotion} & TMLR'24 & & & \checkmark & & & \checkmark & & \checkmark & &  & & \checkmark & \\
\citet{exploiting_map_info} & Arxiv & & & & & \checkmark & \checkmark & \checkmark & & \checkmark & & & \checkmark & \\
\citet{li2023pre} & IROS'24 & \checkmark & & & & \checkmark & \checkmark & & & \checkmark & & & \checkmark & \\

\bottomrule\bottomrule
\end{tabular}
}
\end{table}




Existing SSL approaches for trajectory prediction have mainly adopted two different training paradigms: 1) \textbf{Multi-task learning}: where the self-supervised pretext tasks are jointly optimized with the supervised motion prediction task. 2) \textbf{Pretraining and fine-tuning}: where the feature encoder is first pretrained on self-supervised pretext tasks and then fine-tuned on the supervised downstream task.
One of the main reasons behind the success of SSL in computer vision and natural language processing is the availability of large-scale unlabeled datasets.
In contrast, the scarcity of large-scale motion prediction datasets poses a significant challenge to replicating this success.
To this end, in motion prediction, researchers have focused on maximizing the utility of existing data by designing pretext tasks that capture trajectory dynamics, map topology, and agent–agent as well as agent–map interactions.
In parallel, other efforts have aimed to increase the number and diversity of training samples through data augmentation or synthetic data generation.
We further categorize SSL techniques into 5 main categories based on the type of pretext task leveraged: 1) \textbf{Masking-Reconstruction (MR)}, 2) \textbf{Contrastive Learning (CL)}, 3) \textbf{Redundancy Reduction (RR)}, 4) \textbf{Auxiliary Motion-Related Tasks (AT)}, 5) \textbf{Pretraining on Synthetic Trajectories (ST)}. Some works combine multiple pretext tasks together. Table \ref{tab:ssl} shows a taxonomy of papers related to SSL in trajectory prediction. 
In the following sections, we will briefly introduce each of the pretext tasks and discuss their related works in detail.

\begin{figure}[t]
    \centering
    \includegraphics[width=0.7\textwidth]{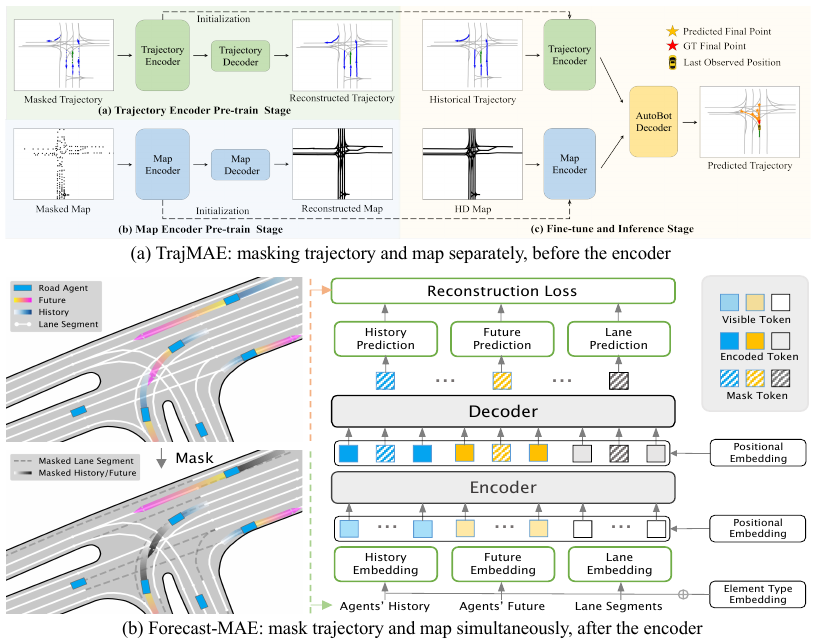}
    \caption{Examples of mask-reconstruction self-supervised-learning for motion prediction: (a) TrajMAE~\cite{traj_mae} masks trajectory and map before the encoder, using individual encoders for trajectory and map respectively. (b) Forecast-MAE~\cite{forecast_mae} masks trajectory and map after the encoder, using a shared encoder. See more examples and discussions in Section~\ref{sec: generalization SSL mae}.}
    \label{fig:mae}
\end{figure}
\paragraph{Masking-Reconstruction}
\label{sec: generalization SSL mae}
A common SSL approach is to employ a masking-reconstruction task, to learn informative and transferable features.
This strategy has proven successful in natural language processing (NLP) by masking and reconstructing text tokens~\cite{devlin2019bertpretrainingdeepbidirectional}, and in computer vision (CV) with image patches~\cite{MAE}.
In motion prediction, it is typically instantiated by masking and reconstructing portions of agent trajectories and/or map features.
Specifically, this pretext task involves an encoder-decoder architecture. 
During pretraining, segments of maps or trajectories can be masked either before the encoder at the input level~\cite{traj_mae}, or after the encoder at the token level~\cite{forecast_mae}.
In the former case, as in Traj-MAE~\cite{traj_mae}, parts of the input data are masked out, and the encoder processes only the remaining visible segments. In the latter case, as in Forecast-MAE~\cite{forecast_mae}, the full input is first encoded, and masking is applied to the encoded features.
In both cases, a small decoder is used to reconstruct the original input during pretraining. After pretraining, the decoder is discarded, and the encoder is fine-tuned with a new decoder for the downstream motion prediction task.



Beyond the two main masking paradigms, a series of finer design choices have been proposed to further enhance the effectiveness of masking-reconstruction pretraining.
For example, Traj-MAE \cite{traj_mae} pretrains the map encoders and trajectory encoders separately via masking/reconstruction, which learns better features for trajectory and maps, but could fall short in encoding relationships between agents and maps during pretraining. 
Recognizing this gap, Forecast-MAE \cite{forecast_mae} utilizes a transformer encoder that takes both agent trajectories and maps as inputs simultaneously, and applies masking/reconstruction pretraining to them for better agent-map reasoning. Besides, Forecast-MAE demonstrates that additionally incorporating future trajectories as input during masking/reconstruction leads to improved performance, and that masking ratio has been crucial for learning good features during pre-training.
Pushing forward, SEPT~\cite{sept} applies masking at a finer granularity. Instead of masking an entire trajectory or map vector, it masks individual trajectory waypoints and short segments of map vectors. This captures more nuanced motion patterns, and outperforms Traj-MAE on the Argoverse dataset and Forecast-MAE on the Argoverse 2 dataset. 
SmartPretrain~\cite{zhou2024smartpretrain} further introduces temporal randomness in masking to encourage the learning of more informative features.
Specifically, given a complete trajectory, SmartPretrain randomly selects a temporal segment as the input and tasks the model with reconstructing the remaining waypoints in the full trajectory.
This strategy compels the model to reason over temporal continuity and motion dynamics based on partial observations, leading to richer and more generalizable spatiotemporal representations.

\begin{figure}[t]
    \centering
    \includegraphics[width=0.9\textwidth]{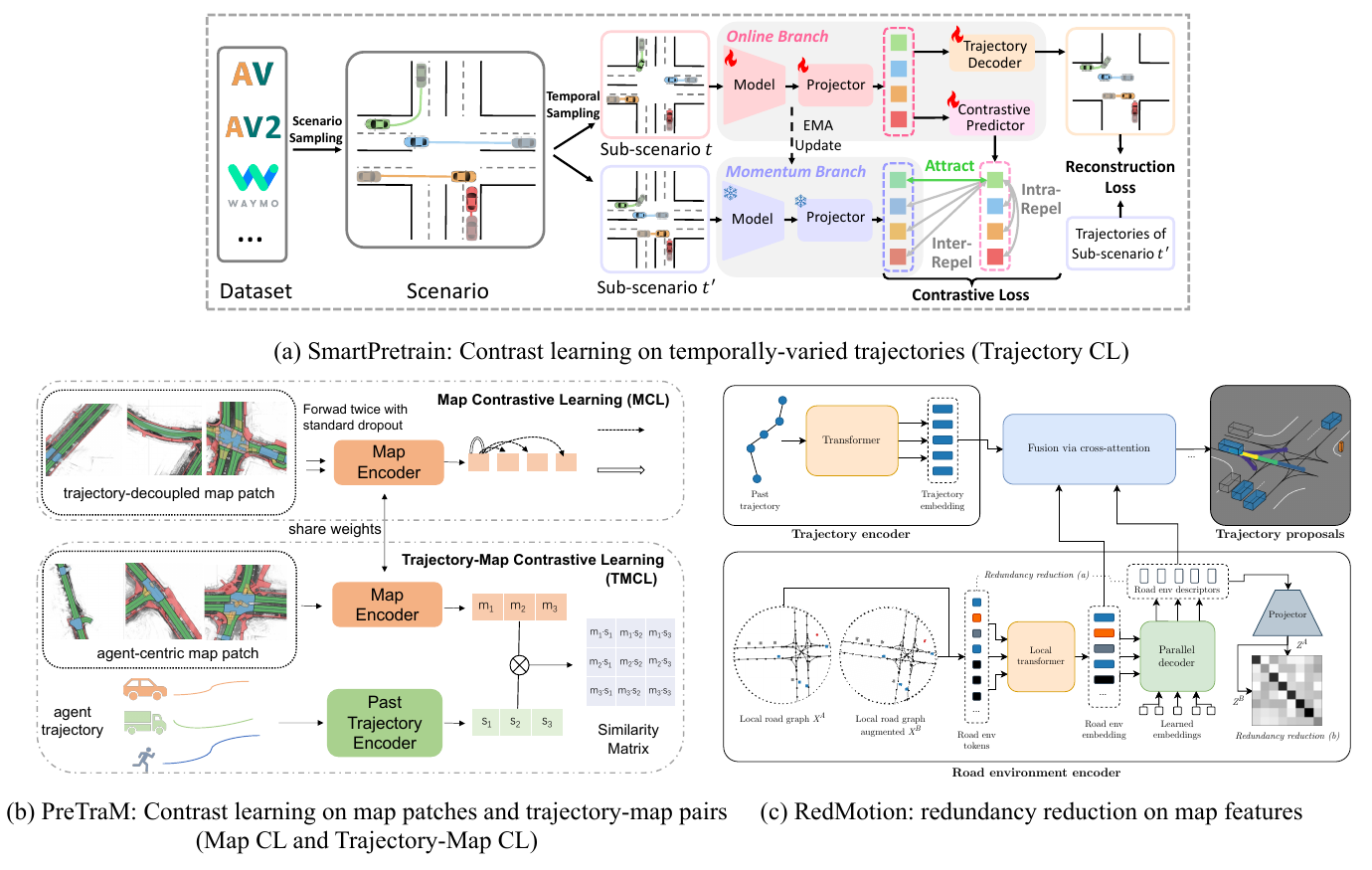}
    \caption{Examples of contrastive learning (CL) for motion prediction: (b) SmarPretrain~\cite{zhou2024smartpretrain} conducts contrastive learning on temporally sampled trajectory segments. (b) PreTraM~\cite{pretram} performs contrastive learning on different local map patches and trajectory-map pairs. (c) RedMotion~\cite{redmotion} exploit redundancy reduction on road features. See more discussions in Section~\ref{sec: generalization SSL contrast leanring}.}
    \label{fig:contrast learning}
\end{figure}
\paragraph{Contrastive Learning}
\label{sec: generalization SSL contrast leanring}
\textbf{Classic Contrastive Learning} Contrastive learning (CL) has emerged as a dominant technique for SSL because of its simplicity and effectiveness \cite{simclr, moco}. 
At its core, CL learns augmentation-invariant and distinguishable representations by contrasting positive and negative pairs within a shared feature space. This is achieved by pulling representations of similar data points (positive pairs) closer while pushing dissimilar ones (negative pairs) apart.
A positive pair, consisting of two semantically close inputs, can be generated either by applying various augmentations to the same input sample or by representing the same input through different modalities~\cite{mahmoud2023self}.
All other inputs form a negative pair with the query input. 
The success of CL critically relies on the correct identification and grouping of positive/negative samples.
For example, if semantically similar samples are mistakenly treated as negatives (i.e., false negatives), the model cannot reliably distinguish true negatives from false negatives, leading to dissimilar embeddings for semantically related inputs.
Such misclassifications, or more generally, the presence of noise in positive/negative pair assignments, can hinder the learning process and result in inconsistent or misaligned feature representations, ultimately degrading model performance.

Recent CL-based works in the trajectory prediction domain, with some of them illustrated in Figure~\ref{fig:contrast learning}, can be categorized into three directions:
(1) \textit{Trajectory CL}, which generates different views of trajectories, via sampling from full trajectories into temporally varied segments~\cite{zhou2024smartpretrain} or categorizing trajectories into different classes~\cite{SSPNFMR};
(2) \textit{Map CL}, which generates different views of maps either by cropping different regions of the global map~\cite{pretram} or by applying rotational perturbations to local maps~\cite{ma2021multi};
(3) \textit{Trajectory-map CL}, which forms positive pairs by associating trajectories with their corresponding local maps~\cite{pretram}.
Building on these paradigms, CL is often combined with other SSL tasks to further enhance representation quality.
For example, SmartPretrain~\cite{zhou2024smartpretrain} and Pre-CLN~\cite{precln} jointly train CL objectives alongside other pretext tasks, such as trajectory prediction and masking-reconstruction.
Similarly, ReLe-GCN~\cite{SSPNFMR} proposes a CL-based framework for human motion prediction, where positive and negative pairs are sampled based on motion categories (e.g., dancing, walking), and the CL objective is jointly optimized with a motion reconstruction task that denoises corrupted input trajectories.
Additionally, SmartPretrain~\cite{zhou2024smartpretrain} explores CL pretraining on multiple large-scale datasets for enhanced generalization and feature learning.
\paragraph{Redundancy Reduction}

\textbf{Redundancy Reduction} A closely related concept to CL is redundancy reduction~\cite{barlowtwins}, which is introduced in the SSL literature to promote feature diversity and avoid learning trivial constant features. 
Specifically, the Barlow Twins objective function \cite{barlowtwins} is introduced to measure the cross-correlation matrix between the embeddings of two identical networks fed with distorted versions of the same sample and tries to make this matrix close to the identity. By trying to equate the diagonal elements of the cross-correlation matrix to 1, the embeddings are made to be invariant to the distortions applied. Whereas by trying to reduce the off-diagonal elements of the cross-correlation matrix to 0, decorrelation is enforced on the different elements of the embedding vector. Hence, each component of the embedding vector is forced to learn a different feature. 
Inspired from this, RedMotion \cite{redmotion} proposes to learn augmentation-invariant features for road environments (road graphs and agents) via the Barlow Twins objective function. 

\begin{figure}[t]
    \centering
    \includegraphics[width=0.9\textwidth]{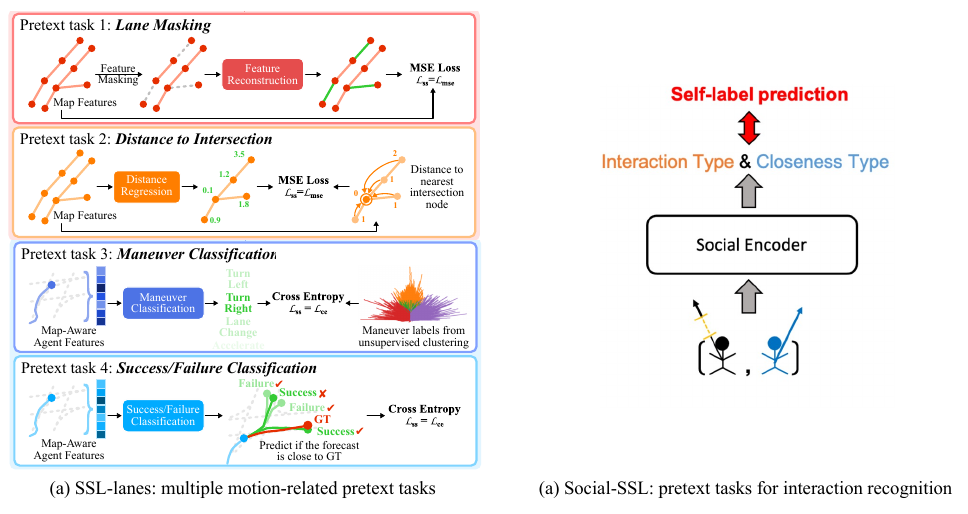}
    \caption{Examples of motion-related pretext for self-supervised learning in motion prediction: (a) SSL-lanes~\cite{ssllanes} introduce four pretexts to train together with the downstream prediction task. (b) Social-SSL~\cite{socialssl} design pretext tasks to recognize interaction types among multiple agents. See more examples and discussion in Section~\ref{sec: generalizable DA/DG auxiliary task}.}
    \label{fig:pretext task}
\end{figure}

\paragraph{Auxiliary Motion-Related Tasks}
\label{sec: generalizable DA/DG auxiliary task}
\textbf{Auxiliary Motion-Related Tasks} SSL-lanes \cite{ssllanes} incorporate multiple novel self-supervised auxiliary tasks, to capture additional map/agent-level information and improves the motion prediction performance. These auxiliary tasks include predicting distance to the intersection, maneuver classification, and classification of the prediction success/failure. Each pretext task, however, is individually optimized, and the authors did not explore training the model with all tasks combined. 
Social-SSL \cite{socialssl} instead pretrains a human trajectory prediction model together with multiple pretext tasks designed to learn interactions among agents, such as interaction type recognition (closing, leaving, and neural), closeness classification, distance prediction, and trajectory masking/reconstruction, showing enhanced performance. 
Focusing on the pedestrian trajectory prediction model, \citet{li2023pedestrian} introduced three auxiliary pretext tasks to learn fine-grained human keypoint features, including keypoints jigsaw puzzle, keypoint prediction, and keypoint contrastiv learning.

\paragraph{Prediction on Synthetic Trajectories}

\textbf{Pretraining on Synthetic Data} Another thread of works aims to exploit pre-training on synthetic data to learn more informative features. Given the abundance of HD maps, \cite{exploiting_map_info,li2023pre} generate synthetic traffic agent trajectories leveraging road geometry in HD maps. The trajectories are generated based on prior knowledge and manually designed rules such as synthetic speeds and constant acceleration. The model is pre-trained on synthetic and fine-tuned on real trajectories using the same task i.e.\ motion prediction. While this method improves motion prediction, the generated trajectories are limited by the sim-to-real gap as they do not model complex motions and scenarios with varying motion styles and intense interaction, and thus the distribution of artificially generated data often differs from real-world distributions. Additionally, generating high-fidelity simulations requires considerable computational resources and careful design to capture complex agent interactions accurately, limiting scalability. We refer readers to Section~\ref{sec:data_synthesis} for a dedicated discussion on data synthesis.



\subsubsection{Point Cloud Forecasting}
\label{sec: SSL - point cloud}

\begin{table}[!htb]
\centering
\caption{Taxonomy of point cloud forecasting methods, based on the input and output representations and the spatio-temporal feature encoder used. Abbreviations used are: Pts: Point Cloud, R-Map: Range Map (i.e. Depth Map), 3DVox: 3D Voxel Grid, Transf: Transformers, GCNN: Graph CNN.}
\label{tab:pcf}
\resizebox{1.0\linewidth}{!}{
\begin{tabular}{@{}ll|ccccc|cccccc}
\toprule\toprule
\multirow{2}{*}{Method} & \multirow{2}{*}{Venue} &\multicolumn{5}{c|}{Input (I) and Output (O)} & \multicolumn{6}{c}{Feature Extractor} \\ \cmidrule(l){3-13} 
 &
 &
  \multicolumn{1}{c}{Pts} &
  \multicolumn{1}{c}{R-Map} &
  \multicolumn{1}{c}{BEVOcc.} &
  \multicolumn{1}{c}{4DOcc.} &
  \multicolumn{1}{c|}{Images} &
  \multicolumn{1}{c}{PointNet} &
  \multicolumn{1}{c}{2D CNN} &
  \multicolumn{1}{c}{3D CNN} &
  \multicolumn{1}{c}{RNN} &
  \multicolumn{1}{c}{Transf.} &
  \multicolumn{1}{c}{GCNN} \\ \midrule
\citet{temporalLiDARFrame_deng2020} & 3DV'20 & I/O & &  &   & &  \checkmark &   &   &   &  & \checkmark \\
SPF2 \cite{SPF2} & CoRL'21 & I/O & O &   & &  & \checkmark & \checkmark  &   & \checkmark  &  &  \\
MoNet \cite{MoNet} & T-ITS'21 & I/O &   &   & & & \checkmark &   &    & \checkmark &  & \\
\citet{mersch2022self} & CoRL'22 & I/O & I/O  & &  &  & &   & \checkmark  &   &  &  \\
S2Net \cite{S2Net} & ECCV'22 & I/O & O  & &  & &  & \checkmark  &   & \checkmark  &  & \\
BEVOcc \cite{bevocc} & ECCV'22 & I &   & O & & & & \checkmark  &   &   &  & \\
4DOcc \cite{khurana2023point} & CVPR'23 & I/O &   &  & O &  & & \checkmark  &   &   &  &  \\
PCPNet \cite{pcpnet} & RA-L'23 & O &  I/O &  & &  & &   &  \checkmark &  & \checkmark & \\
ATPPNet \cite{atppnet} & ICRA'24 & O & I/O  &  & &  & & \checkmark  & \checkmark  & \checkmark &  & \\
Copilot4D~\cite{zhang2023learning} & ICLR'24 & I/O &  & & O & & &  &   &  & \checkmark & \\
OccWorld~\cite{zheng2025occworld} & ECCV'24 &  &  &  & I/O & & &  &   &  & \checkmark & \\
ViDAR \cite{vidar} & CVPR'24 & O &  & O  & & I & &  &   &  & \checkmark & \\
\bottomrule\bottomrule
\end{tabular}
}
\end{table}

SSL techniques discussed so far are limited to the representation of agent trajectory and HD maps obtained from costly annotated bounding boxes, tracks, and maps. Due to reliance on human-annotated agent trajectories and maps, such methods cannot scale to a large completely unlabeled dataset. A promising fully self-supervised task that aims to learn from unannotated LiDAR sequences is 3D point cloud forecasting (PCF). PCF models utilize point cloud as a proxy on how the world evolves, take past point clouds as input, and predict future point clouds as output. Off-the-shelf detection and tracking modules can then be used on predicted future scans to obtain future object trajectories \cite{SPF2, MoNet}. 

As presented in table \ref{tab:pcf}, we categorize PCF by two main criterias: 1) Input and output representation of point cloud, such as raw points, range maps (i.e. depth maps), 3D voxels; and 2) The type of feature encoder used to aggregate spatial-temporal features. In the following, we will present these methods in two parts, based on their input representations and feature encoders, as well as their output representations and training losses, as these components are typically interdependent and closely related.

\textbf{Inputs Representation and Feature Encoder} Various input representation have been exploited in point cloud forecasting, such as raw points, range maps (i.e. depth maps), 3D voxels, and camera image. The input representation/modality is also closely related to the feature encoder applied.
Earlier methods processed unordered point sets with PointNet-based architectures for spatial encoding \cite{SPF2, MoNet, temporalLiDARFrame_deng2020} and utilized RNN for temporal encoding \cite{SPF2, MoNet}. Since RNN-based methods process inputs sequentially and thus result in long training and inference times, more recent methods propose to speed up by encoding spatio-temporal features from range maps \cite{mersch2022self, pcpnet} or voxel grids \cite{bevocc, khurana2023point} using CNNS \cite{mersch2022self, bevocc, khurana2023point} and/or transformers \cite{pcpnet}.
Visual point cloud forecasting has also been explored in ViDAR~\cite{vidar}, which predicts future point clouds from past camera images to pretrain visual encoders.
It demonstrates strong performance when fine-tuned for downstream perception tasks such as 3D detection, map segmentation, semantic occupancy prediction, and multi-object tracking.
Tokenization has also been employed on 3D voxels~\cite{zheng2025occworld} or point clouds~\cite{zhang2023learning}, then GPT-like spatial-temporal generative transformer~\cite{zheng2025occworld} or diffusion models~\cite{zhang2023learning} are used to generate subsequent scenes.



\begin{figure}[t]
    \centering
    \includegraphics[width=0.96\textwidth]{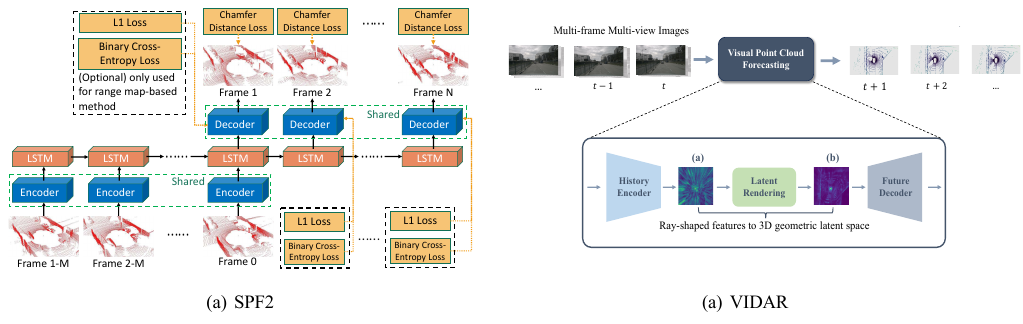}
    \caption{Examples for point cloud forecasting. (a) SPF2 ~\cite{SPF2} adopt LSTM encoder-decoder model to forecast point cloud in a uncertainty-aware manner. (b) VIDAR~\cite{vidar} leverages point cloud forecasting as a pretraining task, taking historic multi-view images as inputs, and predict future occupancy and point cloud. See more examples and discussion in Section~\ref{sec: SSL - point cloud}.
}
    \label{fig:point_cloud_forecast}
\end{figure}

\textbf{Output Representation and Training Losses} PCF methods further differ in using various types of output representation, which are usually associated with the training losses applied.
\begin{enumerate}
    \vspace{-0.5em}
    \item \textit{Point Cloud.} Some methods directly predict the future point cloud and minimize Chamfer distance and/or Earth Mover's distance between the predicted and ground-truth points \cite{SPF2, MoNet, temporalLiDARFrame_deng2020, S2Net}.
    \vspace{-0.5em}
    \item \textit{Range Map.} Range-based methods~\cite{mersch2022self, S2Net} minimize the L1 distance between predicted and ground-truth range maps, and apply a binary cross-entropy loss on the predicted range mask, where pixels without a point are labeled as 0.
    \vspace{-0.5em}
    \item \textit{BEV Occupancy Map.} \citet{bevocc} and ViDAR~\cite{vidar} predict occupancy on the BEV map and then use differentiable raycasting to render future point clouds from the predicted occupancy map. \citet{bevocc} claims that learning the occupancy map in world coordinate frame naturally disentangles the motion of the environment from the motion of the ego-vehicle, which enables learning ego-pose independent representations. Moreover, the predicted occupancy maps can be directly used in motion planning \cite{bevocc}. 
    
    \vspace{-0.5em}
    \item \textit{4D Occupancy Map.} 4DOcc~\cite{khurana2023point} and Copilot4D~\cite{zhang2023learning} further extend the idea of forecasting BEV spacetime occupancy to forecasting 4D spacetime (x, y, z, t) occupancy from unannotated LiDAR sequences, which additionally considers the height dimension. 
    A noteworthy feature of \cite{khurana2023point} is that, unlike other SSL techniques, their method allows for joint training and generalization across different datasets with different lidar patterns. From the same predicted occupancy, one can render point clouds as if they are captured by different LiDAR sensors. This enables zero-shot cross-sensor generalization. OccWorld~\cite{zheng2025occworld} outputs 4D voxel occupancy without rendering point cloud, and apply softmax loss functions to enforce correct classificatiion.
\end{enumerate}

\textbf{Uncertainty-Awareness} S2Net \cite{S2Net} recognized that existing PCF approaches are limited as they forecast future point clouds deterministically, despite the uncertainty of dynamic scenes. 
To predict future point clouds with uncertainty for each point, S2Net extends LSTM encoder-decoder model in \cite{SPF2} with a conditional variational LSTM.

\subsubsection{Video Generation and Scene Reconstruction}
\label{sec: SSL - video}

\begin{figure}[t]
    \centering
    \includegraphics[width=0.9\textwidth]{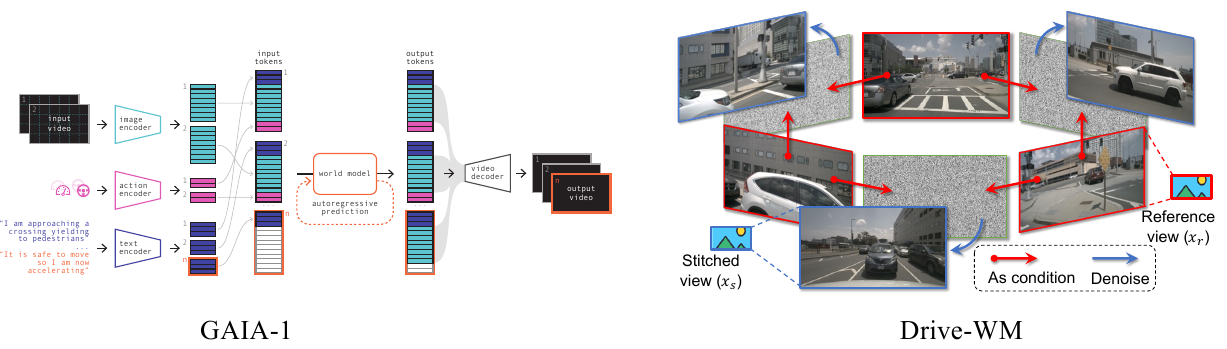}
    \caption{Examples in video generation in driving scenes: GAIA-1~\cite{hu2023gaia} takes the initial image, control action, and text as prompts to autoregressive generate single-front-view future videos. Drive-WM~\cite{wang2024driving} generates consistent multi-view future videos by factorizing the generation to first predict non-overlapping views, and then predict other views conditioned on adjacent views. See more examples and discussion in Section~\ref{sec: SSL - video}.}
    \label{fig:video_gen}
\end{figure}

\begin{table}[t!]
\caption{Taxonomy of video generation methods based on the data setups and model. }
\centering
\scalebox{0.7}{
\begin{tabular}{p{0.20\textwidth} | >{\centering}p{0.1\textwidth} >{\centering}p{0.1\textwidth } >{\centering}p{0.1\textwidth} >{\centering}p{0.1\textwidth} >{\centering}p{0.14\textwidth} | >{\centering}p{0.1\textwidth} >{\centering}p{0.12\textwidth} >{\centering\arraybackslash}p{0.1\textwidth}}
\toprule
\toprule
\multicolumn{1}{l|}{\multirow{2}{*}{{Method}}} & \multicolumn{5}{c|}{{Data Setups}} & \multicolumn{3}{c}{{Model}} \\
\cmidrule(l){2-9}
& Real World & Multi View & Data Scale & Frame Rate & Resolution & GAN & Auto Regressive & Diffusion \\
\midrule
DriveSim~\cite{santana2016learning} &  & & 7h & 5 Hz & 80$\times$160 & \checkmark & & \\
DriveGAN~\cite{kim2021drivegan} & & & 160h & 8 Hz & 256$\times$256 & \checkmark & & \\
DriveDreamer~\cite{wang2023drivedreamer} & \checkmark & \checkmark & 5h & 12 Hz & 128$\times$192 & & & \checkmark \\
ADriver-I~\cite{jia2023adriver} & \checkmark & & 300h & 2 Hz & 256$\times$512 & & & \checkmark\\
MagicDrive~\cite{gao2023magicdrive}  & \checkmark & \checkmark & 5h & 2 Hz & 272$\times$736 & & & \checkmark\\
GAIA-1~\cite{hu2023gaia} & \checkmark & & 4700h & 25 Hz & 288$\times$512 & & \checkmark & \checkmark \\
Drive-WM~\cite{wang2024driving} & \checkmark & \checkmark & 5h & 2 Hz & 192$\times$384 & & & \checkmark \\
WoVoGen~\cite{lu2025wovogen} & \checkmark & \checkmark & 5h & 2 Hz & 256$\times$448 & & & \checkmark \\
GenAD~\cite{zheng2024genad} & \checkmark & & 2000h & 2 Hz & 256$\times$448 & & & \checkmark\\
DrivingDojo~\cite{wang2024drivingdojo}  & \checkmark & & 150h & 5 Hz & 1024$\times$576 & & & \checkmark \\
Vista~\cite{gao2024vista} & \checkmark & & 1740h & 10 Hz & 576$\times$1024 & & & \checkmark \\
\bottomrule
\bottomrule
\end{tabular}
}
\vspace{-3mm}
\end{table}

\textbf{Video Generation} In addition to point cloud forecasting, another self-supervised scene-level temporal forecasting task is video generation. General video generation models take various forms of inputs as prompt, such as text, images, and video, and generate the future evolution in the image space. Regarding the methodology, video generation has been tackled with recurrent neural networks~\cite{babaeizadeh2017stochastic,castrejon2019improved,denton2018stochastic}, autoregressive transformers~\cite{ge2022long,gupta2022rv,hoppe2022diffusion,wu2022nuwa,yan2021videogpt}, Normalizing Flows~\cite{blattmann2021ipoke,dorkenwald2021stochastic}, generative adversarial networks (GANs)~\cite{brooks2022generating,fox2021stylevideogan,kahembwe2020lower,yu2022generating}, and diffusion models~\cite{singer2022make,ho2022imagen,villegas2022phenaki,zhou2022magicvideo,blattmann2023align}. Common techniques include leveraging pre-trained image generators via inserting temporal layers~\cite{singer2022make,ho2022imagen,blattmann2023align}, cascaded models which start with low-resolution models and apply a series of super resolution models~\cite{ho2022cascaded}, latent models which first compress the image data using an autoencoder and learn to generate in the latent space~\cite{rombach2022high,vahdat2021score},
and divide and conquer methods which begins by generating keyframes that outline the main narrative, followed by inpaining/interpolating the intermediate frames to generate cohesive long video~\cite{hong2022cogvideo,ge2022long}. The abundance and easy access of internet-scale videos provide these models the opportunity to scale up with large-scale data, as demonstrated by recent advances in SORA~\cite{sora}.
In the autonomous driving and robotics domain, video generation models are further extended to 1) take more robot-related conditions~\cite{gao2023magicdrive,hu2023gaia,wang2024drivingdojo,wang2024driving,gao2024vista}, such as camera poses, BEV road maps, 3D bounding boxes, textual descriptions, ego actions; 2) generate multi-view consistent videos~\cite{wang2024drivingdojo,wang2024drivingdojo,gao2023magicdrive,gao2024vista}; 3) promote object awareness~\cite{gao2024vista}; 4) inform downstream driving tasks~\cite{wang2024driving}; 5) leverage large-scale internet driving videos~\cite{gao2024vista}.

\textbf{Scene Reconstruction} Compared to point cloud forecasting, video generation inherently provides richer semantics through the synthesis of visual appearances. However, it often lacks 3D grounding, a critically favoured property for autonomous driving, resulting in inconsistent geometry and object motion. To this end, neural scene representations, like NeRFs~\cite{mildenhall2021nerf,muller2022instant} and 3DGS~\cite{kerbl20233d,fan2024instantsplat}, have brought unprecedented success in learning powerful representations of 3D scenes, and have also been successfully applied to challenging driving scenes populated with dynamic objects \cite{rematas2022urban,tancik2022block,guo2023streetsurf,wang2023neural,park2021hypernerf,Wang2024NeRFIR,he2024neural}. Recent works further extend the scene representation to 4D~\cite{yang2023emernerf,zhou2024drivinggaussian}, which considers not only the spatial modeling of the static scenes, but also the temporal modeling of the dynamic objects. 
However, these methods typically require expensive training for each scene, typically in the order of hours or minutes. 
To this end, generalizable methods, such as \cite{yu2021pixelnerf,wang2021ibrnet,liu2022neural,mvsnerf, johari2022geonerf,charatan2024pixelsplat}, adapt the capabilities of conventional NeRFs for 3DGS into a generalizable feedforward model. They replace the costly per-scene optimization with a single feedforward pass through the models, and have also been applied to the driving domain~\cite{zhang2023occnerf,yang2023unipad,huang2023selfocc,ren2024scube,wang2024distillnerf}. Early explorations have also been made toward 4D feed-forward reconstruction methods in autonomous driving domains~\cite{yang2024storm}.

\begin{figure}[t]
    \centering
    \includegraphics[width=0.96\textwidth]{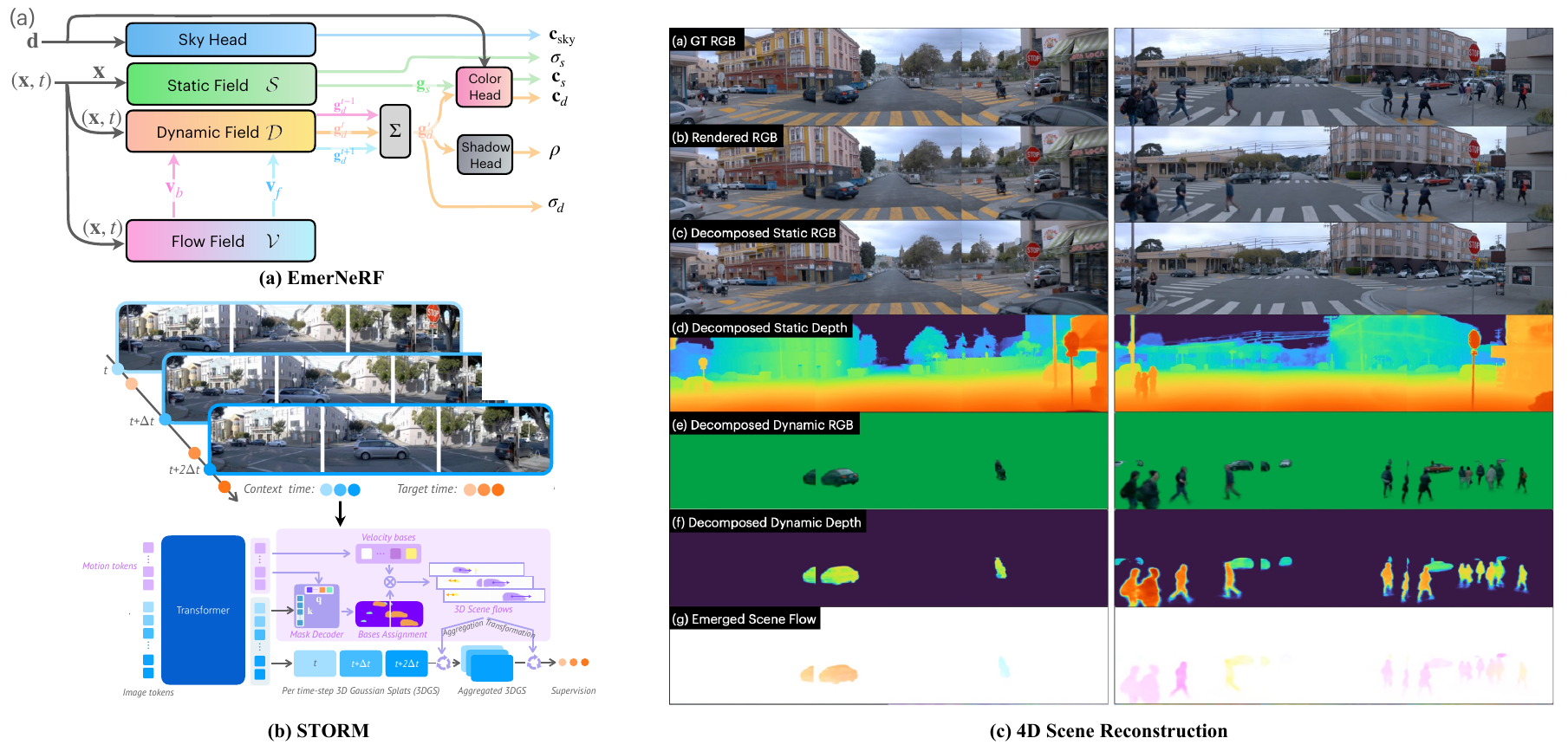}
    \caption{Examples for 4D scene reconstruction. (a) EmerNeRF~\cite{yang2023emernerf} takes an offline NeRF approach, decomposing static and dynamic components into two separate radiance field. (b) STORM~\cite{yang2024storm} adopts an online Gaussian Splatting approach to aggregate multi-frame predictions. (c) An illustration of 4D scene reconstruction, the figure from EmerNeRF~\cite{yang2023emernerf}. See more examples and discussion in Section~\ref{sec: SSL - video}.}
    \label{fig:reconstruction}
\end{figure}

\begin{table}[t!]
\caption{Taxonomy of scene reconstruction methods based, on the representation and model characteristics.}
\centering
\scalebox{0.7}{
\begin{tabular}{p{0.25\textwidth} | >{\centering}p{0.1\textwidth} >{\centering}p{0.1\textwidth } >{\centering}p{0.1\textwidth} >{\centering\arraybackslash}p{0.1\textwidth}}
\toprule
\toprule
\multicolumn{1}{c|}{Method} & \multicolumn{1}{c}{{NeRF}} & \multicolumn{1}{c}{{3DGS}} & \multicolumn{1}{c}{{Dynamic Aware}} & \multicolumn{1}{c}{{Feed Forward}} \\
\midrule
DNMP~\cite{rematas2022urban} & \checkmark & & & \\
Block-NeRF~\cite{tancik2022block} & \checkmark & & & \\
Streetsurf~\cite{guo2023streetsurf} & \checkmark & & & \\
FEGR~\cite{wang2023neural} & \checkmark & & & \\
Hypernerf~\cite{park2021hypernerf} & \checkmark & & & \\
EmerNeRF~\cite{yang2023emernerf} & \checkmark & & \checkmark & \\
DrivingGaussian~\cite{zhou2024drivinggaussian} & & \checkmark & \checkmark & \\

OccNeRF~\cite{zhang2023occnerf} & \checkmark & & & \checkmark \\
UniPAD~\cite{yang2023unipad} & \checkmark & & & \checkmark \\
SelfOcc~\cite{huang2023selfocc} & \checkmark & & & \checkmark\\ 
Scube~\cite{ren2024scube} & & \checkmark & & \checkmark \\
DistillNeRF~\cite{wang2024distillnerf} & \checkmark & & & \checkmark\\
STORM~\cite{yang2024storm} & & \checkmark & \checkmark & \checkmark\\

\bottomrule
\bottomrule
\end{tabular}
}
\vspace{-3mm}
\end{table}

\subsubsection{Perspectives}
\textbf{Limited Size and Diversity of Motion Datasets}
One of the main reasons behind the success of self-supervised learning (SSL) in computer vision and natural language processing is the availability of large-scale pretraining datasets spanning diverse domains.
In contrast, the limited size of publicly available motion prediction datasets has constrained research on scaling model performance with increased data or model capacity.
Moreover, generalization across different motion prediction datasets remains largely unexplored, as variations in dataset settings—such as differing observation and prediction horizons and incompatible data formats—further hinder the unified use of existing data for SSL pretraining. A unified interface and benchmark to multiple trajectory datasets \cite{zhou2024smartpretrain,feng2025unitraj,trajdata} is therefore necessary to enable multi-dataset pretraining or cross-dataset transfer learning. 
Besides, pretraining on a large corpus of data with balanced representation across agent types, motion patterns, and geographic regions is essential for enhancing the generalizability of motion prediction models.
However, existing public datasets are often imbalanced and limited in distributional diversity, posing a significant challenge for effective SSL pretraining.
While the industry has access to substantially larger motion datasets, open questions remain regarding how to quantify motion data distributions, measure data quality and diversity, and improve data balance—factors that are equally critical for efficiently scaling up training data at scale.
In fact, these data limitations are not unique to SSL, but broadly affect various approaches aimed at improving generalization in motion prediction; see Sections~\ref{sec:data_synthesis}, \ref{sec:foundation_models}, and \ref{sec: generalization} for more discussions.

\textbf{Fair Comparison Under Equivalent Compute Budgets}
While some studies~\cite{forecast_mae,zhou2024smartpretrain} have examined the impact of varying pre-training epochs on model performance, to the best of our knowledge, existing research lacks a rigorous comparison between training from scratch and pretrain-finetuning, under an equivalent compute budget. 
For example, the intuitive practice of simply allocating the same number of training epochs to both pretrained and scratch-trained models could introduce bias during evaluation, as the pretrained model leverages additional compute during pretraining and typically converges more quickly~\cite{he2019rethinking}.
Consequently, it remains uncertain whether the observed performance gains stem from the improved learning signals provided by SSL pretext tasks or merely from the additional compute involved in pretraining. 
To establish the true effectiveness of SSL pretext tasks, future work should ensure fair comparisons by evaluating scratch-training and pretrain-finetuning under identical compute budgets. Fundamentally, another key question is how to optimally allocate a fixed compute budget between pretraining and fine-tuning to achieve the greatest performance improvements.


\textbf{SSL in Trajectory vs. Sensor Stream} With the growing interest in world models~\cite{lecun2022path} within the research community, approaches that conduct SSL in sensor streams such as point cloud forecasting, video generation, and 4D scene reconstruction have emerged to capture the temporal evolution of scenes, integrating both geometric and semantic aspects. These methods leverage the advantage of learning directly from sensor streams, bypassing the annotation costs associated with trajectory-based representations, and provide access to high-quality realistic simulations. However, they face challenges such as high computational demands due to direct sensor-level processing and limitations in effectively capturing discrete object entities within the scene. 
Despite these challenges, this remains a vibrant and rapidly evolving field, with these issues continuing to spark numerous research papers.




\begin{figure}[!t]
    \centering
    \subcaptionbox{Training scenario \label{fig:domain_shift1}}[0.4\textwidth]{\includegraphics[width=0.4\textwidth]{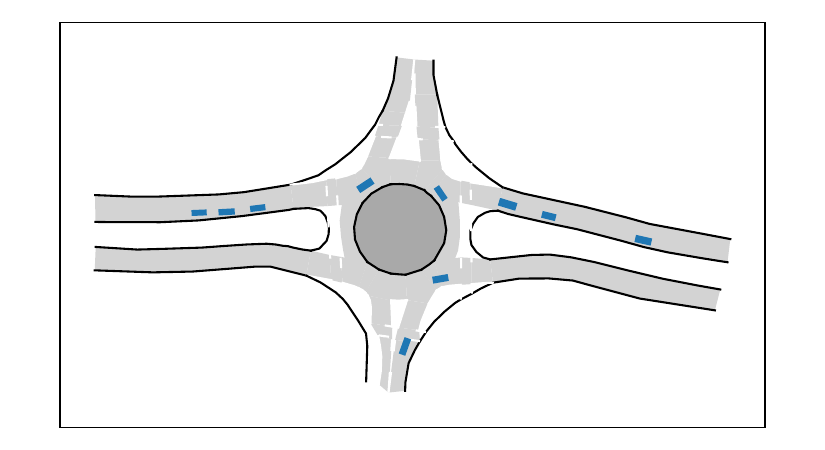}}
    \subcaptionbox{Different road geometry \label{fig:domain_shift2}}[0.4\textwidth][c]{\includegraphics[width=0.4\textwidth]{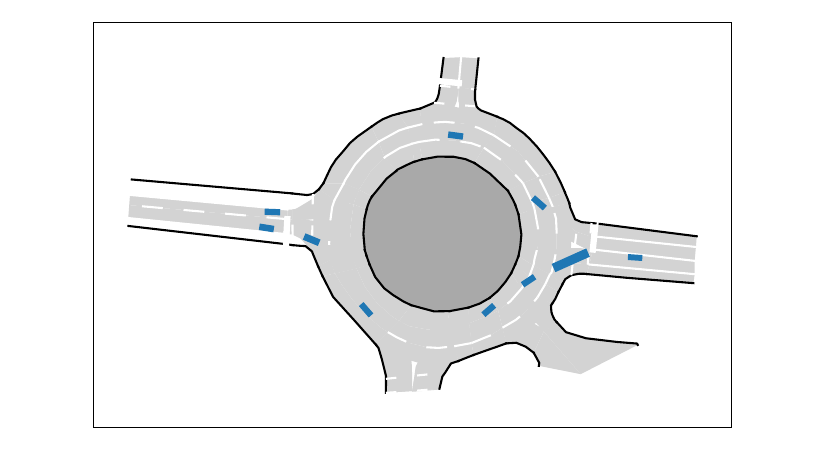}}
    \subcaptionbox{Different traffic regulation (e.g. lights/signs) \label{fig:domain_shift3}}[0.4\textwidth][c]{\includegraphics[width=0.4\textwidth]{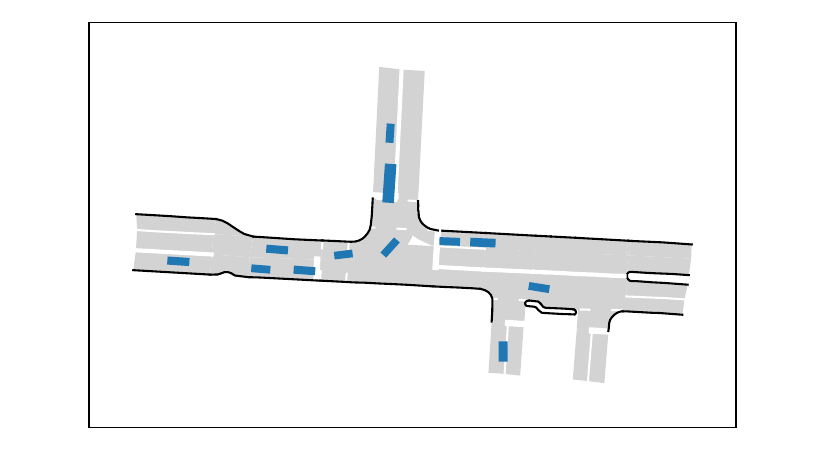}}
    \subcaptionbox{Different agent class \label{fig:domain_shift4}}[0.4\textwidth][c]{\includegraphics[width=0.4\textwidth]{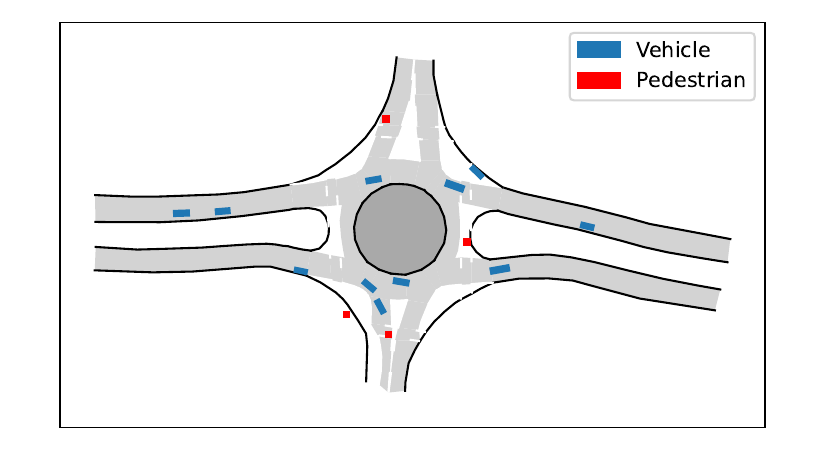}}
\caption{Examples of motion distribution shifts in autonomous driving.}
\label{fig:distribution_shifts}
\end{figure}

\subsection{Domain Generalization and Adaptation}
\label{sec:domain-adaptation-and-generalization}

An important challenge in motion prediction is that algorithms can be very sensitive to the type of agent (e.g. classes, motion patterns) and the operating environment they are trained on. Thus, their performance can be greatly degraded when operating in different conditions. 
To this end, \textbf{domain generalization} and \textbf{domain adaptation} are two tasks proposed to deal with transferring the model from one or many well-sampled and labelled source domains, to a new target domain \cite{pan2009survey,wang2022generalizing}.
These two tasks differ in their assumptions: in domain generalization, the models are trained on source domains and tested on a new target domain, without seeing any data from the new domain. By contrast, in domain adaptation, some target domain data is accessible, but usually this target-domain data is unlabelled or only a few examples are available. 
Representative examples of domain gaps encountered in autonomous driving scenarios are illustrated in Figure~\ref{fig:distribution_shifts}.
In this section, we categorize existing methods into four main categories: 1) \textbf{designing and learning invariant features}, 2) \textbf{domain-specific modules}, 3) \textbf{causal learning}, 4) \textbf{few-shot and test-time adaptation}. In each of the following subsections, we focus on introducing a single category: we first briefly review its underlying principles, and then discuss its application to motion prediction. Table \ref{tab:da_dg_methods} summarizes the discussed approaches. 

\begin{table}[!htb]
\centering
\caption{Taxonomy of domain generalization (DG) and domain adaptation (DA) methods based on the approach used and domain of application. The approaches include invariant features (IF), domain-specific modules (Mod), causal learning (CL), generative and augmentation methods (Gen), fast or test-time adaptation (TTA). The applications are vehicle forecasting (VF), pedestrian forecasting (PF), multi-class forecasting (MMF) and human motion prediction (HMP). Methods are grouped and ordered by the approach used.}
\label{tab:da_dg_methods}
\resizebox{0.8\linewidth}{!}{
\begin{tabular}{@{}l|cc|ccccc|cccc}
\toprule\toprule
\multirow{2}{*}{Method} & \multicolumn{2}{c|}{Task} & \multicolumn{5}{c|}{Approach} & \multicolumn{4}{c}{Application} \\  
 &
  DA & DG  & IF & Mod & CL & Gen & TTA  &  VF & PF & MMF & HMP \\ \midrule
Frenet\textsuperscript{$\dagger$} \cite{ye2023improving}                &   & \checkmark & \checkmark &   &   &   &   & \checkmark &   &   &  \\
STS LSTM \cite{zhang2023spatial}                                        &   & \checkmark & \checkmark &   &   &   &   &   & \checkmark &   &  \\
NSDETraj\textsuperscript{$\dagger$} \cite{park2024improving}            &   & \checkmark & \checkmark &   &   &   &   & \checkmark &   &   &  \\
SingularTrajectory \cite{bae2024singulartrajectory}                     &   & \checkmark & \checkmark &   &   &   &   & \checkmark &   &   &  \\
T-GNN \cite{xu2022t-gnn}                                                & \checkmark &   & \checkmark &   &   &   &   &   & \checkmark &   & \\
HTN \cite{geng2023htn}                                                  & \checkmark & \checkmark & \checkmark &   &   &   &   &   &   & \checkmark & \\
CTSDG \cite{hu2022causalintention}                                      &   & \checkmark & \checkmark &   &   &   &   & \checkmark &   &   & \\             
\midrule
IM\textsuperscript{$\dagger$} \cite{liu2022causalAux}                   &   & \checkmark &   & \checkmark & \checkmark &   &   &   & \checkmark &   & \\
MoSA \cite{kothari2023motionLora}                                       & \checkmark &   &   & \checkmark &   &   &   & \checkmark &   & \checkmark & \\
LaneTransformer++ \cite{wang2023bridging}                               &   & \checkmark &   & \checkmark &   &   &   & \checkmark &   &   & \\
FLN \cite{xu2024adapting}                                               & \checkmark & \checkmark &   & \checkmark &   &   &   & \checkmark & \checkmark &   & \\
GCRL \cite{bagi2023generative}                                          & \checkmark & \checkmark &   &   & \checkmark &   &   &   & \checkmark &   & \\            
\midrule
CILF\textsuperscript{$\dagger$} \cite{causality_ood}                    &   & \checkmark &   &   & \checkmark &   &   &   & \checkmark &   & \\
CFA \cite{chen2021counterfactual}                                       &   & \checkmark &   &   & \checkmark &   &   &   & \checkmark &   & \\
SEAD \cite{ge2023causal}                                                &   & \checkmark &   &   & \checkmark &   &   &   & \checkmark &   & \\
CaDeT \cite{pourkeshavarz2024cadet}                                     &   & \checkmark &   &   & \checkmark &   &   & \checkmark &   &   & \\            
\midrule
AngleAug\textsuperscript{$\dagger$} \cite{kong2024adaptive}             & \checkmark &   & \checkmark &   &   & \checkmark &   &   & \checkmark &   &  \\
GenAdapt \cite{bourached2022generative}                                 &   & \checkmark &   &   &   & \checkmark &   &   &   &   & \checkmark \\
BelFusion \cite{barquero2023belfusion}                                  &   & \checkmark &   &   &   & \checkmark &   &   &   &   & \checkmark \\          
\midrule
HATN \cite{wang2021hierarchical}                                        & \checkmark &   &   &   &   &   & \checkmark & \checkmark &   &   &  \\
MetaLearn\textsuperscript{$\dagger$} \cite{ivanovic2023expanding}       & \checkmark &   &   &   &   &   & \checkmark & \checkmark &   &   &  \\
T4P \cite{park2024t4p}                                                  & \checkmark &   &   &   &   &   & \checkmark & \checkmark &   &   &  \\
GoCNN \cite{li2022online}                                               & \checkmark &   &   &   &   &   & \checkmark &   &   &   & \checkmark \\
FastAdapt\textsuperscript{$\dagger$} \cite{moon2021fastadapt}           & \checkmark &   &   &   &   &   & \checkmark &   &   &   & \checkmark \\
MetaTraj \cite{shi2023metatraj}                                         & \checkmark &   &   &   &   &   & \checkmark &   & \checkmark &   &  \\
MANTRA \cite{marchetti2020mantra}                                       & \checkmark &   &   &   &   &   & \checkmark & \checkmark &   &   &  \\
OATMem \cite{huynh2023online}                                           & \checkmark &   &   &   &   &   & \checkmark &   & \checkmark &   &  \\
OML-PTP \cite{yang2024fast}                                             & \checkmark &   &   &   &   &   & \checkmark &   & \checkmark &   &  \\
\bottomrule\bottomrule
\end{tabular}
}
\end{table}

\subsubsection{Designing and Learning Invariant Feature}
\label{sec: generalization DA/DG invariant feature}

\textbf{Method Recap} The first class of methods focuses on leveraging invariant features whose distributions exhibit minimal variation across domains, thereby enhancing the model's generalizability.
Such features can be either manually designed or automatically learned:



\begin{itemize}[label={\scriptsize$\bullet$}]
    \vspace{-0.5em}
    \item \textbf{Designing Invariant Feature} The first type of methods aims at manually designing data representations that tend to be more invariant under distribution shifts. 
    A fundamental example is the development of hand-crafted features in classical computer vision, such as SIFT~\cite{lowe2004distinctive}, which is designed to be invariant to scale, rotation, and moderate illumination changes.
    By adopting such relatively stable representations, models become less sensitive to environmental variations.
    However, hand-crafted invariant features are constrained by fixed assumptions, lack flexibility to adapt to unseen conditions, have limited representational capacity, and require substantial manual effort for feature engineering.
    
    \vspace{-0.5em}
    
    \item \textbf{Learning Invariant Feature} 
    To avoid manually designing invariant features, another line of work seeks to automatically learn invariant representations by minimizing a distributional discrepancy loss across domains, alongside the main learning objective.
    The goal is to align feature representations across different domains or distributions, ensuring that the model's learned features are invariant to domain shifts. 
    Depending on the actual alignment method applied, existing methods can be broadly categorized into two types:
    \begin{itemize}
        \vspace{-0.5em}

        \item \textbf{Direct alignment methods} learn invariant representations by explicitly enforcing a distributional discrepancy loss across feature representations from different domains.
        A straightforward example is \citet{kong2024adaptive}, which applies an L2 loss to align encoded features from two domains.
        At the sample level, contrastive loss~\cite{oord2018infoNCE, kang2019cdd} is commonly used to encourage alignment by minimizing the angle between normalized feature vectors.
        At the distribution level, several statistical distances have been explored.
        Maximum mean discrepancy (MMD)\cite{gretton2006mmd, long2015mmd, abuduweili2021cvpr} computes the distance between distributions via kernel embeddings.
        Alternatively, the Wasserstein (Earth Mover’s) distance\cite{cuturi2013sinkhorn, courty2017jointOT} measures the minimal cost required to transport one distribution into another within a metric space.
        
        \item \textbf{Adversarial alignment methods} offer an alternative to direct alignment approaches. Inspired by generative adversarial networks (GANs) \cite{goodfellow2014gan}, these methods adopt a similar minimax objective \cite{ganin2016domain, long2018conditional}, where a domain discriminator attempts to identify which domain each encoded feature comes from, while the feature encoder attempts to fool the discriminator.
    \end{itemize}

    

    
\end{itemize}

\begin{figure}[t]

    \subcaptionbox{Designing Invariant Features \label{subfig:design_invariant_feats}}[0.36\textwidth]{\includegraphics[width=0.36\textwidth]{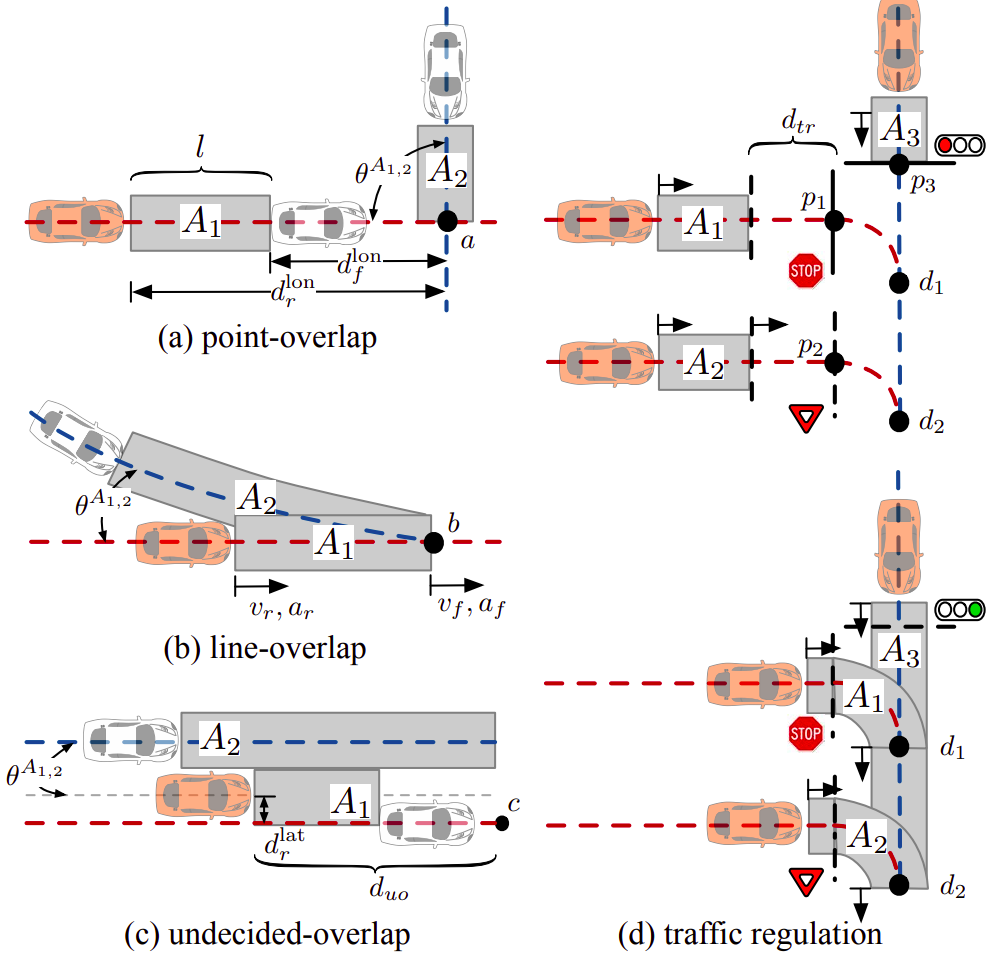}}
    \subcaptionbox{Learning Invariant Features \label{subfig:learn_invariant_feats}}[0.6\textwidth]{\includegraphics[width=0.6\textwidth]{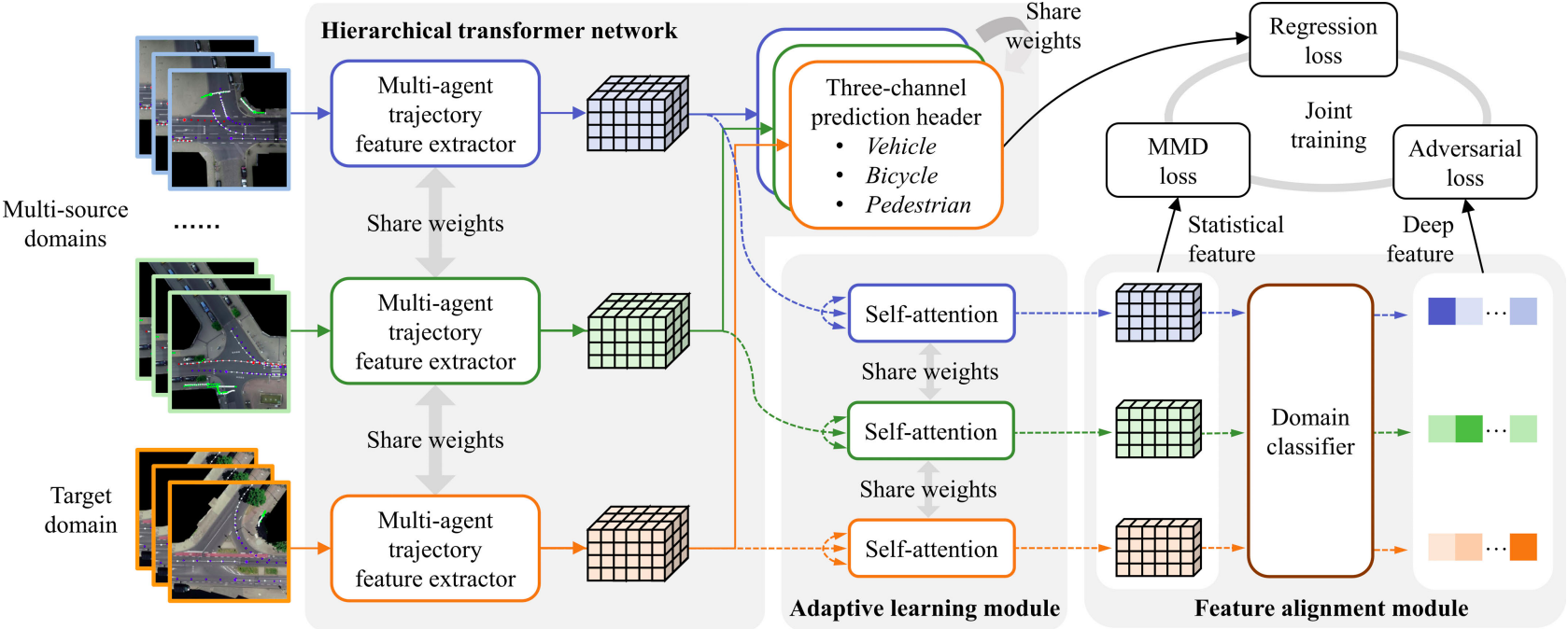}} 
    \caption{Examples of methods that leverage invariant feature for generalization. (a) Designing invariant features: SGN~\cite{hu2020scenario} uses dynamic insertion areas to identify areas that a vehicle could potentially enter, serving as a generic representation across scenarios. (b) Learning invariant features: HTN \cite{geng2023htn} uses both direct feature alignment (MMD loss) and adversarial feature alignment (adversarial losses) to learn robust features across domains. See more examples and discussions in Section~\ref{sec: generalization DA/DG invariant feature}.}
    \label{fig:schematic_align}
\end{figure}

\textbf{Application in Motion Prediction.} Both invariant feature design and invariant feature learning approaches have been used in motion prediction, and we provide some illustrations in Figure~\ref{fig:schematic_align}. 
In addition to variations in feature distributions across domains, differences in trajectory lengths and the number of objects also arise across domains and scenarios in motion prediction, and have been explicitly considered in some works.


\begin{itemize}[label={\scriptsize$\bullet$}]
    \vspace{-0.5em}
    \item \textbf{Designing Invariant Features} There are many methods that improve generalization through feature and representation design, which we group into four categories.

    \begin{itemize}
    \vspace{-0.5em}
        \item 
    \textbf{Frenet Coordinate Representation} A first set of methods leverage the observation that most trajectories follow a reference line. 
    For example, autonomous vehicles typically adhere to lane centerlines, while pedestrians tend to move along sidewalks or footpaths.
    Thus, a conceptually simple approach is to formulate trajectory prediction in a local Frenet reference frame~\cite{hu2019generic, ye2023improving}, which is aligned with the agent’s expected path and thus simplifies motion modeling.
    Compared to Cartesian coordinates, the Frenet-frame representation is generally more invariant and robust to data shift, since 1) the range of the input/output values is reduced, since the lateral coordinate usually maintains a relatively small value; 2) In the longitudinal-lateral coordinate system, when a agent deviates from the reference line in one direction, its position in the orthogonal direction often remains unchanged, which better captures the fact that agents generally follow the reference line in their motion. 
    
\vspace{-0.2em}
        \item 
    \textbf{Hierarchical Representation} Another set of methods follows a ``divide and conquer" approach, decomposing the motion prediction into two hierarchies, namely high-level intention prediction and low-level motion prediction, and invariant representations are designed for each hierarchy to achieve generalization in both intention and motion prediction.
    For example, HATN~\cite{hu2020scenario,wang2021hierarchical,wang2022transferable} proposes dynamic insertion area (DIA), an advanced invariant representation, as the high-level intention representation that capture complex road geometries and interaction patterns. This high-level representation, combined with low-level Frenet Coordinate-based representation, demonstrates strong zero-shot generalization performance across driving scenarios. Simlarly, high-level driving skill~\cite{wang2023efficient} and robot plans~\cite{cheng2020towards} are also explored as the intention representation to facilitate learning and generalization across scenarios and tasks.

\vspace{-0.2em}
        \item 
    \textbf{Trajectory Transformation and Augmentation} Another approach~\cite{kong2024adaptive} observes that the differences between driving scenarios mainly arise from variations in road structure, obstacle positions, and social norms, which are largely reflected by the trajectory heading angle relative to a reference direction. To this end, this approach proposes to augment the orientation of source-domain trajectories to match the distribution of heading angles observed in the target domain. Specifically, it first computes the heading angle distribution from the target domain, then samples angles from this distribution to rotate source trajectories and generate augmented data. By aligning the trajectory orientation distributions between domains, the approach improves the generalization performance of the trajectory predictor on the target domain.

    \vspace{-0.2em}
    \item 
    \textbf{Continous Function Representation} Other methods exploit the fact that, while trajectories are commonly represented as discrete sequences of waypoints, they can alternatively be modeled as continuous functions, which possess multiple advantages:
    1) some approaches transform the trajectory into the frequency domains, via discrete Fourier transform as in STS-LSTM \cite{zhang2023spatial} and discrete cosine transform as in \citet{mao2019learning}, and improved generalization to unseen domains is observed in STS-LSTM \cite{zhang2023spatial}.
    2) Moreover, modeling trajectory in continuous space is also particularly useful when trying to generalize across different input and output horizons.
    For example, \citet{park2024improving} propose to improve a standard RNN model by modeling it as a continuous-time neural stochastic differential equations (NSDE) \cite{li2020scalable}, to help account for the different trajectory horizon and sampling rates across different datasets. Similarly, SingularTrajectory \cite{bae2024singulartrajectory} generalizes across different input sequence lengths by first fitting a continuous spline to the discrete trajectory and re-sampling a fixed number of points along the spline. It then encodes the fixed-length trajectory into latent space of motion primitives, 
    that capture abstract, reusable patterns of motion, enabling generalization across varying contexts and agents.
    3) Beyond handling horizon mismatches, continuous-space modeling can also promote physically consistent predictions.
    For example, PCMP \cite{Tumu2023PhysicsCM} introduces physics-constrained motion prediction, which formulates the trajectory prediction as the integration of a dynamic system given learned control inputs. This ensures that the generated trajectories are smooth and feasible, which improves performance when operating on unseen data.

    \end{itemize}
    

    
    \vspace{-0.5em}
    \item \textbf{Learning Invariant Features} In addition to manually designing invariant features, learning invariant features automatically has also been explored in motion prediction, which is usually accomplished by learning generalizable hierarchical representation, or aligning the features across domains. 

\begin{itemize}
    \vspace{-0.2em}
    \item 
    \textbf{Learning hierarchical representation} In human motion prediction, BeLFusion \cite{barquero2023belfusion} obtains more generalizable predictions by disentangling high-level behavior from low-level motion. The high-level behavior captures an action with semantics, such as waving, sitting, while low-level motion corresponds to the continuous sequence of the most recent joint position and velocity. 
    This disentangled representation is obtained in a more learning approach, namely by training with an adversarial loss: the encoded features for high-level behavior is trained to contain no motion information, while the combination of motion and behavior features should be able to predict the motion sequence.
    Similarly, \citet{bourached2022generative,zhou2023accelerating} augments a prediction model with a variational autoencoder (VAE) -based trajectory reconstruction head, where the prediction model only needs to predict a intention latent, which is then decoded to actual low-level trajectory.

    \vspace{-0.2em}
    \item \textbf{Alignment Methods} As discussed in the method recap of this subsection, learning invariant features is more commonly achieved via alignment methods. 
    In domain generalization, where the target domain data is inaccessible, the alignment is generally done across multiple source datasets, with the expectation that a representation aligned in multiple source domains will have a better chance of extending to unseen target domains. In domain adaptation, which has access to target domain data, the alignment is directly between source and target domain features.

    \textit{Alignment Losses:} multiple alignment losses have been explored: 1) \textit{Direct alignment} methods such as L2 loss~\cite{xu2022t-gnn} and maximum mean discrepancy (MMD) loss~\cite{geng2023htn} have been used. 2) \textit{Adversarial alignment} has also been adopted in HTN \cite{geng2023htn}, where a domain classifier is learned to distinguish source and target scene features with an adversarial loss. The authors find that combining both direct alignment and adversarial alignment yields better performance than using either individually. 
    3) \textit{Distillation alignment:} while the above losses are applied in simultaneous multi-domain training, distillation losses have also been adopted under sequential training paradigms. For example, Laneformer++ \cite{wang2023bridging} first trains a teacher model on a source domain, freezes it, and then trains a student model on a new domain with a similarity loss that aligns its features with those of the frozen teacher, effectively distilling and integrating knowledge from the source domain.
    4) \textit{Class-conditioned alignment:} for prediction problems that consider multiple agent classes \cite{geng2023htn} or maneuver classes \cite{deo2018would}, alignment can be done in a class-conditional way, where trajectories belonging to the same class (e.g., pedestrian with pedestrian, cyclist with cyclist, or the same maneuver or intention class) are matched across domains.
    For instance, in the intention prediction problem, CTSDG \cite{hu2022causalintention} explores the contrastive loss, where it learns domain invariant representations for each intention class by minimizing a contrastive loss, considering samples of the same class in different domains as positives and all other samples as negatives.
    This increases the specificity of the alignment, but requires accurate target domain class labels. In the absence of ground truth class labels in the domain, pseudo labels can be generated using iterative clustering \cite{mahajan2021domainclustering} or confident predictions from the source-trained model \cite{tarvainen2017meanteacher,li2023semisupervised}.


\textit{Choice of To-be-Aligned Feature, and Dealing with Varied Agent Number:} Regarding the instantiation of the alignment, in general, alignment is applied to the encoded features before the trajectory decoding step, corresponding to the richest feature representation. 
However, unlike tasks such as image classification, where the size of the encoded feature is fixed, motion prediction presents the additional challenge that the number of encoded features varies with the number of agents and nodes in a scene.
To address this variability and enable alignment across different scenes and domains, an attention head can be used to aggregate information from all the agent trajectories into a single representative fixed-length feature for each scene \cite{xu2022t-gnn,geng2023htn}, and alignment is applied to this single fixed-length feature.

\end{itemize}
\end{itemize}



\subsubsection{Domain-Specific Modules}
\label{sec: generalization DA/DG domain-specific modules}
\textbf{Method Recap.}
Another class of methods adapts to a new domain by freezing most model weights and introducing small domain-specific modules.
These modules are assumed to capture domain shifts through low-dimensional representations~\cite{li2018intrinsic}, allowing them to be learned with minimal additional parameters.
This approach is similar to techniques like residual adapters \cite{rebuffi2017resadapt} and low-rank adaptation (LoRA) \cite{hu2021lora}, and shares the similar use cases where limited target data is available.



\begin{figure}[!t]
    \centering
    \subcaptionbox{Fine-tuning a low-rank approximation \label{subfig:schematic_mosa}}[0.7\textwidth]{\includegraphics[width=0.7\textwidth]{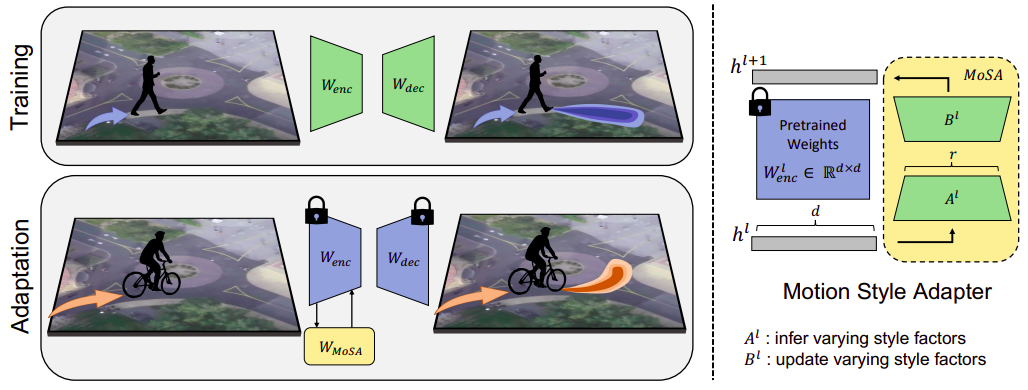}}
    \subcaptionbox{Train domain-specific submodules \label{subfig:schematic_FLN}}[0.7\textwidth][c]{\includegraphics[width=0.7\textwidth]{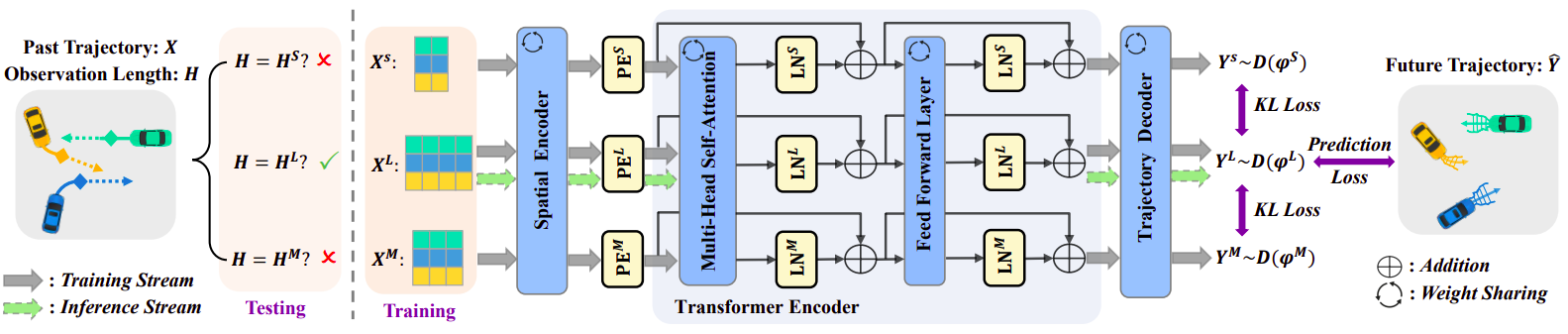}}
\caption{Examples of domain-specific modules in motion prediction. (a) MoSA \cite{kothari2023motionLora} add a low-rank connection layer added on the encoder and finetune the added layer on limited target domain data. (b) FLN \cite{xu2024adapting} routes target domain data to the subnetwork trained with the closest data length. See more examples and discussions in Section~\ref{sec: generalization DA/DG domain-specific modules}.}
\label{fig:domain_specific_modules}
\end{figure}

\textbf{Application in Motion Prediction.} Various types of domain-specific modules have been exploited in the motion prediction field, and we illustrate some examples in Figure~\ref{fig:domain_specific_modules}.

\begin{itemize}[label={\scriptsize$\bullet$}]
    \vspace{-0.5em}
    \item \textit{Style Modulator:} 
    In \citet{liu2022causalAux}, an encoder-decoder model is first trained on both source and target data, obtaining a domain-shared model. To learn domain-specific information, they further train another small encoder, named as style modulator, to extract a domain-specific style feature from each domain, by encoding the full trajectories (input and output) of each domain and applying contrastive loss to maximize cross-domain separation. Finally, they add the style modulator to the original encoder-decoder and finetune the modulator along with the decoder on the original prediction task.
    
    
    \vspace{-0.5em}
    \item \textit{Low-Rank Parallel Connections:} In MoSA \cite{kothari2023motionLora}, the authors propose to train a network on source data, freeze it, add low-rank parallel connections to the frozen encoder layers, and then only train these added weights using a limited amount of target data. 
    Unlike standard fine-tuning, which typically updates the decoder (e.g. final layers of the network) in an encoder-decoder model, this method instead add additional layers to adapt the encoder to align the input feature distribution across the source and target domains.
    They find that this approach outperforms fine-tuning of either the full model or the final layer only. 
    Notably, the authors employ separate encoders for the map and agents, and observe that distribution shifts can originate from either source. When working in a few-shot setting, they find that inserting adapters only into the encoder corresponding to the dominant source of shift—either the map or the agent—leads to better performance than injecting adapters to both encoders, presumably because this avoids overfitting.

    A similar work is LoRD~\cite{diehl2024lord}, which explores multi-task finetuning via low-rank adaptation of a differentiable autonomy stack comprising prediction, planning, and control, under a closed-loop evaluation setting. Notably, unlike methods that freeze weights of the main model, LoRD also trains the main model during fine-tuning.
    It finds that using a 25\% source and 75\% target data ratio yields the best performance, as this balances knowledge retention from the larger source dataset and mitigates catastrophic forgetting.
    

    


    \vspace{-0.2em}
    \item \textit{Velocity Refinement:} LaneTransformer++ \cite{wang2023bridging} improves the original LaneTransformer \cite{wang2023lane} for domain generalization by adding a velocity refinement module to account for large cross-domain variations. 
    \vspace{-0.2em}

    \item \textit{Dealing with varied trajectory length:} To specifically deal with the difference in sequence length when generalizing to unseen datasets, FLN \cite{xu2024adapting} trains multiple subnetworks for 3 different observation sequence lengths in the source datasets. During inference on unseen target domains, the subnetwork trained with the closest sequence length to the observed data is used.
    \vspace{-0.2em}

    \item \textit{Handling dataset-specific ground-truth error and noise:} \citet{park2024improving} finds that training on multiple dataset is hampered by domain-specific errors in the ground truth data (lateral position errors, ID switches, longitudinal errors). To train on multiple datasets at once, they augment their neural stochastic differential equation (NSDE) by training domain-specific noise modules that increase the magnitude of noise when outlier measurements are detected, making learning less sensitive to errors in ground truth values due to imperfect detection or tracking.
    
\end{itemize}

\subsubsection{Causal Learning} \label{sec:DA_causal_learning}


\textbf{Method Recap} Another set of methods approaches domain generalization from the perspectives of causal reasoning \cite{pearl2018bookofwhy}, which categorizes input–output relationships into two types:
1) \textbf{causal relationship}, which reflects the underlying data-generating process and are assumed to remain invariant across domains, and 2) \textbf{spurious correlations}, which capture statistical associations without causal justification and often vary with domain shifts.
For example, a red traffic light is a causal factor that directly influences a vehicle’s decision to stop—it is part of the underlying decision-making mechanism and is likely to be preserved across environments. In contrast, the fact that nearby vehicles are also slowing down may be correlated with the ego vehicle’s braking, but the actual cause is the traffic light, not the surrounding vehicles. This spurious correlation is domain-specific and may introduce undesirable biases into the model.
Focusing on learning causal relationships enables models to rely on stable, invariant signals, thereby improving generalization to unseen or changing environments.
Two methods that try to filter out spurious information are by optimizing for domain invariance through \textbf{invariant risk minimization} or by marginalizing over the possible spurrious effects through \textbf{causal intervention}.

\begin{figure}[!t]
    \centering
    \subcaptionbox{Original Causal Graph \label{fig:causal_orig}}[0.40\textwidth]{\includegraphics[height=0.2\textwidth]{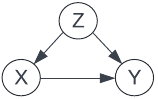}}
    \subcaptionbox{Causal Intervention \label{fig:causal_inter}}[0.40\textwidth][c]{\includegraphics[height=0.20\textwidth]{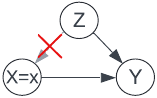}}
    \caption{Illustration of causal intervention for domain generalization and adaptation. (a) An unobserved confounder $Z$ influences both the input $X$ and output $Y$, resulting in learning spurious correlations between $X$ and $Y$ due to the backdoor path $X \leftarrow Z \rightarrow Y$. (b) Applying the causal intervention $do(X = x)$ blocks this path and isolates the true causal effect of $X$ on $Y$, improving generalization across domains. See detailed discussion in Section~\ref{sec:DA_causal_learning}.}
\label{fig:causal_intervention}
\end{figure}

\begin{itemize}[label={\scriptsize$\bullet$}]
    \vspace{-0.5em}
    \item 
\textbf{Invariant Risk Minimization} \label{sec:DA_causal_inter_IRM}
The first approach, Invariant Risk Minimization (IRM) \cite{arjovsky2019irm}, is based on the idea that causal features are stable across different domains, while spurious correlations often vary. Given access to multiple domains during training, IRM encourages the model to rely only on those features that consistently help with prediction in all domains.
Specifically, assume an encoder-decoder model where $x$ is the input, $y$ the output, $g$ the encoder, and $h$ the decoder. For each domain $d \in D$, IRM introduces a constraint: it requires that a \textit{single decoder} $\hat{h}$ should be optimal across all domains, even though the encoded features $g(x_d)$ may vary. This setup encourages the encoder to learn \textit{domain-invariant causal features}, rather than domain-specific spurious patterns. Formally, IRM solves the following problem:
\begin{align}
    \min_{g, \hat{h}} \frac{1}{|D|} \sum_{d}^{D}\mathcal{L}(y_d, \hat{h}(g(x_d))) 
    \quad \text{s.t.} \quad 
    \hat{h} \in \arg\min_h \mathcal{L}(y_d, h(g(x_d))) \quad \forall d \in D.
\end{align}
This means that for all domains, the shared decoder $\hat{h}$ must be optimal.
This contrasts with typical \textit{Empirical Risk Minimization (ERM)}, which minimizes the average loss over all data pooled from different domains without the "single optimal decoder" constraint. Thus, the model could be biased and could exploit spurious correlations that are predictive in a single domain but do not generalize. 
However, enforcing this constraint directly is computationally expensive, as it requires solving a separate optimization for each domain. To make training practical, IRM is often relaxed by \textit{adding a penalty}: it discourages the decoder from changing much when optimizing for individual domains. This gradient-based penalty encourages stability in the decoder across domains and helps the encoder focus on features that are truly invariant.
\begin{align}
    \min_{g,h} \dfrac{1}{\abs{D}} \sum_{d}^{D} \left[ \mathcal{L} \left(y_d,h(\phi_d)\right) + \lambda \norm{\nabla_{h}\mathcal{L} \left(y_d,h(\phi_d)\right)}_2^2 \right] \label{eq:irm_relaxed}
\end{align}

    \vspace{-0.5em}
    \item 
    \textbf{Causal Intervention} A second approach instead considers estimating and removing the effect of non-observed elements (cofounder) in each domain using causal intervention \cite{vanderweele2013causaldecomp,tang2020causalinter}. 
    Consider a causal model where we want to learn the mapping function $p(Y|X)$ between the observed inputs $X$ and outputs $Y$, while some unobserved parameter $Z$ is influencing both the inputs $X$ and outputs $Y$ (see Figure \ref{fig:causal_intervention}). This gives a directed graph $Z \rightarrow X \rightarrow Y \leftarrow Z$, and the actual relationship follows $p(Y|X,Z)$. 
    However, since $Z$ is unobserved, the learned mapping function $p(Y|X)$ can biased due to spurious correlations through $Z$. The learned function may appear predictive in one domain but fail to generalize to others where the distribution of $Z$ is different.
    To remove this spurious effect, causal inference introduces the concept of an \textit{intervention}~\cite{pearl2016causal}, denoted $do(X = x)$, which simulates setting $X$ to a fixed value independently of the unobserved cofounder $Z$. This breaks the backdoor path $X \leftarrow Z \rightarrow Y$, isolating the true effect of $X$ on $Y$.
    Using this intervention, the model instead estimates:
    \begin{align}
        p(Y \mid do(X)) &= \sum_z p(Y \mid X, Z) \cdot p(Z), \label{eq:causal_inter}
    \end{align}
    which marginalizes out the influence of the unobserved $Z$. Although $Z$ is not directly observed, it can be estimated or approximated using techniques such as variational inference or adversarial learning. By removing the effect of $Z$, the resulting model focuses on learning representations that reflect causal relationships, leading to improved generalization across domains with different underlying factors.
    

\end{itemize}

\begin{figure}[t]
    \centering
    \includegraphics[width=0.8\textwidth]{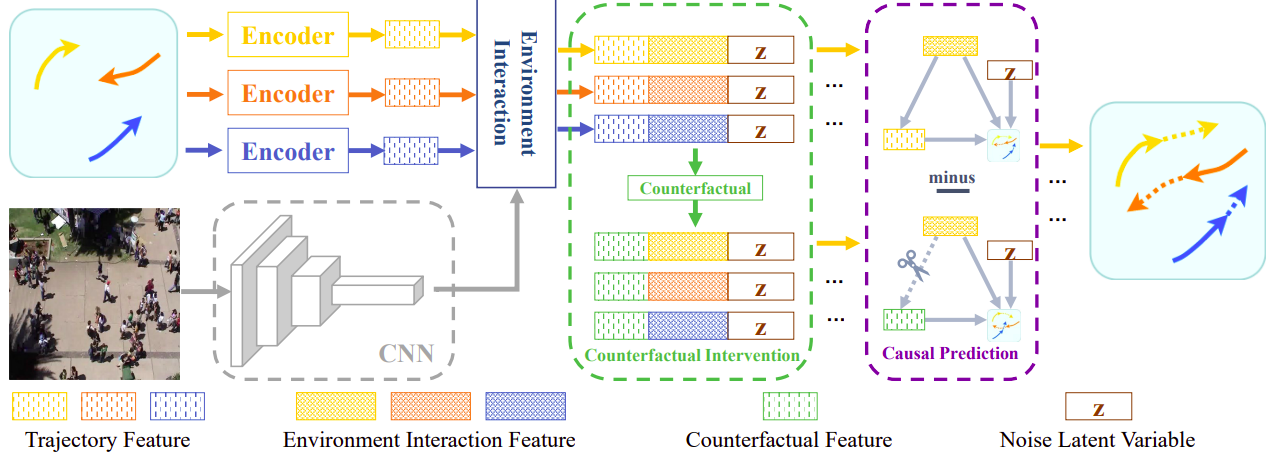}
    \caption{Illustration of counterfactual intervention from \citet{chen2021counterfactual}. To estimate the non-causal environmental bias, the model predicts a counterfactual trajectory by replacing the agent's input with synthetic motion (e.g., zero, mean, or random trajectory). The counterfactual prediction itself is treated as the bias, which is subtracted from the original prediction to produce an unbiased trajectory aligned with ground truth. See more discussions in Section~\ref{sec:DA_causal_learning}.}
    \label{fig:counterfactual_intervention}
\end{figure}

\textbf{Application in Motion Prediction} Both the invariant risk minimization and multiple forms of causal intervention have been used in motion prediction. While the IRM methods are easy to implement, they require access to data from multiple domains. In contrast, many causal intervention methods do not pose such requirements and thus can be used when only a single domain is available.

\begin{itemize}[label={\scriptsize$\bullet$}]
    \vspace{-0.5em}
\item \textbf{Invariant Risk Minimization} \citet{liu2022causalAux} use the relaxed IRM formulation (see Eq.~\ref{eq:irm_relaxed}) to obtain a domain invariant encoder-decoder which learns generalizable features. They then add a domains-specific style modulator to augment the features and improve performance on a specific domain. Similarly, CILF \cite{causality_ood} proposes to learn both invariant features with IRM and domain-specific features through a domain contrastive approach. Compared to \citet{liu2022causalAux}, CILF further separates the domain variant features into a useful causal component and a rejected spurious component. This separation is achieved by minimizing the prediction loss with invariant features augmented with causal variant features, and maximizing the loss when augmented with spurrious variant features.  
At inference, only the causal component is fused with domain-invariant features.




\vspace{-0.5em}
\item \textbf{Causal Intervention.} Recalling the causal intervention equation in Eq.~\ref{eq:causal_inter}, these approaches aim at estimating the unobserved element $Z$ and remove its effect during learning the mapping $p(Y|X)$. In this process, some approaches formulate $Z$ as a deterministic variable, while some approaches formulate it as a probabilistic distribution. 


\textit{Deterministic intervention:} \citet{chen2021counterfactual} assume that the operating environment used for training may significantly bias generating trajectories in ways that do not generalize to new domains, e.g. most pedestrians turning right at an intersection in the dataset does not mean that all pedestrian prefer to turn right in other datasets. They propose to use counterfactual intervention to obtain this environmental bias by replacing all of the agent’s input trajectory with counterfactual features such as uniform rectilinear motion (zero), mean trajectory, or random trajectory. 
Ideally, given zero or random input trajectories, the predicted trajectory should also be zero or random, and anything else would be environmental bias. By subtracting the counterfactual prediction bias from the original predicted trajectory (with original inputs), they obtain an unbiased trajectory prediction, which is then trained to match the ground truth. This method, 
as illustrated in Figure~\ref{fig:counterfactual_intervention}, adopt a deterministic intervention approach where it takes a single counterfactual prediction as the unobserved confounder.

\textit{Probabilistic intervention:} Other works take a probabilistic intervention approach, where they estimate the distribution of the unobserved confounder in different ways, and use sampling to instantiate marginalization.

\begin{itemize}

\item 
GCRL \cite{bagi2023generative} considers that, in the source domain, the unobserved confounder consists of both a domain-invariant component and a domain-variant component, and models them as a single Gaussian distribution and a Gaussian mixture distribution respectively. 
These distributions are approximated by encoding the input trajectories using the Evidence Lower Bound (ELBO) loss. At inference, they obtain latent distributions for both and perform sampling to marginalize over them. When transferred to a new target domain, new parameters for the domain variant latents are learned, while the rest of the network is conserved. 


\item 
SEAD \cite{ge2023causal} approximates non-causal environmental effects as a set of learnable vectors instead of latent distributions. Cross-attention between these vectors and the encoded input yields a set of offset, which are taken as multiple samples of the non-causal features, averaged, and then added to the original encoded trajectory feature, capturing the mean environmental effect. 

    \item 
    CaDeT \cite{pourkeshavarz2024cadet} applies self-attention to generate a weight mask that identifies the most relevant features to the task, which are regarded as casual features, and the non-causal features are identified by simply inverting the mask. Moreover, the non-causal features are formulated as a distribution by measuring the mean and variance in a batch of data. To enforce causal intervention, multiple non-causal features are sampled from the distribution and then combined with causal features as the model inputs, and the model is trained to be robust to non-causal features by minimizing the variance of the predicted output position across multiple sampled non-causal features.

\end{itemize}
\end{itemize}

\subsubsection{Few-Shot and Test-Time Adaptation}
\label{sec: generalizable DA/DG adaptation}
\begin{figure}[!t]
    \centering
    \subcaptionbox{Optimize parameters online, taken from 
    HATN \cite{wang2022transferable} \label{subfig:schematic_HATN}}[0.48\textwidth]{\includegraphics[width=0.48\textwidth]{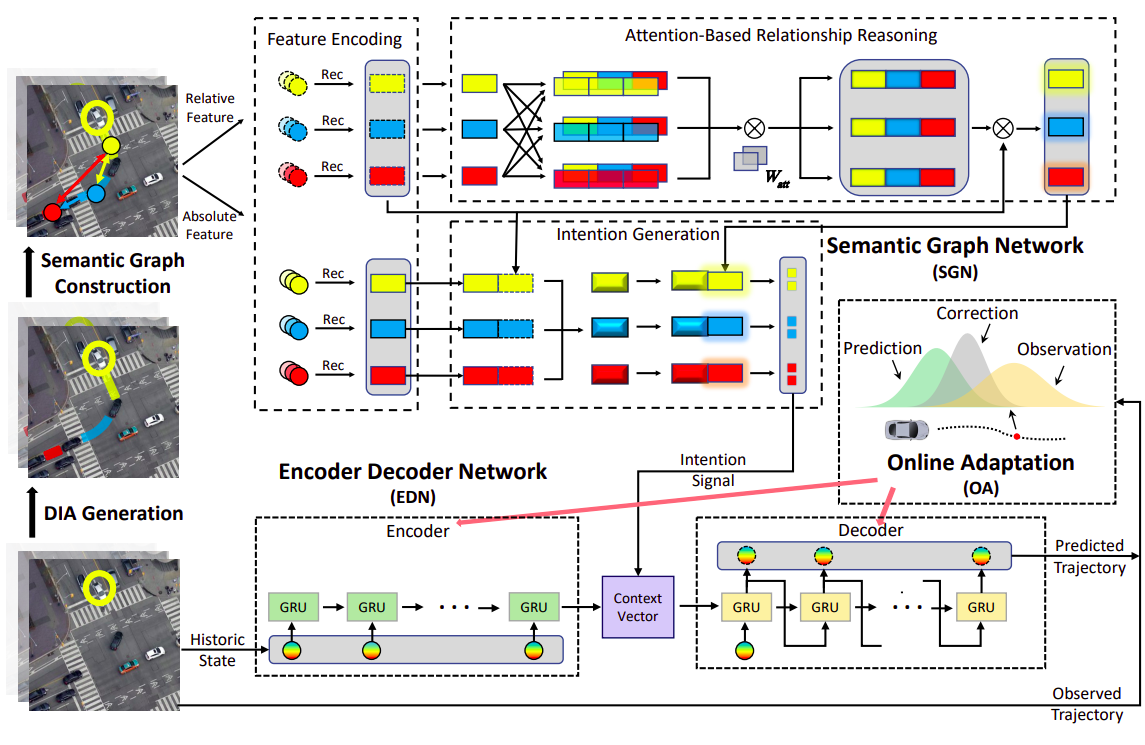}}
    \subcaptionbox{Multi-hop attention with past examples in a memory, taken from 
    OML-PTP \cite{yang2024fast} \label{subfig:schematic_omlptp}}[0.48\textwidth][c]{\includegraphics[width=0.48\textwidth]{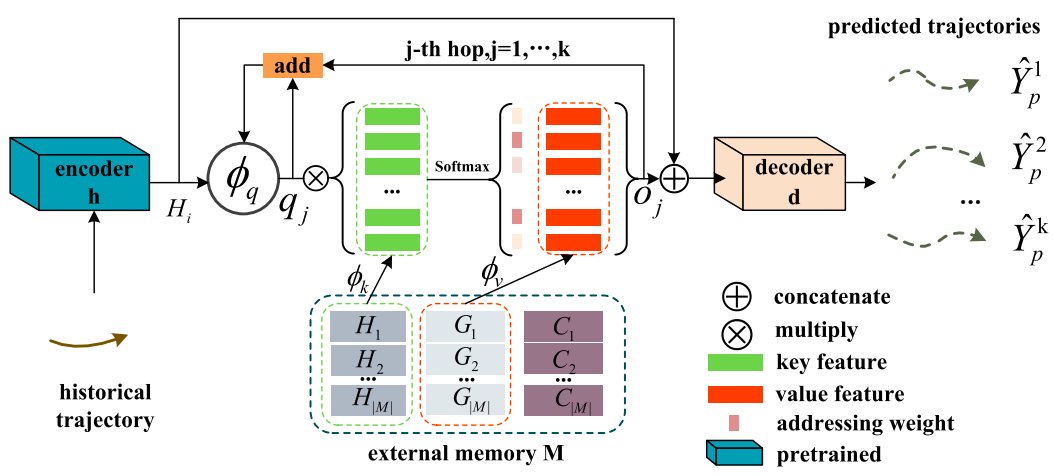}}
\caption{Examples of few-shot and test-time adaptation. \subref{subfig:schematic_HATN} HATN \cite{wang2022transferable} online adapt the model parameters according to the prediction errors in streaming data. \subref{subfig:schematic_omlptp} OML-PTP \cite{yang2024fast} uses a memory bank of past trajectories to inform the current prediction.}
\label{fig:fast_test_time_adapt}
\end{figure}

\textbf{Method Recap}
Another possible test case in domain adaptation involves adapting to few-shot offline data or streaming test-time data. 
This is similar to the continual learning problem that will be discussed in Section \ref{sec:continual-learning}, but with some key differences. 
Domain adaptation methods discussed in this section seek to adapt a model to a new target domain with very limited data, learning a domain-specific model, while 
continual learning deals with training on novel tasks while maintaining performance on old tasks to learn a more general model.
In this section, we discuss three different adaptation approaches: methods that use \textbf{online optimization} to directly update a few parameters, methods that use a \textbf{meta-learning approach} to facilitate adaptation in few-shot and online learning, and methods that keep an \textbf{online memory} of pervious data as references for the current domain. We illustrate examples of there approaches in Figure~\ref{fig:fast_test_time_adapt}.

\begin{itemize}[label={\scriptsize$\bullet$}]
    \vspace{-0.5em}
    \item \textbf{Online Optimization} This approach updates the prediction model during test time using streaming observations. As the streaminig observation arrives, it is taken as the ground-truth for a previous prediction, and an online optimization algorithm, such as stochastic gradient descent (SGD) or the Extended Kalman Filter (EKF)~\cite{abuduweili2020robust, abuduweili2021robust}, is applied to minimize the error~\cite{blum2005line}, thereby improving prediction accuracy on future inputs.
    
    
    \vspace{-0.5em}
    \item \textbf{Meta Learning}
    As an advanced method compared to online optimization, meta-learning~\cite{finn2017maml} poses different objectives during the training stage, to enable the model adapt more quickly to a small amount of data in the new domain: In typical training, the loss is to minimize the errors of the source domain data; In meta learning, this objective is modified to train a model that can rapidly adapt to new tasks using only a small amount of data and a few gradient updates.
    The core idea is to learn an initialization of parameters that are sensitive to task-specific changes—so that small updates, informed by limited task data, yield large improvements in performance. Specifically, during meta-training, the model is exposed to a series of tasks; for each task $i$, it performs an inner-loop optimization using a small data batch, simulating the adaptation process: 
    \begin{equation}
        \theta'_i=\theta-\alpha \nabla_\theta  \mathcal{L}_i(\theta)
    \end{equation}
        
    where $\theta$ and $\theta'_i$ denote the model parameter before and after the adaptation respectively, $\alpha$ denotes the step size, and $\mathcal{L}_i(\theta)$ denotes the loss function for the data in task $i$ under model parameter $\theta$.
    The outer loop then optimizes the parameter initialization $\theta$ to ensure that the inner-loop adaptation is effective across all tasks. 
    \begin{equation}
    \min_\theta \sum_{i} \mathcal{L} ( {\theta_i'}) 
    = \min_\theta \sum_{i}  \mathcal{L} ( {\theta - \alpha \nabla_\theta \mathcal{L}_i(\theta)})
    \end{equation}
    In effect, the model is trained to be easy to fine-tune: it acquires general-purpose representations that can be quickly adapted to new environments.
    Meta learning was initially developed for offline few-shot learning, but can also be used in the online streaming setting.

    \vspace{-0.5em}
    \item \textbf{Memory-Aided Network} Another set of methods assumes that using the previously observed trajectories is helpful to predict at the current time. Thus, they keep a memory buffer of previously observed trajectory input and output. When facing a new trajectory input, they select useful previously seen outputs as a reference by evaluating the similarity of the current input (query) with the previous input (key) in the memory buffer. Both the current input and the selected output trajectories are fed in the prediction encoder, with the assumption that previous output trajectories are informative to of current prediction given similar inputs.
\end{itemize}

\textbf{Application in Motion Prediction}
All these methods have been implemented in motion prediction and have been shown to be effective for adaptation. However, it is interesting to highlight some key differences. Online optimization and meta-learning approaches are used to update model parameters without changing the model architecture, while memory-aided methods change the inputs to the network and thus require changes in the model architecture (a different encoder), introducing more design efforts on the model structure.



\begin{itemize}[label={\scriptsize$\bullet$}]
    \vspace{-0.5em}
    \item \textbf{Online Optimization}
The online optimization approach  HATN \cite{wang2021hierarchical,wang2022transferable,wang2021online} proposes to use a modification of the extended Kalman filter (EKF), which considers the error over multiple past prediction time steps, to find an optimal update of the model weights \cite{abuduweili2021robust} such that the predicted trajectory would have minimal error. The authors propose new metrics to comprehensively analyze adaptation design choices, such as adaptation time steps, and layers to adapt, which shows the adaptation effectively improves performance particularly in short-horizon prediction, when transferring to previously unseen scenarios. 
Similarly, \citet{ivanovic2023expanding} presents another adaptation approach by updating the final layer of the network, using a combination of Kalman filter-like optimal update and gradient descent optimization. Both online and offline adaptation are explored.
Another online method T4P \cite{park2024t4p} adopts the transformer architecture and a test-time training approach. At test time, they add new agent-specific tokens to the original model, and optimize these tokens using streaming data. \
During the adaptation, in addition to typical regression loss, the reconstruction losses is also used, with lane embeddings, past and future trajectory tokens being reconstructed. 
For multi-agent prediction, CoGNN \cite{li2022online} build a graph neural network that explicitly considers the agents as nodes and interaction between pairs of agents as edges. The node encoder and edge weights are updated online.


    \vspace{-0.5em}
    \item \textbf{Meta-Learning}
    \citet{moon2021fastadapt} consider the problem of human motion prediction for human-robot interaction, learning a user-specific embedding vector that individually adapts to the specific human entity.
    In the inner-loop training, the encoder is fixed and the user-specific embeddings is optimized to fit a specific individual's behaviour, while the outer-loop training update the shared encoder across users.
    Another meta-learning approach is MetaTraj \cite{shi2023metatraj}. In the inner loop training, it constructs a sub-task by splitting the input trajectory into input and output subsequence, and updates only the decoder parameters. In the outer loop, the standard prediction task is used to update the entire network. Since MetaTraj relies only on input trajectories to construct sub-tasks, it can perform adaptation even in the absence of ground-truth labels in the target domain.
    

    \vspace{-0.5em}
    \item \textbf{Memory-Aided Networks}
    For memory methods, MANTRA \cite{marchetti2020mantra} modifies the model architecture to also take the previous prediction as additional model inputs. Specifically, it appends the k-nearest prediction in the memory to the current encoded input trajectory as the whole model input, with the goal that the previous prediction could inform the current case.
    The memory is updated with streaming data, namely when the ground truth for a previous prediction is available.
    Notably, they propose to keep trajectories with larger errors in the memory as these represent cases that are poorly modelled. OATMem \cite{huynh2023online} consider a similar approach for motion estimation in videos, but finds that a simple first-in-first-out strategy is sufficient for maintaining the memory queue in that setting. 
    They also use an ensemble of networks trained independently on different domains, and keep a small and consistent memory across the ensemble by k-means clustering. 
    To select more relevant predictions from the memory, OML-PTP \cite{yang2024fast} improves on MANTRA by encoding the previous inputs in the memory and current inputs to features via multi-hop attention \cite{sukhbaatar2015end}, and calculate the similarity in the feature space when selecting previous inputs.
    
    \end{itemize}

\subsubsection{Perspectives}


\textbf{Revisiting the Unlabelled Data Assumption: From Images to Trajectories}
One thread of domain adaptation methods considers the case where only unlabeled data is available in the target domain, which is a common setting in vision tasks such as image classification. 
In such vision tasks, the input images and the output labels have different representations, 
making image collection relatively easy, while obtaining ground-truth labels (e.g., class names) is often time-consuming and expensive.
As a result, it is common to have datasets with abundant unlabeled images, and standard domain adaptation tasks assume access to such unlabeled target-domain data.
Similarly, in motion prediction, there are works~\cite{poibrenski2023uncertainty,mahajan2021domainclustering}, 
 that assumes access to only trajectory inputs in the target domain and uses self-labelling mechanisum, such as Mean Teacher approach \cite{tarvainen2017meanteacher} or input feature matching between source and target domains~\cite{mahajan2021domainclustering}, to generate pseudo trajectories labels for training.
However, in many motion prediction tasks, the input and output share the same representation, time series of states, and are typically available or unavailable simultaneously. Thus the assumption of access to only the unlabeled input motion data is somewhat artificial: for offline data, it is unlikely that input trajectories are available but output trajectories are not, as they are simply different temporal splits of a whole motion trajectory; for online data, if some delay is allowed, the ground truth for earlier predictions is usually accessible, because as the robot keeps operating, the new detected objects state will be the prediction target of a previous time. When using alternative representations for motion prediction other than trajctories—such as occupancy maps, point clouds, or video—the input and output also share the same format, and are therefore again either available or unavailable simultaneously.
Therefore, a more common and meaningful domain adaptation setting for motion prediction would operate either with \textit{limited target data} or \textit{streaming labelled target data}, instead of large scale unlabelled as seen in the vision tasks.
In such cases, the key challenge is how to effectively leverage the available target data to learn transferable features or efficiently adapt the model to new domain, while mitigating the risk of overfitting to the small sample size.

 \textbf{Dataset Discrepancy and the Need for Standardized Evaluation} In principle, domain generalization and adaptation study how models perform when transferred across datasets. While this evaluation paradigm has been well established in computer vision with popular and standardized benchmarks available, it remains underdeveloped in motion prediction, where datasets often differ significantly in settings and formats, such as trajectory horizons and map representations. Therefore, cross-dataset evaluation has been largely infeasible in motion prediction, and most prior works utilize customized data splits for own evaluation. This creates a large barrier in fairly comparing methods and assessing the progress. 
 A few methods \cite{park2024improving,bae2024singulartrajectory,feng2025unitraj,zhou2024smartpretrain} have seeked to unify dataset formats and perform cross-dataset evaluation, but an official benchmark does not yet exist.
 We argue that establishing such a benchmark is essential for advancing domain generalization/adaptation research in motion prediction.
 
 
 

\subsection{Continual Learning}
\label{sec:continual-learning}

Another task that deals with adapting networks to new data distributions is continual learning. Continual learning explores the problem of adapting a network’s parameters as the test distribution changes over time. This is similar to domain adaptation, in which we want to transfer performance from a source domain to a target domain and generally evaluate only on the target data. However, continual learning adopt a life-long learning angle, and aims at dealing with the problem of catastrophic forgetting~\cite{lopez2017gem}, where the a neural network will fit the current new data at the cost of dropped performance on old seen data. That is to say, continual learning aims to \textbf{adapt to new tasks while maintaining performance on previous tasks}.
Moreover, continual learning also differs from domain adaptation as it considers a wider range of new tasks. For example, in typical computer vision classification tasks, continual learning can be split into different scenarios corresponding to how the task changes over time \cite{van2022three}: 1) \textit{Domain-incremental learning} deals with a changing input distribution but the same task and classes, and is the most similar to domain adaptation. 2) \textit{Class-incremental learning} deals with an increase in the output space as new classes are observed in addition to changes in the input, but the actual task remains the same. 3) \textit{Task-incremental learning} deals with learning completely new output spaces and is the most difficult.

In motion prediction, the most common scenario is domain-incremental learning, dealing with changes in the inputs, such as the environments or agent behaviour. While class-incremental learning is relatively common for classification tasks and their derivatives (e.g. detection, segmentation), it is less frequent for motion prediction. 
However, in settings involving multiple agent types—such as cars, pedestrians, and bicycles—extending a model to accommodate new agent categories naturally resembles a class-incremental learning problem.


In this section, we introduce the existing continual learning methods into two broad categories. \textbf{Replay methods} store or generate representative examples of data from old tasks to evaluate on, ensuring that the performance on those older tasks does not decay. \textbf{Regularization} methods attempt to limit the changes to network parameters that are critical to past tasks. 
Notably, another approach, the \textbf{parameter isolation} methods also exist in the literature but have not been used in motion prediction to our knowledge. This approach incrementally expands the network by adding task-specific elements that are not updated when new tasks are seen, or incrementally mask and freeze the weights that are important to past tasks \cite{mallya2018packnet}. The expanding networks are similar to the domain-specific module discussed in Section~\ref{sec: generalizable DA/DG auxiliary task} above.

\begin{figure}
\begin{minipage}{.60\textwidth}
    \centering
    \captionof{table}{Taxonomy of continual learning methods based on the approach used and domain of application. We discuss approaches including example replay (ER), generative replay (GR) and regularization (Reg). The applications are vehicle forecasting (VF) and pedestrian forecasting (PF).}
    \label{tab:cont_learn}
    \begin{tabular}{@{}l|ccc|cc}
    \toprule\toprule
    \multirow{2}{*}{Method} & \multicolumn{3}{c|}{Approach} & \multicolumn{2}{c}{Application} \\  
     & ER & GR & Reg & VF & PF \\ \midrule
    CLTP-MAN \cite{yang2022continual}                       & \checkmark &   &   &   & \checkmark\\
    D-GSM \cite{lin2023continualDGSM}                       & \checkmark &   &   & \checkmark &  \\
    OMBR\textsuperscript{$\dagger$} \cite{kang2024continual}  & \checkmark &   &   & \checkmark &  \\
    CGM\textsuperscript{$\dagger$} \cite{ma2021continual}     &   & \checkmark &   & \checkmark &  \\
    GRTP \cite{bao2023lifelong}                             &   & \checkmark &   & \checkmark &  \\
    Social-GR \cite{wu2022continualsocialGR}                &   & \checkmark &   &   & \checkmark\\
    CTP-UGR  \cite{feng2023continual}                       &   & \checkmark &   & \checkmark &  \\
    TPwF\textsuperscript{$\dagger$} \cite{zhi2023adaptive}    &   &   & \checkmark &   & \checkmark\\
    SCL \cite{knoedler2022improving}                        & \checkmark &   & \checkmark &   & \checkmark\\
    \bottomrule\bottomrule
    \end{tabular}
\end{minipage}
\hfill
\begin{minipage}{.35\textwidth}
    \includegraphics[width=0.8\textwidth]{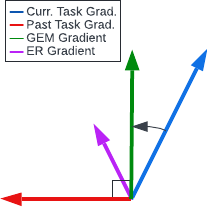}
    \captionof{figure}{Illustrations of data replay methods for continual learning. In gradient episodic memory (GEM) \cite{lopez2017gem}, the gradient is projected to non-negative with past gradients, while in experience replay (ER) \cite{rolnick2019experiencereplay}, the gradient is simply averaged over current and past tasks.}
    \label{fig:gem_er_diff}
    
\end{minipage}
\end{figure}

\subsubsection{Replay Methods}
\textbf{Method Recap} The replay methods include two approaches. First, the \textbf{example replay} methods, also know as rehearsal or memory-based methods, reduce forgetting by training on a mix of both new and old data. One early work was gradient episodic memory (GEM) \cite{lopez2017gem}, which ensures that performance on past tasks is conserved by enforcing an inequality constraint on the loss from old examples in the memory bank.
In practice, this is implemented by projecting the gradient in a direction that does not increase the loss on past tasks. 
This is repeated for each previous task. In an extension of the method, averaged gradient episodic memory (A-GEM) \cite{chaudhry2018agem}, instead of optimizing for each previous task sequentially, it randomly samples a batch of data from all old tasks in the memory, and avoids increasing the average loss during optimizing for new tasks.
Experience replay (ER) \cite{rolnick2019experiencereplay} further relaxes the strict requirements of non-negative gradient updates by simply including samples from both the old and new tasks as a training batch and optimize their averaged loss.
We illustrate the difference between GEM and ER in Figure~\ref{fig:gem_er_diff}. GSS \cite{aljundi2019gss} extends ER by preferentially storing samples that are different from existing samples in the memory, ensuring a good sample diversity.



An important issue of replay methods is that they still require storing past examples in a memory. As the number of seen tasks and data increases, a fixed memory becomes insufficient to retain past performance, while expanding memory would eventually exceed practical limits.
To avoid this memory issue, \textbf{generative replay} methods have been proposed to synthesize examples of old tasks instead of storing them. One example work is Generative replay (GR) \cite{shin2017dgr}, which considers the iterative training of a generator and an annotator. At each iteration, a new generator is trained with a GAN strategy to generate examples similar to both the current task and those generated by a previous generator iteration. Similarly, an annotator is trained to label both current and generated examples. Conditional generative replay (CGR) \cite{lesort2019cgr} removes the need for the annotator by making the generation class-conditional. This also enables control of the sample balance among past tasks during generation. Several methods \cite{liu2020generative_feature,peng2023BioSLAM} further propose to generate only the latent feature embeddings instead of raw data, to reduce the expense when generating large images or the full data distribution.


\begin{figure}[t]
    \centering
    \includegraphics[width=0.8\textwidth]{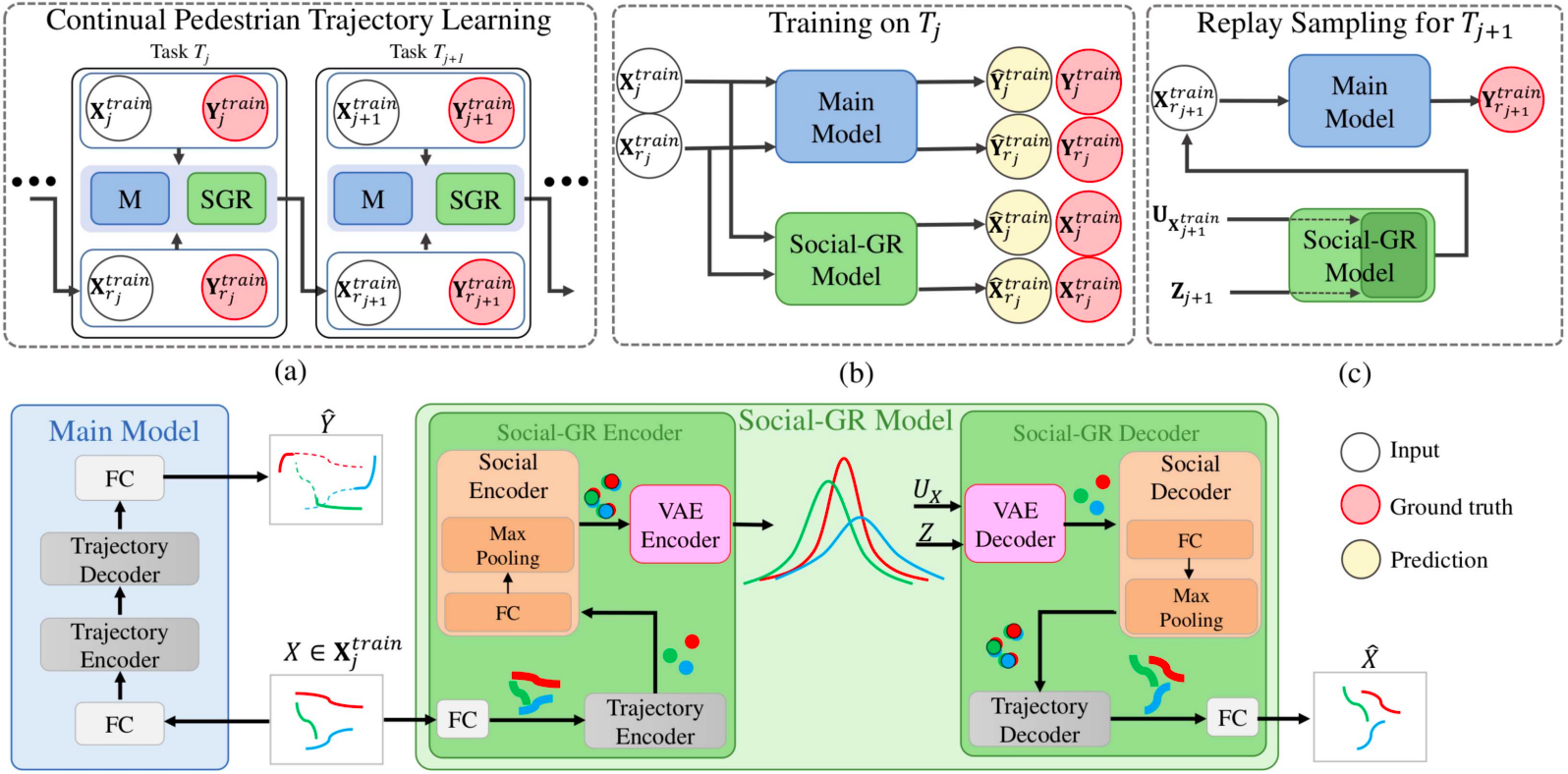}
    \caption{An example of the generative replay method for continual learning, taken from Social-GR \cite{wu2022continualsocialGR}. When generating old-domain data, Social-GR directly conditions on the initial waypoint of new-domain trajectory, avoiding having to keep a memory to store old-domain data.
    }
    
    \label{fig:gen_replay}
\end{figure}

\textbf{Applications in Motion Prediction}
Both example replay and generative replay methods have been used in motion prediction. While there are theoretical storage limitations for the example replay methods, there is very little actual discussion of its importance in the literature.
There is also little discussion on evaluating the quality of generated examples.

\begin{itemize}[label={\scriptsize$\bullet$}]
    \vspace{-0.5em}
    \item \textbf{Example Replay} 
    Based on the GEM approach~\cite{lopez2017gem}, dynamic gradient scenario memory (D-GSM) \cite{lin2023continualDGSM} considers continual learning across multiple motion datasets sequentially, each collected at one driving scenario, and maintains performance on old datasets by keeping a memory of past data and ensures that the training gradient on the current scenario does not cause an increase in the loss for past examples. Considering that the available memory is limited and some scenarios could be similar, the authors also propose to scale the number of past-scenario examples kept in the memory as a function of the divergence \cite{bao2023lifelong} between the past scenarios and the current one, removing similar scenarios in the memory and instead storing more diverse scenarios.
    

    CLTP-MAN \cite{yang2022continual} keeps a coreset of past examples, where $1\%$ of trajectories in each task are stored. In addition to this continual learning design, they also leverage the domain adaptation technique, specifically the memory-aided network from MANTRA \cite{marchetti2020mantra} and OML-PTP \cite{yang2024fast} as discussed in Section \ref{sec: generalizable DA/DG adaptation} above, where they use keep a second memory queue of examples from the current task and modify the network to also take these examples to inform current prediction.
    

    
    
    \citet{kang2024continual} is another approach that combines continual learning and domain adaptation techniques. For continual learning, they employ experience replay by maintaining a buffer of past examples, which is updated when current samples differ significantly from those previously stored.  For the domain adaptation technique, they train the base method using a meta-learning strategy as discussed in Section~\ref{sec: generalizable DA/DG adaptation} above. The motivation is that, the meta-learning approach encourages the model to adapt to new domains with minimal updates, thereby reducing performance degradation on previous domains and improving the stability of continual learning.
    Additionally, they also encourage the model to use different weights for different tasks by adding a mask generator to the backbone, with the motivation that only each portion of the model parameter is responsible for one task and that only a small portion of parameters need to be adapted for a new task, maintaining its performance on old samples. In practice, the mask generator is trained to maximize the number of hidden weights for each given sample. This mask generator design is similar to the parameter isolation methods (e.g. \cite{mallya2018packnet}) and the mixture of experts design~\cite{jacobs1991adaptive,shazeer2017outrageously,artetxe2021efficient}.
    
    
    
    
    
    
    
    \vspace{-0.5em}

    \item \textbf{Generative Replay} To reduce the requirements for storing large amounts of examples, other methods instead use generative replay. \citet{ma2021continual} propose to use a conditional generative memory: for each past data, they only store the start, an intermediate and the end point of the trajectories into the memory bank and generate the full input and output trajectory using a conditional variational autoencoder (CVAE) \cite{kingma2014cvae}. 
    In their ablation studies, the authors find that generative replay trained without such conditions (using purely noise) yields only marginal improvements over the baseline, highlighting the importance of conditional generation.
    Another conditional generation approach is GRTP \cite{bao2023lifelong}, where trajectories are modeled as mixtures of Gaussians \cite{bishop1994mixture}, and generated conditioned on the past trajectory and the relative position of other vehicles. 

    While the above methods reduce the size of the required memory, Social-GR \cite{wu2022continualsocialGR} completely eliminates it, which is illustrated in Figure~\ref{fig:gen_replay}. When generating old-domain data, instead of conditioning on points from old domains and thus requiring a storage memory, they directly condition on the initial waypoint of new-domain trajectory, avoiding having to keep a memory to store old-domain data.

\end{itemize}

\subsubsection{Regularization}
\textbf{Method Recap} Next, the regularization methods add constraints to the update magnitude of the network parameters. This is similar to the gradient constraint in GEM~\cite{lopez2017gem}, but the constraint is enforced by simply limiting changes to the network weights instead of using past examples to navigate the gradient direction. For Learning without forgetting (LwF) \cite{li2017lwf} keeps two versions of the model, where the first model is immediately updated given a new data sample, and the second model is fixed. During training in the new domain, the predictions of the first model are enforced to be similar to those of the delayed model to avoid overly big changes in the parameter.
Another approach is elastic weight constraint (EWC) \cite{kirkpatrick2017ewc}, which assumes that networks are overparametrized and it is possible to improve performance on new tasks by only updating weights that are not critical for previous tasks. The Fisher information is used to determine the importance of network weights to the tasks. However, as the number of tasks grows, the model can suffer from saturation problems—most weights have been used for previous tasks, leaving little capacity for learning new ones.




\textbf{Application in Motion Prediction}
Applying the regularization method to reduce catastrophic forgetting, 
\citet{zhi2023adaptive} consider the case where there is a large amount of training data on a source domain and little data on a target domain. They leverage both continual learning and domain adaptation techniques, using an EWC-like regularizer to avoid large updates to weights that are critical to previous tasks and a maximum mean discrepancy (MMD) loss to align feature distributions across domains. Another method, SCL \cite{knoedler2022improving}, combines both regularization and example replay for continual learning. They use EWC to penalize changes in weights that are critically important for previous tasks. They also keep a memory of old-domain examples, which are trained together with new-domain examples to balance performance across domains. 


\subsubsection{Perspectives}
\textbf{Task Switch Detection} Many of the prediction methods discussed above consider task-aware continual learning, where tasks boundaries (e.g. change in domain) are known during training and testing. However, this is a strong assumption that may not hold in real world settings, where an agent could be exposed to novel scenarios without forewarning. This is know as task-free continual learning \cite{aljundi2019task}. Identifying when this occurs can be related to model uncertainty and out-of-distribution detection model, as discussed in the next Section \ref{sec:OOD_det_gen}.

\textbf{Streaming/Online Continual Learning} Moreover, many of the methods discussed above consider the offline continual learning setting, where the target-domain new data is available in batches and offline training is performed. However, the online or test-time continual learning setting is not considered, where the learning is performed on streaming data.  
The online adaptation methods discussed in section~\ref{sec: generalizable DA/DG adaptation} provide viable solutions for this online setting, but they focus solely on the performance on the new data and ignore the performance on past tasks. Integrating this online adaptability into the continual learning framework via example replay~\cite{aljundi2019gss,aljundi2019online} or constraints on parameter change \cite{lopez2017gem,wang2022continual}
remains an underexplored direction in the motion prediction domain.




\subsection{Out-of-Distribution Detection and Generalization} \label{sec:OOD_det_gen}

In the open-world setting, networks may be exposed to inputs drawn from a data distribution that differs from the one used during training and validation. 
To ensure robust and safe performance in real-world applications, models must be able to handle such out-of-distribution (OOD) inputs. 
There are two main approaches to addressing this challenge: \textbf{OOD detection} identifies whether a given input is in-distribution (ID) or out-of-distribution (OOD), potentially rejects the OOD inputs; in contrast, \textbf{OOD generalization} goes a step further—it aims not only to detect OOD data but also to maintain accurate predictions in their presence, either by enhancing model robustness or by triggering appropriate fallback strategies.
In fact, the definition of OOD generalization is somewhat ambiguous in the literature, and its usage can vary across works. For example, OOD generalization is closely related to the task of domain generalization discussed in Section~\ref{sec:domain-adaptation-and-generalization}, as both aim to ensure strong performance in previously unseen domains. Some works use the two terms interchangeably~\cite{diehl2024lord}.  
In this survey, we adopt a more precise definition of OOD generalization, emphasizing its dual goals of detecting and being robust to distributional shifts, unlike domain generalization, which focuses solely on achieving robust performance yet without explicitly identifying OOD inputs.

In motion prediction, examples of out-of-distribution data can include novel agent type (e.g. animals), novel agent behavior (e.g. pedestrians moving with mobility aids), or novel environmental geometry (e.g. an unseen roundabout), or changes to the traffic regulations (e.g. left-hand driving vs. right-hand driving). Some illustrations are provided in Figure \ref{fig:distribution_shifts}.
Given such diverse sources and varying magnitudes of distributional shift, 
it is critical to accurately and reliably quantify the degree of out-of-distribution (OOD) deviation. 
Such measurement enables principled decisions on whether the observed shift exceeds a threshold warranting rejection via OOD detection, or remains sufficiently small to permit the application of OOD generalization methods, thereby allowing the model to generate reliable predictions despite minor discrepancies from the training distribution.

\begin{table}[!t]
\centering
\caption{Taxonomy of OOD detection and generalization methods. The approaches for OOD detection are \emph{a posteriori} error estimation (Post), inlier-outlier classification (In-Out), evidential deep learning (Evid), Monte Carlo dropout (MC Drop), ensemble disagreement (ED), likelyhood estimation (Likely) and reconstruction by generative methods (Recon), while for OOD generalization, the approaches are routing and fusion.}
\label{tab:odd_methods}
\resizebox{1.0\linewidth}{!}{
\begin{tabular}{@{}l|cc|ccc|cc|cc}
\toprule\toprule
\multirow{3}{*}{Method} & \multicolumn{7}{c|}{OOD Detection} & \multicolumn{2}{c}{OOD Generalization} \\ 
\cmidrule{2-10}
& \multicolumn{2}{c|}{Label-Dependent Met.} & \multicolumn{3}{c|}{Bayesian Met.} & \multicolumn{2}{c|}{Generative Met.} & \multirow{2}{*}{Routing} & \multirow{2}{*}{Fusion}  \\
 & Post & In-Out & Evid & MC Drop & ED & Likely & Recon &  &  \\ \midrule
 \citet{normalizing_flows_anomaly_detection}    &   &   &   &   &   & x &   &   &   \\
 \citet{itkina2022interpretable}                &   &   & x &   &   &   &   &   &   \\
 \citet{Shao2023FailureDF}                      &   &   & x &   & x &   &   &   &   \\
 QAD \cite{farid2023task}                       & x &   &   &   &   &   &   &   &   \\
 \citet{wiederer2023joodu}                      &   &   &   &   &   & x &   &   &   \\ \midrule
 \citet{Sun2021OnCE}                            & x & x &   &   & x &   & x & x &   \\
 \citet{hao2023improving}                       &   & x &   &   &   &   &   & x &   \\
 APE \cite{li2024adaptive}                      &   & x &   &   &   &   &   & x &   \\
 RuleFuser \cite{patrikar2024rulefuser}         &   &   & x &   &   &   &   &   & x \\
\bottomrule\bottomrule
\end{tabular}
} 
\end{table}

\subsubsection{OOD Detection} \label{sec: ood detection}

\textbf{Method Recap}
Out-of-distribution (OOD) detection refers to the task of identifying inputs that lie outside the training distribution, for which the model’s predictions may be unreliable. As previously discussed in Section~\ref{sec: prediction uncertainty}, two types of uncertainty exist: 1) epistemic uncertainty, stemming from limited knowledge and reducible through additional data, comprehensive feature design, and enlarged model capacity, and 2) aleatory uncertainty, resulting from inherent stochasticity in the problem (e.g. human's natural irrationality or errors in track data due to sensor noise) and irreducible by nature. Epistemic uncertainty is often the focus of OOD detection because models naturally have low knowledge of data outside the training distribution. In this section, we discuss three broad categories of OOD detection methods: 1) label-dependent methods that require access to ground-truth or OOD examples, 2) Bayesian methods that estimate epistemic uncertainty by modelling distributions over model parameters or outputs, and 3) generative methods that aim to model the underlying data distribution.

\begin{figure}[!t]
    \centering
    \subcaptionbox{Sampling \label{fig:dirichlet_sampled}}[0.29\textwidth]{\includegraphics[width=0.29\textwidth]{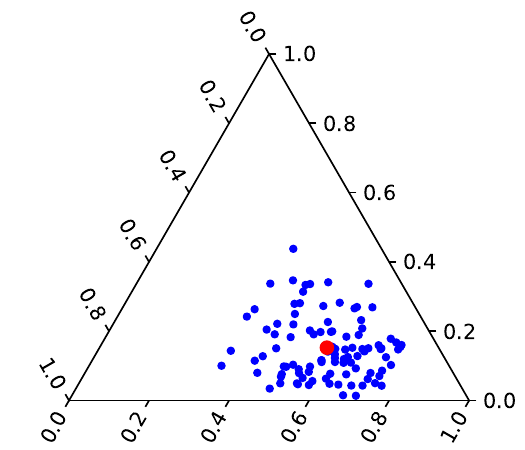}}
    \hspace{20pt}
    \subcaptionbox{Distribution \label{fig:dirichlet_dist}}[0.29\textwidth]{\includegraphics[width=0.29\textwidth]{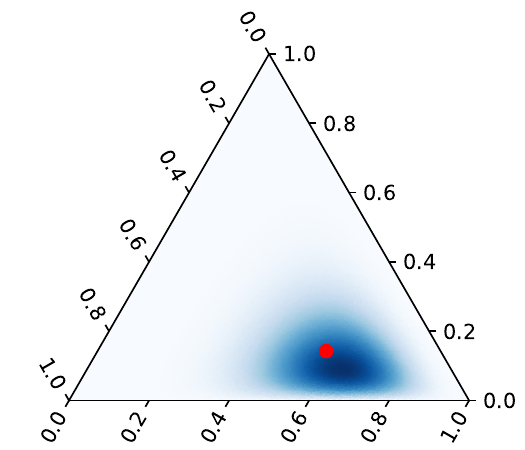}}
\caption{Illustration of common implementation for Bayesian methods. \subref{fig:dirichlet_sampled} Ensembles and Monte Carlo dropout sample/generate multiple outputs for a single input data. \subref{fig:dirichlet_dist} Evidential learning outputs an output distribution of likelihoods for each data input. The red dot is the mean value.}
\label{fig:model_variance}
\end{figure}

\begin{itemize}[label={\scriptsize$\bullet$}]
    \vspace{-0.5em}
    \item \textbf{Label-Dependent Methods} The first type of method adopts an intuitive approach, relying on explicit supervision labels for OOD detection. They either evaluate the errors of past predictions in an online setting, or assume access to out-of-distribution data to train an OOD classifier offline.




    \begin{itemize}
        \vspace{-0.2em}
        
        \item \textbf{\emph{A posteriori} Error Evaluation}
        The simplest method is to identify OOD examples by assuming that data points with large prediction errors are likely out-of-distribution.
        While simple and intuitive, this approach requires access to ground-truth labels, which are often unavailable or expensive to obtain in many tasks.
        For instance, in image classification, offline evaluation requires manual annotation, and labels are typically inaccessible during online inference—limiting this method’s use to offline analysis.
        In contrast, for online motion prediction, future trajectories gradually become available as the robot continues to operate. However, this still introduces a delay, making the method unsuitable for real-time or proactive OOD detection.

        \vspace{-0.2em}
        \item \textbf{Inlier-outlier classification} is another straightforward approach, where a model is trained on both in-distribution and out-of-distribution samples to distinguish between them. 
        A common practice is to define outliers by labeling data points with large prediction errors.
        Once trained, the model does not require labels at inference time, making it more applicable in online settings.
        However, this method relies on the assumption that the training outliers adequately represent the range of potential test-time outliers—an assumption that is often unrealistic and difficult to satisfy in practice. 
    \end{itemize}
    
    \vspace{-0.5em}
    \item \textbf{Bayesian Methods} 
    Instead of requiring access to ground-truth and outliers, an demanding requirement, a more principled and widely adopted approach is to model predictive uncertainty directly, allowing the system to express not just what it predicts, but also how confident it is. This leads to a family of Bayesian methods for uncertainty quantification in deep learning, which treats model parameters or predictions as random variables and aim to infer their posterior distributions given observed data. A number of exact and approximate approaches have been developed:

    \begin{itemize}
        \vspace{-0.5em}
        \item \textbf{Bayesian Neural Network (BNN)}
        place a prior distribution over the model’s weights and learn the posterior distribution over those weights given the dataset, denoted as \( p(\boldsymbol{\theta} \mid \mathcal{D}) \). Given this distribution over weights, the model prediction consists of two parts: a maximum a posteriori (MAP) estimate \( \hat{y} \), and the associated uncertainty quantified by the covariance \( \Sigma_{\hat{y} \mid \mathbf{x}, \mathcal{D}} \) \cite{neal2012bayesian, blundell2015weight}. Although BNNs are statistically principled, their high computational cost often limits their direct application in practice.
        
        \vspace{-0.2em}
        \item \textbf{Markov Chain Monte-Carlo} To address the limitations in Bayesian neural networks — explicitly model parameter distributions and rely on restrictive assumptions on the posterior— Markov Chain Monte Carlo (MCMC)~\cite{chen2014stochastic} offers a more practical alternative for posterior estimation. As an asymptotically exact approach, MCMC generates samples from the true posterior \( p(\boldsymbol{\theta} \mid \mathcal{D}) \) through stochastic exploration of the parameter space over multiple training iterations. Each sample represents a full model, and averaging predictions over multiple samples captures epistemic uncertainty. While MCMC avoids assumptions about the posterior’s form, it remains difficult to scale to large models due to its high computational cost and poor efficiency in high-dimensional spaces~\cite{chen2014stochastic}.


        \vspace{-0.2em}
        \item \textbf{Evidential Deep Learning:} 
        To avoid the computational cost of sampling over the entire set of model parameters, evidential deep learning models the uncertainty directly in the output space by predicting the parameters of a higher-order distribution—such as a Dirichlet distribution in the case of classification~\cite{evidential_uncertainty}. Unlike standard classification models that produce a single estimate of class categorical distributions, evidential models output parameters that define a distribution over class categorical distributions. In this formulation, each sample from the output distribution corresponds to a set of class probabilities, i.e., a full categorical distribution over possible labels. The mean of this distribution represents the predicted class likelihoods, while its spread serves as a measure of epistemic uncertainty. This enables efficient uncertainty estimation in a single forward pass, in contrast to methods that rely on multiple stochastic forward passes to approximate model uncertainty. We illustrate these differences in Figure~\ref{fig:model_variance}.
        \vspace{-0.2em}
        \item \textbf{Variational Inference} While evidential deep learning avoids sampling whole model parameters by modeling the second-order output distributions, variational inference offers another approach by approximating the posterior distribution of model parameters. The core idea is that multiple well-trained models should have similar predictions on the inlier data but should have different errors on outliers. The question becomes how to obtain these multiple models.
        \textbf{Monte Carlo Dropout} \cite{mcdropout_pmlr} is one such method. It approximates Bayesian inference by interpreting dropout as a variational distribution over the weights. During inference, dropout is kept active, enabling multiple stochastic forward passes, which approximates having multiple models. The variance across these passes serves as an estimate of epistemic uncertainty.
        \textbf{Deep Ensembles} \cite{ensembles_uncertainty} provide another practical alternative. Instead of obtaining a distribution over a single model's parameters, this is approximated by a distribution of models. Multiple models are trained with different initializations and data shuffling, which should implicitly capture diverse modes of the posterior. The disagreement between model predictions reflects epistemic uncertainty. Although heuristic, ensembles are simple, effective, and often outperform more theoretically grounded methods.

    \end{itemize}

    \vspace{-0.5em}
    \item \textbf{Generative Methods} While Bayesian methods focus on estimating uncertainty through modelling and approximating the posterior distributions, the generative methods approach out-of-distribution detection by modeling the underlying data distribution itself. These methods assume that in-distribution (ID) data follow a learnable generative process, and that out-of-distribution (OOD) inputs will deviate from it in terms of likelihood or reconstruction accuracy.

    \begin{itemize}
        \vspace{-0.2em}
        \item \textbf{Likelihood Estimation Methods} attempt to build probabilistic models of in-distribution (ID) inputs or latent features, such that these models assign high likelihood to ID data and low likelihood to out-of-distribution (OOD) data. Representative approaches include Gaussian mixture models~\cite{ahujaprobabilistic} and normalizing flows~\cite{HybridModelsForOpenSet}.

        \vspace{-0.2em}
        \item \textbf{Reconstruction-Based Methods} learn to reconstruct the inputs to the model, and assume that ID training data can be accurately modeled by the generative process, while OOD data cannot. How this is evaluated depends on the particular approach: for instance, OOD inputs may be rejected by the discriminator in GAN-based models, or may yield high reconstruction errors in models such as VAEs~\cite{tilted_vae_ood,hierarchical_vae_ood} or diffusion models~\cite{Diffusion_Graham_2023_CVPR_ood,liu2023unsupervised_ood}. 
    \end{itemize}

\end{itemize}

\textbf{Application in Motion Prediction} Although motion prediction has long been a core problem in autonomous systems, works that study its out-of-distribution (OOD) detection problem have only begun to emerge in recent years, and this area still remains relatively underexplored.
\citet{Sun2021OnCE} explores and compares multiple OOD detection methods: direct classification that labels data with large prediction errors as OOD; ensemble disagreement across five models trained from different seeds; GAN discriminator; and Bayesian estimation of error likelihood. 
These methods aim to identify when a learned predictor is likely to fail.
Interestingly, the authors find that the naive classification approach performs best.
They hypothesize that this is because this method directly captures the predictor's failure modes, which may be more relevant than indirect uncertainty estimates.
\citet{Shao2023FailureDF} includes a maneuver classification branch and a trajectory prediction branch, and they propose two OOD methods for each branch, respectively: evidential learning to estimate uncertainty in the classification branch, and ensemble disagreement between multiple replicas of the model to estimate uncertainty in the generated trajectory. QAD \cite{farid2023task} proposes an online anomaly detector that compares whether measured costs (e.g. minimum distance between ego and non-ego vehicles) are unlikely given a distribution of predicted costs from past predicted trajectories. 
Instances where measured costs consistently deviate from the predicted distribution over a time window are treated as indicators of system failure.
Again for vehicle prediction, \citet{itkina2022interpretable} considers different sources of uncertainties: ego historic motion, surrounding agents' historic motion, and the map. Then they use evidential deep learning to estimate epistemic uncertainty for each component. They also aggregate input features into a final feature for future motion prediction, and evidential deep learning is applied to provide an estimate of prediction confidence.
Generative methods have also been applied to the prediction setting: \citet{wiederer2023joodu} uses Gaussian mixtures to model the ID data latent space and identify OOD samples as low-likelihood, while \citet{normalizing_flows_anomaly_detection} uses normalizing flows for density estimation and out-of-distribution detection.

\subsubsection{OOD Generalization}
\textbf{Method Recap} As discussed at the beginning of this section, there is significant conceptual overlap between domain generalization and OOD generalization. Many domain generalization methods, as outlined in Section~\ref{sec:domain-adaptation-and-generalization}, can potentially be applied to OOD generalization.
In this paper, we adopt a more precise definition to distinguish them: domain generalization primarily focuses on achieving strong performance in unseen domains, whereas OOD generalization emphasizes both detecting and being robust to distributional shifts. Similarly, there have also been works that train the model on representative outliers obtained from mining hard examples or from adversarial scenario generation, but we push this discussion to Section~\ref{sec:data_synthesis}, which deals with data augmentation and synthesis.

Thus, we consider the cases where OOD generalization methods combine OOD detection with additional fallback strategies. One typical approach is to integrate a learned model with a rule-based system. The rules-based methods tend to generate safer trajectories, at the cost of reduced accuracy over long horizons and more limited interaction between agents, representing a tradeoff between robustness and performance. The combined system can rely on the learned model to perform well under conditions that are common in the training data while the rule-based model can be used on OOD inputs. 
A central challenge lies in determining when and how to switch between or combine the outputs of multiple models to produce the most reliable prediction.

\begin{figure}[t]
    \centering
    \subcaptionbox{Routing \label{fig:schematic_routing_once}}[0.45\textwidth]    {\includegraphics[width=0.45\textwidth]{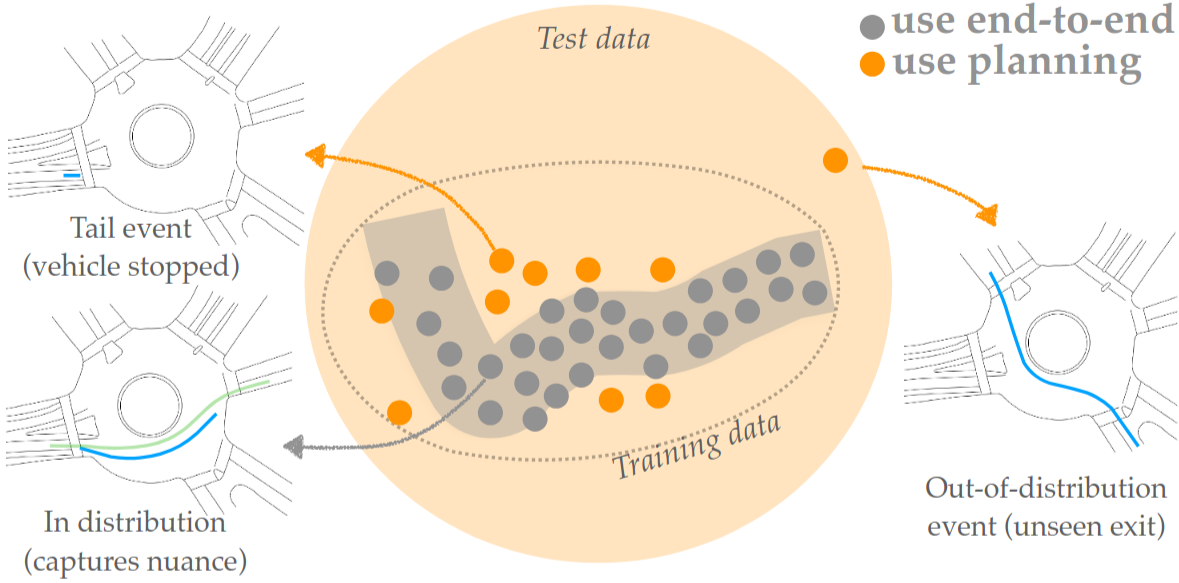}}
    \subcaptionbox{Fusion \label{fig:schematic_rulefuser}}[0.53\textwidth]{\includegraphics[width=0.53\textwidth]{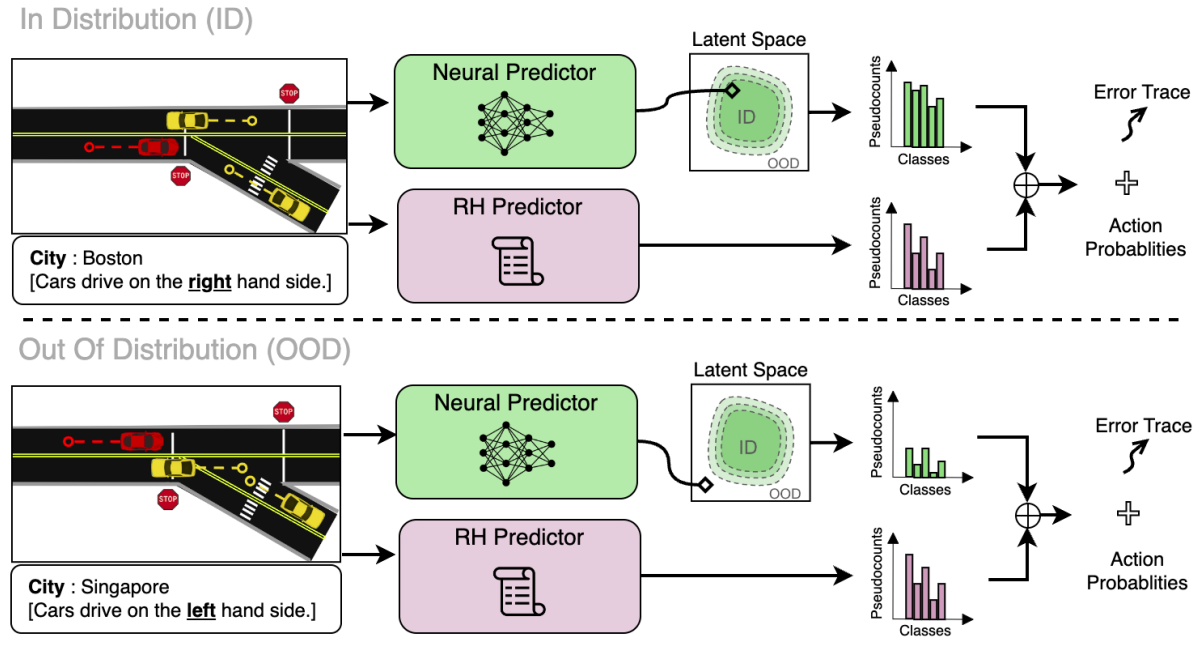}}
    \caption{Example of OOD Generalization. \citet{sun2021complementing} uses a routing approach, which selects between a learning-based model and a rule-based planner, while RuleFuser \cite{patrikar2024rulefuser} fuses the information from both models. }
    \label{fig:rulefuser}
\end{figure}

\textbf{Application in Motion Prediction} 
Several works have explored OOD generalization in the motion prediction domain by incorporating an additional rule-based model or variants of the learned model. The key challenge in these approaches typically lies in selecting an appropriate auxiliary model and designing an effective OOD detection or routing mechanism to decide when and how to rely on each component. An illustrative example is shown in Figure~\ref{fig:rulefuser}. 
\citet{Sun2021OnCE} toggles between a learned and a planner-based method in the presence of detected OOD inputs, where various OOD detection methods are explored and a simple classification method turns to work the best. Similarly, APE \cite{li2024adaptive} maintains a learned and a simple constant-velocity predictor, and train a routing network based on the performance on the source domain to swtich between the two models at test time, highlighting that even such a simple model can significantly improve performance on zero-shot transfer across datasets. Instead of switching between models, RuleFuser \cite{patrikar2024rulefuser} fuses predictions from models. It employs evidential learning (see Section~\ref{sec: ood detection}) to evaluate the likelihood of different trajectory primitives. It incorporates a rule-based trajectory planner (adapted from \citet{leung2023backpropagation}) to provide an initial evidence estimate for each candidate trajectory. A learned model is then used to contribute additional evidence for each primitive before selecting the most likely trajectory. Here, evidence refers to the model’s confidence in a prediction, typically expressed as parameters of a Dirichlet distribution. On in-distribution (ID) data, the learned model tends to produce high evidence, effectively dominating the final decision. In contrast, on out-of-distribution (OOD) data, while the learned model is uncertain and provides low evidence, the rule-based planner could still contribute reasonable evidence estimates.
Another method is \citet{hao2023improving}, which tackled rare driving scenarios by adopting principles from online learning, similar to the test-time adaptation approaches discussed in \cref{sec: generalizable DA/DG adaptation}. They propose augmenting the prediction model with an additional, online-updated replica of the original model, which adapts its weights based on online prediction errors. The system continuously tracks the historical prediction performance of both models and switches to the online model when it consistently outperforms the original within a defined temporal window.
A recent study \citet{zhao2025trajevo} introduces a framework that leverages large language models to design trajectory prediction heuristics automatically by employing an evolutionary algorithm to generate and refine prediction heuristics from past trajectory data, which achieves the state-of-the-art performance in cross-dataset OOD evaluation.

\subsubsection{Perspectives}

\textbf{OOD-Aware Decision Making} There is a fine line between encouraging models to generalize to distribution shifts, and detecting when the distribution has shifted too far and the model cannot be trusted. Solutions which can dynamically fuse OOD detection and OOD generalization will be critical for real world deployment of robotic systems, ensuring that predictions generalize OOD as far as safely possible but no further, and can gracefully alert downstream consumers when predictions can no longer be trusted. This task has been coined ``OOD Aware Decision Making" in \citet{sinha2022system}. Both \citet{Sun2021OnCE} and \citet{patrikar2024rulefuser} provide excellent examples of this, using a learned prediction model but falling back to a planning or rule-based strategy when OOD data is detected. But generally, fusing OOD detection and OOD generalization for real world prediction applications is a valuable yet under-researched area.

\textbf{Rethinking OOD Benchmarking: From Classification to Regression in Motion Prediction}
Much of the out-of-distribution (OOD) literature focuses on visual perception tasks, especially classification problems such as image classification. In such classification tasks, two categories of distribution shifts can be clearly defined: covariant and semantic shift \citep{yang2021generalized}. 
Covariate shift refers to changes in the input distribution—such as variations in lighting or background—while semantic shift involves changes in the label distribution, for example, when new object classes appear that were not present during training.
Accordingly, OOD benchmarks are usually constructed by introducing visual perturbations on the inputs, or more commonly, adding new semantic classes. 


In motion prediction, when the task is formulated as a classification problem, the OOD setting can be framed similarly to image classification tasks. For example, 
the emergence of a new maneuver class constitutes a semantic shift, while a new individual performing a known maneuver reflects a covariate shift.
However, motion prediction is more commonly formulated as a regression problem, predicting future waypoints. 
This shift brings multiple important implications: 
1) First, since the output is a sequence of future waypoints—much simpler and less variable than the input, which reflects complex factors like map structure and multi-agent interactions—covariate shift in the input becomes the primary challenge.
2) Second, in this setting, most OOD detection and generalization methods developed for classification tasks become inapplicable. 
3) Third, constructing benchmarks for regression-based OOD evaluation is more challenging, as class-based definitions of distribution shifts are no longer directly applicable.
As a result, many existing works treat samples from an entirely new dataset as OOD. However, this approach is coarse-grained: a new dataset may include samples that are well covered by the source training data—on which the model may perform well—and the original training domain may contain underrepresented data regions where the model still performs poorly. This highlights the need for more fine-grained and principled definitions of OOD in regression-based motion prediction.



\textbf{Self-Evolving Data Engine} 
The OOD methods discussed above are not only critical for ensuring online safety but also play a vital role in data collection. As illustrated in the generalization lifecycle in Figure~\ref{fig:vision}, identifying diverse, rare, or challenging samples—either generated or observed during deployment, as explored in recent work on long-tail and difficult scenarios~\citep{waymo_long_tail, waymo_difficulty}—and incorporating them into training and validation pipelines is essential for expanding the data distribution, enriching the model’s deployment envelope, and ultimately enabling open-world prediction. This forms a self-evolving data engine that continuously operates throughout the system’s lifecycle. To ensure safe development and data acquisition, it is important to adopt a conservative strategy that gradually broadens the data scope while minimizing risk during deployment-based data collection.

\begin{table}[!t]
\centering
\caption{Taxonomy of data augmentation and synthesis methods.}
\label{tab:data_synthesis}
\resizebox{0.9\linewidth}{!}{
\begin{tabular}{@{}l|c}
\toprule\toprule
Approach & Method \\ \midrule
Data Augmentation & \citet{ngiam2022scene}, \citet{tang2024hpnet}, Trajectron++~\citep{salzmann2020trajectron++}, SceneTransformer~\citep{ngiam2022scene}, \citet{tang2024hpnet}  \\
Rules-Based & MetaDrive~\citep{li2022metadrive}, \citet{li2023pre} \\
Learning Agent & ASAP-RL~\citep{wang2023efficient}, Waymax~\citep{gulino2023waymax}, ScenarioNet~\citep{li2024scenarionet}  \\
Generative Methods & MTG~\citep{ding2018new}, CMTS~\citep{ding2020cmts}, SceneGen~\cite{tan2021scenegen}, TrafficSim~\citep{suo2021trafficsim}, TrafficGen~\citep{feng2023trafficgen}, CTG~\citep{zhong2023guided}, RealGen~\citep{ding2025realgen}    \\
Adversarial Generation & L2C~\citep{ding2020learning}, MMG~\citep{ding2021multimodal}, AdvSim \cite{wang2021advsim}, \citet{zhang2022adversarial}, AdvDO~\citep{cao2022advdo}, KING~\citep{hanselmann2022king}, CAT~\citep{zhang2023cat}         \\
Language Guided & CTG++~\cite{zhong2023language}, LCTGen~\citep{tan2023language}, InteractTraj~\cite{xia2024language}, ProSim~\citep{tan2024promptable} \\
\bottomrule\bottomrule
\end{tabular}
} 
\end{table}

\subsection{Data Augmentation and Synthesis}
\label{sec:data_synthesis}
\subsubsection{Method Recap} 


While previous approaches in this chapter have primarily focused on novel losses and model architectures—such as learning invariant features or rapidly adapting models using limited target data—the last section on OOD detection and generalization marks a mild shift toward an alternative, data-centric perspectives on improving generalization.
In this section, we further discuss about data augmentation and synthesis, two widely adopted strategies to increase data volume/diversity and expand the effective coverage of the data distribution. 
In computer vision, various augmentation techniques have been proposed to simulate input degradation~\cite{hendrycksbenchmarking} in the appearance level, including random color perturbations, blurring, and rotations. On the semantic level, augmentations like Mixup~\citep{zhang2017mixup}, CutMix~\citep{yun2019cutmix}, CutOut~\citep{devries2017improved}, and copy-and-paste strategies~\citep{yan2018second,ghiasi2021simple} inject new semantic information and increase scene complexity. These augmentations create more challenging test cases and allow rare examples to be oversampled when desired. 
Beyond augmentation, an alternative approach is data synthesis/generation. In general, synthesized data could potentially expands the original distribution of the dataset, usually leading to a performance improvement. 
For example, recent works also demonstrate that combining generated data with real-world samples in a bootstrap manner can improve discriminative perception tasks~\cite{azizi2023synthetic}.

\begin{figure}[t]
    \centering
    \subcaptionbox{TrafficGen: Incremental Scenario Generation \label{fig:data_gen_scenario_gen}}[0.53\textwidth]    {\includegraphics[width=0.53\textwidth]{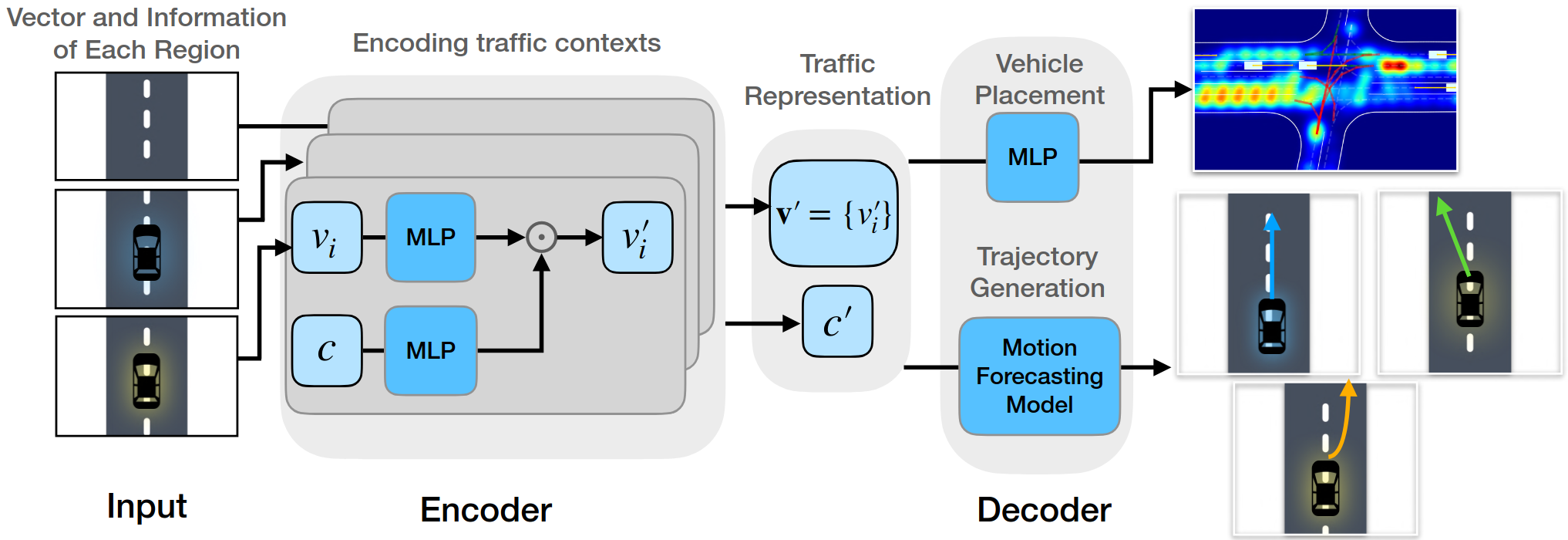}}
    \subcaptionbox{AdvDO: Adversarial Perturbation of Existing Scenarios}[0.45\textwidth]{\includegraphics[width=0.45\textwidth]{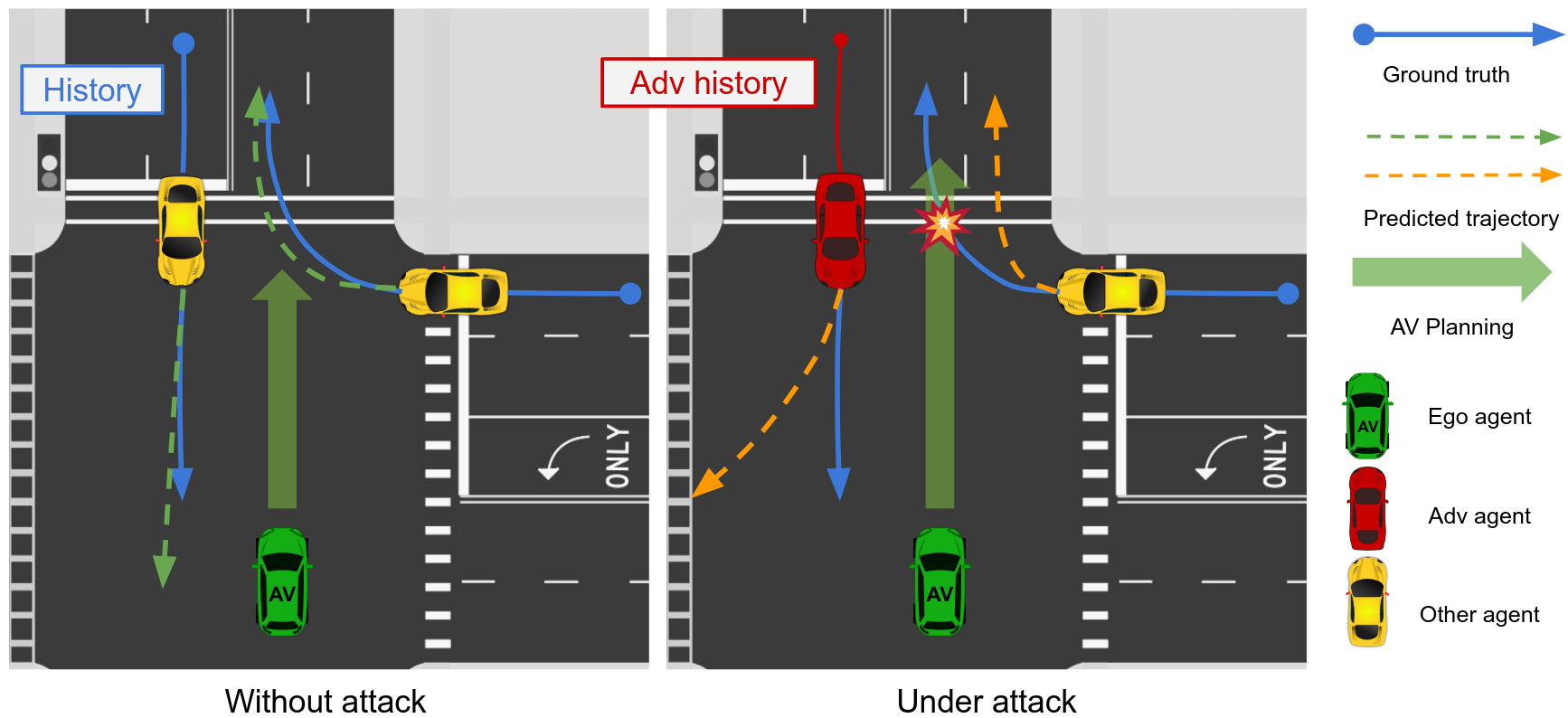}}
    \subcaptionbox{CTG++: Language-Generated Constraints \label{fig:data_gen_lang}}[0.53\textwidth]{\includegraphics[width=0.53\textwidth]{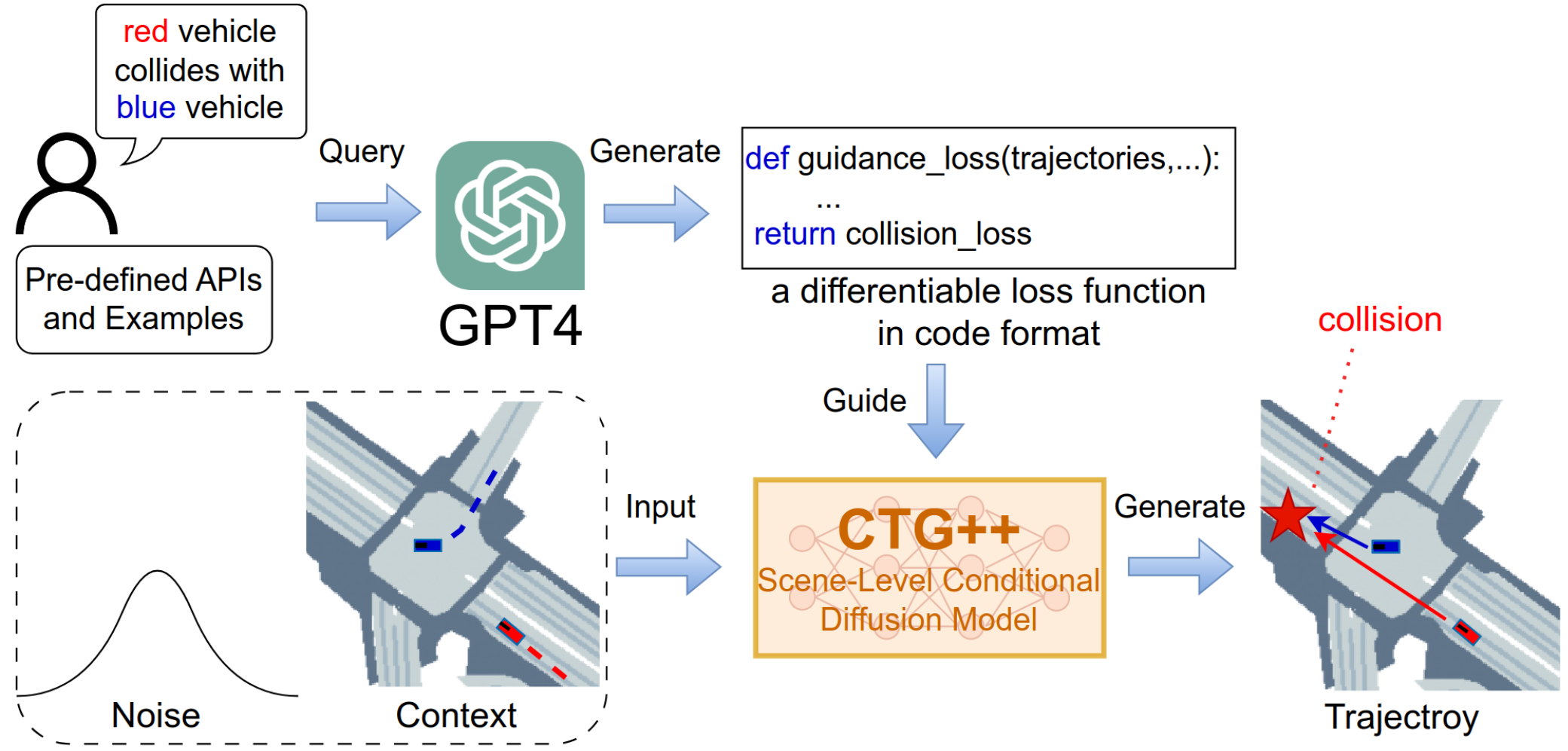}}
    \subcaptionbox{RealGen: Conditioning with Retrieved Examples}[0.45\textwidth]{\includegraphics[width=0.45\textwidth]{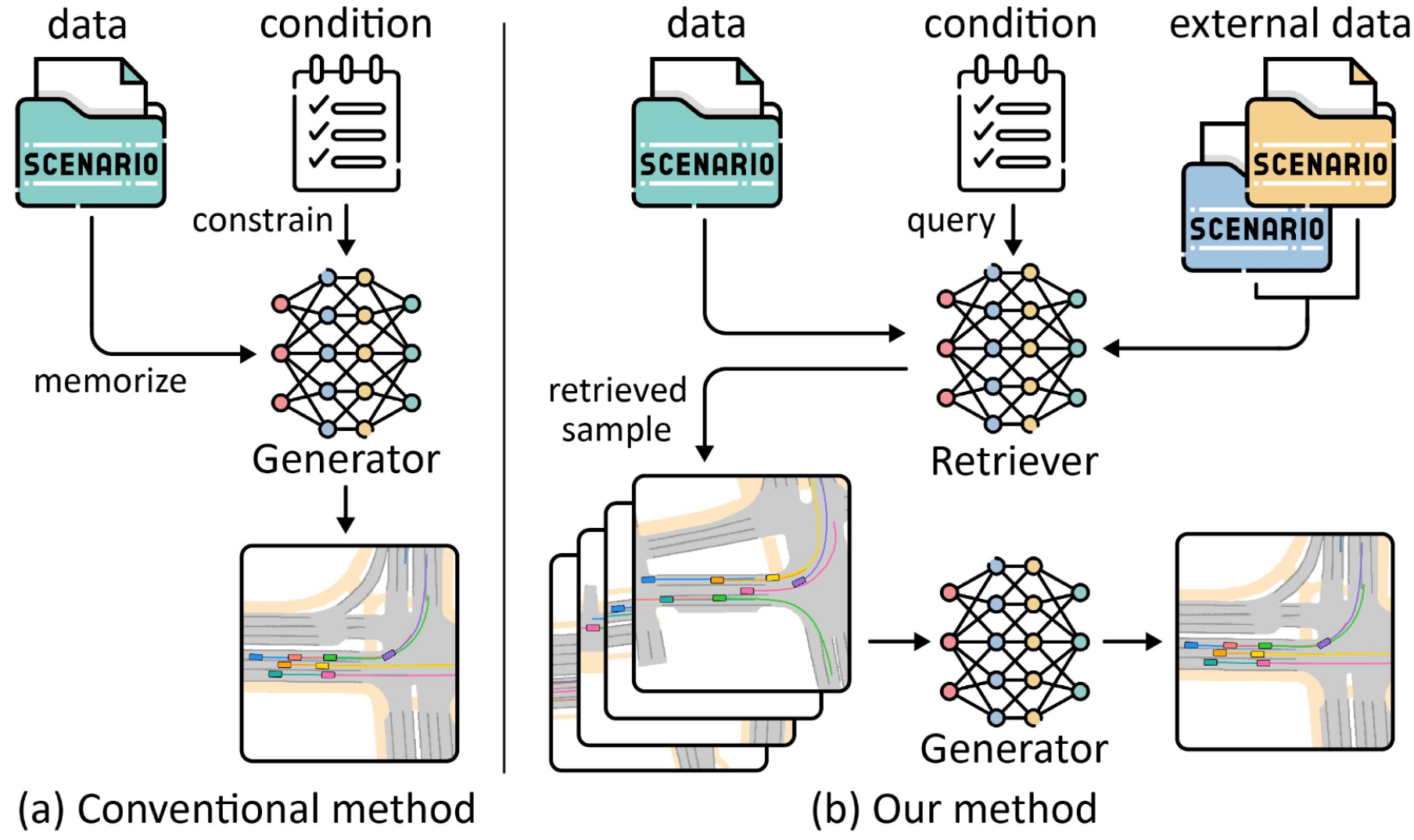}}
    \subcaptionbox{ProSim: Conditioning with Points, Sketches, and Text}[0.85\textwidth]{\includegraphics[width=0.85\textwidth]{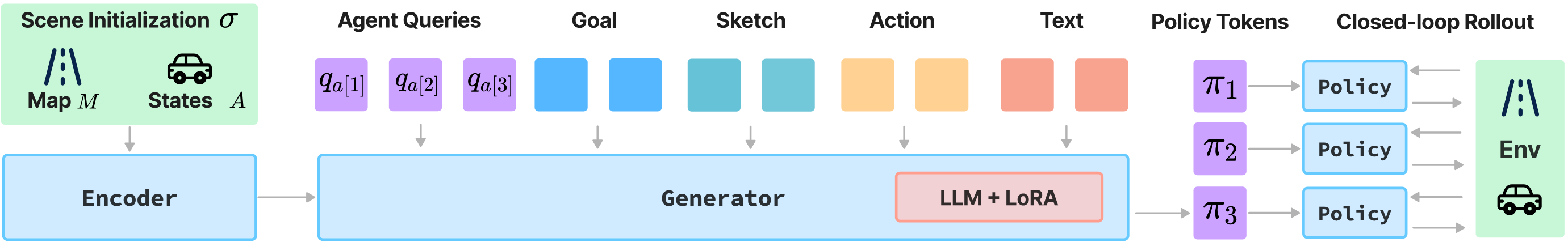}}
    \caption{Example of data generation approaches. TrafficGen \cite{feng2023trafficgen} generates completely new scenarios by incrementally adding agents and predicting trajectories, while AdvDO \cite{cao2022advdo} perturbs existing scenarios with adversarial attacks. Recent works add constraints and conditioning to trajectory generation, which can be generated from natural language as in CTG++~\citep{zhong2023guided} or varied inputs such as points, actions or sketches as in ProSim~\citep{tan2024promptable}, or create varied but realistic scenarios by conditioning the generated behavior on sampled similar scenarios stored in a memory bank, as in RealGen~\citep{ding2025realgen}.}
    \label{fig:data_gen_scenario_vs_adv}
\end{figure}


\subsubsection{Application in the Motion Domain} 
Data augmentation and synthesis techniques have also been studied in the motion domain~\citep{ding2023survey,zhong2021survey}. However, unlike image data that can be manipulated relatively easily, motion data usually needs to satisfy constraints such as the motion dynamics, the map layout, the interaction among agents, etc. In this section, we will review approaches for data augmentation and synthesis in motion prediction tasks, and discuss metrics used to evaluate the quality of generated data. We summarize the existing methods into different categories in Table~\ref{tab:data_synthesis}.

    \textbf{Data Augmentation} Multiple data augmentation techniques for motion prediction have been proposed to improve generalization and mitigate overfitting, with representative examples as that in SceneTransformer~\cite{ngiam2022scene}. We summarize commonly used augmentation strategies into three categories:

    \begin{itemize}[label={\scriptsize$\bullet$}]
        \vspace{-0.5em}
        \item \textbf{Agent and Lane Dropout} A common technique is agent dropout~\cite{ngiam2022scene}, where non-target agents are randomly removed during training. A dropout probability of 0.1 has been shown to work well on both the Argoverse and Waymo Open Motion datasets. Similarly, lane dropout—where certain lane segments are randomly masked—has also been explored~\cite{tang2024hpnet}.
        \vspace{-0.5em}
        \item \textbf{Scene-Level Transformations} It has also been found beneficial to apply random rotations to the entire scene within a range of [$-\pi/2$, $\pi/2$], after centering the map on the target agent. This has been applied in both vectorized representations (e.g., SceneTransformer~\cite{ngiam2022scene}) and rasterized settings (e.g., Trajectron++\cite{salzmann2020trajectron++}). Horizontal flipping of the scene has also been shown to improve performance~\cite{tang2024hpnet}.
        \vspace{-0.5em}
        \item \textbf{Leveraging Contextual Agents} In driving datasets, each scene typically includes many contextual non-target agents that are not the primary prediction target. SceneTransformer~\cite{ngiam2022scene} treats these contextual agents as additional prediction targets during training if they move more than 6 meters within the scene. This augmentation effectively increases the volume of training data and supports better generalization, and thus is widely adopted~\cite{zhou2023query,zhou2024smartrefine,zhou2024smartpretrain}.
    \end{itemize}
      
    \textbf{Data Synthesis} 
    Resonating with the trends of generative AI \cite{wang2025generative}, data synthesis has also been explored in the motion domain, with various approaches explored.
    \begin{itemize}[label={\scriptsize$\bullet$}]
        \vspace{-0.5em}
        \item \textbf{Rule-Based Data Generation}
        One line of work seeks to address data scarcity and expand data coverage in motion prediction by generating synthetic trajectory data using prior knowledge and manually crafted rules.
        \citet{li2023pre} manually augment both map and motion data to diversify existing datasets. Given a real-world scene, they enrich the map by transforming linear lanes into curved ones within a defined range of sharpness and angles. Motion data are further augmented by navigating a rule-based planner across both the original and the modified maps.
        MetaDrive~\cite{li2022metadrive} introduces a procedural generation pipeline that produces diverse maps and traffic scenarios using predefined roadblocks and rule-based traffic agents. In this framework, the attributes of roads—such as type, order, number, and lane configurations—as well as properties of traffic participants—including vehicle type, planning parameters, behavioral style (e.g., aggressive or conservative), spawn points, and destinations—can be either customized or randomized.
        Despite these benefits, a major issue of synthetic data is the so-called \textit{reality gap}: the distribution of artificially generated data often diverges from that of real-world scenarios, resulting in a mismatch between training and deployment. Moreover, using rule-based methods to generate high-fidelity simulations that accurately capture complex agent interactions demands substantial and careful manual design, which limits scalability in practice.
        
        \vspace{-0.5em}
        \item \textbf{Learning Agent-Based Generation} Instead of manually designing rule-based agents—which may deviate from realistic behaviors—another approach is to learn behavior policies directly from data and use them to roll out trajectories.
        To this end, imitation learning~\citep{ho2016generative, chen2019deep} has been applied to train behavior policies that mimic real-world agents and reproduce similar behaviors when presented with the same scenes. However, when deployed in a closed-loop setting, these agents suffer from error accumulation: small inaccuracies at each decision step may compound, leading to out-of-distribution observations not seen during training. Under such conditions, the learned policy may produce unrealistic or unpredictable behaviors.
        To address this, reinforcement learning is often combined with imitation learning to enable agents to handle unseen states and recover from compounding errors~\cite{wang2023efficient}, through trial and error. Building on this idea, ScenarioNet~\citep{li2024scenarionet} and WayMax~\citep{gulino2023waymax} further extend the framework to multi-agent settings, where multiple agents are jointly controlled to generate realistic interaction-rich motion data in simulation environments.

        \vspace{-0.5em}
        \item \textbf{Generative Methods} With the rapid advancement of deep generative models—such as Generative Adversarial Networks (GANs)\citep{goodfellow2020generative}, Variational Autoencoders (VAEs)\citep{kingma2013auto}, and diffusion models~\citep{ho2020denoising}—recent research has explored their application in modeling and generating motion data. Early work utilized GANs to generate single-agent lane change trajectories in highway settings~\citep{haakansson2021driving}, and VAEs to synthesize two-agent encounter scenarios~\citep{ding2018new}. CMTS~\citep{ding2020cmts} extended this by incorporating rasterized maps as inputs, enabling the generation of safety-critical two-agent scenarios via latent space interpolation.

         More advanced methods then emerged to handle complex, multi-agent interactions. SceneGen~\cite{tan2021scenegen} considers generating initial states of multi-agent scenarios that comply with the scene layout. Given the ego-vehicle’s state and an HD map, SceneGen applies a ConvLSTM model to autoregressively insert actors of various types onto the map, synthesizing each actor’s location, bounding box, and velocity.
        TrafficSim~\citep{suo2021trafficsim} builds on this, where given the map layout and initial actor placement, it leverages an implicit latent variable model to learn a joint policy that generates socially-consistent trajectories for all actors. To ensure robustness over long horizons, the model unrolls the policy during training and optimizes it through a fully differentiable simulation.
        TrafficGen~\citep{feng2023trafficgen} integrates both initial layout and trajectory generation using an autoregressive transformer-based encoder-decoder architecture. At each iteration, it encodes the current traffic context with attention mechanisms and decodes an agent’s initial state followed by its full trajectory.

        To enable more controllable scenario generation, CTG~\citep{zhong2023guided} introduces a conditional diffusion framework that allows users to specify desired trajectory attributes—such as goal-reaching or speed constraints—while preserving realism and physical plausibility. It generates actions step-by-step using diffusion, which are then propagated using a bicycle model. To enhance realism further, RealGen~\citep{ding2025realgen} adopts a transformer architecture and a Retrieval-Augmented Generation approach, using retrieved scenarios as templates to better guide the generation of novel, plausible traffic scenes.

        In addition to methods that generate object-level motion, approaches that generate directly in the scene level, such as point cloud forecasting, video generation, and scene reconstruction have also been explored. We refer authors to Section~\ref{sec: SSL - point cloud} and Section~\ref{sec: SSL - video} for more discussions.

        \vspace{-0.5em}
        \item \textbf{Adversarial Generation} 
        While generative models aim to capture and sample from the underlying distribution of realistic driving behaviors, another complementary approach is adversarial generation, which focuses on synthesizing rare and safety-critical scenarios. These scenarios—such as collisions, aggressive maneuvers, or traffic rule violations—represent critical edge cases where models are most prone to failure, yet are typically extremely scarce in real-world datasets. Adversarial generation addresses this scarcity by creating examples that are explicitly optimized to induce failure in the prediction/planning model, often by triggering collisions.


         The first category of approaches generates adversarial scenarios by introducing adversarial agents into the scene, which are trained to pose risks and induce unsafe situations for the ego agent. Early works focus on inserting new agents into the environment, often with limited consideration of how these agents move. For example, L2C~\citep{ding2020learning} considers a simple two-agent setup and learns to insert adversarial agents via reinforcement learning to maximize collision risk. The greater the risk posed by the generated scenario, the higher the reward received by the adversarial agent. MMG~\citep{ding2021multimodal} improves upon L2C by replacing the unimodal Gaussian adversarial policy with a multimodal one using normalizing flows, enhancing the diversity and realism of generated behaviors.

        In contrast to inserting new agents, another line of work focuses on manipulating the planning trajectories of existing agents to induce unsafe interactions. KING~\citep{hanselmann2022king} perturbs the state sequences of $N$ context agents to cause collisions with the ego vehicle. A kinematic bicycle model is used as a proxy for the agents’ true dynamics, allowing gradients to be optimized through this differentiable proxy to encourage collisions with ego-agent, discourage collisions among context-agent, and prevent off-road driving. After fine-tuning on this new adversarial data, they show that the policy becomes better at avoiding collisions. CAT~\citep{zhang2023cat}, on the other hand, leverages off-the-shelf motion forecasting models as learned priors to efficiently generate high-fidelity, diverse adversarial scenarios.

         Instead of perturbing planning trajectories to induce failures in downstream planning modules, another class of methods focuses on perturbing the past trajectories of context agents to specifically fool motion prediction models.
        AdvSim \cite{wang2021advsim} seeks the optimal perturbation of an agent's past trajectory to fool the prediction algorithm such that it misses a collision. Because the target agent's prediction model is not accessible, optimization is expensive as it requires sampling of responses or approximation through a surrogate model. A similar method is \citet{zhang2022adversarial}, which generates position offsets in the past trajectory of the adversarial agent to maximize changes in the target's prediction. In contrast, AdvDO~\citep{cao2022advdo} assumes access to the target prediction model, which simplifies training by enabling direct gradient-based optimization without the need for expensive sampling or surrogate modeling. It further imposes constraints on the maximum allowable changes in acceleration and curvature to ensure the physical realism of the generated trajectories. When considering white-box attacks of generative models, RobustTraj~\citep{cao2023robust} finds that the adversarial optimization is made more effective by attacking the maximum likelihood of the latents instead of sampling them. They also find that training with adversarial examples tends to reduce performance on clean data. likely due to overfitting \cite{rebuffi2021fixing}, but that this can be mitigated by using additional augmentations: they propose to maximize the offset in predicted trajectories (left, right, forward, backward) by regenerating them with updated acceleration and change of curvature, which diversity the generated adversarial data.

        Another line of work aims to simulate adversarial events that are more representative of real-world deployment conditions. 
        For example, ReasonNet~\citep{shao2023reasonnet} explicitly design occlusion scenarios to examine the model's occlusion reasoning capability in partially observable environments. 
        Moreover, as discussed in Section~\ref{sec:deployable_perception_dealing_with_uncertainty}, various studies have shown that uncertainty and errors in the perception pipeline can significantly degrade the performance of motion prediction models. Perception errors such as sensor degradation due to adverse environmental conditions (e.g. nighttime, snow), missed detections, track ID switches, and mapping inaccuracies are all common in practice.
        To study the impact of these factors, several works inject perception-related errors into the inputs of motion prediction models. These include sensor-level corruptions (e.g., masked pixels or degraded point clouds), detection noise or failures~\citep{ivanovic2022propagating, ivanovic2022heterogeneous}, tracking errors~\citep{weng2022whose, wengMTPMultihypothesisTracking2022}, and mapping inaccuracies~\citep{gu2024producing}. The goal is to evaluate and improve the robustness of motion prediction systems under such degraded upstream conditions. 

        While adversarial generation enhances robustness to known failure modes, models often remain vulnerable to unseen or worst-case perturbations. To address this, recent work has explored theoretical robustness verification \citep{liu2021algorithms,abuduweili2024estimating}, which provides formal guarantees by ensuring model predictions remain invariant within a bounded input region~\citep{zhang2018efficient}. Unlike empirical adversarial testing, these certified training methods offer provable safety but often come with high computational costs and limited scalability.

        \vspace{-0.5em}
        \item \textbf{Language-Guided Generation} As discussed in the generative methods above, controllability is another key desired capability for scenario generation methods (e.g. CTG \cite{zhong2023guided} allows users to specify desired attributes of the generated scenario). 
        However, such approaches for controlling learning-based traffic models require significant domain expertise and are difficult for practitioners to use.
        To remedy this, CTG++~\cite{zhong2023language} improves on the original CTG~\citep{zhong2023guided} formulation to propose a scene-level conditional diffusion model that can be guided by language instructions. 
        Specifically, they harness a large language model (LLM) to convert a user’s language description into a loss function, which replaces the gradient guidance process of CTG to guide the diffusion model towards instruction-compliant generation.
        Despite this advance, CTG++ still requires access to past agent trajectories as conditioning input, which introduces a degree of domain dependency. LCTGen~\citep{tan2023language} mitigates this by only requiring high-level scene information—including the number of agents, their initial states, and a summary of map features— in the LLM prompt. 
        Interestingly, it is able to generate safety-critical scenarios from traffic crash report.
        A further extension is InteractTraj~\cite{xia2024language}, which explicitly encodes desired agent-to-agent interactions in the LLM code.
        In ProSim~\citep{tan2024promptable}, the authors further extend the prompt from language to other modalities, including goal position, trajectory sketch, and actions.
\end{itemize}

    \textbf{Evaluation metrics} 
    The synthesized data should be utilized for training and evaluation only after a comprehensive assessment of its realism, diversity, and controllability. In most prior studies~\cite{tan2021scenegen, feng2023trafficgen, tan2023language}, realism is typically evaluated by measuring the distributional divergence between generated and real-world scenarios, often using the maximum mean discrepancy (MMD)\cite{gretton2012kernel} score based on features such as position, heading, speed, and size. Additionally, in self driving and robotics, adherence to traffic regulations, map compliance and vehicle dynamics is used to assess the plausibility of generated scenarios~\cite{zhong2023guided}. Diversity is generally quantified by the coverage of key attributes~\cite{zhou2024diffroad, sun2023drivescenegen}, including the number of surrounding agents, speed and acceleration ranges, and road geometries. Controllability requires that synthesis methods generate data in accordance with given instructions; however, these instructions are often difficult to verify, complicating the evaluation process. LCTGen~\cite{tan2023language} addresses this challenge by employing human judgment to compare preferences between generated scenarios, which demands significant manual effort. Overall, assessing the quality of synthesized data remains a critical yet unresolved issue that directly impacts downstream applications.

\subsubsection{Perspective}
\textbf{Quantifying the Distribution of Motion Data}
One of the objectives of data generation is to sample rare or challenging scenarios that can be used to train motion prediction methods to be more robust. However, 
if the generated data is still within the training distribution and contain little additional information not included in the original data, it may not generalize the prediction model to new unseen distributions. 
In part, this is because of the absence of diversity and distance metrics to quantify data distribution among datasets. The recent work ScenarioNet \cite{li2024scenarionet} visualizes the t-SNE plots of scenario embeddings from multiple datasets and finds they are disjoint, but significant efforts are still needed to study the reasons behind, and explore the different ways of quantifying data distribution, and discuss 
whether generation could be used to bridge the gap.
A more effective metric for characterizing datasets is essential—one that can capture rare or missing scenarios, quantify distribution shifts across datasets, and decompose these shifts into interpretable dimensions such as interaction complexity or road geometry. 


\textbf{Generalization for Motion Generation Methods}
Beyond quantifying data distributions to benefit motion prediction models, such measures are also valuable for characterizing the boundaries of generative models trained on motion data. As these generation models are trained on specific datasets, they inherently reflect the underlying data distribution and therefore encounter similar generalization challenges. Distributional metrics could help address several fundamental questions: How broadly can a generative model trained on one dataset produce plausible motion data under novel conditions? How can we systematically expand its generative coverage? What are effective strategies for adapting generative models to new datasets or previously unseen data distributions? Furthermore, an exciting direction is to explore whether motion generation and motion prediction can be integrated in a bootstrapped framework—where synthetic data improves prediction, and predictive feedback refines generation.





\textbf{Map Generation}
While there are many works that consider creating agents and trajectories on existing maps (whether HP maps or polylines), the generation of new maps is rarely considered. For instances, MetaDrive \cite{li2022metadrive} generates new maps by manually designing road blocks and connecting these predefined building blocks, but this limits types of scenarios that can be explored. This is particularly problematic because the map is a significant constraint on the trajectories and is often the source of agent interaction. Complex behaviors like yielding in narrow streets or handling emergency road restrictions are common situations that may not be present in datasets. ScenarioNet \cite{li2024scenarionet} encodes maps and scenarios from multiple datasets into a single searchable database, but does not consider generation. Efficiently generating HD maps remains an active area of research \cite{xie2023mv,zhang2024online}.

\textbf{Combining controllable and adversarial generation}
Recent developments of controllable scenario generation that extent to the closed-loop setting, such as ProSim \cite{tan2024promptable}, allow models to update their prediction and control models given generated scenarios. Modeling adversarial agents and integrating adversarial policies into this framework would allow the generation of both representative and adversarial scenarios with the same simulator. Just like how InteractTraj adds interactions to the LLM prompt used to condition the generation, adding driving styles or intentions to generate adversarial agents would be interesting, and a closed-loop setting could allow learning of and adaptation to adversarial attacks.

\subsection{Foundation Model - Scale Up the Data and Model} \label{sec:foundation_models}
\textbf{Method Recap} While the previous section explored data augmentation and synthesis as means to expand training data distributions, a natural next step from the data perspectives is to more intensively scale both the amount of data and the capacity of models. This direction aims to develop foundation models that generalize across a wide range of scenarios. Such an approach has proven especially transformative in natural language processing (NLP) and computer vision (CV), where training and pretraining of large models on large-scale datasets have significantly advanced the capabilities and generalization of modern AI systems.
Empirical studies in these domains have also revealed empirically-validated \textit{scaling laws}~\citep{kaplan2020scaling,hoffmann2022training}, which show that model performance improves predictably and consistently with increased compute, dataset size, and model parameters, providing further motivation for this direction.

\textbf{Application in Motion Prediction} However, despite the demonstrated success of large-scale datasets in NLP and CV, the field of motion prediction has largely remained in a \textit{small-data regime}, due to the expensive and labor-intensive process of collecting and annotating motion data. For instance, in autonomous driving, widely used motion prediction datasets such as Argoverse~\citep{Argoverse}, Argoverse 2~\citep{wilson2023argoverse}, and the Waymo Open Motion Dataset (WOMD)\citep{sun2020scalability} contain only 320K, 250K, and 480K data sequences respectively—orders of magnitude smaller than the datasets commonly used in NLP and CV (e.g., GPT-3 trained on the 400B-token CommonCrawl~\citep{raffel2020exploring}, or Vision Transformer pre-trained on the 303M-image JFT dataset~\citep{sun2017revisiting}). This scarcity of trajectory data limits the ability of models to learn rich, transferable scene representations, thereby constraining their performance and generalizability.
To address this, recent work in motion prediction has begun to explore data scaling by unifying and integrating trajectory data from multiple sources, aiming to emulate the success of foundation models in other domains. 
Although the total amount of public data remains small compared to other fields—largely because NLP and CV can scrape massive datasets from the internet, while motion data requires costly sensor setups and annotation—this multi-dataset integration marks an important first step toward scaling up and diversifying motion data, which is critical for enabling better generalization and broader deployment.
However, unlike in NLP and CV, where multi-dataset integration is relatively straightforward, leveraging multiple datasets simultaneously in motion forecasting remains non-trivial. Each dataset often comes with its own data format and development APIs, along with numerous discrepancies—such as differences in past/future prediction horizons, map resolutions, sampling rates, and types of semantic annotations. These inconsistencies make it cumbersome for researchers to train and evaluate models across multiple datasets in a unified manner. Some recent efforts have attempted to bridge this gap by unifying pedestrian and vehicle motion datasets, laying the groundwork for broader-scale integration.

\begin{figure}[t]
    \centering
    \includegraphics[width=0.9\textwidth]{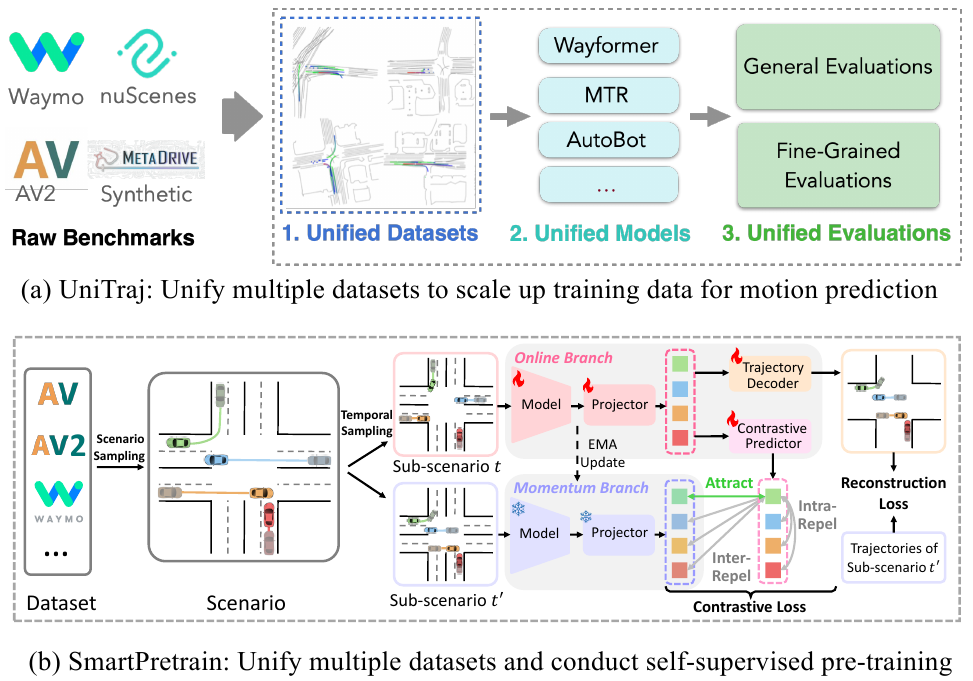}
    \caption{Early exploration toward data and model scaling in motion prediction: Unitraj~\cite{feng2025unitraj} unifies multiple driving datasets to scale up training data for motion prediction; SmartPretrain~\cite{zhou2024smartpretrain} unifies multiple driving datasets, conducts self-supervised pretraning to learn informative features, and then fintune the model to actual motion prediction task.}
    \label{fig:data_scale}
\end{figure}

\begin{itemize}[label={\scriptsize$\bullet$}]
    \vspace{-0.5em}
    \item In the pedestrian motion data domain, OpenTraj~\cite{amirian2020opentraj} created dataloaders for different pedestrian motion datasets as part of its effort to evaluate and compare motion complexity across pedestrian datasets. More recently, TrajNet++~\cite{kothari2021human} and Atlas~\cite{rudenko2022atlas} present multi-dataset benchmarks to systematically evaluate human motion trajectory prediction algorithms in a unified framework. 

    \vspace{-0.5em}
    \item Another thread of works focuses on cross-dataset development on vehicle motion data. 
    trajdata~\cite{ivanovic2024trajdata} provides a unified interface to multiple human/vehicle trajectory datasets. A related work for the task of vehicle trajectory planning is ScenarioNet~\cite{li2024scenarionet}, a simulator aggregating multiple real-world datasets into a unified format and providing a planning development and evaluation framework. Based on ScenarioNet, UniTraj~\cite{feng2025unitraj} unified the data formats of multiple driving datasets, where multiple performant models are trained and evaluated. Several interesting findings are demonstrated: 1) models trained on larger and more diverse dataset (Waymo) exhibit the highest generalization ability; 2) larger models exhibit stronger capabilities in scaling with more data. Such results underscore the potential and need for larger, more diverse datasets in the trajectory prediction field. SmartPretrain~\cite{zhou2024smartpretrain} proposes a general and scalable pretraining strategy that can be flexibly applied to various SOTA models, and a dataset-agnostic scenario sampling strategy that integrates multiple datasets, enhancing data volume, diversity, and robustness. Strong experiments are demonstrated where the proposed pre-training consistently improves the performance of state-of-the-art prediction models across datasets, data splits and main metrics. Interestingly, the results also demonstrate that, in the pretraining strategy, the proposed contrastive and reconstructive task significantly outperform the standard motion prediction tasks in terms of learning informative features.

\end{itemize}

\subsubsection{Perspectives}
\textbf{Unexplored Challenges in Scaling Up - Pretraining Strategy and Data Distribution} 
While unifying and mixing various motion data sources have shown promising improvement and potential, multiple important questions remain unexplored in scaling up motion prediction models. For example, 
While some works have made promising progress in exploring pretraining strategies (see Section~\ref{sec:gen_pred_ssl}), the optimal approach for learning informative features that generalize across diverse data sources remains unclear. Furthermore, although industry players possess vast amounts of data—such as the large-scale driving logs collected by autonomous vehicle companies—challenges persist in quantifying the underlying data distribution to ensure diversity and efficiency when scaling up data usage.
For example, when a new batch of data is collected, how can we determine whether this batch contains new information that the model needs to absorb, or this batch has been seen previously and covered by the model. Bootstrapping would be a common practice, where ranking data value by evaluating the model's prediction error against them, while other alternatives could potentially be developed.

Notably, a very recent work~\cite{baniodeh2025scaling} provides promising empirical insights into this direction. Using 500,000 hours of driving data, the authors demonstrate that, similar to language models, performance improves predictably with training compute budget, following a power-law scaling. They also observe that closed-loop metrics—crucial for downstream deployment—consistently improve with scaling, reinforcing their value over open-loop metrics for model selection. Additionally, their study suggests that to maintain compute-optimality, model size should increase 1.5× faster than dataset size as compute scales. Interestingly, at inference time, strategies such as sampling and clustering outputs from smaller models can remain competitive up to a crossover point. These findings highlight that scaling properties—both in training and inference—play a critical role in driving performance improvements, and further emphasize the importance of understanding data utility and distribution when building large-capacity motion models.





\textbf{Adapting Existing Foundation Models} Due to the high cost of collecting and annotating motion data, the public motion data, even when unified and mixed, is still limited, especially compared to the internet-scale text and vision data used to train foundation models in language and vision. To this end, how to adapt existing foundation models (e.g., large language models, video generation models), that are trained on internet-scale data and thus have emergent reasoning capabilities and common-sense knowledge, into the motion domain, has been a promising but understudied direction. Early efforts~\cite{seff2023motionlm,philion2024trajeglish} have introduced concepts from large language models, like tokenization, into motion prediction, but this remains a nascent field with significant potential for further exploration and development.

\subsection{Outlooks}
\label{sec: generalization}

As motion prediction systems are deployed in increasingly diverse and unpredictable environments, generalization becomes a central challenge—requiring models not only to perform well on known domains but also to adapt, recover, or remain robust in the face of distributional shifts and open-world settings. In this chapter, we presented a comprehensive survey of methods that address this challenge across different stages of the generalization lifecycle, including self-supervised learning, domain adaptation and generalization, continual learning, out-of-distribution (OOD) detection and generalization, data-centric strategies (augmentation and synthesis), and foundation model scaling. Several key themes emerge from this survey:

\textbf{Data is Fundamental but Limited} 
As discussed throughout this chapter, despite the demonstrated success of large-scale datasets in NLP and computer vision, the field of motion prediction remains largely in a small-data regime due to the costly and labor-intensive process of collecting and annotating motion data. For instance, leading autonomous driving datasets such as Argoverse~\citep{Argoverse}, Argoverse 2~\citep{wilson2023argoverse}, and the Waymo Open Motion Dataset (WOMD)\citep{sun2020scalability} contain only 320K, 250K, and 480K data sequences, respectively—orders of magnitude smaller than datasets in NLP (e.g., GPT-3's 400B-token CommonCrawl\citep{raffel2020exploring}) or vision (e.g., JFT's 303M images~\citep{sun2017revisiting}). This scarcity of data constrains models’ ability to learn rich and transferable representations, ultimately limiting their robustness and generalization. 
Beyond limited volume, dataset discrepancy poses another significant barrier. Unlike image data, which is typically represented in standardized formats (e.g., RGB arrays with class labels), motion prediction datasets often differ in key structural properties—including trajectory horizon, sampling frequency, coordinate system, map modality (e.g., HD map, polyline), and agent taxonomy (e.g., vehicle categories, pedestrian types). These differences make it difficult to directly merge multiple datasets to big one.
As a result, each dataset often remains siloed, and models trained on one do not generalize well to others, hampering the development of universal motion models.
In fact, many of the limitations in motion prediction generalization ultimately stem from the small size, narrow coverage, and inconsistent formats of existing datasets. Scaling up high-quality datasets—either through unifying existing sources or leveraging generation—is a clear priority.

\textbf{Dataset Discrepancy and the Absence of Standardized Cross-Dataset Benchmarks}
In motion prediction, dataset inconsistency poses a major challenge not only for model training but also for benchmarking generalization.
In principle, evaluating how well models generalize across datasets—i.e., cross-dataset evaluation—is a core metric of generalization. This paradigm is well-established in computer vision, supported by standardized benchmarks that facilitate fair comparison and reliable progress tracking.
However, unlike computer vision, where datasets typically follow consistent formats and support large-scale integration, motion prediction datasets often differ widely in trajectory horizon, sampling rate, coordinate system, map representation, and agent definitions.
These discrepancies make cross-dataset evaluation largely infeasible, and most existing works utilize customized data splits for their own evaluation. 
This fragmentation has slowed progress in the field, as practitioners face significant barriers in conducting fair comparisons, reusing others’ methods and results, and conveniently assessing the progress.
Recent studies~\citep{park2024improving, bae2024singulartrajectory, feng2025unitraj, zhou2024smartpretrain} have made promising progress by unifying formats and enabling preliminary cross-dataset evaluations. Still, a widely accepted benchmark suite remains absent. 
Establishing such a benchmark is crucial for advancing generalization research in motion prediction, enabling rigorous and reproducible evaluations, and providing a practical framework for fair comparison across methods—ultimately making it easier for practitioners to select approaches suited to their specific scenarios.

\textbf{Rethinking Cross-Dataset Generalization Benchmarking: From Classification to Regression}
Even with standardized formats, the dominant benchmarking paradigms from classification tasks do not directly translate to regression settings like motion prediction, further complicating evaluation efforts.
Specifically, in computer vision, particularly in classification tasks like image recognition, benchmarking model generalization often centers around clearly defined distribution shifts. These include covariate shifts, which alter the input distribution (e.g., lighting or texture changes), and semantic shifts, which introduce new output classes not seen during training \citep{yang2021generalized}.
Among these, semantic shifts have become the most common and direct way to construct OOD benchmarks. This is because the presence of entirely new classes—such as transferring a model trained on one dataset to another that includes previously unseen categories—offers a clear signal of distributional shift. In contrast, covariate perturbations (e.g., noise or occlusion) are often more ambiguous and gradual in their effect. As a result, most perception benchmarks rely on introducing novel semantic classes to evaluate generalization under distribution shift.

In motion prediction, when the task is formulated as a classification problem, the OOD setting can be framed similarly to image classification tasks. For instance, the appearance of a new maneuver class—such as an unseen driving intention or pedestrian action—naturally constitutes a semantic shift.
However, motion prediction is more commonly cast as a regression problem, where the model predicts future waypoints. In this setting, class-based definitions of distribution shifts no longer apply directly, making OOD benchmark construction substantially more challenging. Consequently, many existing works resort to a coarse-grained approach by treating entire datasets unseen during training as out-of-distribution. Yet this strategy lacks nuance: new datasets often contain scenarios already well-represented in the source data (where the model performs well), while underrepresented or rare cases may exist within the source domain itself (where the model still fails). This underscores the need for more fine-grained, principled definitions of OOD in regression-based motion prediction, potentially grounded in trajectory diversity, map complexity, or interaction density.

\textbf{Fragmented Evaluation and the Need for Unified Protocols}
Compounding the challenge, the evaluation of generalization methods remains fragmented even when the tasks share similar mechanisms or algorithms.
Specifically, generalization methods in motion prediction are often developed in isolation for specific tasks, data regimes, or evaluation metrics. However, in practice, the boundaries between tasks and methods are increasingly blurred, and as shown in Figure~\ref{fig:methods_gen}, many tasks actually share the same underlying methods. For example, regularization techniques originally designed for continual learning are also used in domain adaptation; buffer-based example replay appears in both adaptation and test-time learning; and OOD detection techniques inform both robust training and safe deployment. This convergence highlights the need for unified frameworks that flexibly support multiple generalization objectives within a single paradigm.
At the same time, existing evaluation protocols often fall short of capturing this generality. Most approaches are tested under narrow assumptions—e.g., a fixed data scale or a single type of distribution shift—without investigating how well they generalize across settings. For example, methods developed for offline few-shot learning are rarely tested in online streaming scenarios. Likewise, distinctions between test-time adaptation and continual learning are sometimes artificial, as both involve learning under evolving distributions.
In addition, standard metrics such as minADE and minFDE focus solely on accuracy and ignore uncertainty estimation,
although confidence-aware metrics—such as probabilistic ADE/FDE or Brier scores—have been proposed, they remain insufficient in capturing how uncertainty should be interpreted and used by downstream modules like planning or control, as we discussed in Section~\ref{sec: app - Planning-Aware Prediction Evaluation}. This calls for further investigation into evaluation metrics that are not only statistically sound but also aligned with the operational semantics of motion forecasting in real-world autonomy systems.
Together, these observations suggest that advancing generalization in motion prediction requires not only stronger algorithms, but also a rethinking of how we evaluate them—under unified, realistic protocols, across multiple tasks, data scales, and metrics that reflect both accuracy and confidence.




 \textbf{Understanding and Quantifying Motion Data Distribution}
A fundamental but often overlooked challenge in motion prediction is the lack of tools to effectively understand and quantify the underlying distribution of motion data. This limitation affects nearly every aspect of the generalization pipeline discussed in this chapter—including benchmarking, training, data recollection or synthesis, and development of generalizable approaches.

\begin{itemize}[label={\scriptsize$\bullet$}]
    \vspace{-0.5em}
\item From a \textit{benchmarking} perspective, without a clear understanding of how different datasets relate to one another in terms of coverage, diversity, and scenario difficulty, it becomes difficult to design fair and informative cross-dataset evaluations. For instance, as just discussed, a new dataset may contain samples well within the support of the training set, or conversely, may include entirely novel distributions. Without metrics to characterize such differences, benchmarking remains coarse and often uninformative.

    \vspace{-0.5em}
\item In \textit{training}, better characterization of motion data distributions could enable more efficient learning paradigms such as active learning~\cite{settles2009active} and dataset distillation~\cite{feng2024tarot}, which aim to select the most representative or informative samples, thereby reducing the amount of data required while maintaining or even improving model performance. 


A deeper understanding of data distributions also helps mitigate overfitting due to data imbalance. While computer vision foundation models (e.g., SAM) often visualize geographic coverage to assess diversity, such approaches may overlook latent similarities across regions—images from different countries may exhibit comparable semantics or photometric features, leading to hidden redundancy despite geographic balance.
In motion prediction, researchers commonly rely on low-level statistics (e.g., speed, acceleration, agent density~\cite{trajdata,rudenko2022atlas}) or handcrafted metrics (e.g., interaction intensity~\cite{tolstaya2021identifying}) to characterize data diversity. More recently, some have explored latent space analyses to identify distributional gaps~\cite{li2024scenarionet}. However, there remains no unified framework or systematic understanding of motion data distributions. Developing such tools could inform better data selection and curriculum design, opening promising directions for future research.




    \vspace{-0.5em}
\item In \textit{data collection}, even though industry players have access to vast driving logs, questions remain: when new data is collected, how do we know whether it contributes new coverage or simply duplicates existing data? Methods like bootstrapping—where prediction error is used to rank data value—are one approach, but a principled, distribution-based framework is still lacking.

    \vspace{-0.5em}
\item Moreover, in \textit{data synthesis}, quantifying the difference between generated and real data distributions is crucial. One of the key goals of data generation is to introduce rare or challenging scenarios that are underrepresented in the original dataset. However, without proper metrics, it is difficult to evaluate whether synthetic data truly expands the training distribution or merely replicates it. Recent work such as ScenarioNet~\cite{li2024scenarionet} shows that different datasets often occupy disjoint regions in embedding space, but more interpretable and actionable measures are needed—ones that capture semantic aspects like road topology, interaction complexity, or rare maneuvers.

    \vspace{-0.5em}
\item Beyond these applications, understanding the data distribution is instrumental for \textit{designing generalizable models} themselves. Most methods discussed in this chapter—ranging from self-supervised learning and domain adaptation to continual learning and OOD detection—implicitly rely on assumptions about how data is distributed and how it shifts. A more effective metric for characterizing datasets—one that can capture rare or missing scenarios, quantify distributional shifts across datasets, and decompose these shifts into interpretable dimensions such as interaction complexity or road geometry—would provide a solid foundation for guiding and evaluating these approaches. This is particularly important because real-world datasets often differ in multiple entangled factors, making it difficult to isolate which components drive model performance gaps. Without such insights, it remains challenging to develop models that are truly generalizable across domains.
\end{itemize}

\textbf{Generalization Research Remain Immature}
Although this chapter reviews a broad range of methods—spanning self-supervised learning, domain adaptation, continual learning, OOD detection, and foundation models—the field of generalizable motion prediction remains in an early stage. Progress has been made, but many methods are still exploratory, fragmented, and lack standardized evaluation frameworks, largely due to limitations in data availability, inconsistencies across datasets, the absence of unified benchmarks, and a lack of understanding motion data distribution, as discussed above. Recently, a few works~\cite{zhou2024smartpretrain,feng2025unitraj,trajdata,li2024scenarionet} have started to explore large-scale, cross-dataset unified evaluation protocols—marking an important step forward—but systematic comparisons across methods remain lacking, and their relative advantages are still not well understood.

\begin{itemize}[label={\scriptsize$\bullet$}]
    \vspace{-0.5em}
\item A central challenge is learning robust and transferable representations. While self-supervised learning, domain generalization, and domain adaptation have shown promising results, there is little consensus on what pretraining/training objectives or architectures best support cross-environment and cross-task generalization. Improvements are often confined to specific domains or tasks, and rarely translate to broad generalization across diverse scenarios. This underscores the need for better pretext tasks, clearer understanding of motion data distributions, and more consistent evaluation protocols.

\vspace{-0.5em}
\item Test-time, online, and continual adaptation remain significantly underexplored in motion prediction. While real-world deployment demands models that can continuously adapt to dynamic environments, most existing works still assume offline training settings or access to labeled target data. Continual learning, which addresses the challenge of adapting to a sequence of new domains without catastrophic forgetting, is especially relevant here but remains rarely explored. This limits their applicability in open-world, streaming scenarios. Moreover, current benchmarks and datasets are not well suited for studying long-term adaptation. Most data sequences span only a few seconds of motion, making it difficult to evaluate how well models adapt over extended time horizons or evolving distributions. Supporting long-term, online adaptation will require both methodological advances and new datasets designed with sustained, temporally extended evaluation in mind. 

\vspace{-0.5em}
\item Similarly, out-of-distribution (OOD) detection and generalization are critical for real-world safety, but remain loosely defined and inconsistently evaluated in motion prediction. Most existing methods treat new datasets as entirely OOD, a coarse approximation that fails to account for intra-dataset diversity or partial overlap with training data. Better definitions of OOD, improved uncertainty calibration, and principled evaluation metrics—especially ones aligned with downstream task requirements—are necessary to advance OOD-aware prediction.
\end{itemize}

In summary, advancing toward mature, generalizable prediction systems calls for not only methodological innovation, but also community-level efforts to address foundational issues in data, evaluation, and benchmarking.

\textbf{Foundation Models: A Path Toward Open-World Generalization}
As methods such as self-supervised learning, domain adaptation, OOD detection, and data understanding continue to mature, foundation models are emerging as a promising direction to unify and extend these capabilities—offering a path toward robust generalization in the open world. This paradigm builds on transformative advances in computer vision and natural language processing, where a large-parameter model trained with a large amount of data shows strong zero-shot and few-shot generalization across diverse tasks and modalities.
In such a era of foundation models, we naturally wonder, can we develop a foundation model for motion prediction, that achieve generalized performance and have many downstream usages?
However, motion prediction remains relatively underexplored in this regard. Literature on training large motion models still lags behind, and has yet to fully leverage the generalization power of scalable learning paradigms or model architectures suited for the structured, dynamic nature of motion data. Achieving similar success in motion prediction that can be adapted efficiently to diverse agents, geographies, and scenarios presents unique challenges across data, model, and learning objective dimensions.

\begin{itemize}[label={\scriptsize$\bullet$}]
    \vspace{-0.5em}
    \item From the \textit{data} perspective, one priority is to unify existing datasets—despite discrepancies in format, resolution, and label conventions—so that models can be pretrained across diverse environments and agent types. Alongside this, collecting additional high-quality data and leveraging data synthesis methods remain essential to expanding the operational coverage. 
    Meanwhile, as datasets scale, simply adding more data is not sufficient. Understanding and quantifying the underlying data distribution becomes essential to ensure diversity, avoid redundancy, and guide efficient data usage. Moreover, to ensure meaningful progress, fair comparison protocols under equal compute budgets should be established to disentangle the benefits of scale from those of architecture or algorithm design.

    \vspace{-0.5em}
    \item On the \textit{modeling} side, motion prediction introduces unique structural and temporal challenges that set it apart from vision and language tasks. Unlike static images or discrete text, motion data consists of continuous trajectories shaped by physical laws, spatial constraints, and multi-agent interactions. Foundation models in this domain must therefore learn representations that generalize across diverse scene layouts, agent types, behavioral patterns, and geographies—while respecting the causal and temporal dependencies inherent in real-world motion.
    To this end, several key questions remain open. What inductive biases, tokenization strategies, or architectural modules are best suited to represent agents, maps, and interactions effectively? Should representations be built on trajectory-level abstractions or raw sensor streams like point clouds and videos—which bypass annotation and offer richer context, but introduce high computational costs and struggle to capture discrete agent behaviors? More broadly, the optimal approach for learning informative, transferable features across heterogeneous motion datasets remains unclear—highlighting the need for further research into motion-specific pretraining objectives, scalable architectures, and adaptation strategies.
    Ultimately, developing such models will require rethinking not only training but also \textit{deployment}—encompassing how to fine-tune foundation models for specific downstream tasks, and how to ensure robust generalization under distribution shifts through test-time OOD detection, generalization, and adaptation.
    

    
    \vspace{-0.5em}
    \item \textit{Adapting Existing Foundation Models}.
    In parallel, given the high cost and limited availability of large-scale motion datasets, especially when compared to the internet-scale corpora available in vision and language, adapting existing foundation models offers a compelling but still nascent direction for motion prediction. Models such as large language models and video generation architectures, pretrained on vast and diverse data, already exhibit strong generalization, reasoning, and commonsense capabilities. Leveraging these capabilities for motion-related tasks—such as intent inference, scene understanding, or trajectory prediction—could offer significant benefits, particularly in rare or ambiguous scenarios. Early efforts~\cite{seff2023motionlm,philion2024trajeglish} have explored this space by introducing language-model-style tokenization and autoregressive modeling to motion trajectories. However, the field remains largely unexplored, and many challenges lie ahead in bridging modality gaps, aligning spatial-temporal reasoning with physical constraints, and integrating these pretrained models into safety-critical, real-time systems. 
    For example, when properly prompted, large language models (LLMs) have demonstrated impressive reasoning capabilities across language and vision tasks. However, how to bridge language-based representations with the structured nature of motion—characterized by spatial, temporal, and physical constraints—remains an open and promising research direction.
    Moreover, one of the core strengths of foundation models lies in their ability to generalize to new tasks and data distributions with little or no additional training—yet discussions on enabling and evaluating zero-shot or few-shot generalization in the motion domain remain scarce, representing a key opportunity for future exploration.
    \vspace{-0.5em}
    
\end{itemize}


Continued scaling of large foundation models has led to striking advances across domains. While autoregressive transformers dominate in language and diffusion models in vision, both architectures are increasingly being applied to planning and prediction tasks~\citep{zhong2023guided, zhong2023language, jiang2023motiondiffuser, ngiam2022scene, shao2023lmdrive, shazeer2017outrageously}.
Particularly exciting is the emergence of human-aligned behaviors in large-scale generative models. These models not only match human-level accuracy under distribution shifts, but their errors are strongly correlated with errors observed in human subjects \citep{2023intriguing}. In essence: the unique robustness to open world distribution shifts that is natural for humans may be an emergent property of large scale models, and the path to achieving human-like performance in prediction systems may be a matter of further scaling up large foundation models. This aligns with \citet{sutton2019bitter}'s "Bitter Lesson", which emphasizes the simple importance of scaling up resources over custom-tailored algorithms and architectures.

\textbf{Open-World and Lifelong Generalization} Open-world scenarios are often surprisingly broad and difficult to anticipate. From rare behaviors and cross-domain shifts to adversarial conditions and previously unseen classes, these settings reveal not only coverage gaps in training data but also limitations in the modeling assumptions. For example, how should a system predict the behavior of entities never encountered during training—such as animal-drawn carts on rural roads, construction vehicles driving against traffic flow, or police officers manually overriding map-based traffic rules? These scenarios fall far outside the training distribution and illustrate the limits of static, closed-world assumptions. Therefore, the surprising breadth of open-world scenarios often necessitates not just model retraining, but a fundamental reconsideration of the problem formulation itself.
To this end, generalization is not just a model-centric challenge—it is a system-level, continuously lifecycle problem, as illustrated in Figure~\ref{fig:vision}. Emerging approaches such as OOD detection, self-evolving data engines, and foundation models reflect a growing recognition that generalization must be treated as a continuous lifecycle problem—spanning data acquisition, learning signals, runtime uncertainty estimation, and online adaptation.
With each round of real-world deployment, models inevitably encounter new out-of-distribution data, triggering the next cycle of the generalization process. This process involves not just fine-tuning but potentially rethinking the model architecture, retraining with enriched data, and updating evaluation protocols. Such a self-evolving loop forms the foundation of a truly robust open-world motion prediction system.

%% file: section/5_discussion.tex
\newpage
\section{Conclusion}
\label{sec: discussion}


In this survey, we revisited motion prediction—a cornerstone of intelligent autonomy—from the dual perspectives of \textit{deployability} and \textit{generalizability}, two critical yet under-explored dimensions for real-world adoption. While remarkable progress has been made in existing benchmarks, many state-of-the-art approaches still fall short when integrated into deployed systems or faced with open-world variability. This gap underscores the importance of rethinking problem formulations, modeling assumptions, and evaluation protocols to bridge the divide between academic progress and practical utility.

We began with a comprehensive taxonomy (Section~\ref{sec: taxonomy}) of motion prediction methods, spanning representation choices, modeling paradigms, domain-specific applications, and evaluation strategies. This provided foundational clarity to analyze how current practices map—or fail to map—to real-world demands. In Section~\ref{sec:applicable}, we focused on \textit{deployability}, dissecting the role of motion prediction within closed-loop autonomy stacks. We highlighted key issues such as interface mismatches between modules, the neglect of uncertainty propagation, and evaluation schemes that fail to capture full-system performance. Addressing these concerns calls for tighter \textit{integration across modules}, \textit{uncertainty-aware learning}, \textit{joint learning across modules}, and \textit{holistic evaluation} frameworks.
In Section~\ref{sec:generalizable}, we turned to the challenge of \textit{generalization}, which is essential for scaling motion prediction to diverse and unstructured environments. We explored methods across a rich landscape: self-supervised learning, domain generalization/adaptation, continual learning, out-of-distribution (OOD) detection/generalization, data augmentation/synthesis, and the emerging role of foundation models.

Throughout each section and chapter, we provide perspectives and insights, and highlight future challenges. We emphasized that prediction within autonomy stacks requires more than just accuracy—it must feature well-defined interfaces, compatible learning signals, and an awareness of both upstream inputs and downstream requirements. Similarly, generalization should be treated as a primary objective. It is not simply a byproduct of scale, but a property that must be explicitly pursued through deliberate design. 
In discussing deployability and generalizability individually, we highlighted the shortcomings of current benchmarks and called for more realistic, integrated alternatives. 
When zooming out to consider both challenges together, the difficulty becomes even more pronounced: the two domains often rely on fundamentally different datasets—deployability research uses motion data embedded in full autonomy stacks, while generalization studies focus on clean, decoupled datasets. These differences in format, structure, and assumptions make unified benchmarking especially complex.

By bringing \textit{deployability} and \textit{generalizability} to the forefront, we hope this survey serves as a roadmap for the next phase of motion prediction research—one that moves beyond idealized benchmarks and embraces the full complexity of real-world autonomy and are truly ready for the open-world.

\newpage